\documentclass[journal]{IEEEtran}
\usepackage{amsmath,amsfonts}
\usepackage{algorithmic}
\usepackage{array}
\usepackage[caption=false,font=normalsize,labelfont=sf,textfont=sf]{subfig}
\usepackage{textcomp}
\usepackage{stfloats}
\usepackage{url}
\usepackage{verbatim}
\usepackage{graphicx}
\usepackage{cite}
\usepackage{capt-of}
\usepackage{hyperref}
\usepackage{colortbl}
\usepackage{pifont}
\newcommand{\cmark}{\ding{51}}%
\newcommand{\xmark}{\ding{55}}%
\usepackage{xcolor}
\definecolor{best}{HTML}{ff9999}  % Light yellow
\definecolor{second}{HTML}{ffcc99}  % Soft orange
\definecolor{third}{HTML}{fff6b2}  % Light yellow

\usepackage{multirow}
\newcommand*\circled[1]{\raisebox{.5pt}{\textcircled{\raisebox{-.9pt}{#1}}}}
\usepackage[ruled, vlined]{algorithm2e}
\usepackage{booktabs}
\usepackage{romannum}
\usepackage{etoolbox}
\AtBeginDocument{\pagenumbering{arabic}}

\begin{document}
\title{Geometry-Aware Scene Configurations for Novel View Synthesis}
\author{Minkwan Kim, %~\IEEEmembership{Staff,~IEEE,} % Be sure to use the \IEEEmembership command to identify IEEE membership status.
        Changwoon Choi,
        Young Min Kim
\IEEEcompsocitemizethanks{
    \IEEEcompsocthanksitem Minkwan Kim, Changwoon Choi, and Young Min Kim are with the Department of Electrical and Computer Engineering, Seoul National University, Seoul 08826, South Korea. E-mail: \{mkjjang3598, zzzmaster, youngmin.kim\}@snu.ac.kr.
    \IEEEcompsocthanksitem Young Min Kim is the corresponding author.
}
\thanks{Manuscript received December 19, 2025.}
}

\markboth{IEEE TRANSACTIONS ON VISUALIZATION AND COMPUTER GRAPHICS,~Vol.~XX, No.~XX, AUGUST~202X}%
{Kim \MakeLowercase{\textit{et al.}}: Geometry-Aware Scene Configurations for Novel View Synthesis}

\maketitle
\begin{abstract}

We propose scene-adaptive strategies to efficiently allocate representation capacity for generating immersive experiences of indoor environments from incomplete observations.
Indoor scenes with multiple rooms often exhibit irregular layouts with varying complexity, containing clutter, occlusion, and flat walls.
We maximize the utilization of limited resources with guidance from geometric priors, which are often readily available after pre-processing stages.
We record observation statistics on the estimated geometric scaffold and guide the optimal placement of bases, which greatly improves upon the uniform basis arrangements adopted by previous scalable scene representations.
We also suggest scene-adaptive virtual viewpoints to compensate for geometric deficiencies inherent in view configurations in the input trajectory and impose the necessary regularization.
We present a comprehensive analysis and discussion regarding rendering quality and memory requirements in several large-scale indoor scenes, demonstrating significant enhancements compared to baselines that employ regular placements.
Project page is available at: \url{https://mkjjang3598.github.io/Geo-Scene-Config}.
\end{abstract}    

\begin{IEEEkeywords}
Neural Radiance Fields, Novel view synthesis, Free view synthesis, Indoor scene reconstruction
\end{IEEEkeywords}

\begin{figure*}
    \centering
    \includegraphics[width=\linewidth]{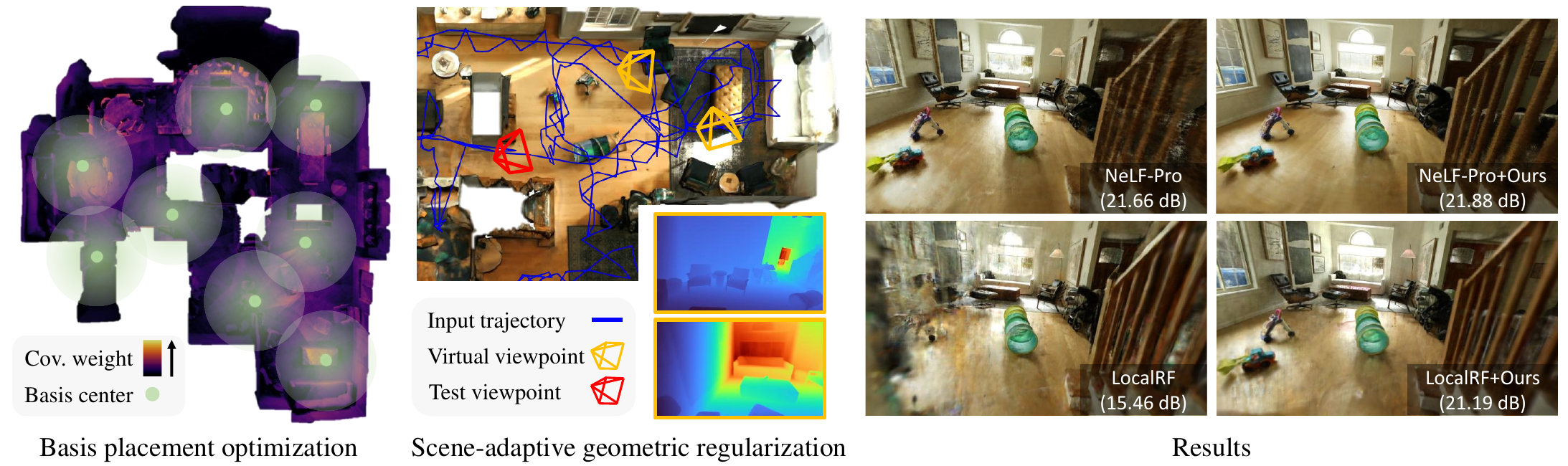}
    \caption{\textbf{Geometry-aware basis configuration and novel view synthesis results.} After recording measurement statistics as coverage weights on geometry scaffold, we find optimal basis positions and training-view configurations in indoor environments.}
    \label{fig:teaser}
\end{figure*}

\section{Introduction}
\label{sec:intro}

\IEEEPARstart{R}{ecent} advances in Neural Radiance Fields (NeRF)~\cite{nerf} enable casual users to obtain an immersive experience of observing 3D objects or scenes only from image observations. 
The geometric information is implicitly contained in a pre-defined volumetric structure that encodes the spatial distribution of density and radiance in a highly efficient manner. 
The typical benchmark datasets assume regular distribution of measurement rays, e.g., forward-facing views of an object in front, or surrounding views of objects within a bounding box.
With sufficient regularization from multi-view consistency, the novel-view synthesis results demonstrate impressive performance in nearby views. 
However, hand-held trajectories capturing everyday environments yield highly irregular patterns of observations, which often lead to degraded performance.
Modeling large-scale indoor scenes with sparse observations from casual capture, therefore, entails unconventional geometric configurations for novel-view synthesis, motivating a systematic approach that accounts for the interplay between the underlying geometry and the pre-captured camera viewpoints.

% Using geometry is natural and good choice, adaptive to scenes, significantly increase efficiency
% Given the sparse observation of hand-held cameras, modeling large-scale indoor scenes may benefit from a systematic approach that accounts for the interplay between the geometric prior and camera viewpoints.
In many real-world scenarios, an approximate scene geometry is readily available. 
The common practice for novel-view synthesis is to first acquire measurements and extract camera poses via a structure-from-motion (SfM)~\cite{colmap_mvs} pipeline to train the neural volume offline.
During the geometric optimization, we can obtain geometric scaffolds using existing methods, such as neural implicit surface reconstruction methods~\cite{volsdf, monosdf} or feed-forward models~\cite{vggt, MAPAnything}.

\begin{figure}
    \centering
    \includegraphics[width=\linewidth]{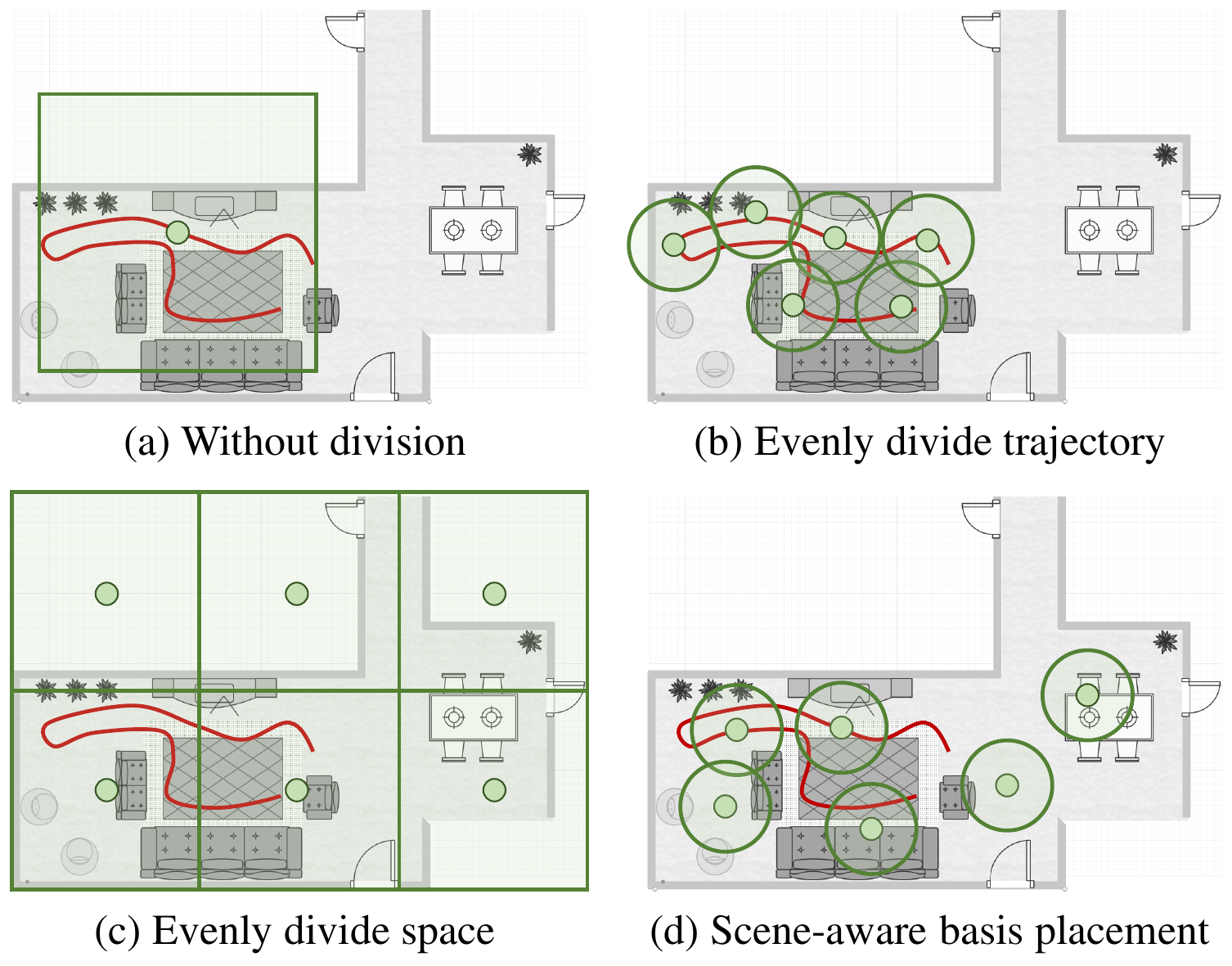}
    \caption{\textbf{Scene-splitting strategies.} 
    (a) The original NeRF represents the entire scene with a single block. Scalable scene representations set multiple bases (b) evenly along the camera trajectory or (c) uniformly dividing the scene's spatial extent. (d) We propose an adaptive approach based on scene configurations. Red curves denote camera trajectories and green dots mark the bases' centers. 
    }
    \label{fig:scene_divide_strategies}
\end{figure}

In this work, we exploit such geometric priors to develop scene-adaptive strategies that significantly enhance reconstruction quality under resource constraints. 
NeRF representations, along with various data structures~\cite {instant-ngp, plenoxels, plenoctrees}, have limited expressive power to encompass the infinite resolution of large-scale scenes.
Simply expanding the spatial range to unbounded scenes~\cite{nerf++, mip-nerf360, egonerf} cannot efficiently account for the complex layouts. 
As illustrated in Figure~\ref{fig:scene_divide_strategies}, several previous works demonstrated distributing spatial entities to represent the whole scene~\cite{block-nerf, mega-nerf, nerf-xl, localrf, nelf-pro}.
We refer to spatial entities, such as NeRF blocks or feature grids, as bases, which may suffer from limited expressiveness in isolation. 
They have been allocated 1) based on camera trajectory~\cite{localrf, nelf-pro} or 2) by dividing scenes evenly in space \cite{block-nerf, mega-nerf, nerf-xl}, which considers neither the non-uniform nature of indoor geometry nor the measurements.
Instead, we propose a scene-adaptive distribution of bases optimized to cover the input trajectory on the estimated geometry scaffolds. 
The resulting spatial arrangements of bases account for both scene geometry and measurement statistics, yielding an efficient scene representation. 
We demonstrate our experimental setup and method with example results in Figure~\ref{fig:teaser}.

%Optimal reconstruction to overcome missed views using geometry
Geometric configuration can further guide us in applying additional regularization to suppress artifacts in regions with insufficient measurements.
% Given the geometric scaffold and camera measurements, we further introduce a scene-adaptive geometric regularization strategy to guide optimization for desired regions.
In practice, observations collected along hand-held trajectories rarely cover the entire scene exhaustively, especially when navigating through complex layouts. 
Furthermore, indoor scenes often contain flat, homogeneous walls, which are highly susceptible to prominent artifacts in novel-view synthesis.
As a result, large portions of the scene remain weakly constrained, often leading to severe artifacts in novel-view synthesis.
To address this issue, we leverage the obtained geometric scaffold to assess the deficiency within the acquired representation and locate virtual views that efficiently eliminate the prominent artifacts, as suggested by next-best-view planning \cite{activenerf, progressive_camera_placement, NeRF_Director}.
Then, we extract depths in the selected virtual views to provide plausible supervision in unobserved regions \cite{regnerf, vip-nerf}.
The regularization provides additional guidance in sparsely observed regions, leading to improved reconstruction quality with stable optimization.

While recent 3D Gaussian Splatting (3DGS) methods~\cite{3dgs, Mip-Splatting, Scaffold-GS, IndoorGS, PlanarGS} offer compelling rendering speed, our work targets offline reconstruction quality for large-scale indoor scenes, where NeRF-based representations retain distinct advantages. As scene complexity scales, Gaussian-based methods must store a large number of primitives, leading to significant memory overhead, whereas NeRF models maintain a more compact representation. Furthermore, the discrete nature of Gaussian primitives often introduces floaters or blurring artifacts in textureless or sparsely observed regions—conditions that are prevalent in large-scale indoor environments and become especially pronounced when rendering views that extrapolate beyond the observed camera trajectory. The continuous volumetric formulation of NeRF provides more stable and geometrically consistent rendering under such challenging settings, motivating our choice of NeRF-based scalable representations as the foundation of this work.

%  (Contribution) 
Our scene-adaptive strategies are applicable to various representations for novel-view synthesis and greatly enhance the quality of results of the given measurement trajectory.
Our key contributions are summarized as follows:
\begin{itemize}
\item We propose efficient scene adaptation and training techniques for novel-view synthesis from pre-defined camera trajectories in large-scale indoor environments.
\item We successfully accumulate the statistics of input observations within energy functions defined on the geometric scaffold from which we can optimize the positions of a limited set of bases.
\item We guide optimization for sparsely captured regions using scene-aware virtual viewpoints, seamlessly blending them into a coherent representation. % environment-adaptive / scene-aware / context-aware / spatially adaptive
\end{itemize}
We evaluate our approaches using two representative basis sets~\cite{nelf-pro, localrf} on real-world indoor scenes from the ScanNet++~\cite{scannetpp} and Zip-NeRF~\cite{zip-nerf} datasets and demonstrate effective performance enhancements in practical setups.

\begin{figure*}[t]
    \centering
    \includegraphics[width=\linewidth]{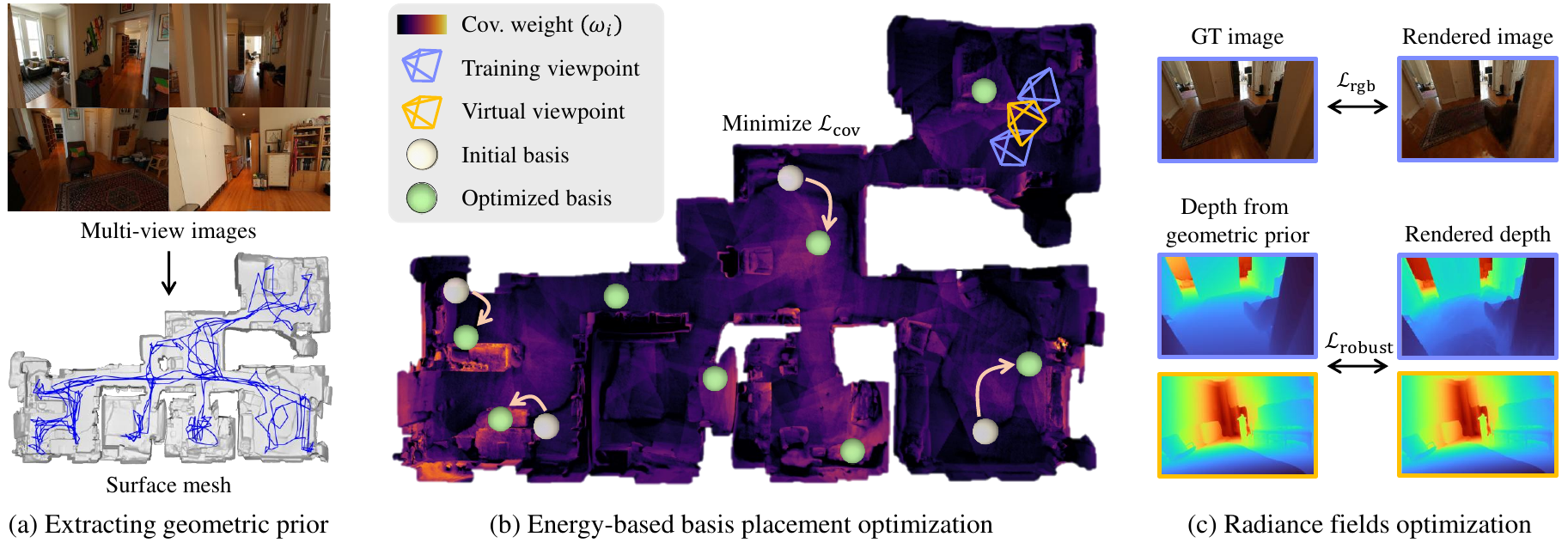}
    \caption{
    \textbf{Method overview.} (a) We first extract the geometric scaffold from multi-view image observation.
    (b) Then we define coverage weights $w_i$ for surface points $\mathbf{x}_i$ on the obtained mesh surface, which consider both scene geometry and observation statistics.
    Starting from bases sampled along the camera trajectory using the FPS algorithm, we optimize their positions by minimizing the energy function $\mathcal{L}_{\text{cov}}$.
    (c) Finally, we optimize radiance fields with RGB supervision, guided by geometric regularization on training and virtual viewpoints.
    }
    \label{fig:method_overview}
\end{figure*}
    
\section{Related Work}
\label{sec:related_work}

\subsection{Scalable Scene Representation} \label{related_scalable_scene_representation}
While NeRF representations demonstrate remarkable performance in representing 3D scenes, they are often limited in their ability to express large-scale scenes accurately.
To model large-scale, unbounded scenes, recent approaches constitute the scene with several units of spatial structure, where each unit is modeled by a separate neural network \cite{block-nerf, mega-nerf, nerf-xl, localrf}.
To further support efficient training and rendering as the representation scales, other recent works employ compact data structures for each local unit, such as feature grids~\cite{nelf-pro} or hash-based encodings~\cite{f2nerf}.
Despite these advances, relatively little work explores how to locate subsets of radiance fields, which we refer to as bases. 
Common approaches either regularly place bases along the camera trajectories~\cite{localrf, nelf-pro} or uniformly divide the scene region \cite{block-nerf, mega-nerf, nerf-xl}.
On the other hand, recent novel-view synthesis methods employ explicit 3DGS representations~\cite{3dgs} which achieve faster rendering speed at the expense of a larger memory footprint.
Variants of 3DGS also explore scalable scene representations by organizing Gaussians in hierarchical structures or spatial partitioning, enabling efficient culling and level-of-detail rendering for large-scale scenes~\cite{hierarchical_3dgs, LongSplat}.
However, these approaches typically rely on predefined spatial hierarchies or density-driven subdivision, with limited consideration of camera–geometry correlations when allocating representational units.
Instead, we adaptively consider the correlation between the scene geometry and camera observation to enhance the performance given limited resources.

\subsection{Scene-Adaptive Capture and Reconstruction} \label{related_next-best-view planning}
The reconstruction quality of 3D representation depends heavily on the viewpoints from which the measurements are collected.
Motivated by 3D scanning applications, next-best-view planning designs the best viewpoints to scan a 3D scene~\cite{NeRF_Director}.
While pursuing uniform coverage can be a general approach~\cite{progressive_camera_placement}, more advanced methods achieve high-quality reconstruction by minimizing the uncertainty based on the estimated underlying scene geometry~\cite{activenerf, neural_visibility_field}.
%Several works also demonstrate performance enhancement in neural 3D scene representation by allocating more resources to regions with finer details \cite{acorn, plenoctrees, plenoxels, nsvf}.
Beyond viewpoint selection, several works have shown that reconstruction fidelity can be further improved by allocating representation capacity adaptively across the scene.
By dedicating more resources to geometrically complex or high-frequency regions, these methods achieve improved efficiency and quality in neural scene representations~\cite{acorn, plenoctrees, plenoxels, nsvf}.
Similarly, recent 3DGS variants~\cite{Scaffold-GS, LongSplat, SGGS} employ anchor-based representations to adaptively allocate Gaussian primitives from an initial sparse voxelized point cloud. 
This design improves memory efficiency by avoiding redundant primitives while maintaining high reconstruction fidelity.
In this work, we suggest a scene-adaptive strategy for free-camera trajectory input, where we optimize the spatial arrangements for both scene representation and training views.

\subsection{Geometric Priors for Novel-View Synthesis} \label{related_geometric_regularization}
Fitting 3D scene structure is prone to overfitting and under-constrained conditions, leading to significantly degraded rendering quality for viewpoints far from input observations.
Recent approaches notably reduce floating artifacts with monocular depth estimation \cite{midas, dpt}, which may not be multi-view consistent.
Possible misalignments can be adjusted by modeling scale ambiguities \cite{nope-nerf, monosdf} or multi-modality of estimations \cite{scade}. 
On the other hand, coarse depths obtained from SfM~\cite{colmap_sfm} or depth sensors are noisy but can guide building the coherent scene representation.
Groups of works~\cite{ddp, dsnerf} model uncertainty for these settings, while SparseNeRF \cite{sparsenerf} enables  
regularization by only comparing the relative depth.
NeRFVS \cite{nerfvs} introduces the robust depth loss term for noisy and dense depth supervision.
In parallel, recent 3DGS representations have also explored geometric regularization strategies.
Similar to NeRF-based approaches, several works~\cite{dngaussian, FSGS} incorporate depth-based regularization to stabilize training and improve geometric fidelity.
Another line of work~\cite{PlanarGS, IndoorGS} target structured indoor environments by leveraging strong geometric priors, such as planar constraints to improve reconstruction quality.
Beyond regularization with multi-view inconsistent depth maps, neural implicit surface reconstruction methods~\cite{volsdf,monosdf} leverage signed distance fields to enforce geometric consistency, providing strong priors for novel-view synthesis.
More recently, feed-forward models~\cite{vggt, MAPAnything} offer scalable and consistent geometric priors by exploiting large-scale pretraining, which can serve as an effective geometric scaffold for downstream rendering tasks.
Building on these insights, we propose a geometric regularization method that leverages a 3D scaffold and adaptive virtual viewpoints, enabling robust training under limited views.
\section{Method}
\label{sec:method}

Given frames of a hand-held video capturing a large indoor scene, our approach first extracts geometric scaffolds (Section~\ref{subsec: obtaining_geometric_scaffolds}).
% Given free-trajectory images capturing a large indoor scene, our approach first extracts geometric scaffolds (Section~\ref{subsec: obtaining_geometric_scaffolds}).
We use geometric scaffolds to optimize basis placements (Section~\ref{subsec:method_basis_placement}), where each basis comprises a local scene representation as described in Section~\ref{related_scalable_scene_representation}.
Then, based on geometry and observation statistics, we sample virtual views to be regularized and train radiance fields with depth supervision on both input and virtual views (Section~\ref{subsec:method_geometric_regularization}).
Importantly, rather than introducing a new scene representation, our approach focuses on a principled formulation that couples geometric scaffolds, adaptive basis placement, and geometry-aware regularization. 
This systematic integration is crucial for addressing the non-trivial challenges of allocating representation capacity in large-scale scenes.

%% 3.1) Geometric priors : preliminaries of NeRF and neural implicit surface representation (monosdf)
\subsection{Preliminaries}
\label{subsec: preliminaries}
NeRFs train a neural network that estimates a mapping function from 3D location $\mathbf{x}\in \mathbb{R}^3$ and viewing direction $\mathbf{d}\in \mathbb{R}^2$ to volume density $\sigma \in \mathbb{R}^+$ and color $\mathbf{c} \in \mathbb{R}^3$. 
Given $N$ discrete points sampled along the ray $\mathbf{r}$, the color is rendered by the volume rendering equation given as follows:
\begin{equation}
    \hat{C}(\mathbf{r})=\sum_{i=0}^{N-1}T_i(1-\text{exp}(-\sigma_i\delta_i))\mathbf{c}_i,
    \label{eq:volume_rendering}
\end{equation}
where $T_i=\text{exp}(-\sum_{j=0}^{i-1}\sigma_j\delta_j)$ and $\delta_i$ denotes the distance between ray samples.
The training loss for the network is the color %discrepancy 
difference between the rendered color $\hat{C}$ and the pixel value of the training image $C$ as follows:
\begin{equation}
    \mathcal{L}_{\text{rgb}}= \lVert C(\mathbf{r})-\hat{C}(\mathbf{r})\rVert^2_2 .
    \label{eq:rgb_loss}
\end{equation}

In addition to color, depth information can also be estimated via the volume rendering formulation.
Similar to Equation~\ref{eq:volume_rendering}, we compute the expected depth along each ray as:
\begin{equation}
     \hat{D}(\mathbf{r})=\sum_{i=0}^{N-1}T_i(1-\text{exp}(-\sigma_i\delta_i))t_i,
    \label{eq:depth_rendering}
\end{equation}
where $t_i$ is the distance from the sample to the origin.
With input depths, we can apply additional depth loss as
\begin{equation}
    \mathcal{L}_{\text{depth}}= \lVert D^*(\mathbf{r})-\hat{D}(\mathbf{r})\Vert^2_2.
    \label{eq:depth_loss}
\end{equation}
$D^*=\alpha D+\beta$ denotes undistorted depth, where $\alpha$ and $\beta$ are scale and shift factors defined per frame. 

\subsection{Obtaining Geometric Scaffolds}
\label{subsec: obtaining_geometric_scaffolds}

To train the representation for novel-view synthesis, the preprocessing stage must estimate the camera poses of the input image sequences. 
Our framework leverages camera poses to derive geometric scaffolds, which allow the subsequent scene-adaptive strategies.
We demonstrate the framework using geometric scaffolds from two approaches: an implicit surface model and a feed-forward depth predictor.
MonoSDF~\cite{monosdf} is an implicit surface model, trained similarly to NeRFs but predicts signed distance fields (SDFs) and colors.
After training, we extract the surface mesh using the Marching Cubes algorithm \cite{marching_cubes} as illustrated in Figure~\ref{fig:method_overview}(a).
The geometric scaffold is a smooth, consistent estimate of the underlying surface, but requires additional optimization of the implicit representation.
Additionally, we employ MapAnything~\cite{MAPAnything}, a pre-trained feed-forward model that generates globally consistent depth estimates within tens of seconds.
This approach directly reconstructs dense meshes without requiring per-scene optimization, significantly reducing computational overhead. %; although they may not be globally consistent, they provide a useful geometric prior.
As will be discussed later, our framework consistently improves efficiency and reconstruction quality across both approaches, demonstrating versatility across various representations of geometric scaffolds. 

\begin{figure}[t]
    \centering
    \includegraphics[width=\linewidth]{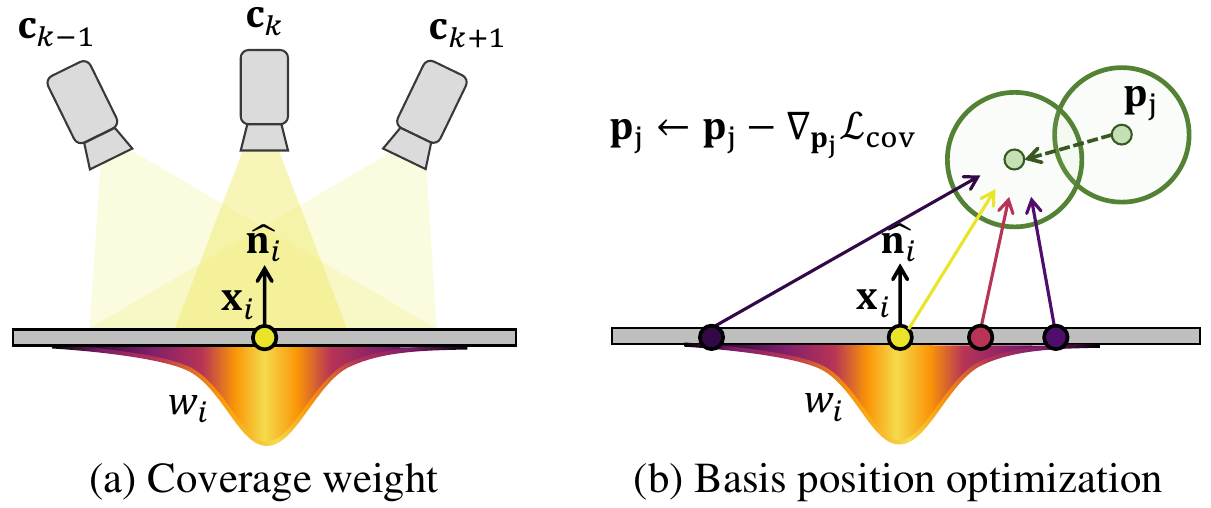}
    \caption{\textbf{Basis placement pipeline.} (a) After accumulating coverage weights $\mathbf{w}_i$ along surface points $\mathbf{x}_i$, (b) bases positions are updated using these weights and their distance from the surface.}
    \label{fig:basis_optimization}
\end{figure}

%% 3.2) Basis Placement Optimization : we consider correlated information of geometry and camera trajectory to optimize basis position
\subsection{Basis Placement Optimization}
\label{subsec:method_basis_placement}

This section introduces adaptively optimizing the basis distribution based on the scene geometry and camera trajectories, where the basis contributes to modeling local radiance fields.
We define the coverage weights $w_i$ that summarize the observation statistics on the geometric scaffold obtained from Section~\ref{subsec: obtaining_geometric_scaffolds}.
At a point $\mathbf{x}_i \in \mathbb{R}^3$ sampled from the scaffold, the corresponding coverage weight $w_i$ is calculated as 
\begin{equation}
    w_i=\sum_{k}\left(\frac{\hat{\mathbf{n}_i}\cdot(\mathbf{c}_k-\mathbf{x}_i)}{\left\Vert \mathbf{c}_k-\mathbf{x}_i 
    \right\Vert^3}\right) \, \text{Vis}(\mathbf{c}_k\rightarrow \mathbf{x}_i).
    \label{eq:coverage_weight}
\end{equation}
Vis$(\mathbf{c}_k\rightarrow \mathbf{x}_i)\in\{0,1\}$ denotes the point $\mathbf{x}_i$'s visibility from the camera $\mathbf{c}_k$.
% (CVPR version) denotes the visibility from the camera $\mathbf{c}_k$ to the observed point $\mathbf{x}_i$ considering occlusion, camera frustum, and surface normal directions.
The weight term accumulates observations of the surface points from cameras, whose weights are inversely proportional to the squared distance to cameras and proportional to the inner product between the viewing vector and the surface normal $\hat{\mathbf{n}}_i$. Figure~\ref{fig:basis_optimization} (a) illustrates the process of obtaining the coverage weight. 

Then, we optimize bases positions $\mathbf{p}_j \in \mathbb{R}^3$ to best describe the accumulated coverage.
We define the coverage-based loss term as follows: 
\begin{equation}
    \mathcal{L}_{\text{cov}} = \sum_{i}w_i\left(\frac{\left\Vert \mathbf{p}_{\mathcal{C}(i)}-\mathbf{x}_i 
    \right\Vert^3}{\hat{\mathbf{n}_i}\cdot(\mathbf{p}_{\mathcal{C}(i)}-\mathbf{x}_i)+\epsilon}\right),
    \label{eq:coverage_loss}
\end{equation}
where $\epsilon\approx0$ was added to the denominator for stable optimization.
The loss term penalizes distances between the basis and the mesh points, weighted by the observation frequency using Equation~\ref{eq:coverage_weight}.
We only consider the nearest basis position $\mathbf{p}_j$ from the surface point $\mathbf{x}_i$, where the $ \hat{\mathbf{n}_i}\cdot(\mathbf{p}_j-\mathbf{x}_i)>0$, indicated as $j=\mathcal{C}(i)$.
See the supplementary material for detailed information on Equation~\ref{eq:coverage_loss}.
During the optimization process illustrated in Figure~\ref{fig:basis_optimization}(b), each basis element shifts toward the frequently observed regions 
such that the fixed capacity of bases is distributed to balance covering densely populated ray samples.
Therefore, our basis optimization strategy has advantages in covering measurements of indoor environments with limited resources.

The optimized basis positions $\mathbf{p}_j$ serve as the position of spatial elements that constitute global radiance fields in parts.
We demonstrate results with two representative basis configurations by mapping $\mathbf{p}_j$ as the centroid of each NeRF block~\cite{block-nerf, mega-nerf, nerf-xl, localrf} or neural light field probe position of the NeLF-Pro \cite{nelf-pro} representation.
In both cases, a set of bases is assigned to regional partitions of the scene extent, and we can benefit from adaptive placement that achieves collaborative coverage.
See Figure~\ref{fig:method_overview}(b) for initial and optimized basis positions, and the supplementary material contains additional details and illustrations.

%% LocalRF 
\subsubsection{NeRF blocks}
We modify the block locations to evaluate our placement strategy for multiple NeRF blocks.
We set centroids of NeRF blocks at scene-adaptive basis positions while LocalRF~\cite{localrf} sequentially assigns NeRF blocks along the camera trajectory as shown in Figure~\ref{fig:scene_divide_strategies}(b).
Other works suggest uniformly partitioning the spaces to distribute the blocks~\cite{block-nerf, mega-nerf, nerf-xl}, which are demonstrated in Figure~\ref{fig:scene_divide_strategies}(c).
The blocks are trained independently from rays emanating from the original input camera poses.
We do not incorporate joint pose optimization to focus on the effect of block placements that respect ray distributions.
As discussed in the result section, our proposed basis configuration can better describe necessary scene details contained within the input measurements.

While the detailed implementation for individual blocks may vary \cite{block-nerf, mega-nerf, nerf-xl, localrf}, we follow the original implementation in LocalRF \cite{localrf} and represent the radiance as the sum of factorized tensor products of TensoRF \cite{tensorf} with the scene contraction of Mip-NeRF 360 \cite{mip-nerf360}.
Radiances in the overlapping area are blended from multiple blocks whose blending weights are inversely proportional to the distance between the camera position and centroid, following the Block-NeRF \cite{block-nerf} implementation.
We blend radiance values from, at most, two nearest NeRF blocks.

\subsubsection{Neural lightfield probes}
NeLF-Pro~\cite{nelf-pro} is another approach that leverages local radiance fields to describe large-scale scenes.
They decompose the scene representation into vector-matrix-matrix (VMM) factorization, also motivated by TensoRF~\cite{tensorf}.
Specifically, core factors (i.e., VM) encode the global features and are combined with basis factors (i.e., M) that encode local features.
These factors are assigned as probes distributed in 3D space, whose locations are uniformly selected from input camera poses.
The core factors are located at clustered positions resulting from K-means clustering among input poses, whereas the basis factors are positioned by the farthest point sampling (FPS).
We propose the basis probe factors to be located at the optimized basis positions $\mathbf{p}_j$, which jointly considers the scene layout and the measurement statistics.
We allocate core factors to be at the cluster centers after applying K-means clustering on the basis probe locations.

When rendering, NeLF-Pro aggregates features from a set of probes near the rendering view, which are decoded to volume densities and colors by a shallow MLP.
In our setting, we find the closest set of probes for every sample on the ray, which may change within the same ray.
We exploit the local radiance distribution specialized to the probes with sorting operations along the ray samples.
Unlike FPS-based probe allocation, simply calculating nearby probes based on camera positions does not ensure full usage of probes on rendering. 
Several probes are not selected if the probe is located near the surface region far from the camera position.
Hence, to fully utilize the given resources, determining nearby probes per each ray sample is necessary.
% With PyTorch sorting implementation, only about an additional 0.3ms per optimization step is required for this modification on the NVIDIA GeForce RTX 4090 GPU.

\begin{figure}[t]
    \centering
    \includegraphics[width=\linewidth]{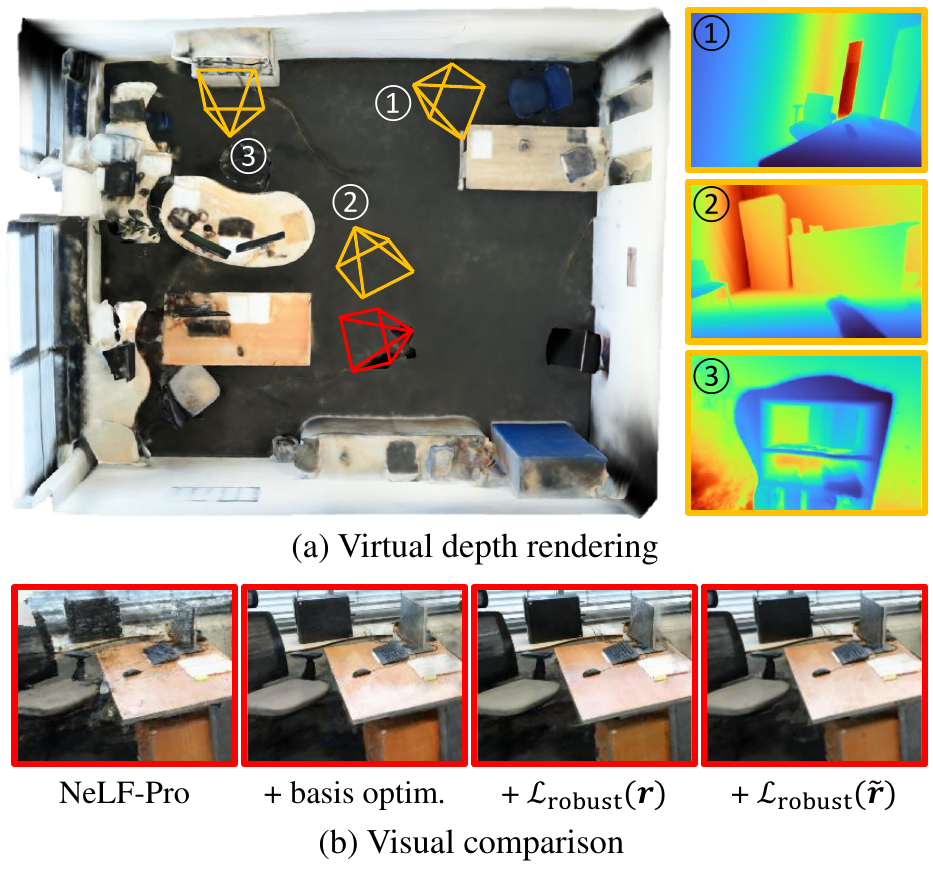}
    \caption{\textbf{Novel view geometric regularization.} (a) The illustration of selected virtual views (yellow) with rendered depth images on the right. (b) Test viewpoint images (red) show the effect of each component, added sequentially from left to right. 
    }
    \label{fig:novel_depth_regularization}
\end{figure}

%% 3.3) Scene-Adpative Geometric regularization : as camera trajectories are not able to cover all the desired region for novel view synthesis, we employ an additional geometric regularization.
\subsection{Scene-Adaptive Geometric Regularization} \label{subsec:method_geometric_regularization}
We additionally incorporate geometric regularization to reduce floating artifacts, which are prominent in extreme viewpoint changes.
While noisy, the estimated geometric scaffold from Section~\ref{subsec: obtaining_geometric_scaffolds} can impose depth regularization to improve reconstruction quality.
We use robust depth loss in Yang et al.~\cite{nerfvs} instead of the L2 loss introduced in Equation~\ref{eq:depth_loss}: 
\begin{equation}
    \mathcal{L}_{\text{robust}}= 
    \begin{cases}
        0.5 \, \Delta D(\mathbf{r})^2, & \text{if } \Delta D(\mathbf{r}) < \gamma\\ 
        \gamma^2 \left(0.5 + \log (\frac{\Delta D(\mathbf{r})}{\gamma}) \right), & \text{otherwise}
    \end{cases}
    \label{eq:robust_depth}
\end{equation}
where $\Delta D(\mathbf{r})=|D^*(\mathbf{r})-\hat{D}(\mathbf{r})|$ and $D^*(\mathbf{r})$ is undistorted depth defined in Equation~\ref{eq:depth_loss}.

% Minkwan modification not covered -> rarely covered 
For each scene, we need additional regularization to estimate regions rarely covered by the input camera trajectories.
We include supervision from novel views to cover the underlying scene efficiently.
Starting from the original training cameras, we add viewpoints to build the set of training views $S$.
The new views are progressively added to $S$ from random samples $V\in\{v_1, v_2, \ldots,v_n\}$ extracted from empty space in the indoor region.
Motivated by Xiao et al.~\cite{NeRF_Director}, we define the selection criteria as follows:
\begin{equation}
    v^*=\operatorname*{arg\,max}_{v \in V \setminus S} \left[ \min_{s \in S} \, d(v, s) \right],   
\end{equation}
where $d(v,s)$ is the view distance defined as 
\begin{equation}
    d(v,s)=\left\Vert \mathbf{c}_v-\mathbf{c}_s\right\Vert^2+ \, \kappa\left(1-\frac{\mathcal{A}_{vs}}{\text{max}(\mathcal{A}_{vs})}\right).
    \label{eq:view_distance}
\end{equation}
The first term in Equation~\ref{eq:view_distance} considers the distance between the next view candidate $\mathbf{c}_v$ and the existing camera location $\mathbf{c}_s$, considering the spatial expansion.
The second term employs similarity matrix $\mathcal{A}_{vs}$ to account for scene diversity.
The similarity matrix $\mathcal{A}_{vs}$ contains the number of commonly observed feature points between view candidates $v \in V $ and sampled camera sets $s \in S$ such that the distance in Equation~\ref{eq:view_distance} increases between dissimilar viewpoints.
The features from the unseen views are estimated as projections of triangulated feature points from initial training sets, which are visible after considering the occlusion and camera frustums. The resulting depths from virtual viewpoints are demonstrated in Figure~\ref{fig:novel_depth_regularization}(a).

We use the combined set of viewpoints in $S$ to provide additional supervision with the robust depth loss defined in Equation~\ref{eq:robust_depth}.
The total loss function to train each basis representation is given as follows:
\begin{equation}    
    \mathcal{L}_{\text{total}}(\Tilde{\mathbf{r}})=\mathcal{L}_{\text{rgb}}(\mathbf{r})+\lambda_{\text{robust}}\mathcal{L}_{\text{robust}}(\Tilde{\mathbf{r}})+\lambda_{\text{reg}}\mathcal{L}_\text{reg}(\Tilde{\mathbf{r}}),
    \label{eq:total_loss}
\end{equation} 
where $\mathbf{r}$ are the rays in the training views and $\Tilde{\mathbf{r}}$ includes the sampled virtual views.
$\mathcal{L}_\text{reg}$ denotes other regularization terms, e.g., interlevel loss or distortion loss \cite{nelf-pro,mip-nerf360}.
Figure~\ref{fig:method_overview}(c) illustrates our geometric regularization, whose effects are compared in Figure~\ref{fig:novel_depth_regularization}(b) and the supplementary material.

\section{Experiments}
\label{sec:experiments}

\subsection{Experimental Setting}
\label{sec:experimental_setting}
\subsubsection{Dataset}
We evaluate our framework using two recent indoor scene datasets. 
(1) ScanNet++ \cite{scannetpp} includes high-quality training and test image datasets with camera poses.
We use images captured from DSLR cameras and undistorted them using the OpenCV python package~\cite{opencv}. 
Test sets are predefined and include both interpolation and extrapolation settings. 
We selected six scenes for evaluation: [\textsc{785e7504b9, ef69d58016, bb87c292ad, 0d2ee665be, e91722b5a3, 281ba69af1}].
(2) The Zip-NeRF dataset \cite{zip-nerf} comprises large-scale indoor and outdoor scenes with camera poses sampled from trajectories. 
We uniformly downsampled the dataset by a factor of 3, as the original contains extremely dense and redundant viewpoints, which substantially increase computational cost without providing proportional performance gains.
%We uniformly downsampled the dataset by a factor of 3 for evaluating multi-room indoor scenarios, as the original Zip-NeRF dataset contains extremely dense and highly redundant viewpoints, which substantially increase computational cost without providing proportional performance gains.
Test views are taken from every eighth frame from the sampled sequences.
Among the entire dataset, we evaluated four undistorted scenes: [\textsc{Berlin, NYC, Alameda, London}].

\subsubsection{Baselines}
We compare our method against recent approaches for 3D scene modeling and novel view synthesis, including neural implicit representation methods and 3DGS variants.
MonoSDF~\cite{monosdf} reconstructs neural implicit surfaces using RGB, depth, and normal supervision.
NeRFacto~\cite{nerfstudio} combines hash encoding and proposal sampling; depth-supervised and larger-capacity variants were also evaluated.
Zip-NeRF~\cite{zip-nerf} applies anti-aliasing on multi-scale hashed feature grids for fast, high-fidelity training.
3DGS~\cite{3dgs}, FSGS~\cite{FSGS}, and DNGaussian~\cite{dngaussian} represent state-of-the-art Gaussian splatting methods, with the latter two incorporating geometric regularization for sparse-view settings.
Mip-Splatting~\cite{Mip-Splatting}, Scaffold-GS~\cite{Scaffold-GS}, and PlanarGS~\cite{PlanarGS} extend 3DGS with anti-aliasing, structured primitives, and planar priors, respectively.
We further compare against scalable multi-block representations:
NeLF-Pro~\cite{nelf-pro}, LocalRF~\cite{localrf}, and Mega-NeRF~\cite{mega-nerf} partition scenes into multiple neural units for large-scale modeling.
Hierarchical-3DGS~\cite{hierarchical_3dgs} and LongSplat~\cite{LongSplat} scale Gaussian splatting to large scenes and long trajectories, respectively.
See the supplementary material for detailed configurations.

\subsubsection{Models}
In addition to details that were described in Section~\ref{subsec:method_basis_placement}, we elaborate on model configurations that were used throughout our experiments.
(1) To use the neural feature grids as basis, we modify the configurations of the original NeLF-Pro~\cite{nelf-pro}.
While optimizing NeLF-Pro models with novel view regularization, we sample additional rays from the novel viewpoint, so that the number of rays used in RGB supervision is maintained identical.
We used 64 basis and 3 core probes to model the ScanNet++ scenes while 16 near basis and 3 core probes are employed for volume rendering. 
For Zip-NeRF dataset, 128 basis and 3 core probes are used, while the same number of probes that were used in the ScanNet++ experiment contribute to image rendering.
All of the models, including the baselines, are trained for a total of 30000 steps.
(2) For multiple NeRF block optimization, we have tuned several parameters used in LocalRF~\cite{localrf}. 
We first removed Laplacian weights that were multiplied in the original implementation. 
As we optimize radiance fields without camera pose optimization, simple RGB supervision without Laplacian weighting worked better. 
We further remove flow loss and per-frame progressive optimization for all experiments, as those produce worse results for our dataset.
Next, we do not set the maximum number of frames per NeRF block which is set to 100 frames for the original LocalRF implementation. 
In other words, we only consider drifts when allocating new NeRF blocks. 
We finally adjusted the scene scale to allocate the number of blocks between 6 and 10 blocks, and increased iterations per frame from 600 to 1000 for experiments to ensure convergence.

\subsubsection{Hyperparameters} % Implementation Details
After obtaining geometric scaffolds, 8192 surface points were batchified, each contributing to the position updates of their nearest bases.
We used an ADAM optimizer with an initial learning rate of 0.0016, where a standard exponential scheduling strategy was used similar to that of Plenoxels~\cite{plenoxels} and 3DGS~\cite{3dgs}.
% geometric regularization
During geometric regularization, we use $\gamma=0.1$ in Equation~\ref{eq:robust_depth} for all experiments following Yang et al.~\cite{nerfvs}.
To regularize geometry using unobserved viewpoint, we sample $N=5000$ views for candidates and select $100$ views in addition to the initial training views, where we use $\kappa=0.1$ for view distance calculation in Equation~\ref{eq:view_distance}.
% Total loss
In Equation~\ref{eq:total_loss}, we used $\lambda_{\text{robust}}=0.005$ for NeLF-Pro and $\lambda_{\text{robust}}=0.01$ for LocalRF experiments. 
To determine these values, we searched $\lambda_\text{robust}$ for values between [0.001, 0.2], to maximize the validation image quality metrics. 
We set $\lambda_\text{robust}$ within a specific range ([0.001, 0.01] for NeLF-Pro and [0.005, 0.02] for LocalRF), where image quality does not change significantly.
For $\lambda_{\text{reg}}$, we follow the weights that were used in the original implementations, as $L_\text{reg}$ is not our contribution.

\begin{figure*}[p]
    \centering
    \begin{tabular}{c@{\,}c@{\,}c@{\,}c@{\,}c@{\,}c}
    \small GT& \small NeRFacto & \small Depth-NeRFacto & \small 3DGS & \small NeLF-Pro & \small NeLF-Pro+Ours$_{MA}$\\

    \includegraphics[width=0.158\linewidth]{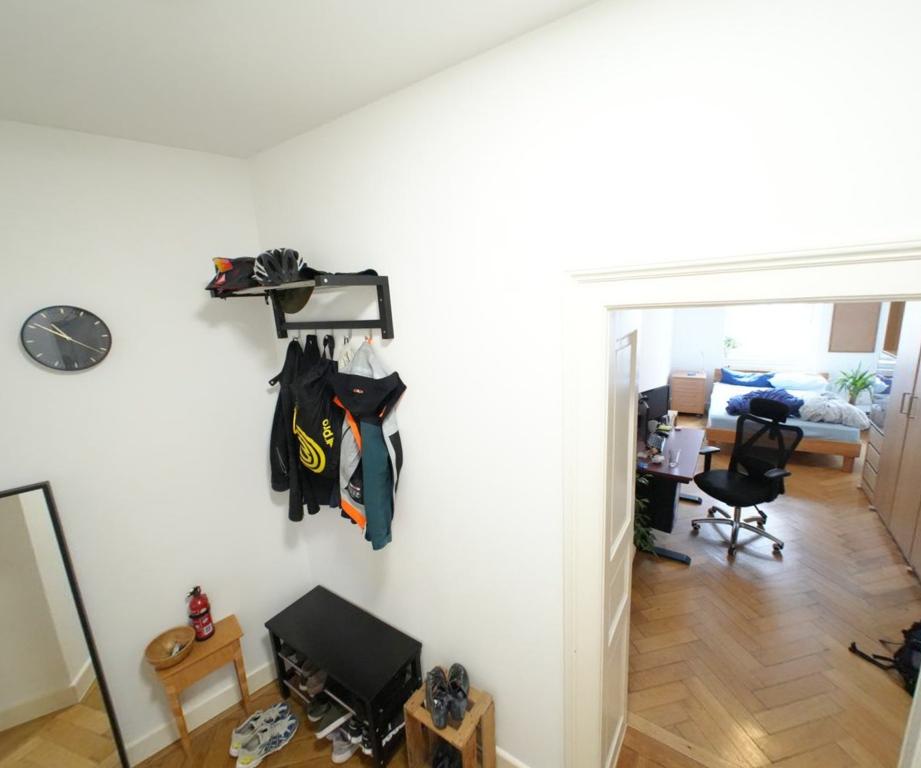}& 
    \includegraphics[width=0.158\linewidth]{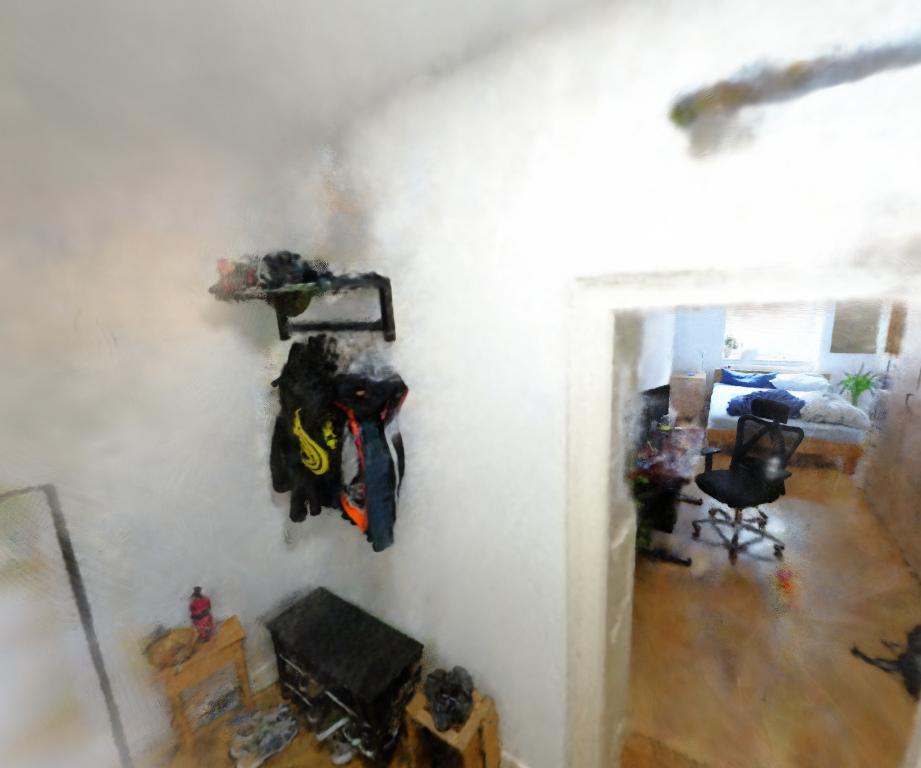}& \includegraphics[width=0.158\linewidth]{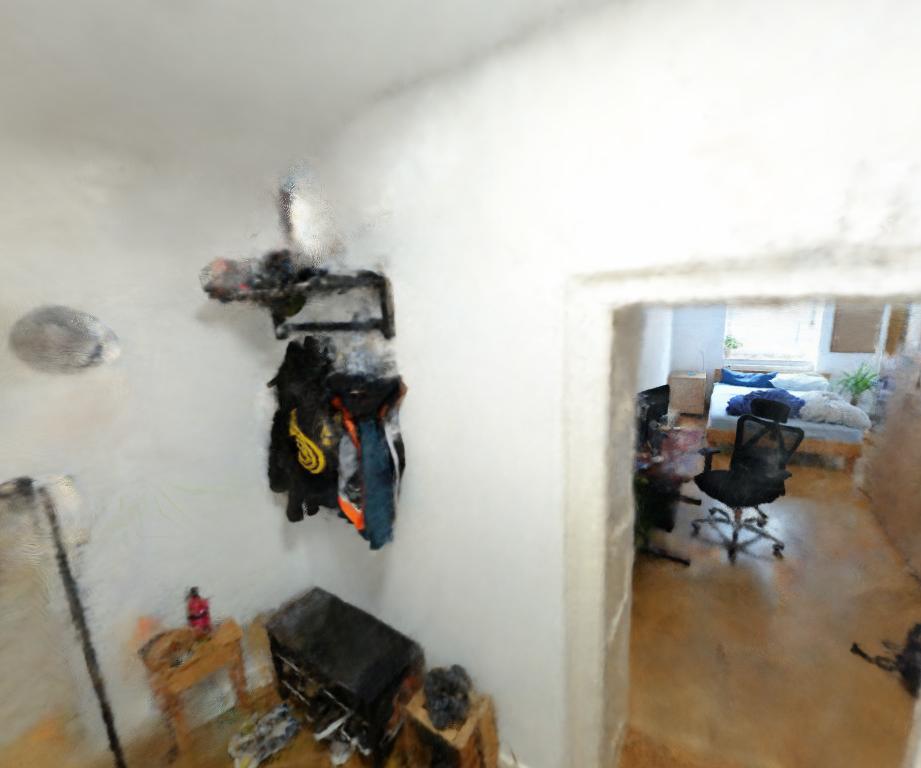}& \includegraphics[width=0.158\linewidth]{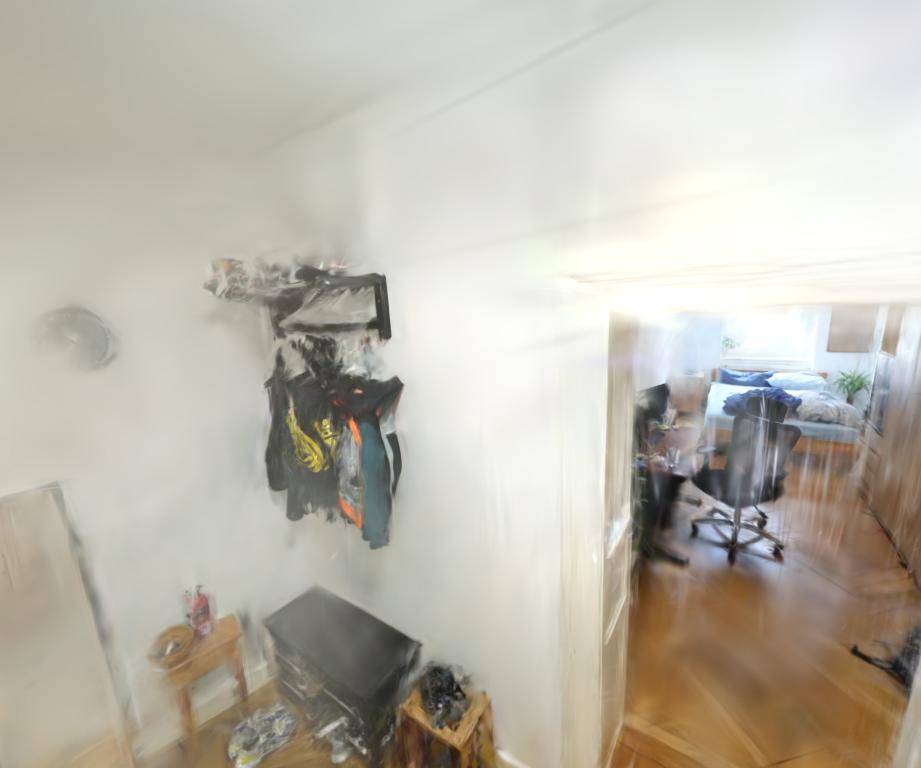}&
    \includegraphics[width=0.158\linewidth]{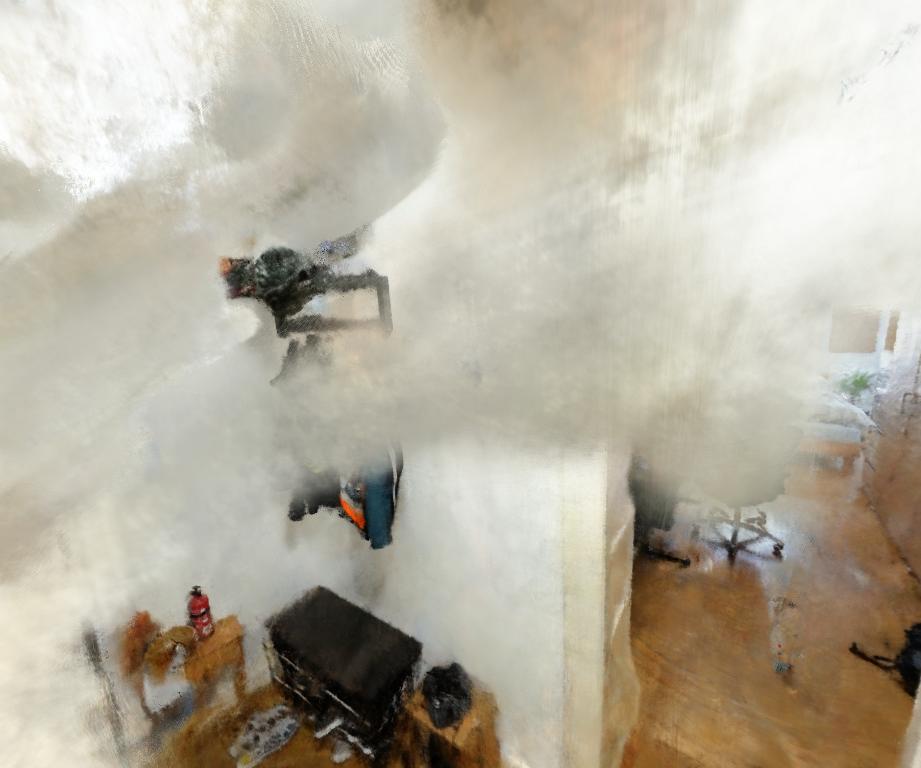}&
    \includegraphics[width=0.158\linewidth]{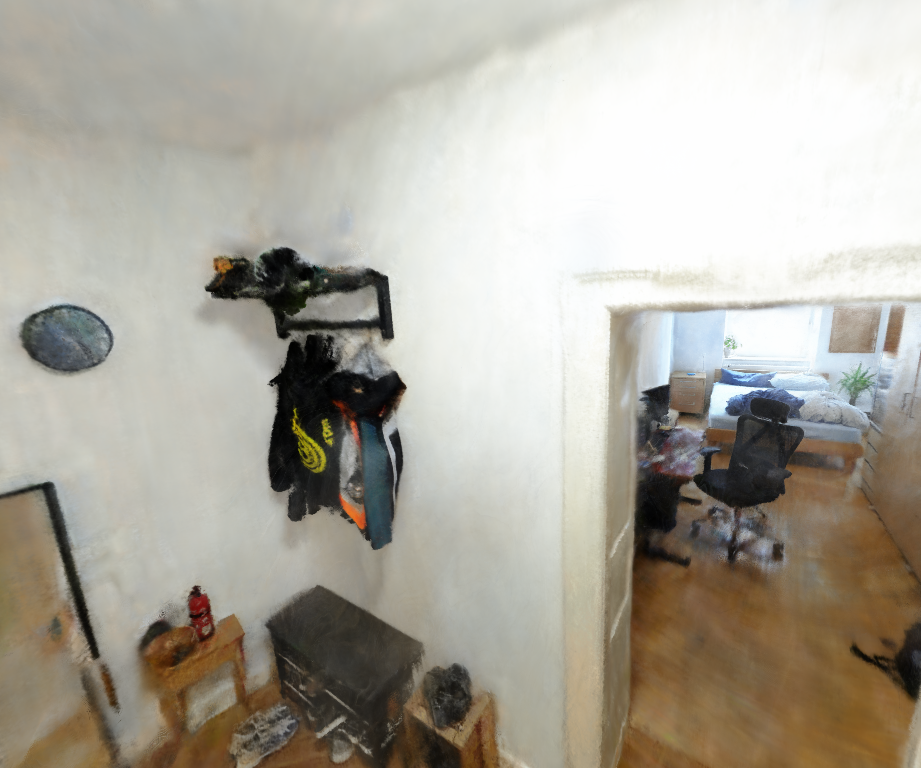}\\

    \includegraphics[width=0.158\linewidth]{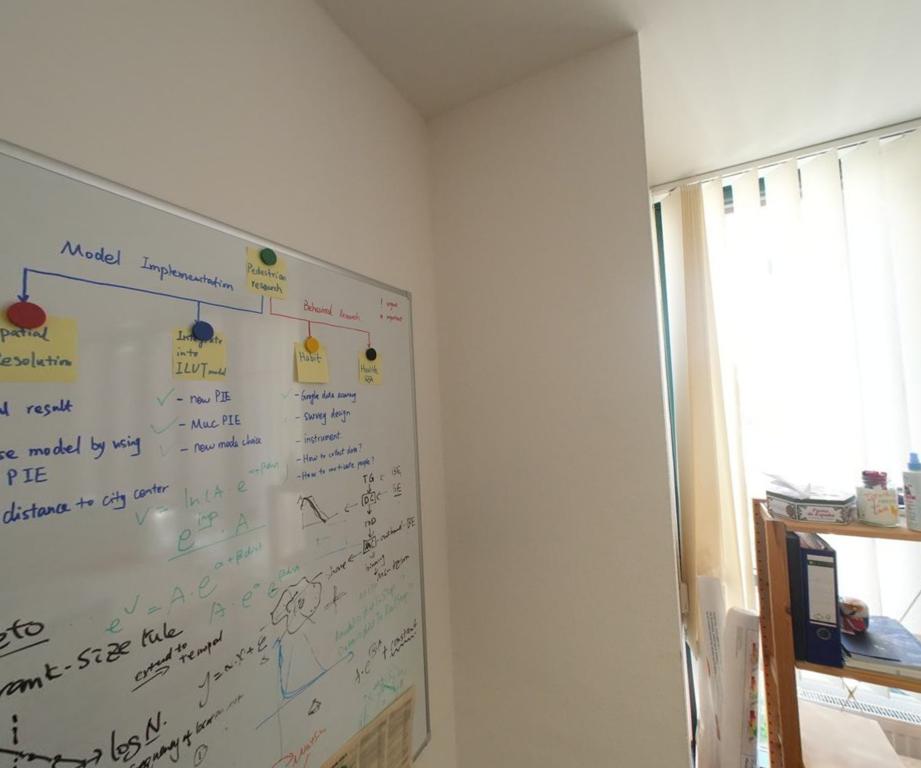}& 
    \includegraphics[width=0.158\linewidth]{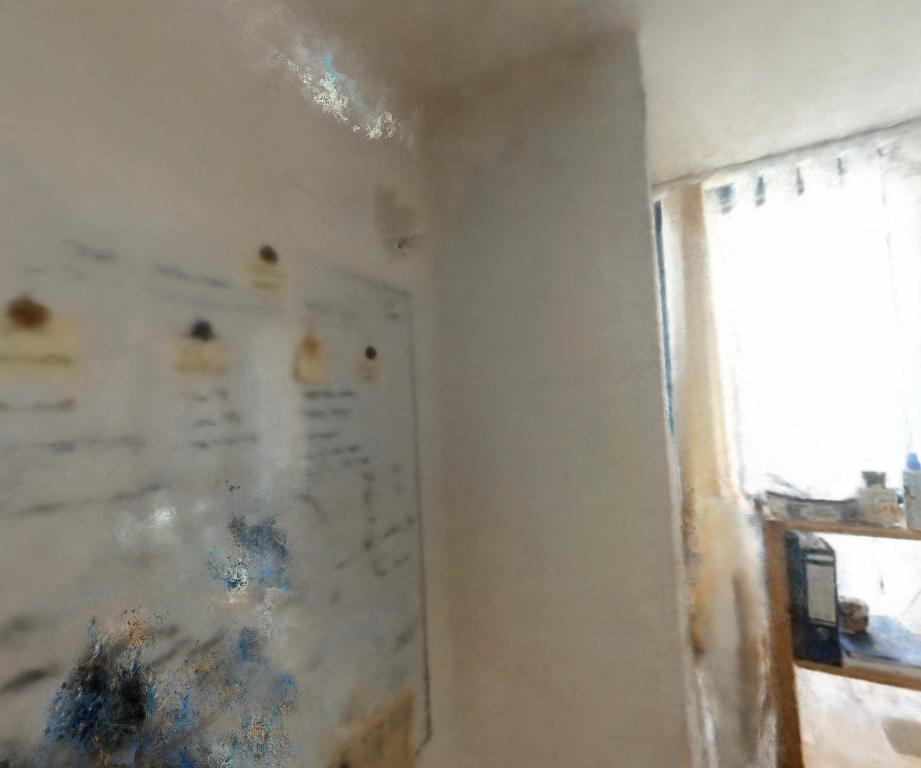}& \includegraphics[width=0.158\linewidth]{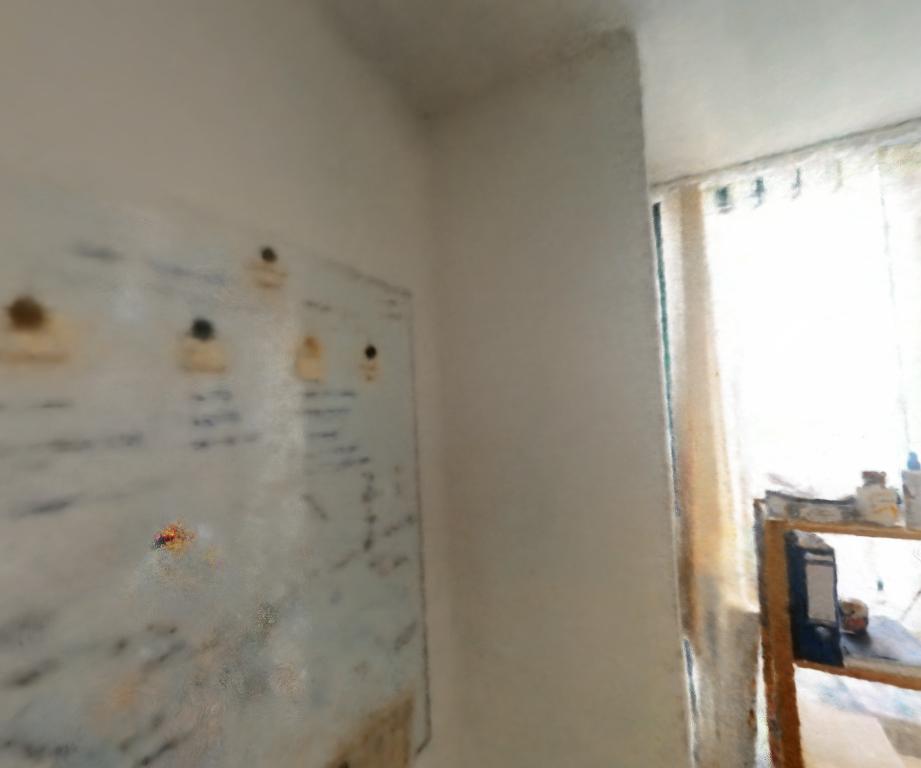}& \includegraphics[width=0.158\linewidth]{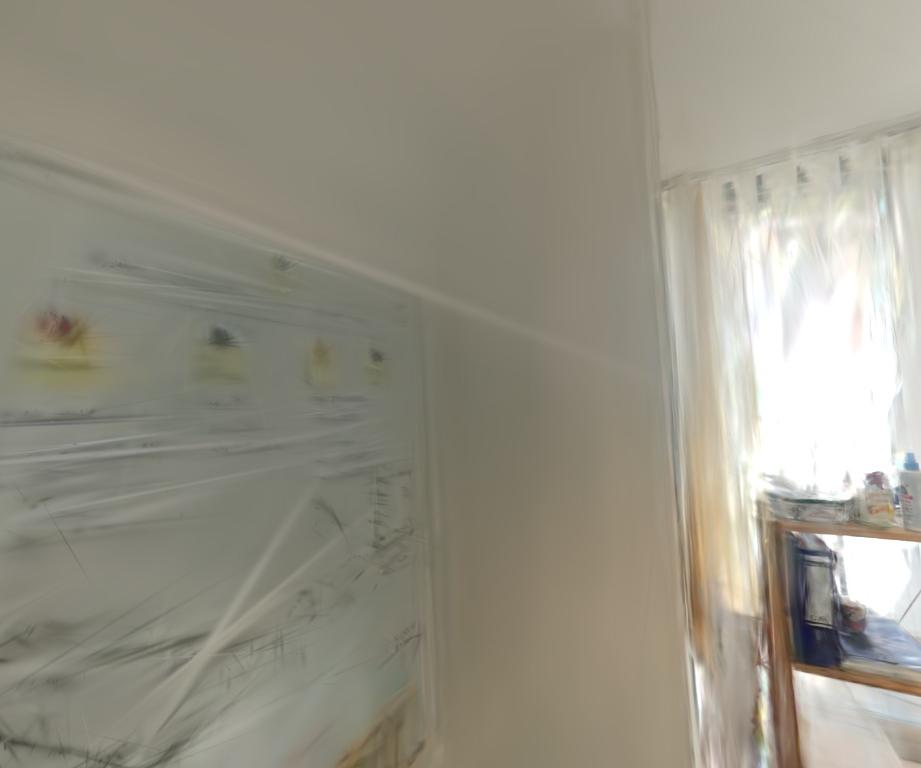}&
    \includegraphics[width=0.158\linewidth]{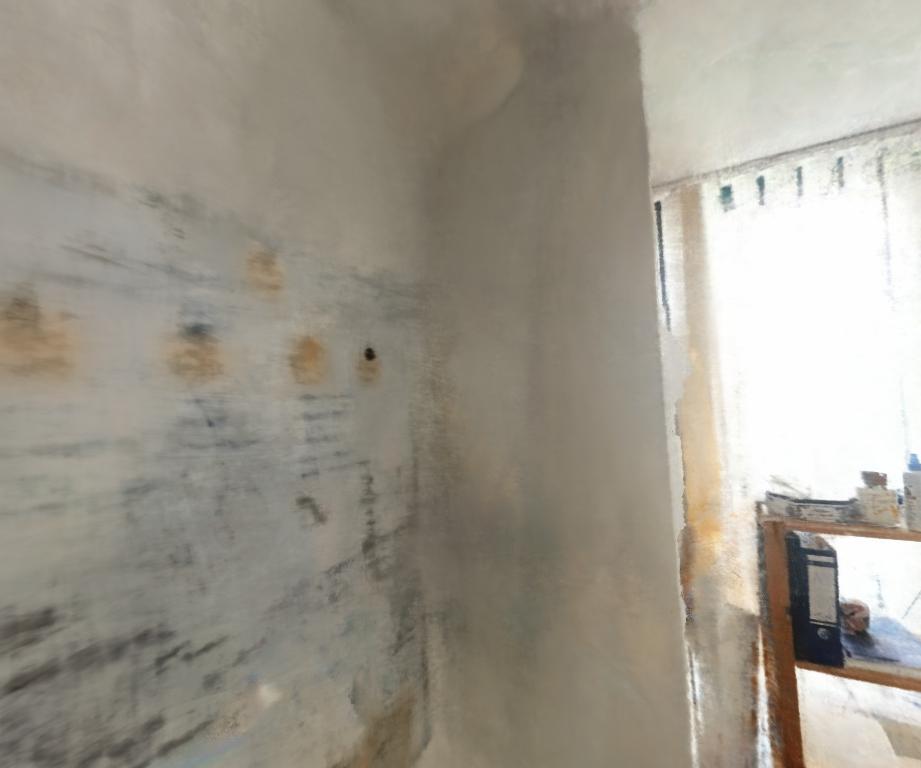}&
    \includegraphics[width=0.158\linewidth]{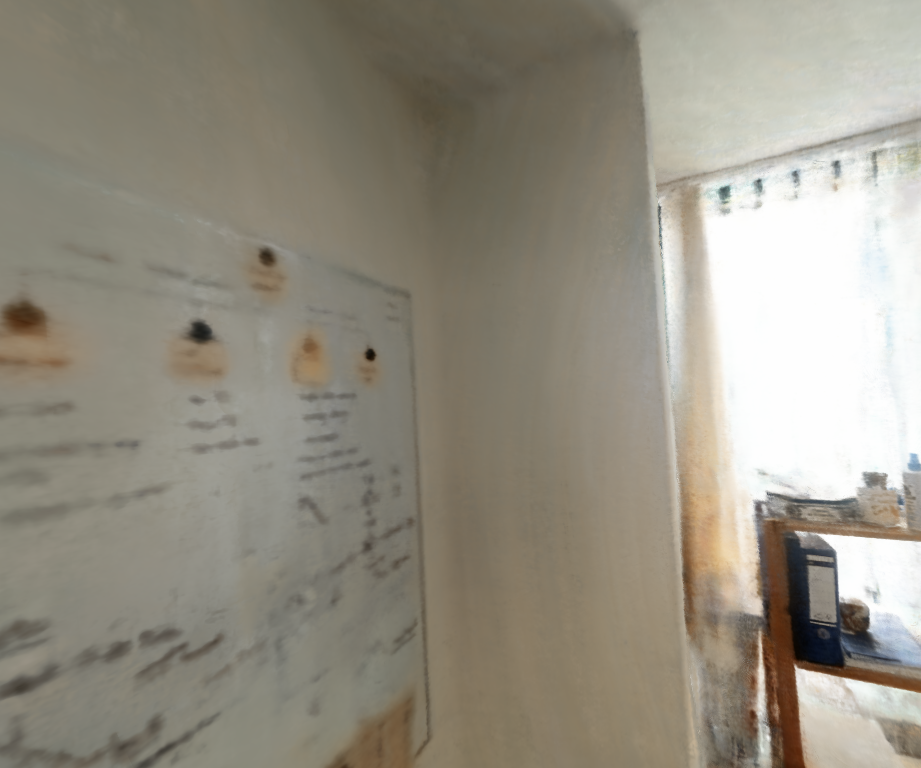}\\

    \includegraphics[width=0.158\linewidth]{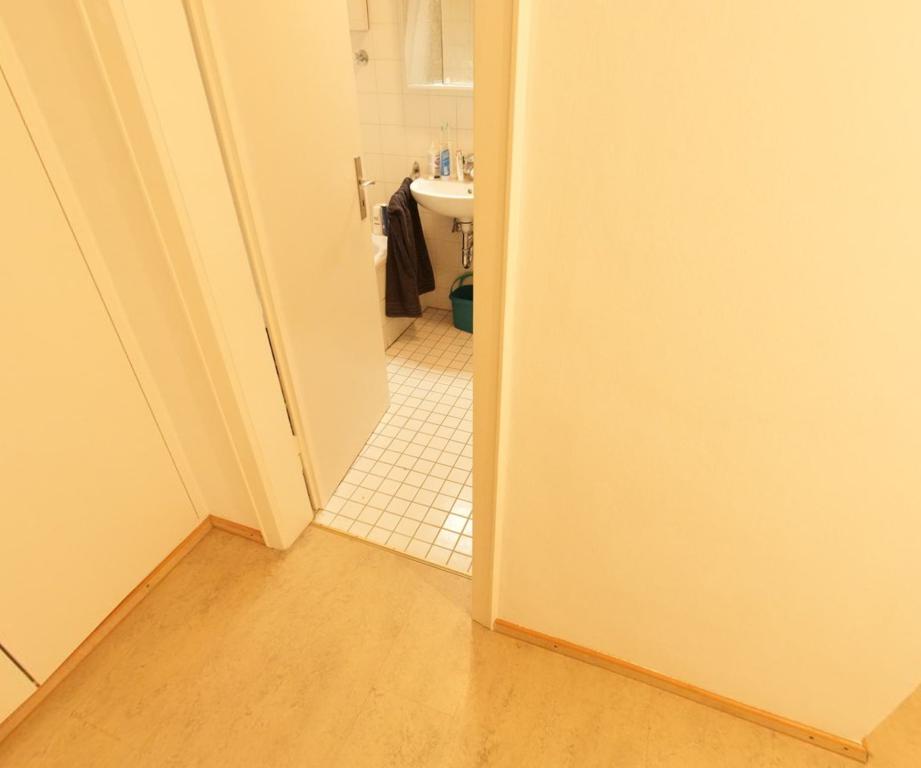}& 
    \includegraphics[width=0.158\linewidth]{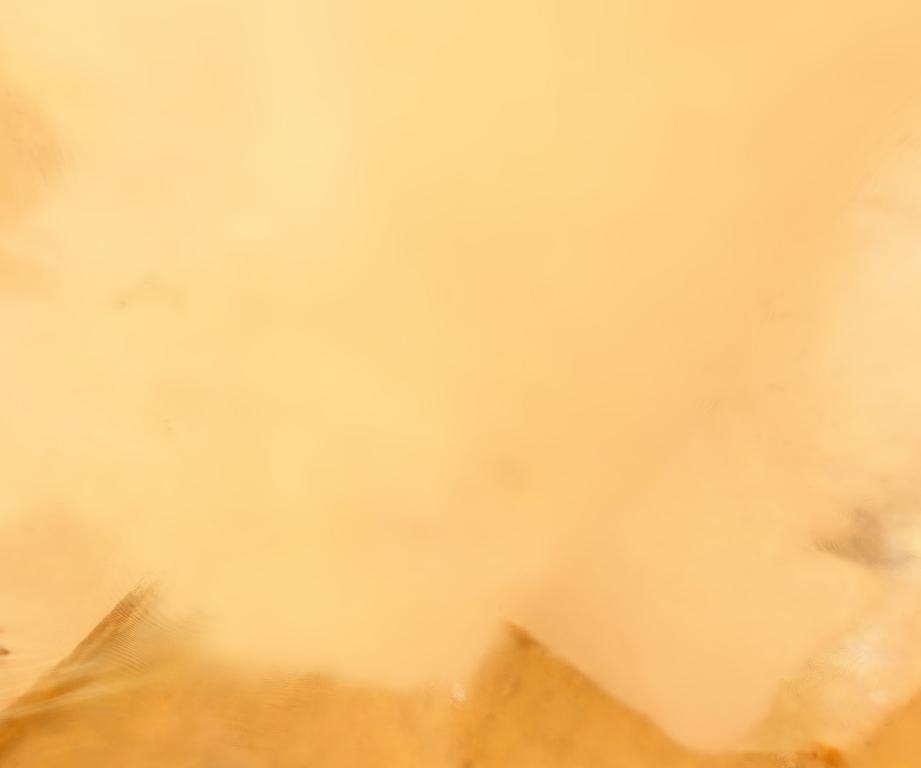}& \includegraphics[width=0.158\linewidth]{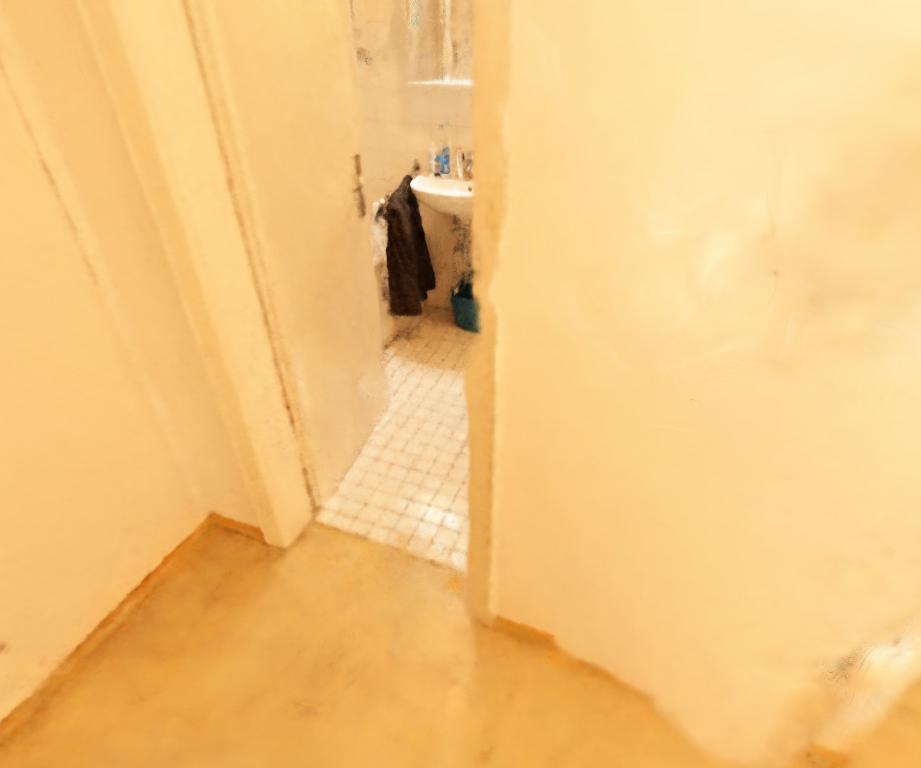}& \includegraphics[width=0.158\linewidth]{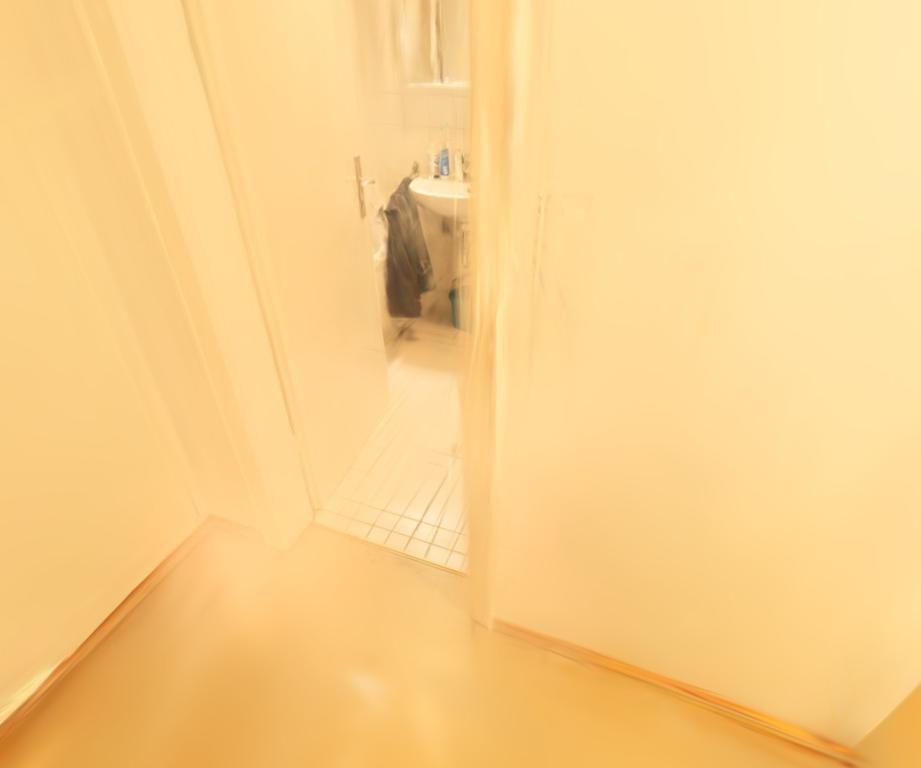}&
    \includegraphics[width=0.158\linewidth]{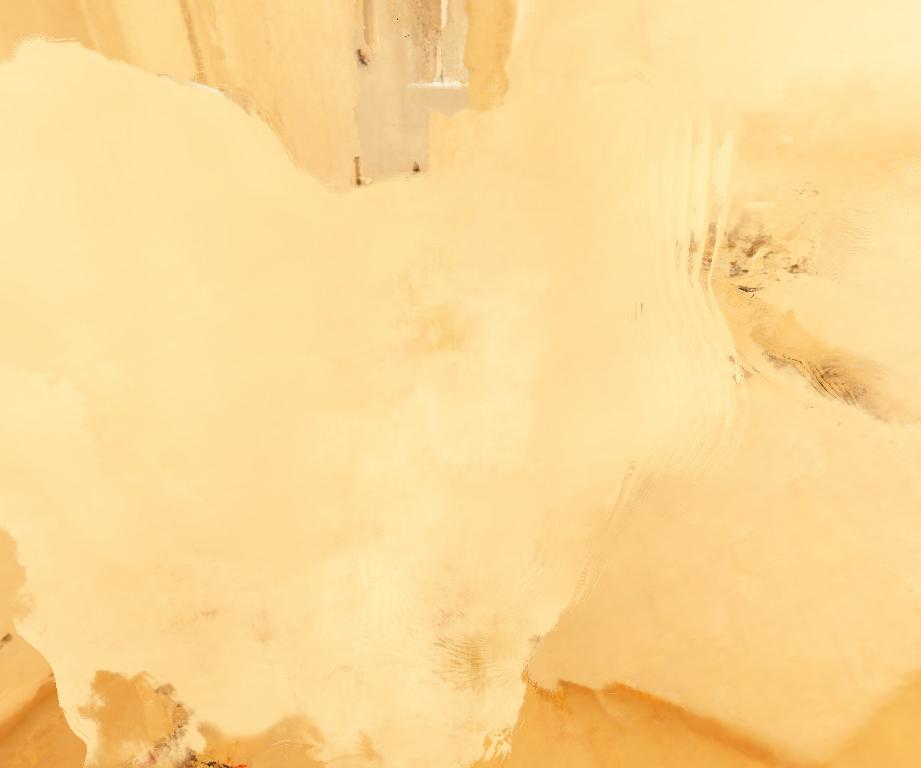}&
    \includegraphics[width=0.158\linewidth]{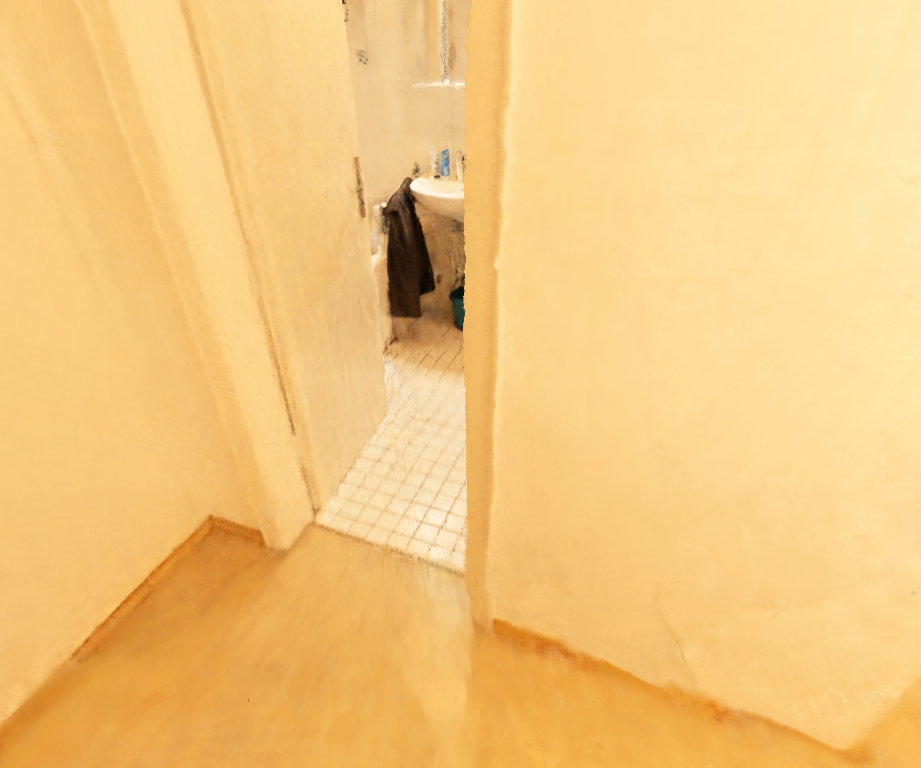}\\

    \includegraphics[width=0.158\linewidth]{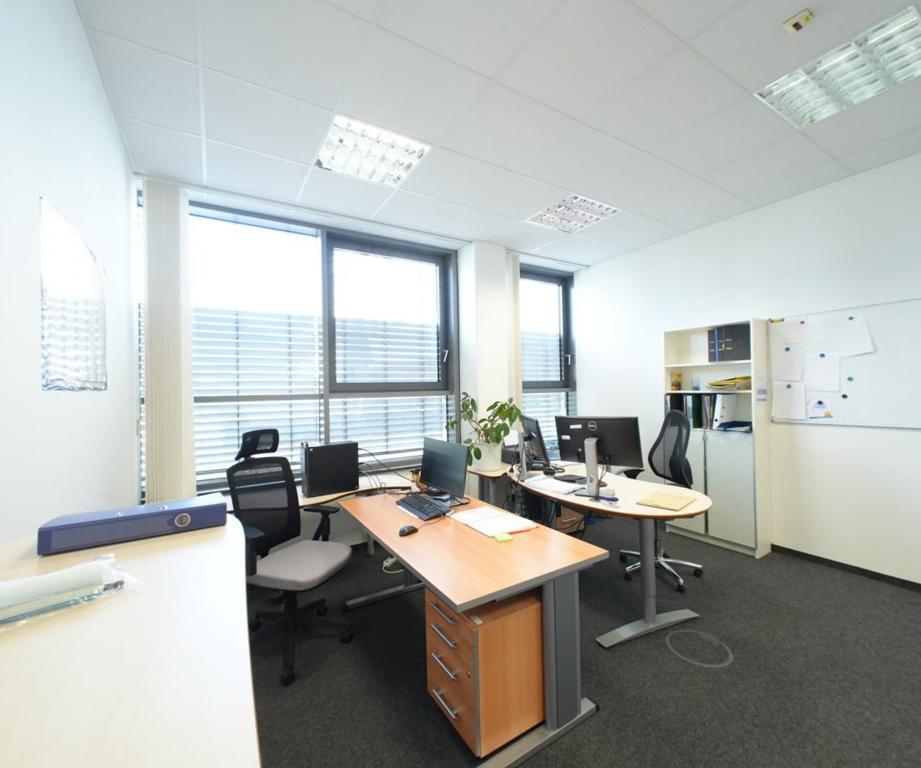}& 
    \includegraphics[width=0.158\linewidth]{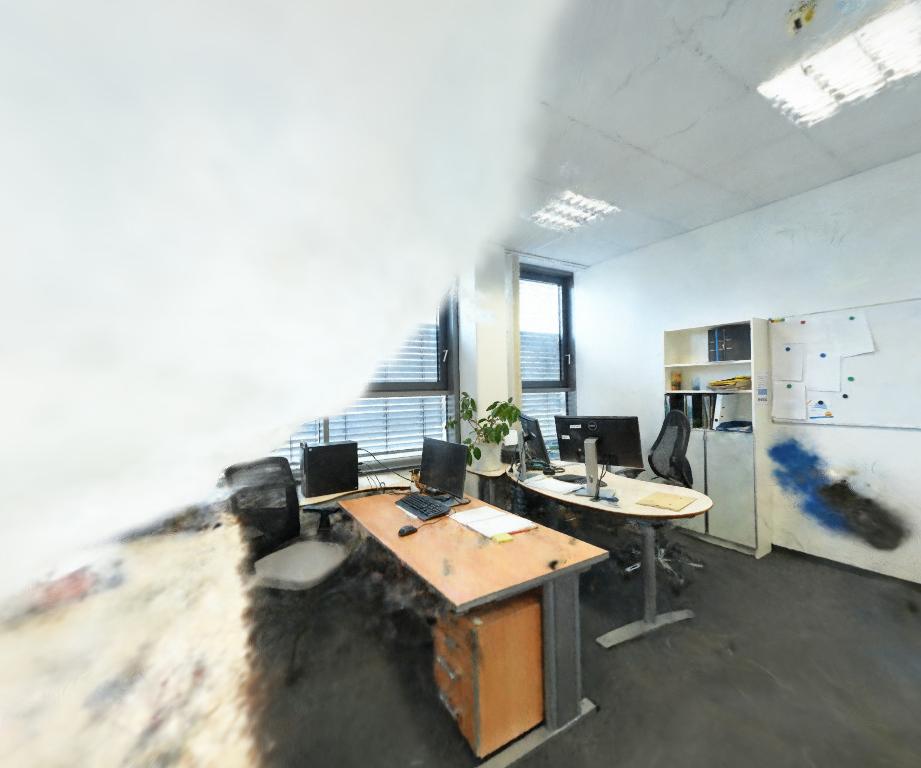}& \includegraphics[width=0.158\linewidth]{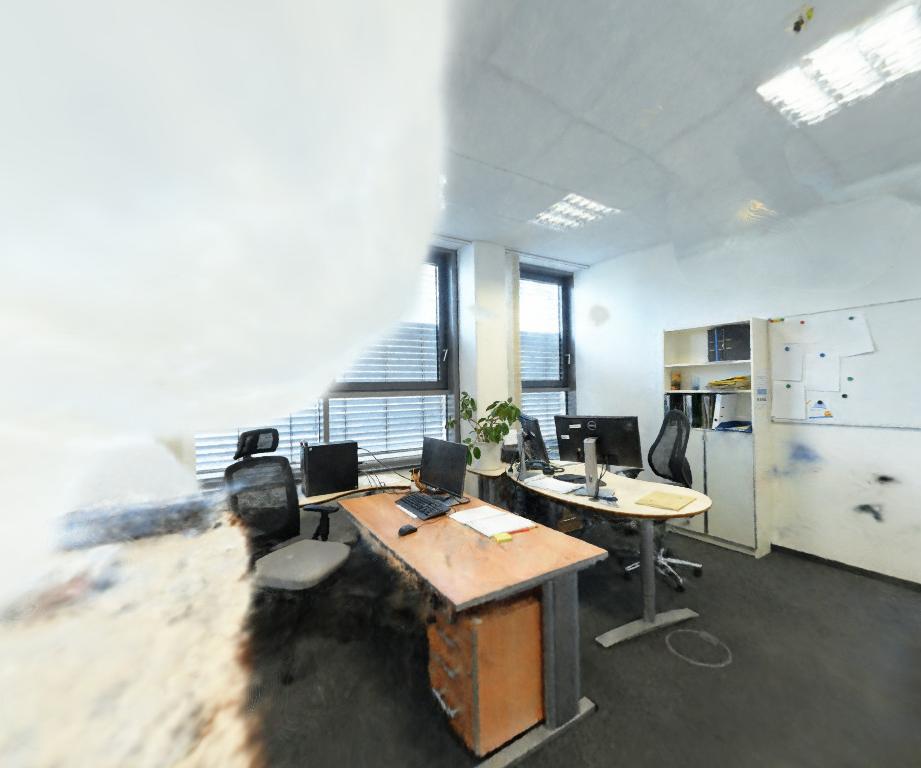}& \includegraphics[width=0.158\linewidth]{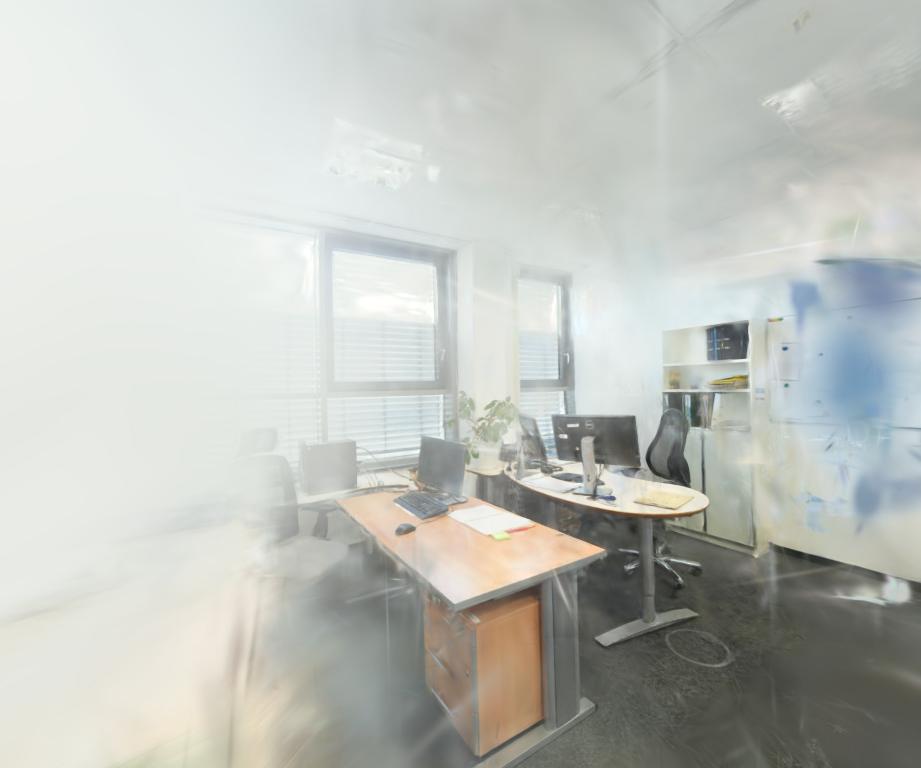}&
    \includegraphics[width=0.158\linewidth]{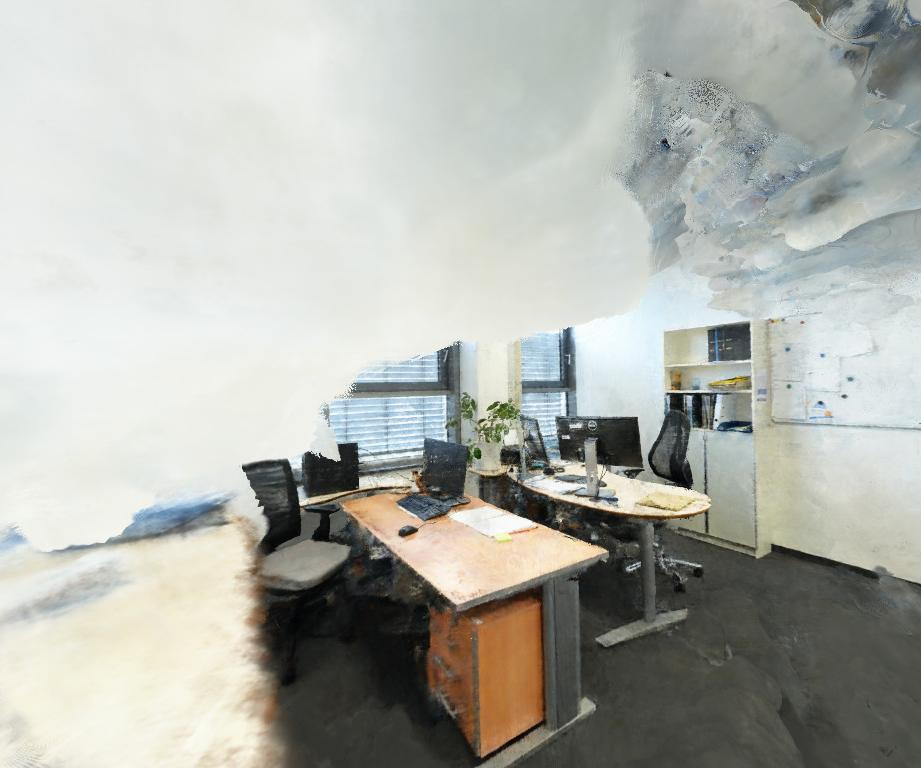}&
    \includegraphics[width=0.158\linewidth]{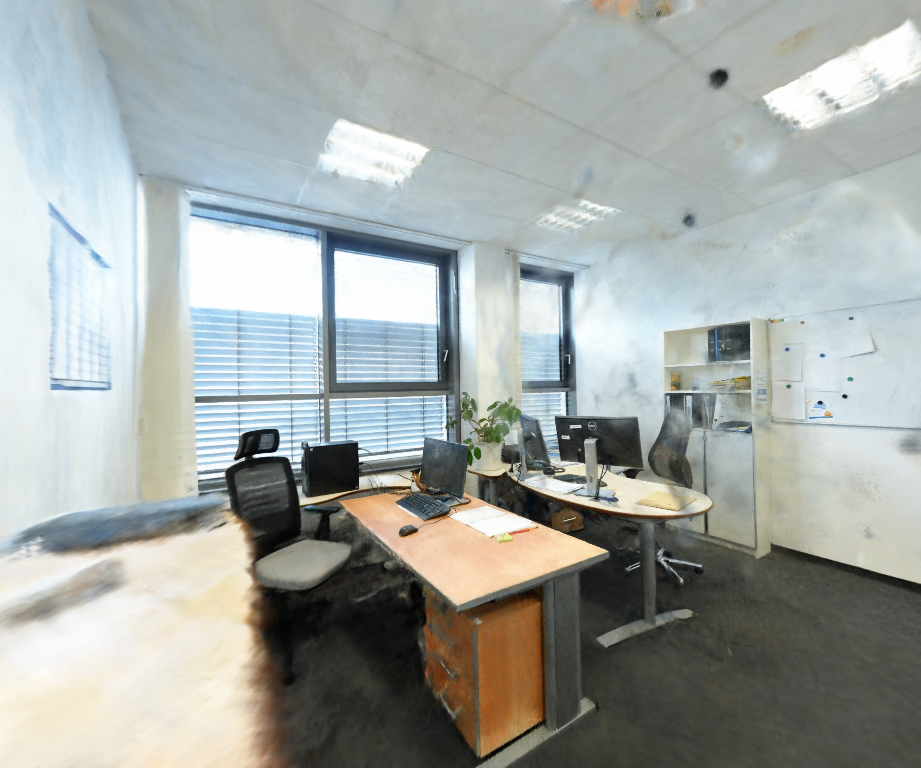}\\

    \end{tabular}
    \caption{\textbf{Qualitative results of novel view synthesis on ScanNet++~\cite{scannetpp} dataset.} We compare our framework against NeRFacto~\cite{nerfstudio}, depth-regularized NeRFacto~\cite{sparsenerf}, 3DGS~\cite{3dgs}, and NeLF-Pro~\cite{nelf-pro} under severe extrapolation scenario. %NeRFacto exhibits blurry geometry and view-inconsistent artifacts under sparse views, where its depth-regularized variant does not fully address these issues. 3DGS produces renderings with high-frequency details, but often suffers from sharp floating artifacts. %While NeLF-Pro similarly suffer from artifacts at under-constrained regions, our method yields cleaner geometry and mitigates artifacts, resulting in notably improved renderings across viewpoints.
    }
    \label{fig:qualitative_scannet++}
\end{figure*}
\begin{figure*}[p]
    \centering
    \begin{tabular}{c@{\,}c@{\,}c@{\,}c@{\,}c@{\,}c}
    \small GT& \small Mega-NeRF & \small NeLF-Pro & \small NeLF-Pro+Ours$_{MA}$& \small LocalRF & \small LocalRF+Ours$_{MA}$\\

    \includegraphics[width=0.158\linewidth]{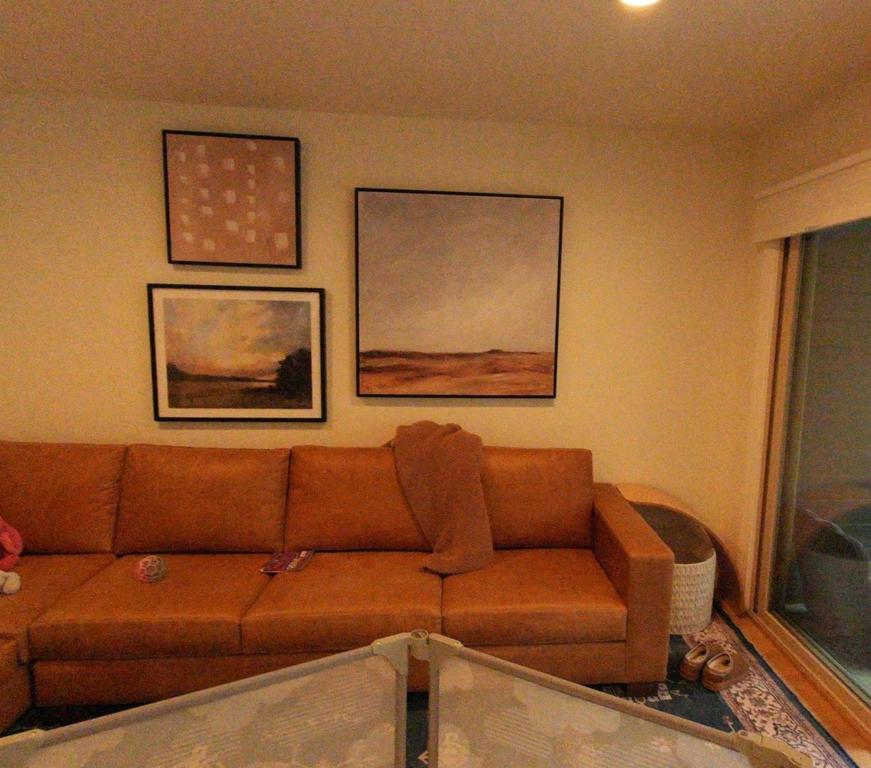}& 
    \includegraphics[width=0.158\linewidth]{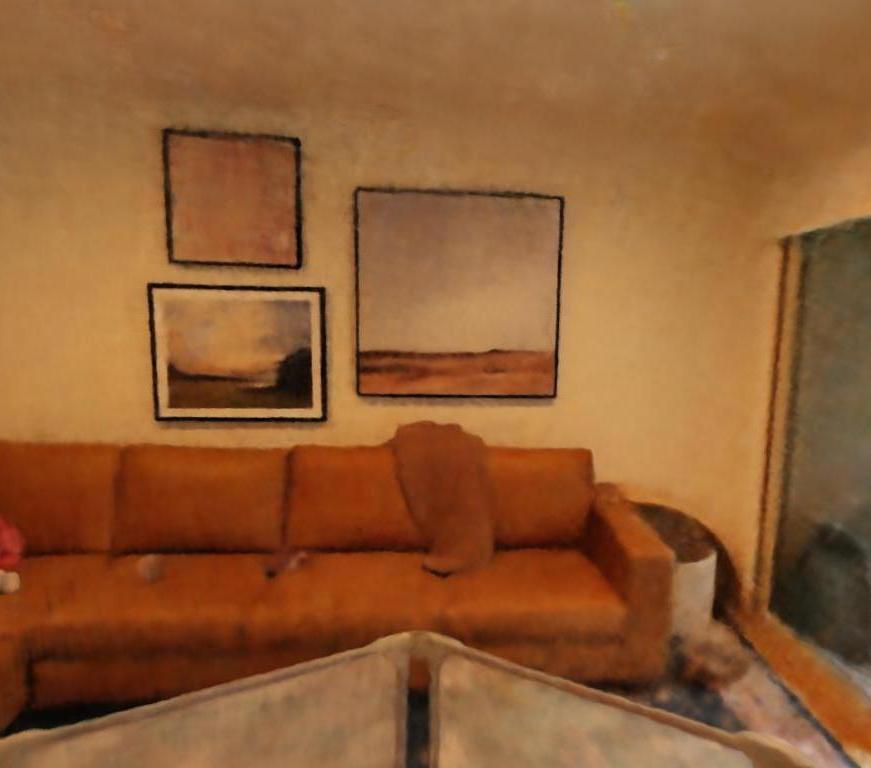}& 
    \includegraphics[width=0.158\linewidth]{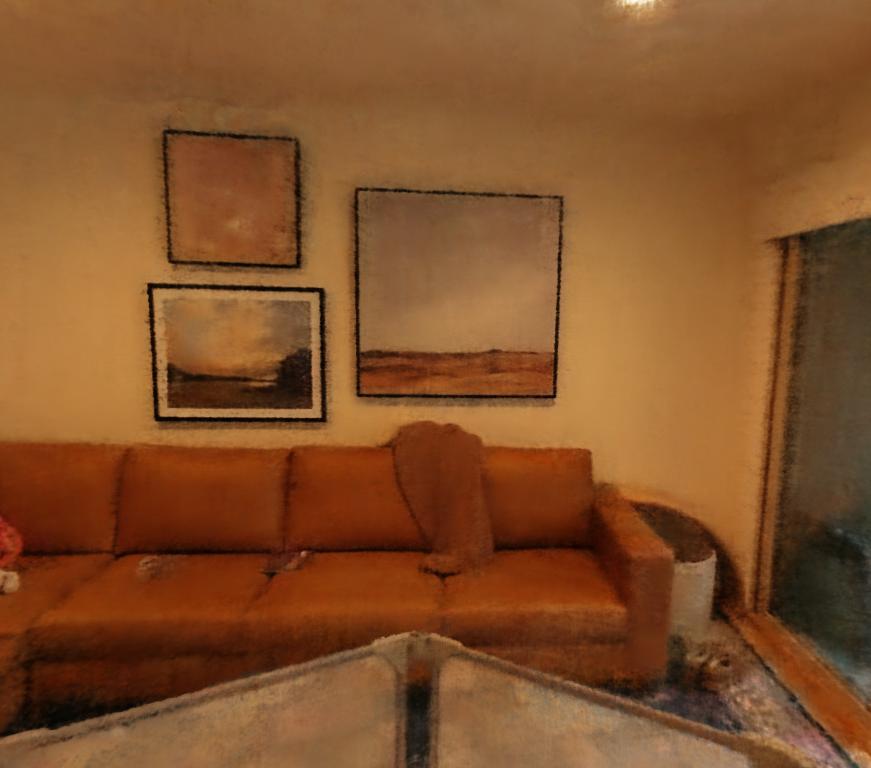} & \includegraphics[width=0.158\linewidth]{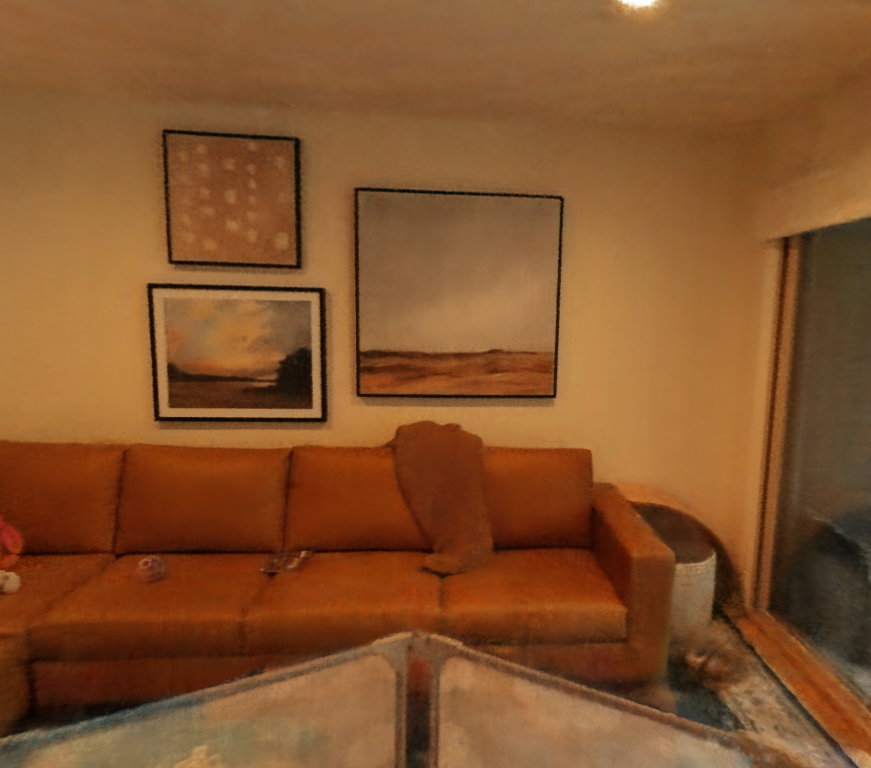}&
    \includegraphics[width=0.158\linewidth]{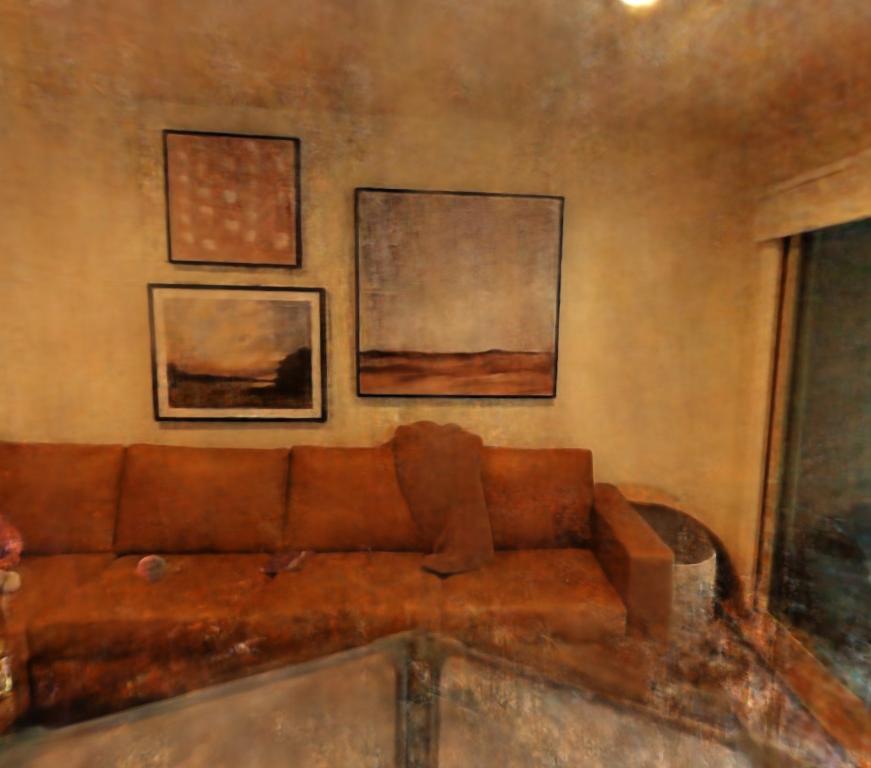}&
    \includegraphics[width=0.158\linewidth]{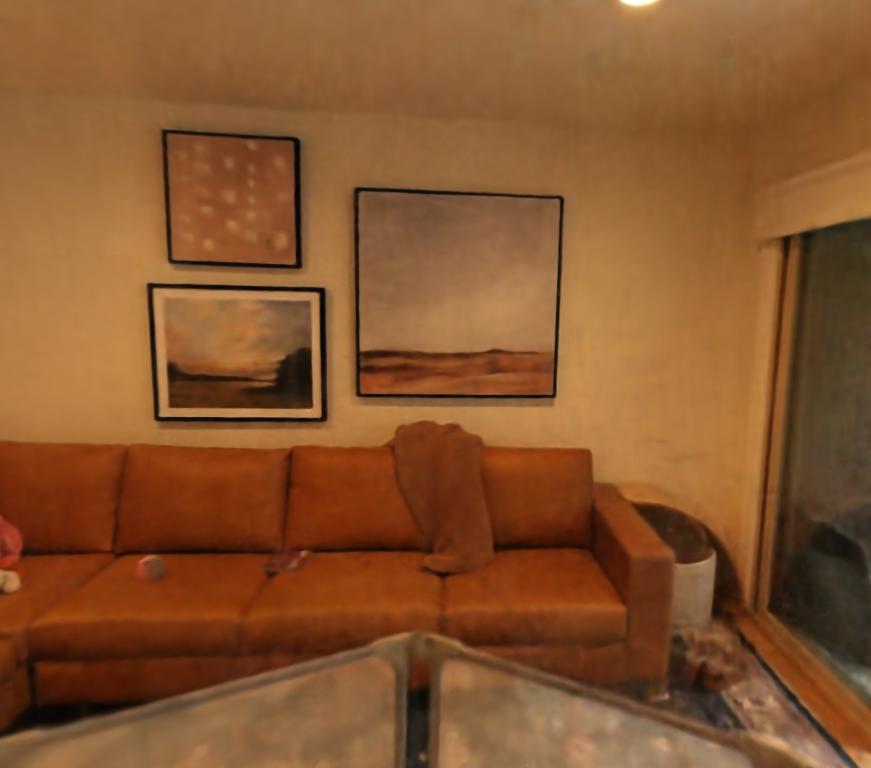}\\

    \includegraphics[width=0.158\linewidth]{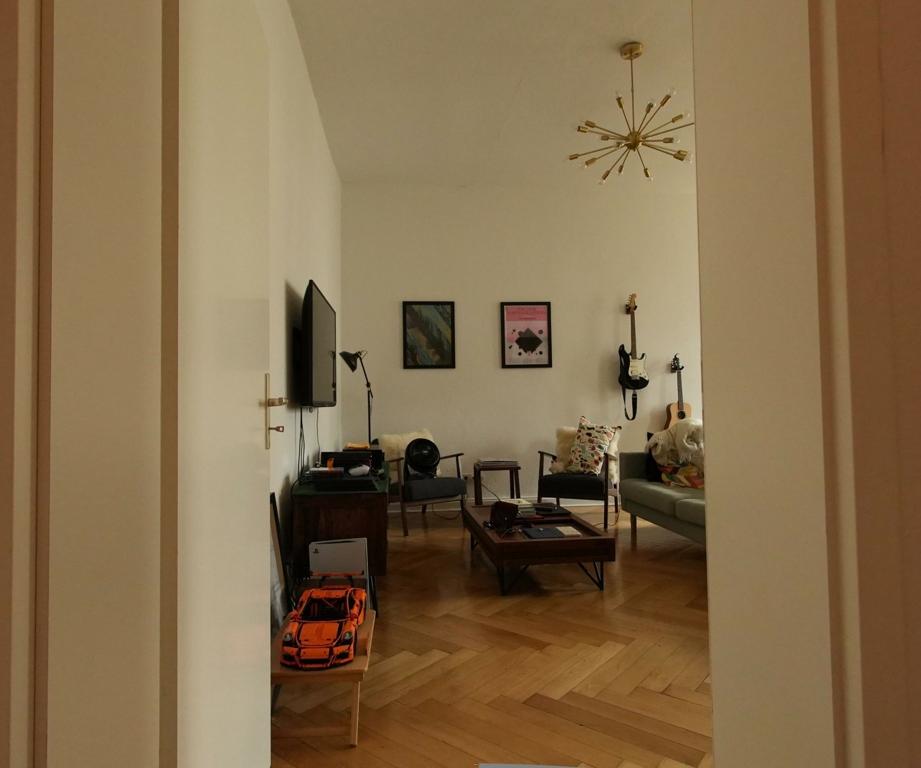}& 
    \includegraphics[width=0.158\linewidth]{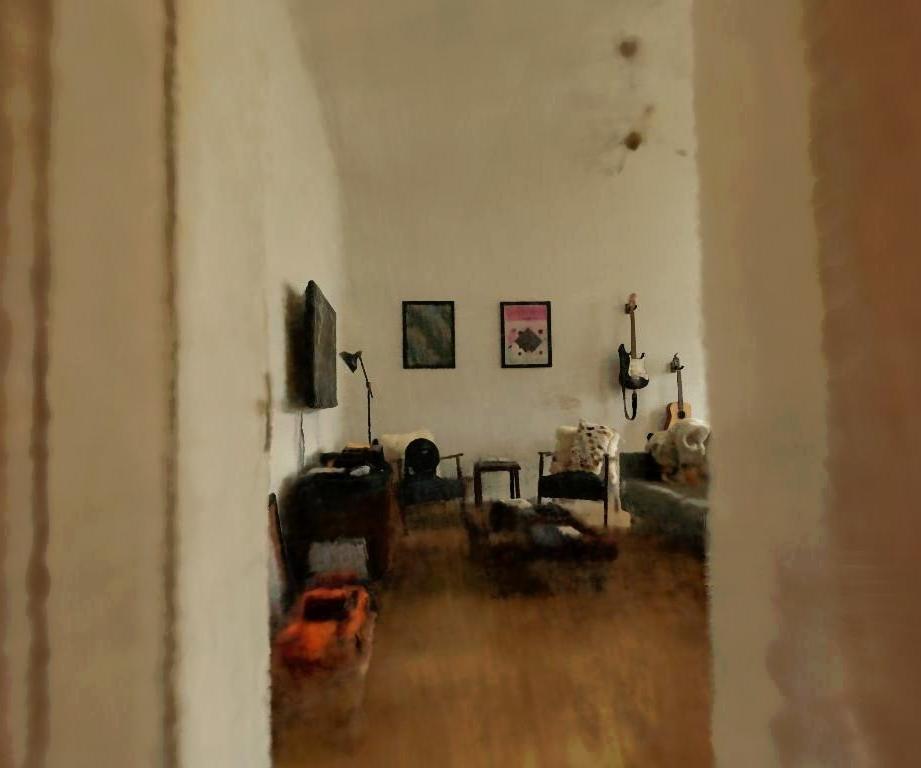}& 
    \includegraphics[width=0.158\linewidth]{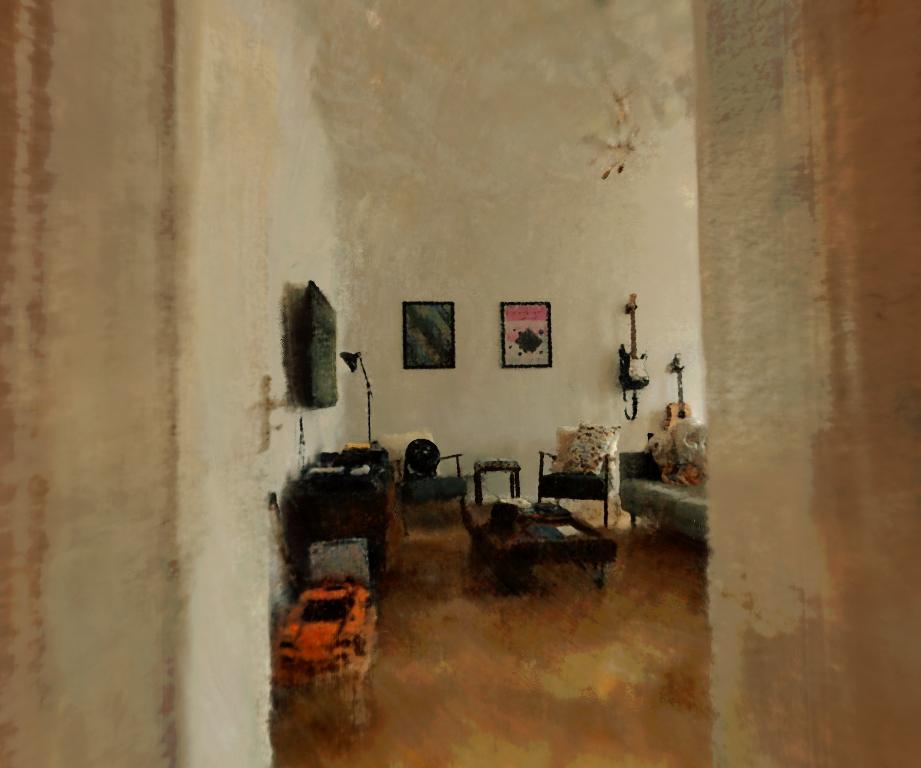}& 
    \includegraphics[width=0.158\linewidth]{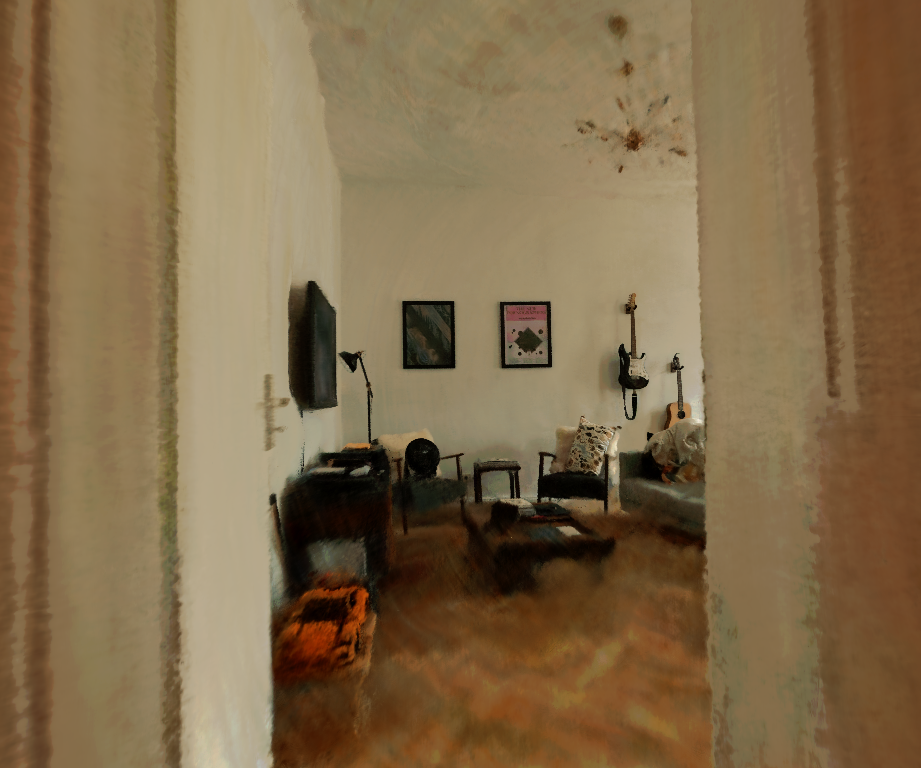}&
    \includegraphics[width=0.158\linewidth]{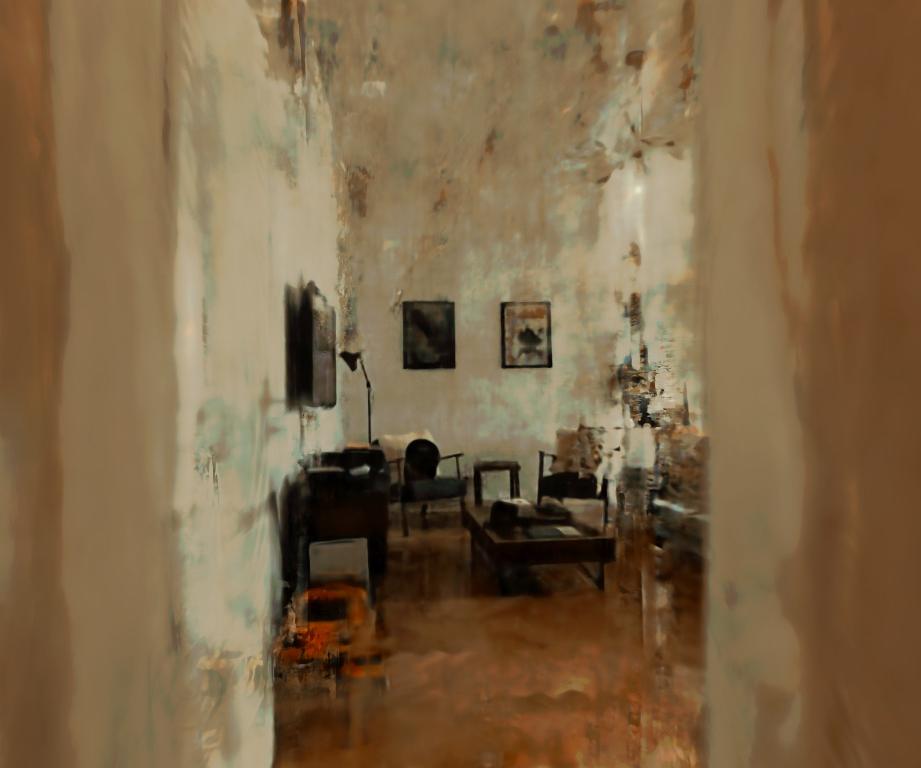}&
    \includegraphics[width=0.158\linewidth]{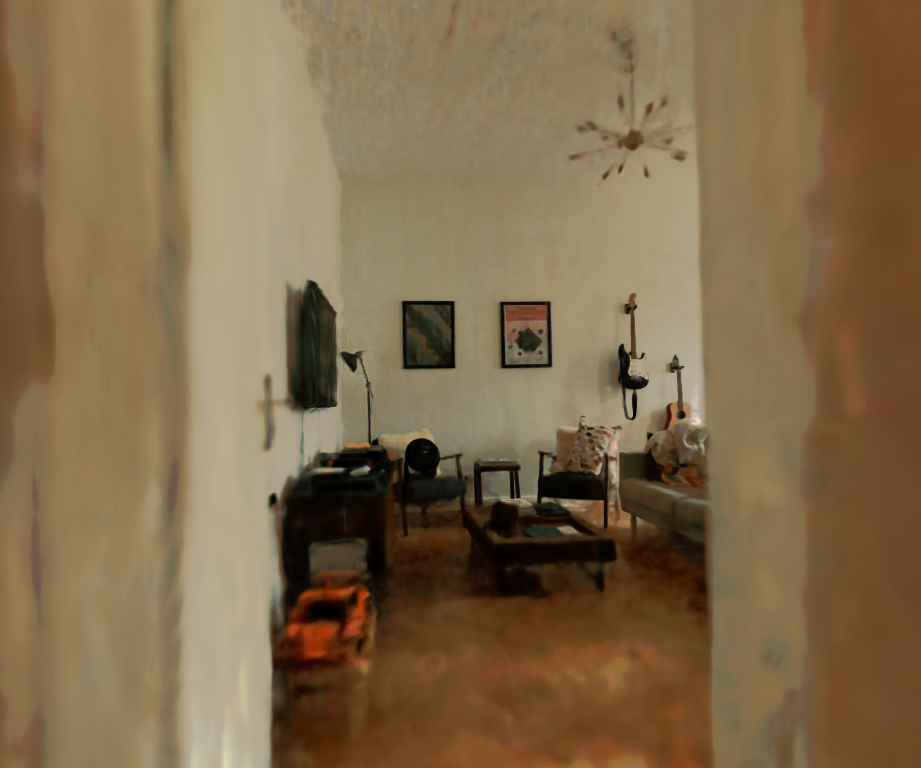}\\

    \includegraphics[width=0.158\linewidth]{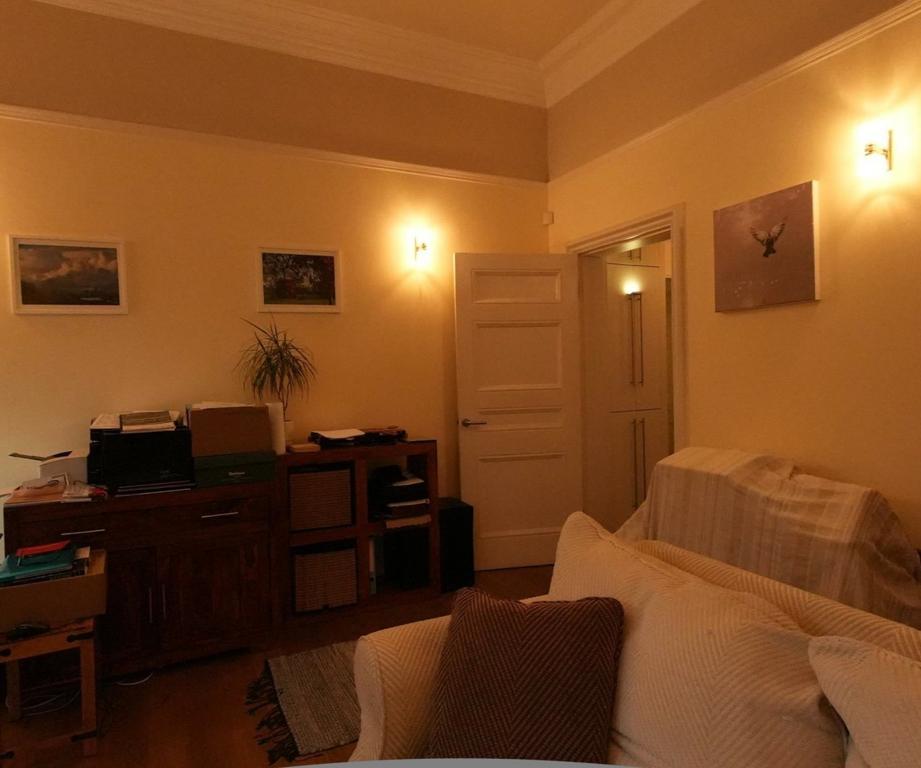}& 
    \includegraphics[width=0.158\linewidth]{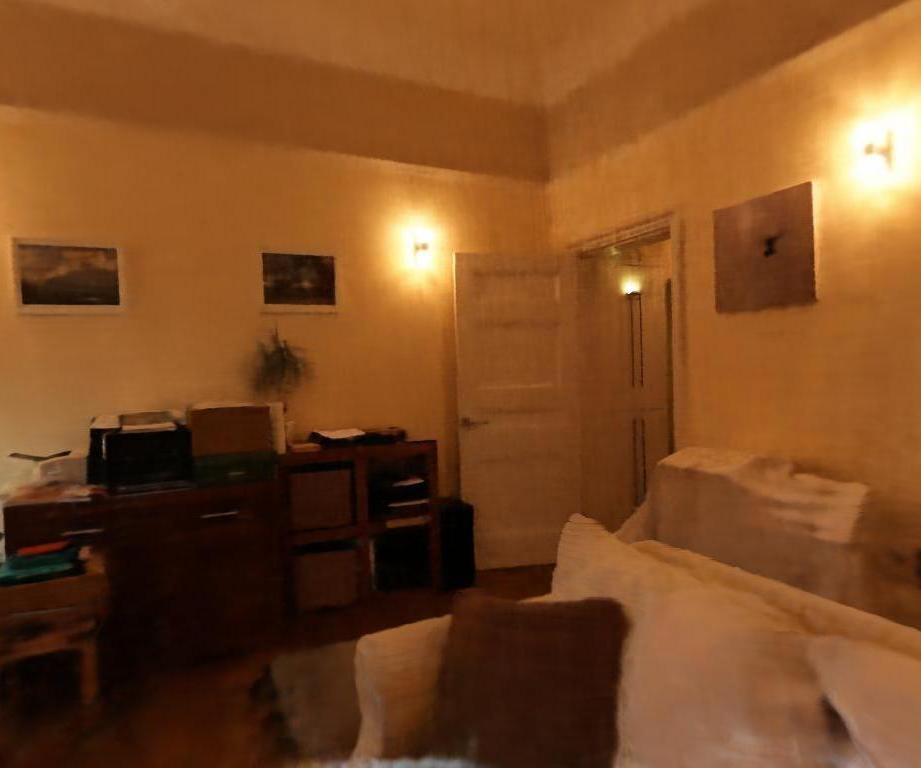}& 
    \includegraphics[width=0.158\linewidth]{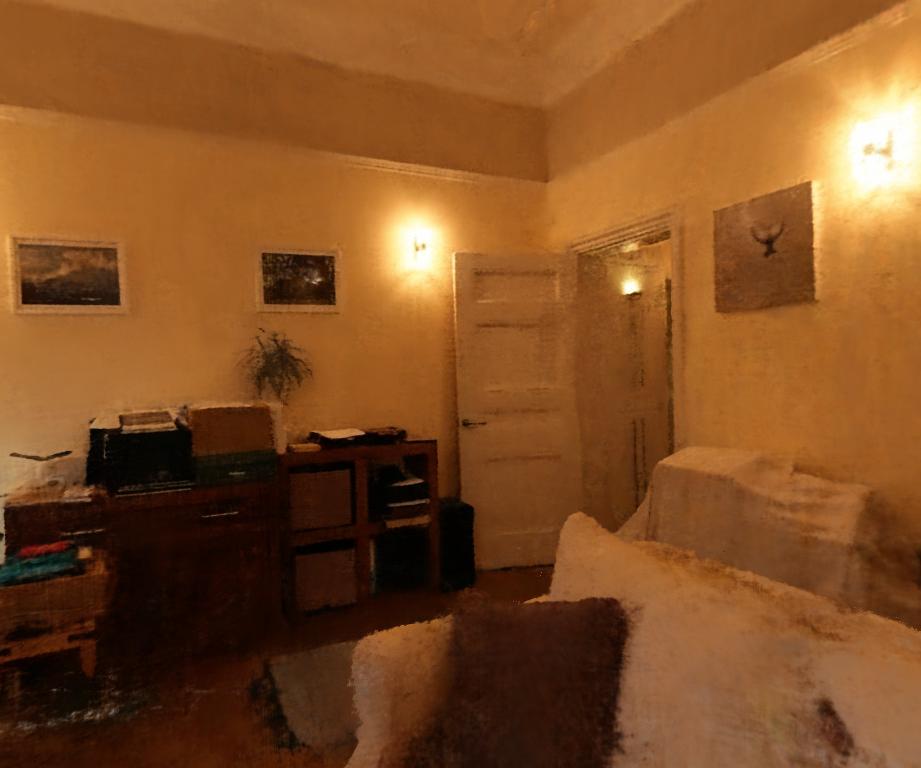}& 
    \includegraphics[width=0.158\linewidth]{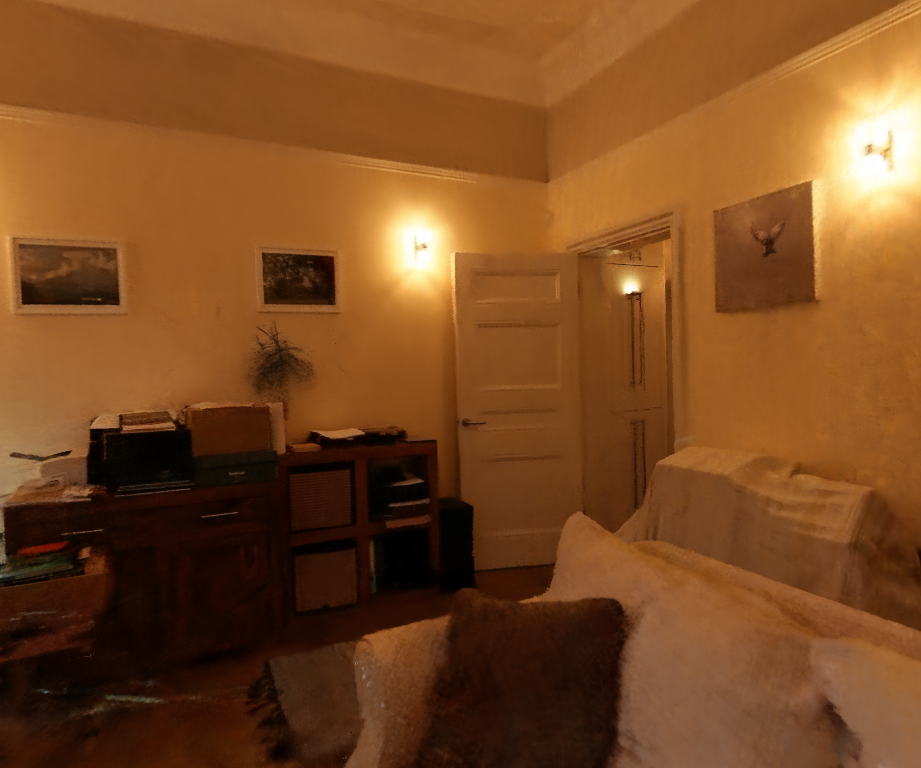}&
    \includegraphics[width=0.158\linewidth]{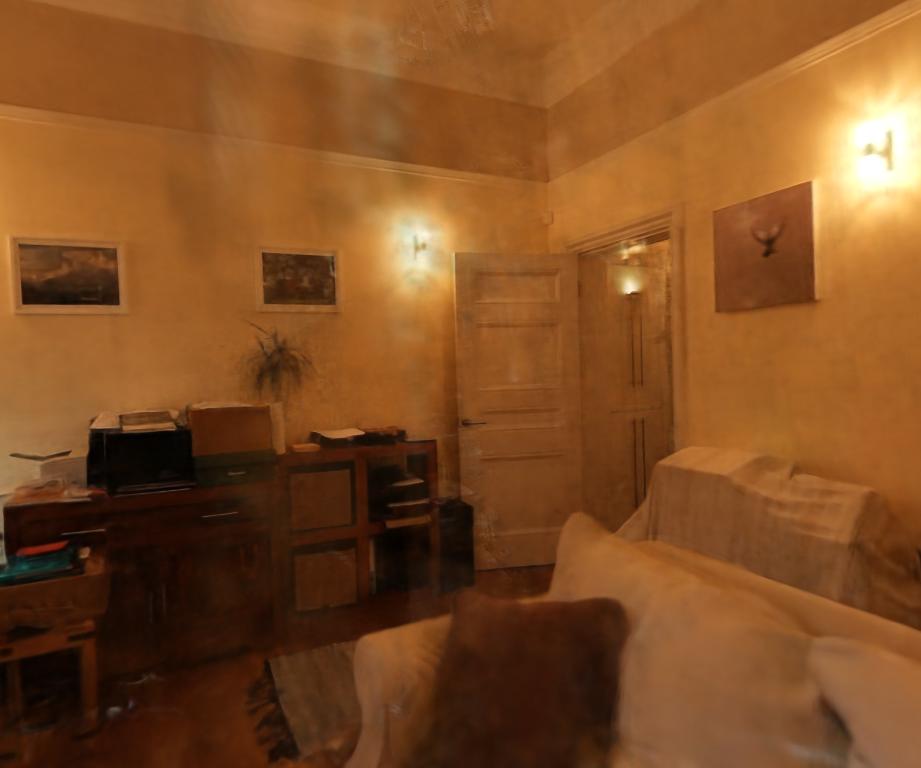}&
    \includegraphics[width=0.158\linewidth]{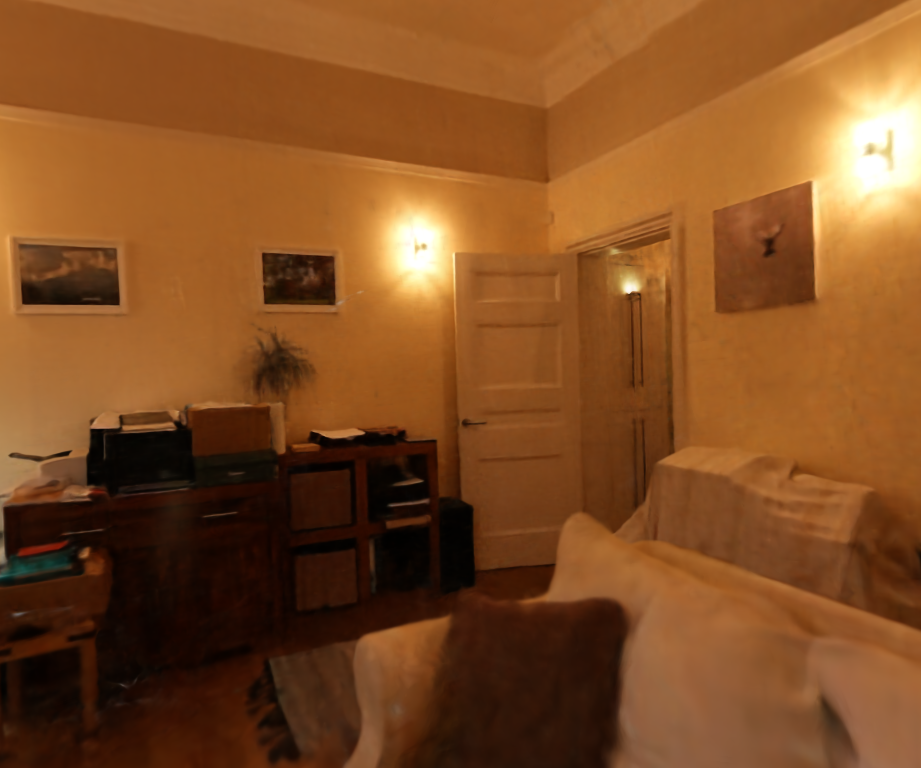}\\

    \includegraphics[width=0.158\linewidth]{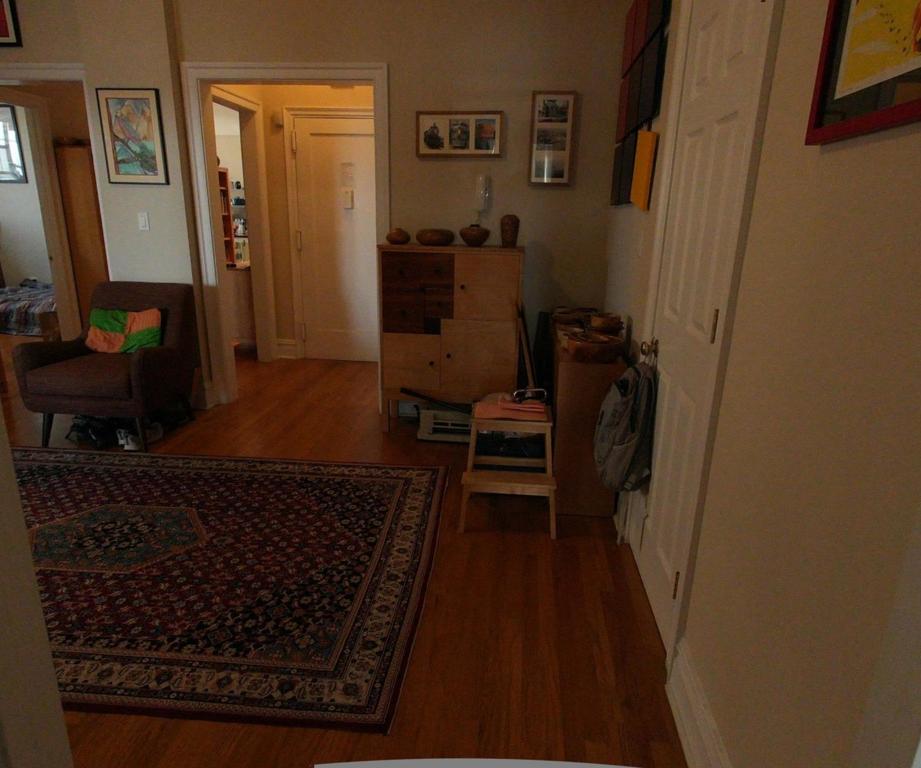}& 
    \includegraphics[width=0.158\linewidth]{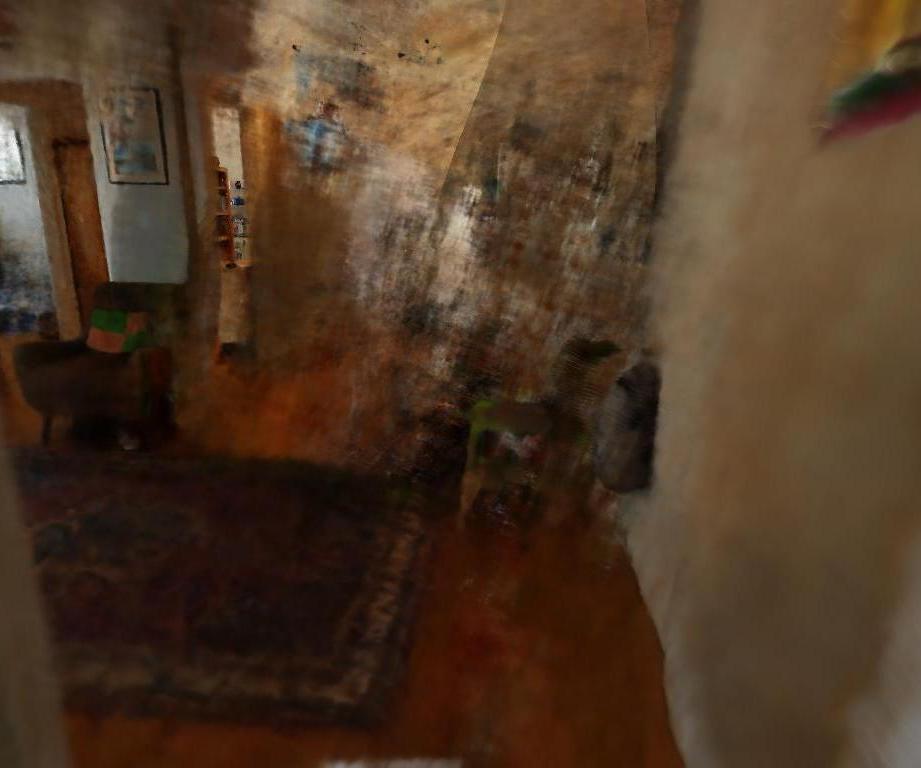}& 
    \includegraphics[width=0.158\linewidth]{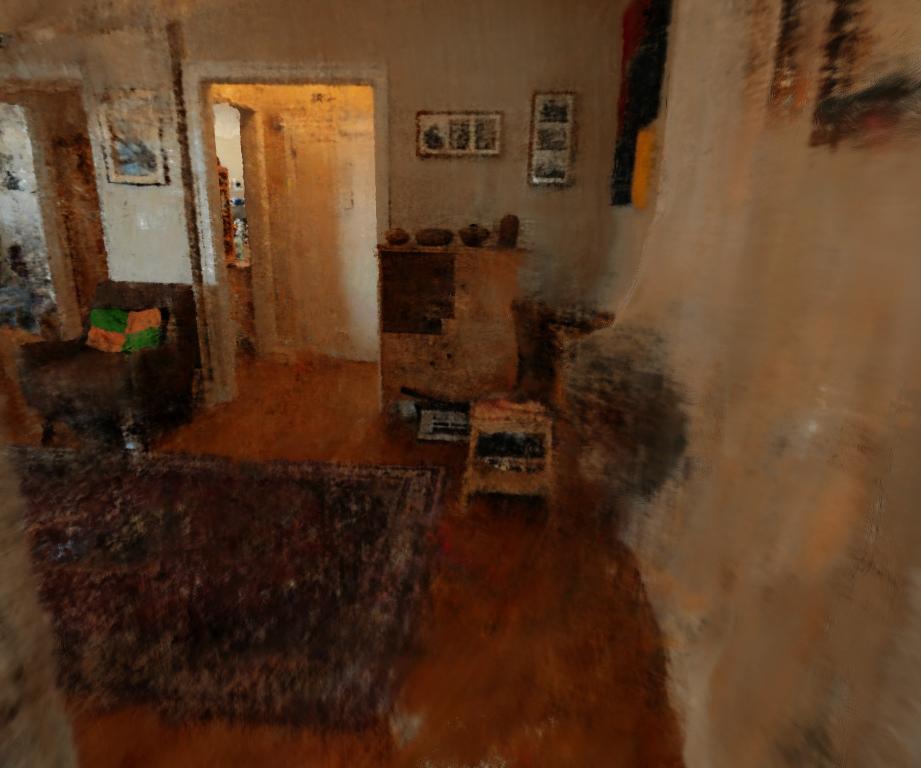}& 
    \includegraphics[width=0.158\linewidth]{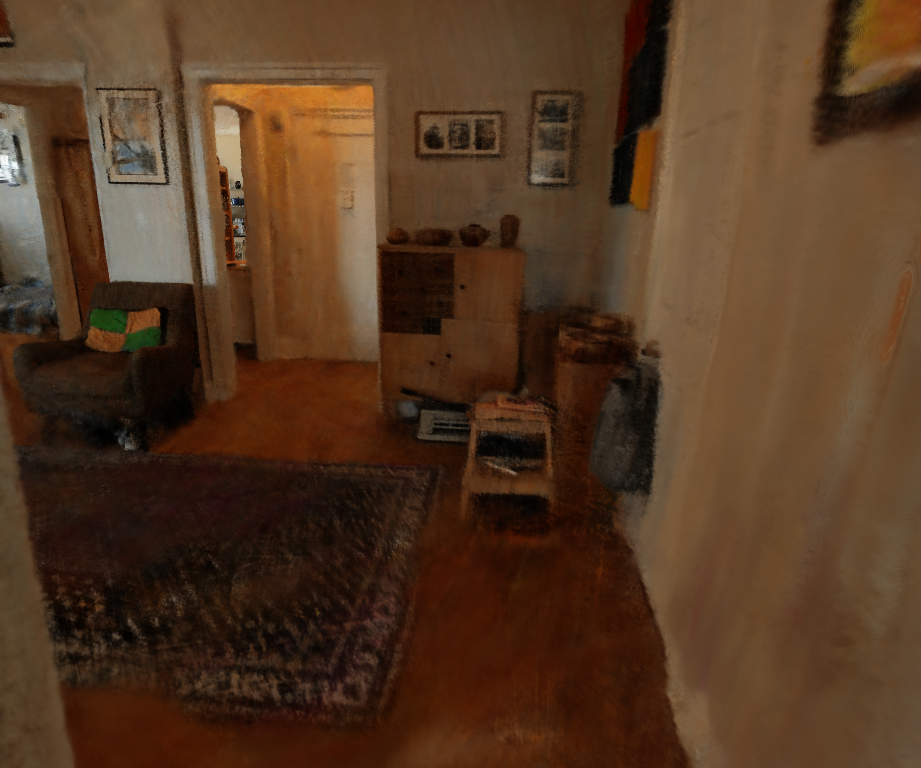}&
    \includegraphics[width=0.158\linewidth]{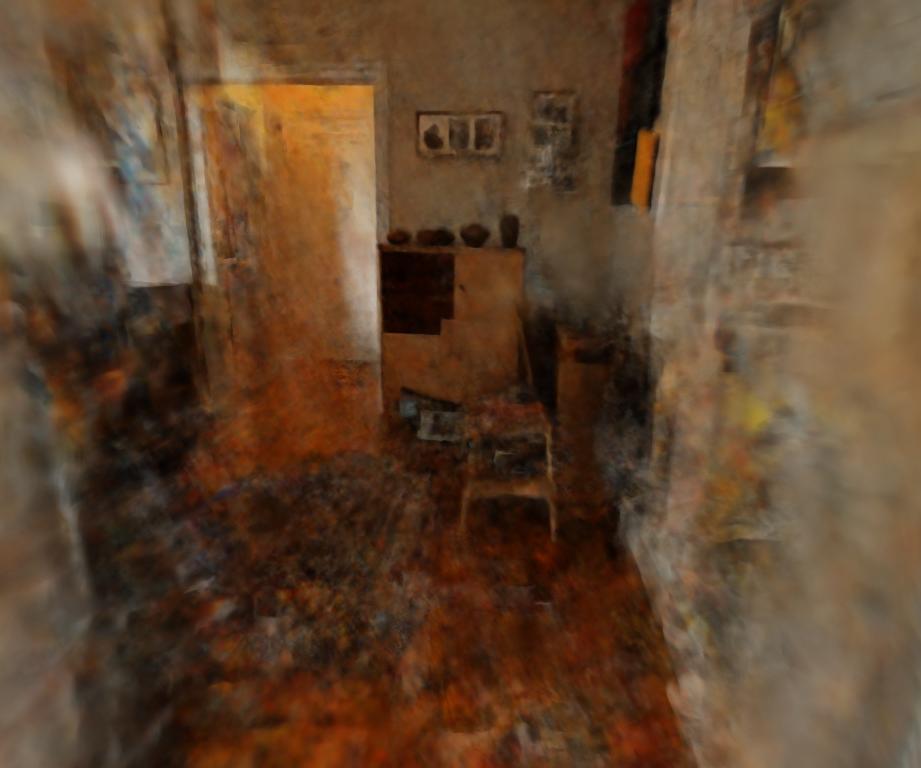}&
    \includegraphics[width=0.158\linewidth]{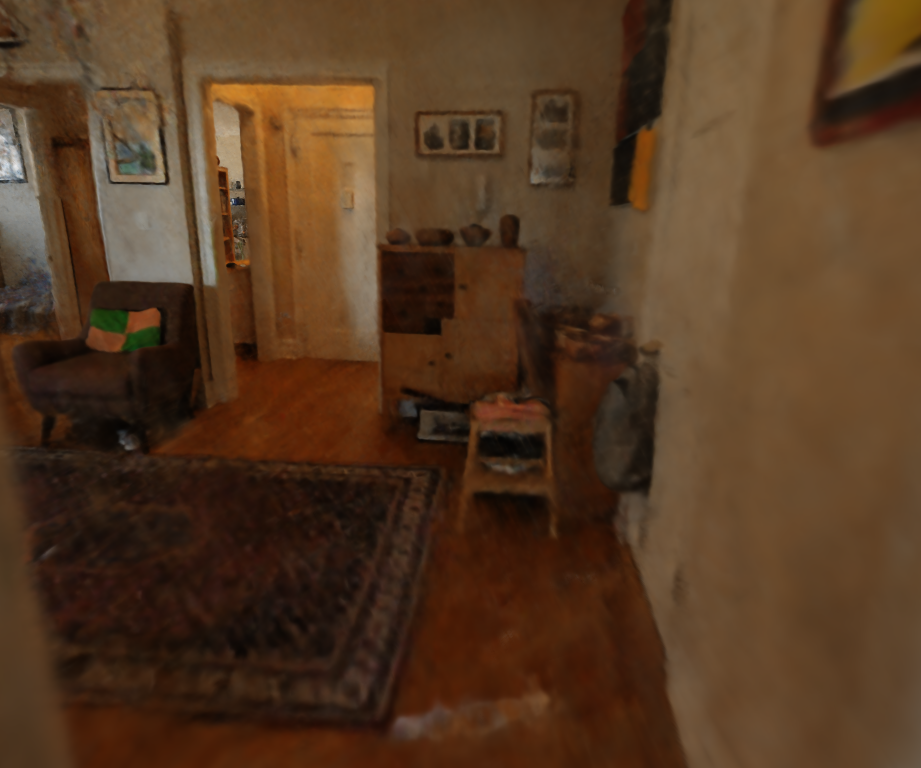}\\
    \end{tabular}
    \caption{\textbf{Qualitative results of novel view synthesis on Zip-NeRF~\cite{zip-nerf} dataset.} We compare our method against  Mega-NeRF~\cite{mega-nerf}, NeLF-Pro~\cite{nelf-pro}, and LocalRF~\cite{localrf} which leverage multiple bases for large-scale indoor scenes. %While Mega-NeRF~\cite{mega-nerf} preserves global geometry, it tends to suffer from loss of fine details and blurred artifacts. By employing scene-adaptive basis placement and geometric regularization, our method significantly outperforms baselines that utilize multiple NeRF blocks~\cite{localrf} and neural light field probes~\cite{nelf-pro}.
    }
    \label{fig:qualitative_zipnerf}
\end{figure*}

\subsection{Comparative Study}
In this section, we report the qualitative and quantitative results of our framework with comparative studies.
For quantitative studies, we evaluated reconstructed image qualities using the peak-signal-to-noise ratio (PSNR), structure-similarity index measure (SSIM), and learned perceptual image patch similarity (LPIPS). 

We first evaluate the performance of the novel view synthesis of our framework on the ScanNet++ dataset.
We reported results for our method using a geometric scaffold obtained from MonoSDF~\cite{monosdf} and MAPAnything~\cite{MAPAnything}, denoted as Ours$_{MS}$ and Ours$_{MA}$, respectively. 
In the results with the ScanNet++ dataset in Table~\ref{table:scannet++_nelf-pro_comparison}, our work significantly outperforms other baselines in PSNR and SSIM without increasing memory usage.
While our method uses the same number of parameters as NeLF-Pro, it shows better reconstruction results with basis placement and geometry guidance accounting for scene configurations.
Since 3DGS and its variants optimize scenes using primitives with extensive parameters, they overfit to training images and demonstrate good performance on densely observed regions. 

% ScanNet++ dataset
% \begin{table}[htb!]
\begin{table}
\centering
\caption{\textbf{Performance comparison on ScanNet++~\cite{scannetpp} dataset.} Our framework is tested using multiple neural feature grids. The upper table includes single-basis architectures while methods in the bottom leverage multiple basis. The first and second best results are highlighted in \textbf{bold} and \underline{underlined}, respectively.}
\resizebox{\linewidth}{!}{
    \begin{tabular}{l|c|c|c|c}
        \toprule
        Method & PSNR $\uparrow$ & SSIM $\uparrow$ & LPIPS $\downarrow$ & Memory \\
        \midrule
        MonoSDF~\cite{monosdf} & 20.41 & 0.800 & 0.351 & 61 MB \\
        NeRFacto~\cite{nerfstudio} & 20.52 & 0.793 & 0.352 & 65 MB \\
        Depth-NeRFacto~\cite{nerfstudio} & 21.75 & 0.808 & 0.333 & 65 MB \\
        ZipNeRF~\cite{zip-nerf} & 21.72 & 0.804 & 0.332 & 1.2 GB \\
        3DGS~\cite{3dgs} & 21.26 & 0.797 & \textbf{0.305} & 164 MB \\
        Mip-Splatting~\cite{Mip-Splatting} &  20.49 &  0.791 &  0.390 &  274 MB \\
         Scaffold-GS~\cite{Scaffold-GS} &  20.72 &  0.799 &  0.372 &  91 MB \\
         PlanarGS~\cite{PlanarGS} &  21.15 &  0.801 &  0.347 &  119 MB \\
        FSGS~\cite{FSGS} & 20.69 & 0.803 & 0.352 & 124 MB \\
        DNGaussian~\cite{dngaussian} & 19.26 & 0.780 & 0.483 & 156 MB \\
        \midrule
        Hierarchical-3DGS~\cite{hierarchical_3dgs} & 20.57 & 0.771 & 0.335 & 793 MB \\
        NeLF-Pro~\cite{nelf-pro} & 21.88 & 0.795 & 0.367 & 89 MB \\
        NeLF-Pro~\cite{nelf-pro}+Ours$_{MS}$ & \underline{22.52} & \underline{0.814} & 0.329 & 89 MB \\
        NeLF-Pro~\cite{nelf-pro}+Ours$_{MA}$ & \textbf{22.61} & \textbf{0.815} & \underline{0.326} & 89 MB \\
        \bottomrule
    \end{tabular}
}
\label{table:scannet++_nelf-pro_comparison}
\end{table}

However, the significance of our geometric regularization is more noticeable in Figure~\ref{fig:qualitative_scannet++}, which presents results on view extrapolation scenarios.
The quantitative evaluation in Table~\ref{table:scannet++_nelf-pro_comparison} follows a conventional practice in neural rendering~\cite{llff, nerf, mip-nerf360}, where test views are frames uniformly sampled from training trajectories, which mainly contain view-interpolation settings.
However, the ScanNet++ dataset contains challenging test views that require view extrapolation, for which many conventional approaches, including 3DGS, have limited capability.
The renderings of baseline methods in Figure~\ref{fig:qualitative_scannet++} contain severe artifacts, especially in textureless walls, which are significantly suppressed with the proposed techniques.

%% Zip-NeRF dataset
To validate our framework using more challenging multi-room indoor scenarios, we demonstrate results with the Zip-NeRF dataset.
Here, we validate our scheme using both NeLF-Pro and LocalRF. 
Similar to experiments conducted on ScanNet++ dataset, we also tested our methods using geometric scaffolds from MonoSDF and MapAnything.
As demonstrated in Table~\ref{table:zipnerf_comparison}, our method greatly enhances both frameworks using multiple NeRF blocks and neural lightfield probes.
Notably, combining LocalRF with our strategy outperforms Mega-NeRF as shown in Figure~\ref{fig:qualitative_zipnerf}.
While LocalRF variants require more memory due to the TensoRF framework, Mega-NeRF uses less memory with the original NeRF implementation, taking over 2 days to train for up to 200k iterations.  
Meanwhile, Hierarchical-3DGS enables very fast rendering using 3D Gaussians within multiple chunks, leading to excessive memory consumption as the scene gets larger.
While LongSplat also targets scalability for large scenes, its design fundamentally relies on continuous video inputs and a built-in pose estimation mechanism. In our experimental setup, this built-in module exhibited limited robustness, ultimately resulting in degraded rendering quality.

%% Add extrapolation here
To evaluate extrapolation tasks on the Zip-NeRF datasets, which only include interpolation test settings, we rendered novel views along free camera trajectories.
In this scenario, 3DGS suffers from degradation with sharp artifacts as demonstrated in Figure~\ref{fig:zipnerf_extrapolate}.
The additional results in the supplementary material demonstrate that our framework produces cleaner reconstructions in under-constrained regions, where even the SOTA 3DGS methods still exhibit artifacts.

\begin{table}[t!]
\caption{\textbf{Performance comparison of methods on Zip-NeRF~\cite{zip-nerf} dataset.} Our framework is both tested on neural lightfield probes and NeRF blocks. %Same highlighting and sectioning rules are applied as in \cref{table:scannet++_nelf-pro_comparison}.
The first and second best results are highlighted in \textbf{bold} and \underline{underlined}, respectively.
}
\centering
\resizebox{\linewidth}{!}{
    \begin{tabular}{l|c|c|c|c}
        \toprule
        Method & PSNR $\uparrow$ & SSIM $\uparrow$ & LPIPS $\downarrow$ & Memory \\
        \midrule
        MonoSDF~\cite{monosdf} & 20.61 & 0.733 & 0.459 & 61 MB \\
        NeRFacto~\cite{nerfstudio} & 19.78 & 0.692 & 0.491 & 65 MB \\
        NeRFacto~\cite{nerfstudio}-Big & 17.31 & 0.626 & 0.556 & 197 MB\\
        NeRFacto~\cite{nerfstudio}-Huge & 17.41 & 0.627 & 0.514 & 211 MB \\
        Depth-NeRFacto~\cite{nerfstudio} & 20.64 & 0.710 & 0.486 & 65 MB \\
        ZipNeRF~\cite{zip-nerf} & 23.53 & 0.758 & 0.415 & 1.2 GB \\
        3DGS~\cite{3dgs} & 23.70 & \textbf{0.794} & \textbf{0.347} & 225 MB \\
         Mip-Splatting~\cite{Mip-Splatting} &  22.18 &  0.744 &  0.435 &  323 MB \\
         Scaffold-GS~\cite{Scaffold-GS} &  22.76 &  0.753 &  0.398 &  104 MB \\
         PlanarGS~\cite{PlanarGS} &  22.38 &  0.749 &  0.420 &  220 MB \\
        FSGS~\cite{FSGS} & 22.55 & 0.751 & 0.451 & 134 MB \\
        DNGaussian~\cite{dngaussian} & 19.38 & 0.694 & 0.679 & 170 MB \\
        \midrule
        Mega-NeRF~\cite{mega-nerf} & 22.65 & 0.743 & 0.496 & 80 MB \\
        Hierarchical-3DGS~\cite{hierarchical_3dgs} & 22.30 & 0.757 & 0.419 & 1.5 GB \\
         LongSplat~\cite{LongSplat} &  19.95 &  0.714 &  0.488 &  180 MB \\        
        NeLF-Pro~\cite{nelf-pro} & 23.42 & 0.750 & 0.421 & 160 MB \\
        NeLF-Pro~\cite{nelf-pro}+Ours$_{MS}$ & \underline{23.89} & 0.762 & 0.400 & 160 MB \\
        NeLF-Pro~\cite{nelf-pro}+Ours$_{MA}$ & \textbf{23.96} & \underline{0.768} & \underline{0.393} & 160 MB \\
        LocalRF~\cite{localrf} & 21.39 & 0.730 & 0.484 & 1.9 GB \\ 
        LocalRF~\cite{localrf}+Ours$_{MS}$ & 23.02 & 0.761 & 0.471 & 1.9 GB \\
        LocalRF~\cite{localrf}+Ours$_{MA}$ & 23.06 & 0.763 & 0.468 & 1.9 GB \\
        
    \bottomrule
    \end{tabular}
}
\label{table:zipnerf_comparison}
\end{table}
\begin{figure}[t]
    \centering
    \includegraphics[width=\linewidth]{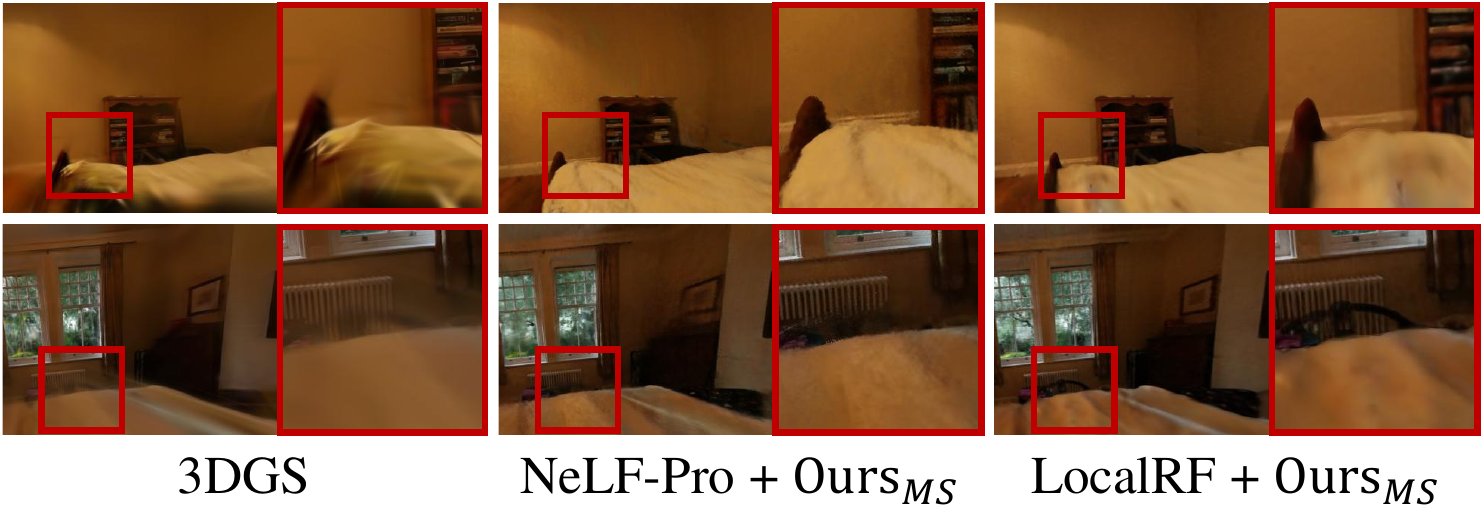}
    \caption{\textbf{View extrapolation on Zip-NeRF~\cite{zip-nerf} datasets.} 3DGS~\cite{3dgs} exhibits noticeable artifacts when rendering extrapolated views due to its limited generalization beyond observed camera trajectories.
    In contrast, our scene-adaptive strategies effectively mitigate these artifacts, resulting in more stable and visually consistent novel-view synthesis under extrapolation.}
    \label{fig:zipnerf_extrapolate}
\end{figure}

In experiments with both datasets, our scene-adaptive strategy enables improved novel view synthesis in both interpolation and extrapolation scenarios with optimal basis placement and geometric regularization.
We demonstrate consistent performance improvements with geometric scaffolds derived from both MonoSDF and MAPAnything.
As our method is built upon the original NeLF-Pro and LocalRF framework, the memory usage is identical to that of the original implementation, given the same number of bases. 
While obtaining geometric scaffolds using MonoSDF requires several hours of optimization, leveraging MapAnything allows us to extract geometric scaffolds in approximately 30 seconds.
The subsequent optimization of basis placement takes 13 minutes, followed by 50 seconds for selecting virtual viewpoints.
Optimizing the radiance fields using our strategy requires 24 minutes for NeLF-Pro and 6.5 hours for LocalRF, corresponding to a 15\% and 8\% additional increase in computational cost, respectively.
See supplementary material for additional results.

\subsection{Ablation Study}
\label{subsec:analysis}
\begin{table}
\caption{\textbf{Ablation studies on ScanNet++~\cite{scannetpp} and Zip-NeRF~\cite{zip-nerf} dataset.} We used multiple neural lightfield probes and nerf blocks for each case. Components marked by circled number denote the following:  $\circled{1}$: basis optimization, $\circled{2}$: robust depth loss for training views, $\circled{3}$: novel view depth regularization, and $\circled{4}$: L2 depth loss for monocular depth estimation model.}
\centering
\resizebox{\linewidth}{!}{
    \begin{tabular}{l|c|c|c|c|c|c|c}
        \toprule
        Method & \circled{1} & \circled{2} & \circled{3} & \circled{4} & PSNR $\uparrow$ & SSIM $\uparrow$ & LPIPS $\downarrow$ \\
        \midrule
        \multirow{5}{*}{NeLF-Pro~\cite{nelf-pro}} & \xmark & \xmark & \xmark & \xmark & 21.88 & 0.795 & 0.367 \\ 
          & \xmark & \cmark & \xmark & \xmark & 22.23 & 0.799 & 0.352  \\ 
          & \xmark & \cmark & \cmark & \xmark & 22.26 & 0.800 & 0.353 \\ 
          & \cmark & \xmark & \xmark & \xmark & 22.22 & 0.807 & 0.349 \\ 
          & \cmark & \cmark & \xmark & \xmark & 22.42 & 0.810 & 0.342 \\ 
        \midrule
        + Ours (full) & \cmark & \cmark & \cmark & \xmark & \textbf{22.52} & \textbf{0.814} & \textbf{0.339} \\
        \midrule
        \multirow{5}{*}{LocalRF~\cite{localrf}} & \xmark & \xmark & \xmark & \cmark & 21.39 & 0.725 & 0.484 \\
         & \xmark & \cmark & \xmark & \xmark & 21.95 & 0.740 & 0.482 \\
         & \xmark & \cmark & \cmark & \xmark & 22.74 & 0.752 & 0.477 \\
         & \cmark & \xmark & \xmark & \cmark & 22.81 & 0.756 & 0.478 \\
         & \cmark & \cmark & \xmark & \xmark & 22.90 & 0.759 & 0.472 \\
        \midrule
        + Ours (full) & \cmark & \cmark & \cmark & \xmark & \textbf{23.02} & \textbf{0.761} & \textbf{0.471} \\
        \bottomrule
    \end{tabular}
}
\label{table:ablation}
\end{table}

In Table~\ref{table:ablation}, we compare image metrics for different cases to analyze the effect of our basis placement and geometric regularization strategy. 
For NeLF-Pro and LocalRF variants, we tested on ScanNet++ and Zip-NeRF dataset each.
To check the effect of our main components, we conduct experiments for basis optimization, robust depth loss in Equation~\ref{eq:robust_depth} on training viewpoint, and robust depth loss for novel views.  
% We observe that each component of our strategy contributes to the increase in reconstruction quality as shown in Table~\ref{table:ablation}.
As demonstrated in Table~\ref{table:ablation}, each component contributes to reconstruction quality improvement. 
Furthermore, with LocalRF variants, we compare the effect of depth supervision using a monocular depth~\cite{dpt} with the supervision using depth from a geometric scaffold.
For monocular depth supervision, we add the regularization term defined in Equation~\ref{eq:depth_loss}.
Interestingly, our depth supervision strategy that uses geometric scaffold outperforms the original depth supervision using DPT~\cite{dpt}, in the original LocalRF implementation. See supplementary materials for qualitative comparisons.

\begin{table}[t!]
\centering
\caption{\textbf{Ablation study for different basis placements.} We compare optimal basis placement against uniform-in-geometry and trajectory-based placements, reporting PSNRs for different numbers of bases.}
% \footnotesize
\resizebox{\linewidth}{!}{
    \begin{tabular}{@{}l|c|c|c|c|c|c}
        \toprule
        \# Basis & 4 & 8 & 16 & 32 & 64 & 128 \\
        \midrule
        trajectory & 19.89 & 19.69 & 19.97 & 19.90 & 20.15 & 20.29 \\
        uniform & 19.91 & 19.87 & 19.95 & 20.22 & 20.11 & 19.94 \\
        optimal (ours) & \textbf{20.21} & \textbf{20.07} & \textbf{20.37} & \textbf{20.29} & \textbf{20.46} & \textbf{20.31} \\
        \bottomrule
    \end{tabular}
}
\label{suppl_table:basis_ablation}
\end{table}

\begin{table}[tb!]
\centering
\caption{\textbf{Controlled comparison against different geometric regularizations.} We compare the effects of our geometric regularization with prior approaches while maintaining the same optimized basis placement.}
\small
\resizebox{\linewidth}{!}{
    \begin{tabular}{l|ccc|ccc}
        \toprule
        {} & \multicolumn{3}{c|}{ScanNet++~\cite{scannetpp}} & \multicolumn{3}{c}{Zip-NeRF~\cite{zip-nerf}}\\ 
        {Method} & PSNR $\uparrow$ & SSIM $\uparrow$ & LPIPS $\downarrow$ 
        & PSNR $\uparrow$ & SSIM $\uparrow$ & LPIPS $\downarrow$ \\
        \midrule
        SparseNeRF~\cite{sparsenerf} & 21.75 & 0.808 & 0.332 & 20.64 & 0.710 & 0.486 \\ 
        RegNeRF~\cite{regnerf} & 18.39 & 0.728 & 0.482 & 18.26 & 0.637 & 0.567 \\
        DiffNeRF~\cite{diffnerf} & 18.57 & 0.736 & 0.473 & 18.54 & 0.649 & 0.544 \\
        Ours & \textbf{22.52} & \textbf{0.814} & \textbf{0.329} & \textbf{23.89} & \textbf{0.762} & \textbf{0.400} \\
        \bottomrule
    \end{tabular}
}
\label{suppl_table:regularization_comparison}
\end{table}

We conducted additional ablation studies to analyze the effect of basis placement strategies on reconstruction quality. Specifically, we compared three configurations: (1) uniform distribution within the estimated geometric scaffold, (2) uniform placement along the camera trajectory, and (3) optimal placement based on scene geometry and observation statistics. These experiments were conducted across varying numbers of bases to assess consistency and performance under different representation budgets. We demonstrate the results using the  ScanNet++~\cite{scannetpp} \textsc{e91722b5a3} scene in Table~\ref{suppl_table:basis_ablation}, clearly indicating that optimized placement consistently outperforms both uniform-in-geometry and trajectory-based strategies. 
These results highlight the importance of scene-adaptive placement for achieving efficient and expressive representations.

Furthermore, we demonstrate the result of additional controlled experiments to verify the effectiveness of our geometry-aware regularization strategy.
While keeping the basis placement are fixed after optimization, we tested on different regularization methods.
We compared our depth regularization strategy against regularization methods in RegNeRF~\cite{regnerf}, DiffNeRF~\cite{diffnerf} and SparseNeRF~\cite{sparsenerf} in Table~\ref{suppl_table:regularization_comparison}.
Because the depth maps used in our setting are noisy and the training observations are relatively dense, these prior regularization strategies do not consistently improve reconstruction quality in our experiments.
In contrast, the robust depth loss combined with our scene-adaptive geometric regularization using the correlated information of the geometry and camera demonstrates the significant performance gains across all evaluation metrics. 

\begin{figure}[t]
    \centering
    \includegraphics[width=\linewidth]{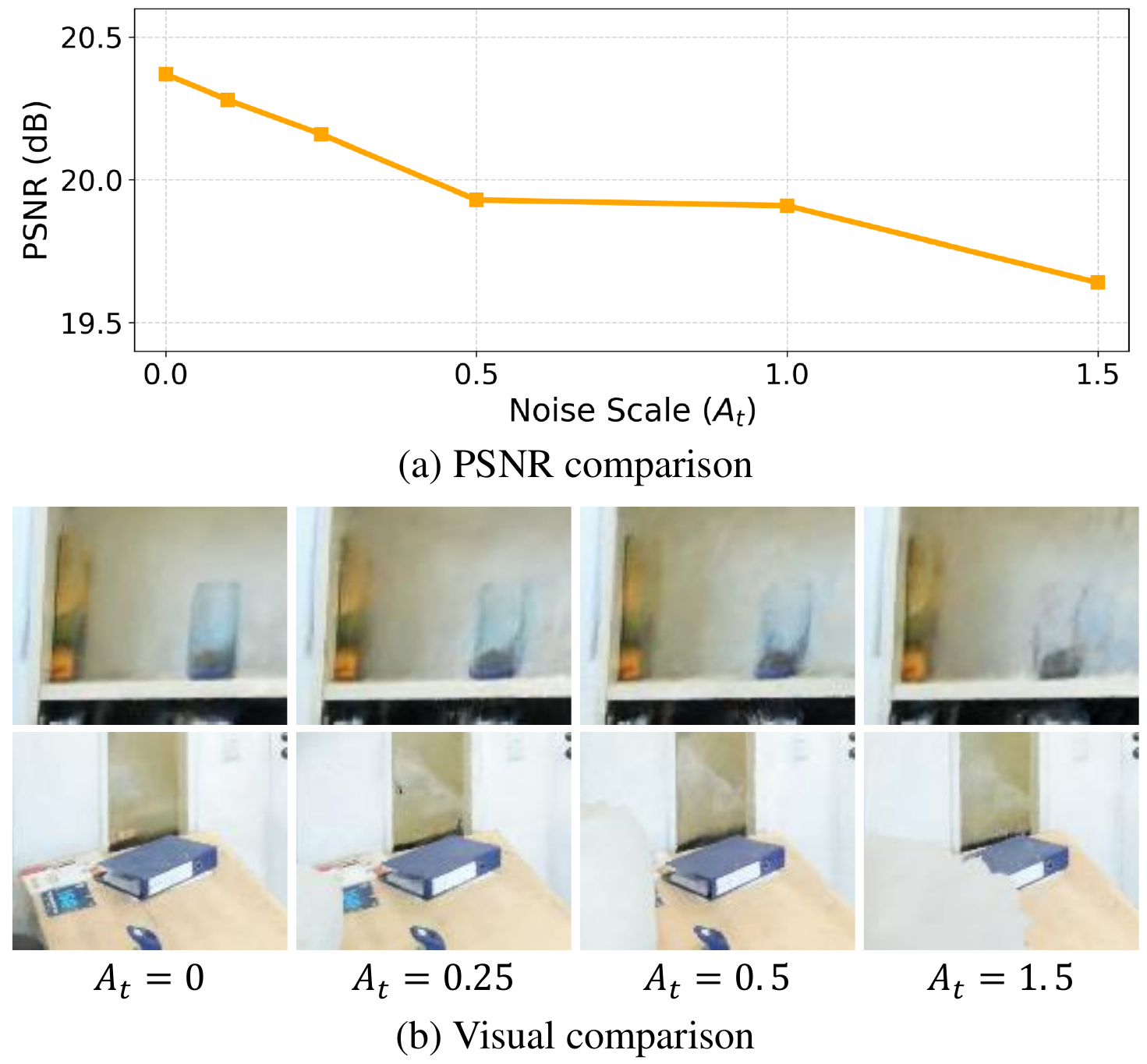}
    \caption{\textbf{Shifting optimized basis positions.} (a) PSNR comparisons with different levels of perturbation noise scales. (b) Enlargements of rendered images for visual comparison of sampled noise levels.}
    \label{fig:shifting_basis}
\end{figure}

%% Added after ICCV rebuttal
\subsection {Perturbing Optimized Basis}
To understand the robustness of our optimized basis configuration, we explore how sensitive the system is to perturbations in basis placement.
We conducted experiments to verify the effect of shifting bases from the optimized positions.
Using the optimal set of basis positions, we have gradually added the random noise to the basis positions with varying noise scales from 0.1 to 1.5 and measured the PSNR of the rendering results compared to the validation set.
Here, the noise scale corresponds to the amplitude of isotropic random perturbations applied to the probe positions defined in the NeRF camera-to-world coordinate system.
To clearly analyze the effects of basis perturbation, we used neural lightfield probes, reducing the number of basis probes to 16, with 4 nearby probes contributing to rendering.
We have tested on ScanNet++~\cite{scannetpp} \textsc{e91722b5a3} scene for this experiment. 
As the bases are perturbed, they deviate from the regions with the high-density training rays, hence making it difficult to model highly informative regions.
% Perturbed bases deviate from high-density training ray regions, degrading modeling of informative areas.
The rendering quality degrades gradually as the basis placement perturb from the optimal as shown in Figure~\ref{fig:shifting_basis}.

\subsection{Robustness to Inaccurate Geometry}
\label{subsec:robustness}
\begin{figure}[t!]
    \centering
    \includegraphics[width=\linewidth]{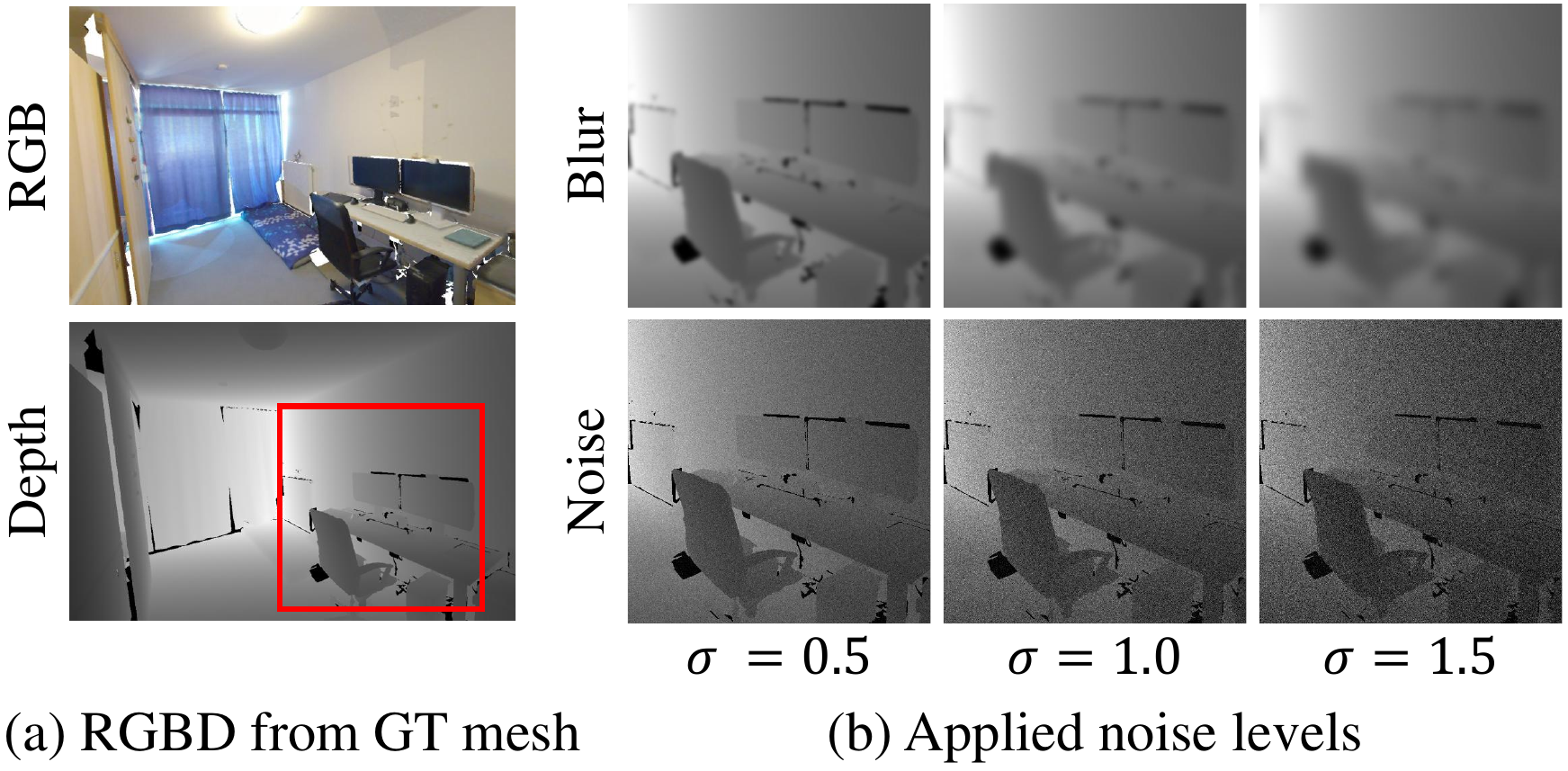}
    \caption{\textbf{Examples of inaccurate geometries.} (a) Rendered RGBD from GT mesh and (b) example depth inputs with different noise levels.
    }
    \label{fig:inaccuate_geometries}
\end{figure}

%% Additional experiments
To further evaluate the robustness of our method across diverse scenarios, we analyze its performance under inaccurate geometry. 
As we use geometry scaffolds from the implicit surface reconstruction model and the pretrained feed-forward model, predictions could be inaccurate. 
To test the robustness of inaccurate geometry, we examine the performance of our method when the geometric scaffolds are contaminated.
% For these experiments, we directly use the ground truth geometry to simulate the scaffold and apply different degrees of noise to observe the model's dependency on the accuracy of geometry prediction.
For these experiments, we use ground-truth geometry as the scaffold and introduce controlled levels of noise to evaluate the sensitivity of the proposed method to inaccuracies in the geometric prior.
For ScanNet++~\cite{scannetpp} scenes, ground truth scanned meshes are given. We rendered depth using the mesh for \textsc{ef69d58016} scene and used it as the ground truth depth. 
We have perturbed the ground truth depth by adding Gaussian blur and random noise with different scales as shown in Figure~\ref{fig:inaccuate_geometries}.
%Note that the RGB image in Figure~\ref{fig:inaccuate_geometries}(a) is only given for visualization and not used for our experiments.
We report the performance comparisons with different noise levels in Table~\ref{suppl_table:inaccurate_depth}. 
Even when we used some severely noised depth inputs, the reconstruction still shows improved results compared to the case without depth supervision. 

\begin{table}[t!]
\centering
\caption{\textbf{Performance comparisons for experiments using inaccurate depth inputs.} Even with imperfect geometric information, our geometric regularization strategy remains effective and improves performance.}
\resizebox{\linewidth}{!}{
    \begin{tabular}{l|c|l|c|l|c}
        \toprule
        \textbf{Blur} & PSNR & \textbf{Noise} & PSNR & \textbf{Reference} & PSNR \\
        \midrule
        $\sigma = 0.25$ & 27.61 & $\sigma = 0.25$ & 27.61 & NeLF-Pro & 26.83 \\
        $\sigma = 0.5$ &  27.49 & $\sigma = 0.5$ & 27.52 & Ours w/o depth & 26.96\\
        $\sigma = 1.0$ &  27.40 & $\sigma = 1.0$ & 27.27 & Ours w/ GT depth ($\sigma=0$) & 27.67\\
        $\sigma = 1.5$ &  27.21 & $\sigma = 1.5$ & 27.23 & Ours w/ MonoSDF depth & 27.20\\
        \bottomrule
    \end{tabular}
}
\label{suppl_table:inaccurate_depth}
\end{table}

\subsection{Limitation}
\label{subsec:limitation}
\begin{figure}[t!]
    \centering
    \includegraphics[width=\linewidth]{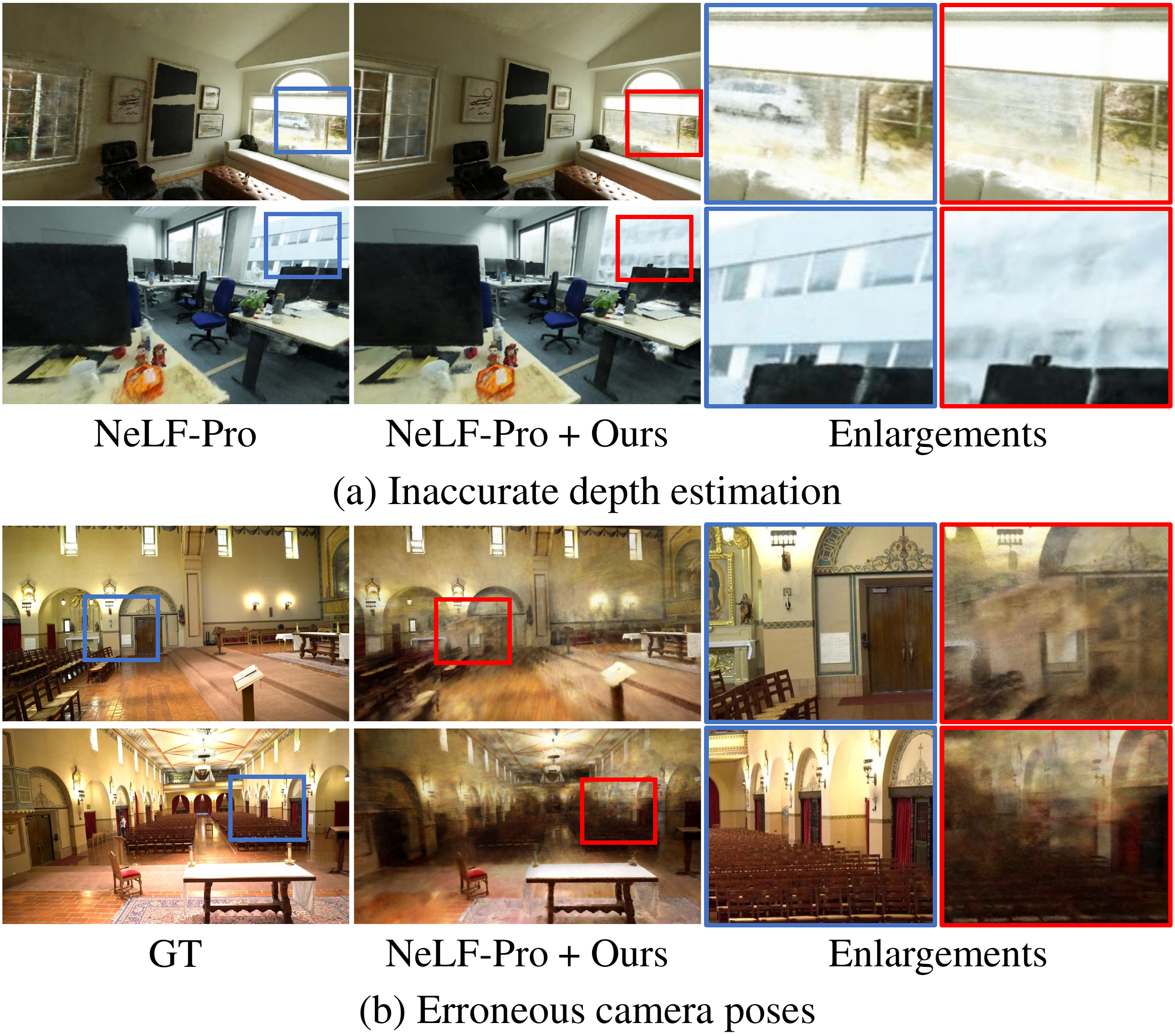}
    \caption{\textbf{Examples of limitations due to the failure of upstream preprocessing steps.} (a) Degradation from inaccurate geometry in distant outdoor regions. (b) Poor reconstruction from failed SfM calibration.}
    \label{rebuttal_fig:limitation}
\end{figure}

We present several limitations of our framework. 
First, our method degrades in regions where the geometric scaffold is severely inaccurate, as illustrated in Figure~\ref{rebuttal_fig:limitation}. Window regions are a representative example: the scaffold inherently struggles to model outdoor areas with large depth variation and ambiguous geometry, and although these regions are visible from indoor viewpoints, their appearance and depth remain poorly constrained. 
Second, our framework, like the vast majority of novel-view synthesis pipelines, assumes a baseline level of successful camera calibration and SfM reconstruction. 
Catastrophic failures in these upstream steps remain beyond the scope of the current work, and extending our framework to in-the-wild scenarios with more robust upstream reconstruction would be a valuable direction for future work.

\section{Conclusion} 
In this work, we propose scene-adaptive strategies to allocate basis using correlated information of geometry and camera trajectories to fully leverage limited resources.
From the preprocessed geometric scaffold, we find optimal positions of bases to represent large-scale indoor scenes.
We further regularize geometries that are under-constrained, using scene-aware virtual views and geometric priors.
Through extensive experiments, we have verified that our framework significantly enhances baselines for challenging scenarios, including extrapolation settings. 
% Moreover, we have demonstrated the result for varying numbers of bases, where we could represent scenes with fewer bases while maintaining reconstruction quality comparable to the original implementations.
We additionally evaluate different basis placement strategies, including perturbation experiments, which further highlight the importance of optimal basis placement. 
Moreover, we show that our method remains robust under imperfect geometric scaffolds, consistently yielding performance improvements.
% Moreover, we analyzed that our method can scale to large outdoor scenes in well-constrained regions and exhibits robustness under imperfect geometric scaffolds.
Although our approach shows performance improvements in general indoor environments, our framework does not account for view-dependent reflectance and may create artifacts in regions with complex lighting or strong reflectance.
Also, our results degrade in regions with inevitable errors in geometric scaffold, such as windows. 
Scaling our work to in-the-wild scenarios with complex appearance and lighting changes would be an interesting future work.

\section*{Acknowledgments}
This work was supported by Institute of Information \& communications Technology Planning \& Evaluation (IITP) grant funded by the Korea government (MSIT) (RS-2023-00216821, Development of Beyond X-verse Core Technology for Hyper-realistic interactions by Synchronizing the Real World and Virtual Space), the BK21 FOUR program of the Education and Research Program for Future ICT Pioneers, and Creative-Pioneering Researchers Program through Seoul National University in 2026.

\bibliographystyle{IEEEtran}
\bibliography{main}
\vspace{-1.5em}
\begin{IEEEbiography}[{\includegraphics[width=1in]{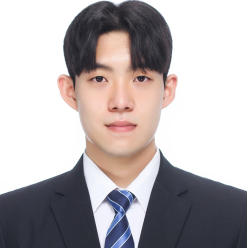}}]{Minkwan Kim}
% \begin{IEEEbiographynophoto}{Minkwan Kim}
received his B.S. in electrical
engineering from Seoul National University in
2021, where he is currently pursuing a Ph.D.
in the department of Electrical and Computer Engineering, Seoul National University. His research interests include 3D scene reconstruction, novel-view synthesis, and 3D scene understanding.
\end{IEEEbiography}
\vspace{-0.5em}
\begin{IEEEbiography}[{\includegraphics[width=1in]{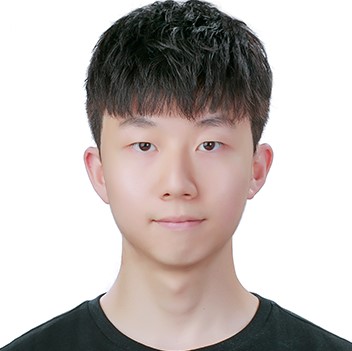}}]{Changwoon Choi} received B.S. degree from Seoul National University.
He is currently pursuing Ph.D. degree in the department of Electrical and Computer Engineering, Seoul National University.
His research interests include 3D reconstruction and neural rendering, ranging from photorealistic reconstruction to more conceptual and abstract representations.
\end{IEEEbiography}
\vspace{-0.5em}
\begin{IEEEbiography}[{\includegraphics[width=1in]{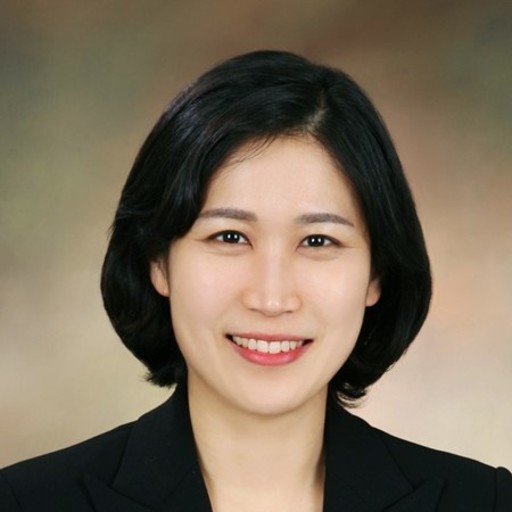}}]{Young Min Kim}
is an Associate Professor in the Department of Electrical and Computer Engineering at Seoul National University, Seoul, Korea, where she is leading a 3D vision lab. She received a B.S. from Seoul National University in 2006 and an M.S. and Ph.D. in electrical engineering from Stanford University in 2008 and 2013, respectively. Before joining SNU, she was a Senior Research Scientist at the Korea Institute of Science and Technology (KIST). She serves as an area chair in CVPR, ICCV, ECCV, ACCV, program committee for Pacific Graphics, AAAI, and technical papers committee for SIGGRAPH and SIGGRAPH Asia. She is also a program chair for 3DV 2026. Her research interest lies in 3D vision, where she combines computer vision, graphics, and robotics algorithms to solve practical problems. 
\end{IEEEbiography}
\vfill

% \end{document}
% \fi
%%%%%%%%%%%%%%%%%%%%%%%%%%%
%% Supplementary
% \if 0
% \begin{document}
\clearpage
% \clearpage
% \setcounter{page}{1} % For arXiv
\if 0 % arXiv
\setcounter{figure}{0}
\setcounter{table}{0}

\title{Supplementary Material for Geometry-Aware Scene Configurations for Novel View Synthesis}
\author{}
\vspace{-10pt}

\fi

\twocolumn[
\begin{center}
{\Huge Supplementary Material for
Geometry-Aware Scene Configurations for Novel View Synthesis}
\vspace{3em}
\end{center}
]

\maketitle   
\section {Basis Placement Optimization}
\subsection {Algorithms and Optimization Results}
In this section, we demonstrate the results of the obtained coverage weights and corresponding basis positions.
We outline the detailed process of obtaining a coverage weight map in Algorithm~\ref{alg:coverage_weight}.
To obtain the coverage weight for each point in the geometry scaffold, we accumulate the observation statistics of all training cameras, with the consideration of the visibility of each point. 
We calculate visibility by projecting the points sampled from the geometry scaffold into the camera frustum, and discard those whose orientations face away from the camera according to the scaffold normals.
Additionally, we leverage depth map which is directly obtained from geometry scaffold to further filter out points located behind the depth surface.
Once the coverage weight map is constructed, it serves as an input to the basis optimization process as described in Algorithm~\ref{alg:basis_optimization}.

\DontPrintSemicolon
\SetCommentSty{algcommentfont}
\SetKwFunction{KwFn}{Fn}
\SetKw{Continue}{continue}
\SetNlSty{}{}{}

\begin{algorithm}[b!]
    \SetKwInOut{Input}{Input}
    \SetKwInOut{Output}{Output}
    \small
    \Input{Geometry scaffold $\mathcal{G}$, Cameras $\mathcal{C}$
    }    
    \Output{Coverage weight $w_i$}
    \BlankLine
    \textbf{Initialize:} \;
    $\mathbf{x}_i, \hat{\mathbf{n}}_i \gets$ point, normal from $\mathcal{G}$\;
    $\mathbf{c}_k$, $\mathcal{M}_k \gets$ camera center, proj matrix from $\mathcal{C}$\;
    $\mathcal{D}_k \gets$ Depth map from $\mathcal{G}$\;
    $w_i \gets 0$\;

    \ForEach{$\{\mathbf{c}_k, \mathcal{M}_k\} \in \mathcal{C}$}{
        % Calculate visibility mask
        Vis$(\mathbf{c}_k \to \mathbf{x}_i) \leftarrow 1$\;
        $(u_i, v_i, z_i) \leftarrow$ project $\mathbf{x}_i$ with  $\mathcal{M}_k$\;
        \If{$|u_i|>1$ \textbf{or} $z_i<0$}{
            Vis$(\mathbf{c}_k \to \mathbf{x}_i) \leftarrow 0$\;
        }       
        \If{$z_i - \mathcal{D}_k > \epsilon$}{
            Vis$(\mathbf{c}_k \to \mathbf{x}_i) \leftarrow 0$\;
        }
        \If{$\hat{\mathbf{n}}_i \cdot (\mathbf{c}_k - \mathbf{x}_i) < 0$}{
            Vis$(\mathbf{c}_k \to \mathbf{x}_i) \leftarrow 0$\;
        }
        % Update coverage weight
        $w_i \gets w_i + \displaystyle\frac{\hat{\mathbf{n}}_i \cdot (\mathbf{c}_k - \mathbf{x}_i)}{\|\mathbf{c}_k - \mathbf{x}_i\|^2} \cdot \text{Vis}(\mathbf{c}_k \to \mathbf{x}_i)$
    }     
    \Return $w_i$
    \caption{Coverage weight calculation}
    \label{alg:coverage_weight}
\end{algorithm}

\begin{figure*}[p]
    \centering
    \includegraphics[width=0.9\linewidth]{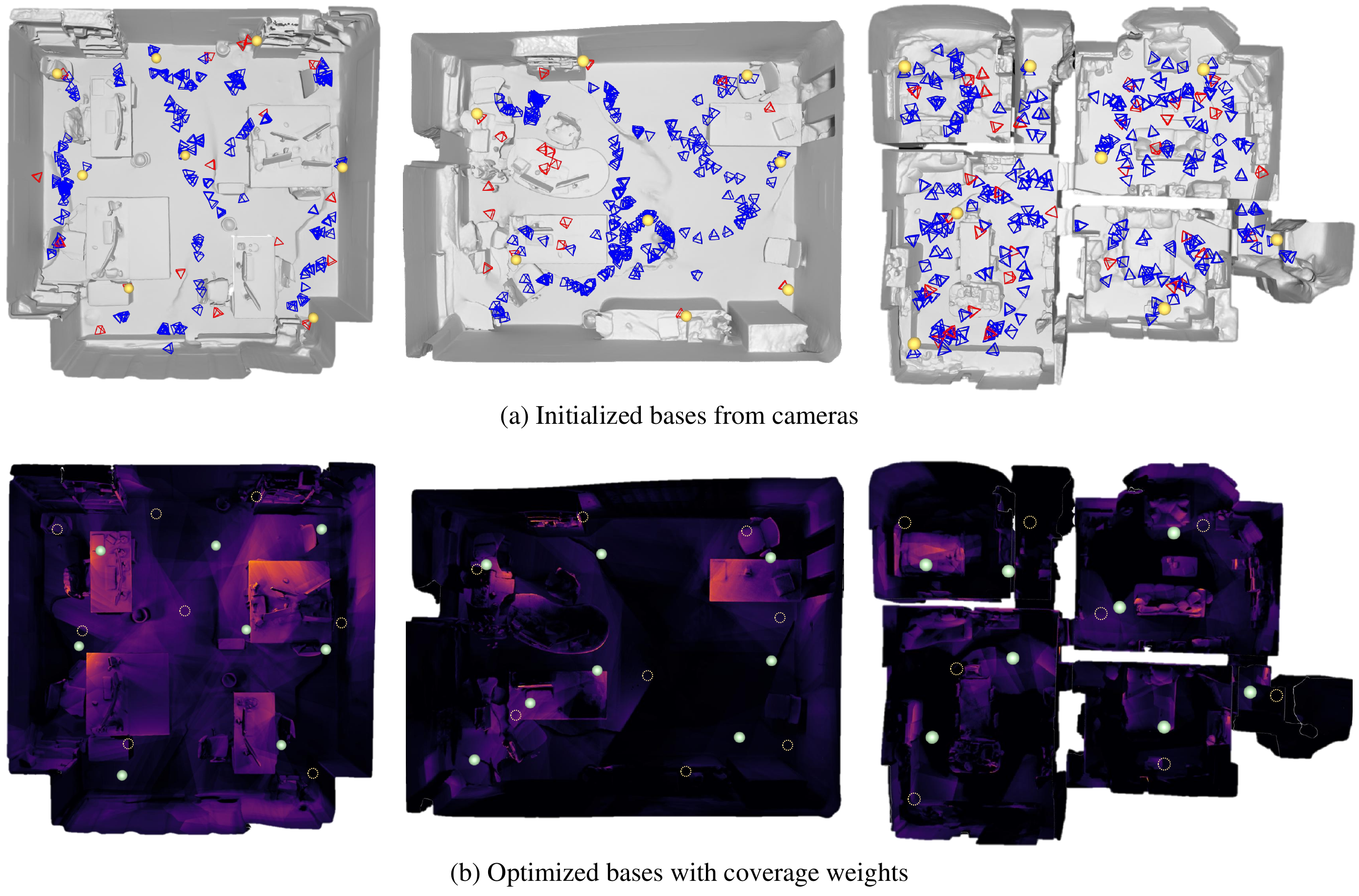}
    \caption{\textbf{Coverage weights and optimized basis positions.} (a) Given training (blue) and test (red) cameras, initial bases (yellow spheres) are obtained using FPS on cameras. 
    (b) Optimized bases are distributed at balanced positions regarding the coverage weight. Initial bases locations are indicated with sphere outlines, while the optimized bases are solid spheres in a bright green color.
    }
\label{suppl_fig:coverage_weights_and_basis}
\end{figure*}

We illustrate coverage weights and optimized basis positions in Figure~\ref{suppl_fig:coverage_weights_and_basis} for two example scenes, each from ScanNet++~\cite{scannetpp} and Zip-NeRF~\cite{zip-nerf} datasets.
To help understand of our optimization scheme, we optimized eight basis positions, while 64 and 128 bases are used in real experiments for NeLF-Pro~\cite{nelf-pro} variants. We visualize the initial basis positions, with training and evaluation cameras in Figure~\ref{suppl_fig:coverage_weights_and_basis}(a), while the coverage weight map and optimized basis positions are depicted in Figure~\ref{suppl_fig:coverage_weights_and_basis}(b). 
The basis positions are effectively drawn to regions with frequent measurements.

\DontPrintSemicolon
\SetCommentSty{algcommentfont}
\SetKwFunction{KwFn}{Fn}
\SetKw{Continue}{continue}
\SetNlSty{}{}{}

\begin{algorithm}[t!]
    \SetKwInOut{Input}{Input}
    \SetKwInOut{Output}{Output}
    \small
    \Input{
        Point set $\mathcal{X} = \{\mathbf{x}_i | \mathbf{x}_i \in \mathbb{R}^3\}_{i=1}^B$, \\
        Normal set $\mathcal{N} = \{\hat{\mathbf{n}}_i | \hat{\mathbf{n}}_i \in \mathbb{R}^3\}_{i=1}^B$, \\
        Coverage weight set $\mathcal{W} = \{w_i | w_i \in \mathbb{R}\}_{i=1}^B$
    }
    \Output{Optimized basis positions $\mathcal{P} = \{\mathbf{p}_j | \mathbf{p}_j \in \mathbb{R}^3\}_{j=1}^{N_p}$}

    \textbf{Initialize:} \;
    $\mathcal{P} \gets$ Initial basis positions using FPS algorithm\;

    \For{iter = 1 \textbf{to} maxiter}{
        $\mathbf{p}_{\mathcal{C}(i)} \leftarrow \arg\min_{\mathbf{p}_j} \|\mathbf{p}_j - \mathbf{x}_i\| \quad \forall \mathbf{x}_i \in \mathcal{X}, \mathbf{p}_j \in \mathcal{P}$ 
        
        $\mathcal{L}_{\text{cov}} \leftarrow \displaystyle\sum_{i=1}^Bw_i\left(\frac{\left\Vert \mathbf{p}_{\mathcal{C}(i)}-\mathbf{x}_i 
        \right\Vert^3}{\hat{\mathbf{n}_i}\cdot(\mathbf{p}_{\mathcal{C}(i)}-\mathbf{x}_i)+\epsilon}\right)$

        $\mathbf{p}_{\mathcal{C}(i)} \leftarrow \displaystyle\mathbf{p}_{\mathcal{C}(i)} - \alpha\nabla_{\mathbf{p}_{\mathcal{C}(i)}}\mathcal{L}_{\text{cov}}$  
    }
    
    \Return $\mathcal{P}$

    \caption{Basis placement optimization}
    \label{alg:basis_optimization}
\end{algorithm}

\begin{figure}[t]
    \centering
    \includegraphics[width=\linewidth]{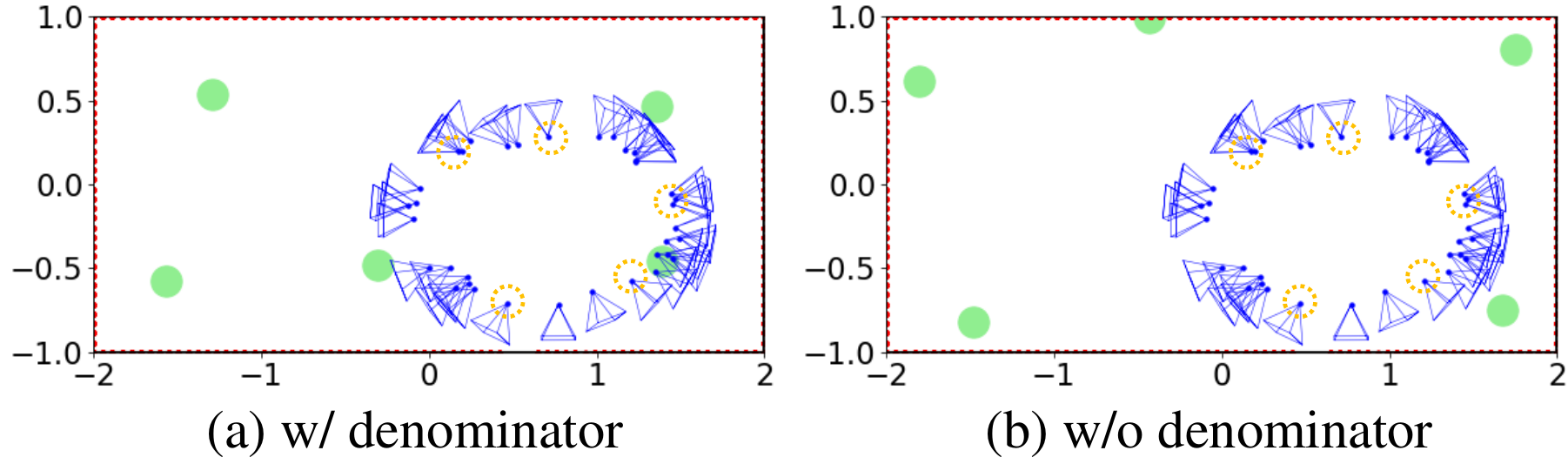}
    \caption{\textbf{2D toy examples of basis placement.} We evaluate basis placement using \( \mathcal{L}_{\text{cov}} \) (a) with and (b) without the denominator. Initial basis (circle outlines), optimized basis (solid green circles), and observation views (blue frustums) are visualized.}
    \label{suppl_fig:toy_example}
\end{figure}

\begin{figure*}[p]
    \centering
    \includegraphics[width=0.9\linewidth]{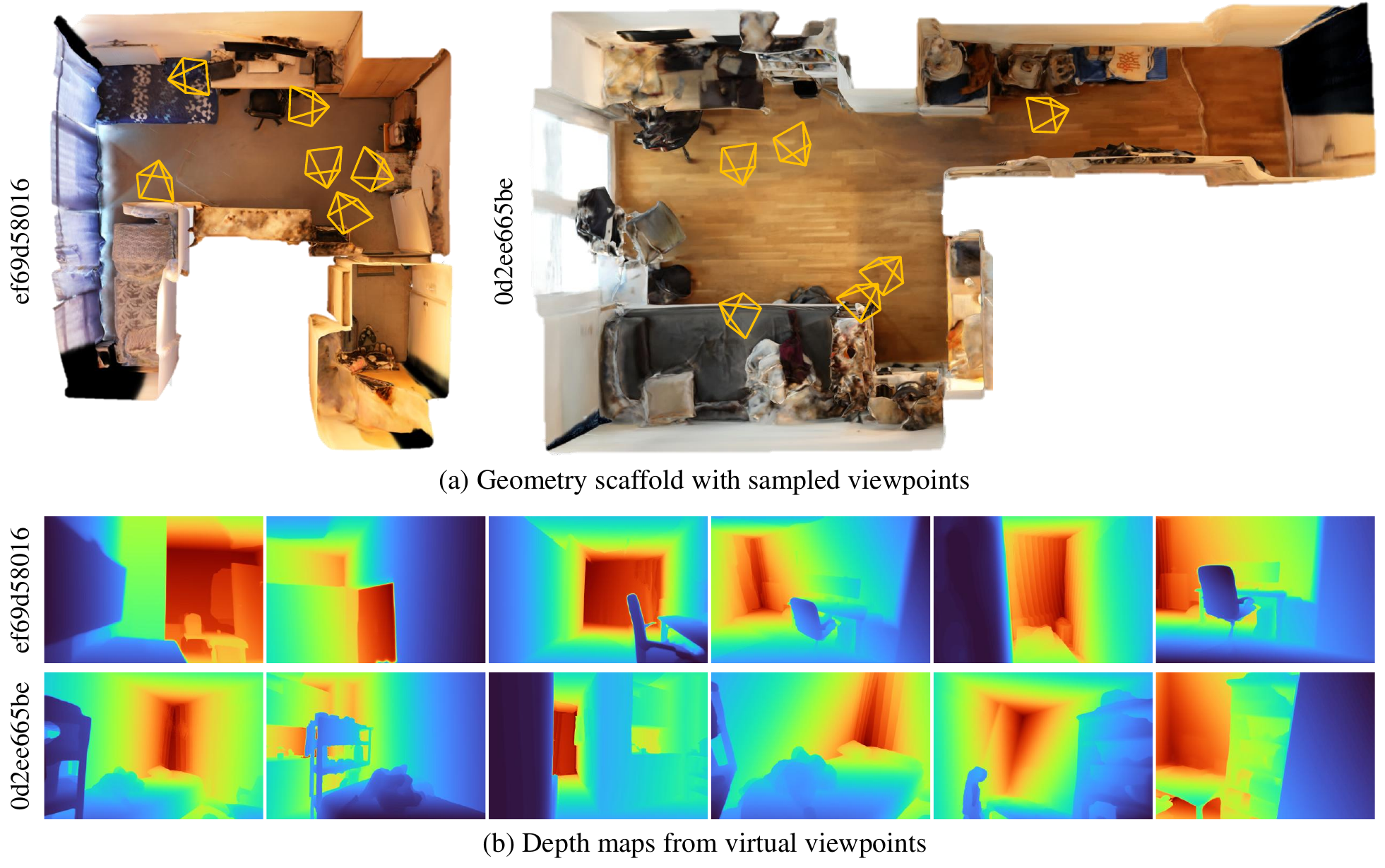}
    \caption{\textbf{Sampled virtual depths from ScanNet++~\cite{scannetpp} dataset.} (a) Sampled virtual viewpoints with the geometry scaffold and (b) the corresponding rendered depth maps.}
    \label{suppl_fig:virtual_depths} 
\end{figure*}
\subsection {Analysis of Coverage Loss}
The coverage loss $\mathcal{L}_{\text{cov}}$ in Equation 6 of the main manuscript is designed to ensure that the basis positions effectively accumulate the coverage while considering the geometry of the scene.
In the coverage loss term, we can decompose the component multiplied by the coverage weight as follows:
\begin{equation}
    \frac{\left\Vert \mathbf{p}_{\mathcal{C}(i)}-\mathbf{x}_i 
    \right\Vert^3}{\hat{\mathbf{n}_i}\cdot(\mathbf{p}_{\mathcal{C}(i)}-\mathbf{x}_i)+\epsilon}
    =\frac{\left\Vert \mathbf{p}_{\mathcal{C}(i)}-\mathbf{x}_i 
    \right\Vert^2}{\hat{\mathbf{n}_i}\cdot(\widehat{\mathbf{p}_{\mathcal{C}(i)}-\mathbf{x}_i)}+\tilde{\epsilon}},
    % =\frac{\left\Vert \mathbf{p}_{\mathcal{C}(i)}-\mathbf{x}_i 
    % \right\Vert^2}{\cos\theta_i + \epsilon},
\end{equation}
where $\widehat{(\mathbf{p}_{\mathcal{C}(i)} - \mathbf{x}_i)}$ denotes the normalized directional vector.
The term $\Vert\mathbf{p}_{\mathcal{C}(i)}-\mathbf{x}_i\Vert$ in the numerator represents the Euclidean distance between a point $\mathbf{x}_i$ and its closest basis position $\mathbf{p}_{\mathcal{C}(i)}$.
The denominator \( \hat{\mathbf{n}_i} \cdot\widehat{(\mathbf{p}_{\mathcal{C}(i)} - \mathbf{x}_i)} + \tilde{\epsilon} \) incorporates the normal vector \( \mathbf{n}_i \) at \( \mathbf{x}_i \) to enforce alignment between the basis position and the underlying surface geometry.
Without the denominator, the loss would only encourage closeness in Euclidean distance, disregarding the local surface orientation as illustrated in the 2D toy example of Figure~\ref{suppl_fig:toy_example}.
The denominator ensures that the basis positions are not only close to the points but also aligned with the surface normal, resulting in more accurate placement within structured environments.
The small term \( \epsilon \) is added to the denominator to prevent numerical instability. % when the dot product \( \hat{\mathbf{n}_i} \cdot{(\mathbf{p}_{\mathcal{C}(i)} - \mathbf{x}_i)}\) approaches zero.

\begin{figure}[t]
    \centering
    \includegraphics[width=0.95\linewidth]{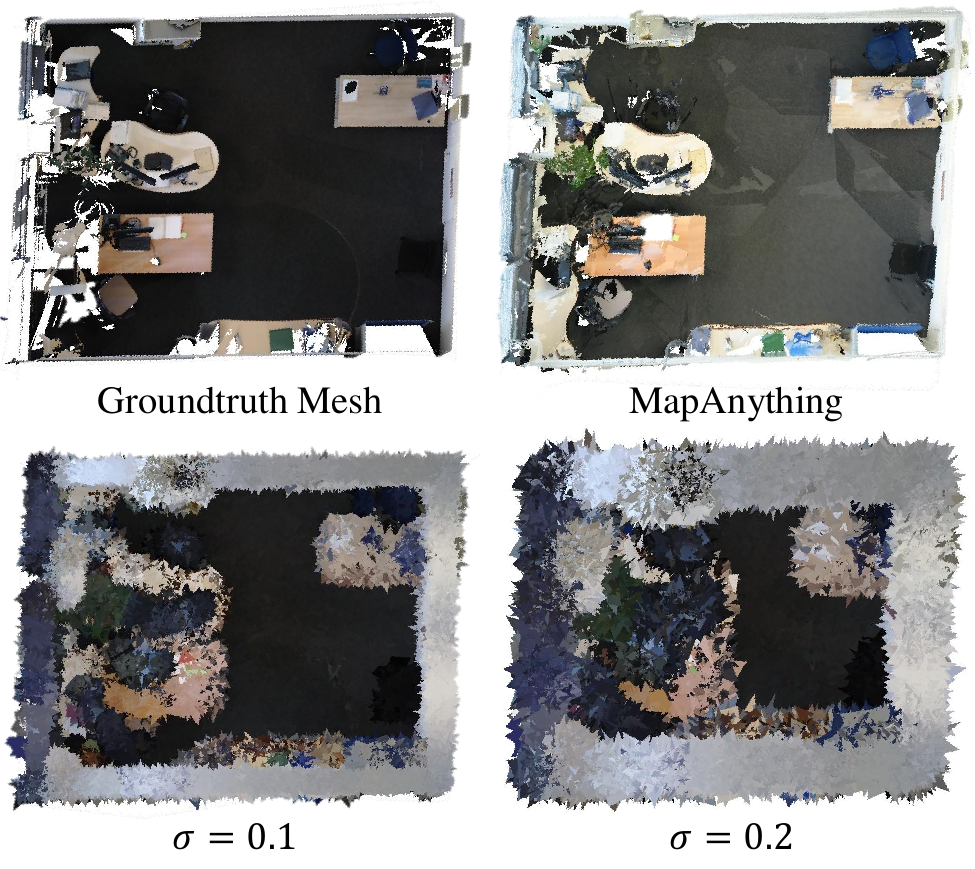}
    \caption{\textbf{Mesh visualization with different noise levels on depth maps.} Noise levels are defined by the standard deviation $\sigma$ in metric units.}
    \label{suppl_fig:noisy_mesh}
\end{figure}
\begin{table}[t!]
\centering
\caption{\textbf{Effect of geometric noise levels on basis placement.} We measure the sensitivity across different noise levels applied to depth and normal inputs on ScanNet++ \textsc{E91722b5a3} scene. 
}
\resizebox{\linewidth}{!}{
    \begin{tabular}{l|c|l|c|l|c}
        \toprule
        \textbf{Depth} & PSNR & \textbf{Normal} & PSNR & \textbf{Reference} & PSNR \\
        \midrule
        $\sigma = 0.01$ & 20.29 & $\sigma = 0.01$ & 20.32 & GT mesh & 20.38 \\
        $\sigma = 0.05$ &  20.30 & $\sigma = 0.05$ & 20.25 & NeLF-Pro & 20.09\\
        $\sigma = 0.1$ &  20.16 & $\sigma = 0.1$ & 20.23 & Ours$_{MS}$(w/o depth) & 27.36\\
        $\sigma = 0.2$ &  20.07 & $\sigma = 0.2$ & 20.17 & Ours$_{MA}$(w/o depth) & 20.41\\
        \bottomrule
    \end{tabular}
}
% \vspace{-4em}
\label{suppl_table:basis_sensitivity}
\end{table}

\subsection{Systemic Sensitivity Analysis on Basis Placements}
We provide additional analysis to evaluate the sensitivity of the proposed adaptive basis placement strategy to (1) the quality of the geometric scaffold and (2) the weighting of observation statistics in the optimization.

\subsubsection{Robustness to Geometric Noise}
To analyze the influence of the geometric prior, we conduct robustness experiments by injecting Gaussian noise into the depth and surface normal inputs used to construct the scaffold. We perturbed the depth by adding Gaussian noise, where the standard deviation is defined in metric scale (meters). For normal noise, we applied unitless Gaussian perturbation directly to the 3D components of the surface normal vectors, followed by L2 re-normalization.
Specifically, we perturb the depth and normal maps with Gaussian noise of standard deviations $\sigma \in \{0.01, 0.05, 0.1, 0.2\}$ on the ScanNet++ \textsc{E91722b5a3} scene.
Quantitative results are reported in Table~\ref{suppl_table:basis_sensitivity}, and visualization of the noisy geometric scaffolds are illustrated in Figure~\ref{suppl_fig:noisy_mesh}.
Under moderate noise levels, our method consistently outperforms the trajectory-based baseline (NeLF-Pro), indicating that the proposed adaptive basis placement can effectively leverage coarse geometric cues.
At higher noise levels (e.g., $\sigma = 0.2$), the geometric scaffold becomes severely corrupted, which reduces the effectiveness of geometry-aware allocation.
Nevertheless, our method remains competitive with the baseline, demonstrating robustness to imperfect and noisy geometry.
These results suggest that our approach does not require watertight geometric priors, but instead benefits from approximate structural information that captures the overall scene layout.

\subsubsection{Sensitivity to Coverage Weights}
We further analyze the role of observation statistics in the energy-based optimization.
To this end, we perform an ablation study where the coverage weights in $\mathcal{L}_{cov}$ are replaced with uniform weights ($w_i = 1$), effectively removing the influence of observation statistics and relying solely on geometric features.
The results are summarized in Table~\ref{suppl_table:coverage_sensitivity}.
When coverage weight is removed, the optimization becomes purely geometry-driven, which leads to suboptimal basis placement, particularly in regions with uneven observation density.
In contrast, incorporating coverage statistics enables the model to account for observation redundancy and under-constrained areas, resulting in a more balanced allocation of representation capacity.
Overall, while geometric priors provide useful spatial structure, the integration of observation statistics is critical for aligning representation capacity with the underlying data distribution.
This highlights the importance of the proposed coupled formulation in achieving robust and effective basis placement.

\begin{table}[t!]
\centering
\caption{\textbf{Effect of coverage statistics on the energy-based basis allocation.} We compare the basis placement with geometry-only allocation ($w_i = 1$) and with coverage weight-based allocation.}
\resizebox{0.9\linewidth}{!}{
    \begin{tabular}{@{}l|c|c|c}
        \toprule
        Method & PSNR↑ & SSIM↑ & LPIPS↓ \\
        \midrule
        Geometry-only ($w_i=1$) & 19.98 & 0.744 & 0.398 \\
        NeLF-Pro & 20.11 & 0.752 & 0.378 \\
        NeLF-Pro+Ours & \textbf{20.41} & \textbf{0.770} & \textbf{0.367} \\
        \bottomrule
    \end{tabular}
}
\label{suppl_table:coverage_sensitivity}
\end{table}

\section{Scene-Adaptive Depth Regularization}
\subsection{Algorithms and Sampled Virtual Depths}

In this section, we clarify the objective used for virtual viewpoint selection.
In our framework, the selection objective consists of two complementary components: (1) spatial diversity of camera poses and (2) minimization of feature co-visibility redundancy.
First, to encourage spatial diversity, we measure the Euclidean distance between camera centers, prioritizing viewpoints that observe the scene from distinct spatial locations.
Second, to avoid redundant observations, we penalize candidate viewpoints that share a high degree of co-visibility with already selected views. 
Rather than relying on 2D image correspondences, we evaluate this co-visibility based on shared 3D feature points, where the technical details are elaborated below.

We present a detailed process for sampling virtual depths in Algorithm~\ref{alg:viewpoint_sampling}. In addition, we provide several supplementary examples of depth maps rendered from virtual viewpoints in Figure~\ref{suppl_fig:virtual_depths}.
Unlike common next-best-view planning setups~\cite{activenerf, progressive_camera_placement, NeRF_Director}, we do not have images for candidate viewpoints. 
In other words, feature correspondences between candidate viewpoints are not available.
Therefore, to calculate a scene similarity matrix $\mathcal{A}_{vs}$ in Section 3.4 of the main paper, we utilized the triangulated feature points from COLMAP~\cite{colmap_sfm}.
For the given 3D feature points and cameras, we computed the occlusion on camera rays using the \textsc{RaycastingScene} class in Open3D~\cite{open3d}. 
If the points are visible in multiple cameras, we update only the elements of the similarity matrix corresponding to those camera indices.
Depth maps sampled from two scenes of the ScanNet++ dataset are illustrated in Figure~\ref{suppl_fig:virtual_depths}.
The relative importance between spatial diversity and co-visibility redundancy is controlled by the parameter $\kappa$ in Equation 9 of the main manuscript, which is fixed to 0.1 in all experiments.
We provide a parameter sweep over $\kappa$ in Table~\ref{suppl_table:parameter_sweeping}.

\DontPrintSemicolon
\SetCommentSty{algcommentfont}
\SetKwFunction{VisibilityCheck}{VisibilityCheck}
\SetKwInOut{Input}{Input}
\SetKwInOut{Output}{Output}
\SetNlSty{}{}{}

\begin{algorithm}[tb!]
    \small
    \Input{
    Geometry scaffold $\mathcal{G}$,\\
    Cameras $\mathcal{C}\in\mathbb{R}^{N_c\times3\times4}$,\\
    Feature points $\mathcal{X}=\{\mathbf{x}_i|\mathbf{x}_i\in\mathbb{R}^3\}_{i=1}^{N_{fp}}$
    }
    \Output{Sampled cameras $\mathcal{S}=\{s_i|s_i\in\mathbb{R}^{3\times4}\}_{i=1}^{n}$}
    
    \textbf{Initialize:} \;
    $\mathcal{V} \gets \{v_i | v_i \in \mathbb{R}^{3 \times 4} \}_{i=1}^N$ (random candidate cameras)\;
    $\mathcal{S} \gets \mathcal{C}$ \;
    $\mathcal{A}_{vs} \in \mathbb{R}^{(N+N_c) \times (N+N_c)} \gets 0$ \tcp*{Similarity matrix}
    
    \tcp{Build similarity matrix} 
    \ForEach{$\mathbf{x}_i \in \mathcal{X}$}{
        $i_{\text{visible}} \gets 0 \in \mathbb{R}^{(N + N_c) \times 1}$\;
        \ForEach{$(j, c)\in \textbf{enumerate}(\mathcal{V} \cup \mathcal{S})$}{
            \If{\VisibilityCheck{$\mathbf{x}_i$, $c$}}{
                $i_{\text{visible}}[j] \gets 1$\;
            }
        }
        \ForEach{$(i_p, i_q) \in \text{Pairs}(i_{\text{visible}})$}{
            $\mathcal{A}_{vs}[i_p, i_q] \gets \mathcal{A}_{vs}[i_p, i_q] + 1$\;
        }    
    }     
    \tcp{Sample virtual viewpoints} 
    \While{$|\mathcal{S}| < n$}{
        $c_v, c_s \gets$ positions of $v, s$ \;
        $i_v, i_s \gets$ indices of $v, s$\;
        $d_{\text{euc}} \gets \|c_v - c_s\|$ \;
        $d_{\text{feat}} \gets 1 - \mathcal{A}_{vs}[i_v, i_s]$\;
        $v^* \gets \arg\max_{v \in \mathcal{V} \setminus \mathcal{S}} \min_{s \in \mathcal{S}}(d_{\text{euc}} + \kappa d_{\text{feat}})$\;
        $\mathcal{S} \gets \mathcal{S} \cup \{v^*\}$\;
    }
    
    \Return $\mathcal{S}$
    \caption{Novel viewpoint sampling}
    \label{alg:viewpoint_sampling} 
\end{algorithm}

\begin{table}[t!]
\centering
\caption{\textbf{Effect of $\kappa$ in Equation 9 of the main manuscript.} ScanNet++ \textsc{E91722b5a3} scene is used for this experiment.}
\scriptsize
\setlength{\tabcolsep}{7pt}
\resizebox{0.9\linewidth}{!}{
    \begin{tabular}{@{}l|c|c|c|c|c}
        \toprule
        $\kappa$ & 0.01 & 0.05 & 0.1 & 0.25 & 0.5 \\
        \midrule
        PSNR & 21.32 & 21.38 & \textbf{21.41} & 21.35 & 21.29 \\
        \bottomrule
    \end{tabular}
}
\label{suppl_table:parameter_sweeping}
\end{table}

\begin{table}[t]
\centering
\caption{\textbf{Additional comparisons with next-best-view strategies.} We compare coverage maximization, uncertainty minimization, and our method on ScanNet++ and Zip-NeRF datasets.}
\resizebox{\linewidth}{!}{
    \begin{tabular}{@{}l|ccc|ccc@{}}
        \toprule
         & \multicolumn{3}{c|}{ScanNet++} & \multicolumn{3}{c}{Zip-NeRF} \\
         & PSNR↑ & SSIM↑ & LPIPS↓ & PSNR↑ & SSIM↑ & LPIPS↓ \\
        \midrule
        Maximizing Coverage & 22.31 & 0.810 & 0.343 & 23.53 & 0.750 & 0.422 \\
        Minimizing Uncertainty & 22.25 & 0.806 & 0.345 & 23.43 & 0.744 & 0.434 \\
        Ours & \textbf{22.61} & \textbf{0.815} & \textbf{0.326} & \textbf{23.96} & \textbf{0.768} & \textbf{0.393} \\
        \bottomrule
    \end{tabular}
}
\label{suppl_table:nbv_comparison}
\end{table}
% \vspace{-2em}
\subsection{Comparison with Next-Best-View Algorithms}
To further validate the efficacy of our viewpoint selection, we compared our method against representative next-best-view planning strategies.
(1) Visibility coverage maximization~\cite{progressive_camera_placement}: We select viewpoints that maximize the spatial coverage of visible regions based on accumulated visibility scores.
(2) Uncertainty-based view selection~\cite{FisherRF, activenerf}:
We use depth errors as a proxy to estimate uncertainty from the current reconstruction, following prior work in active NeRF view selection (e.g., ActiveNeRF~\cite{activenerf}).
For both methods, we adopt a greedy selection strategy from a set of randomly sampled candidate viewpoints.
At each step, the next viewpoint is selected according to the corresponding objective until the desired number of virtual views is reached.
All other components, including basis placement and training settings, are kept identical for fair comparison.
As shown in Table~\ref{suppl_table:nbv_comparison}, our geometry-aware viewpoint selection achieves more stable reconstruction and better rendering quality.
This is because our method explicitly accounts for both scene geometry and observation statistics, leading to more effective allocation of representation capacity.

\section{Scalability Analysis}
\subsection{Outdoor Experiments}
\label{subsec:outdoor_experiments}
\begin{figure}[t]
    \centering
    \includegraphics[width=\linewidth]{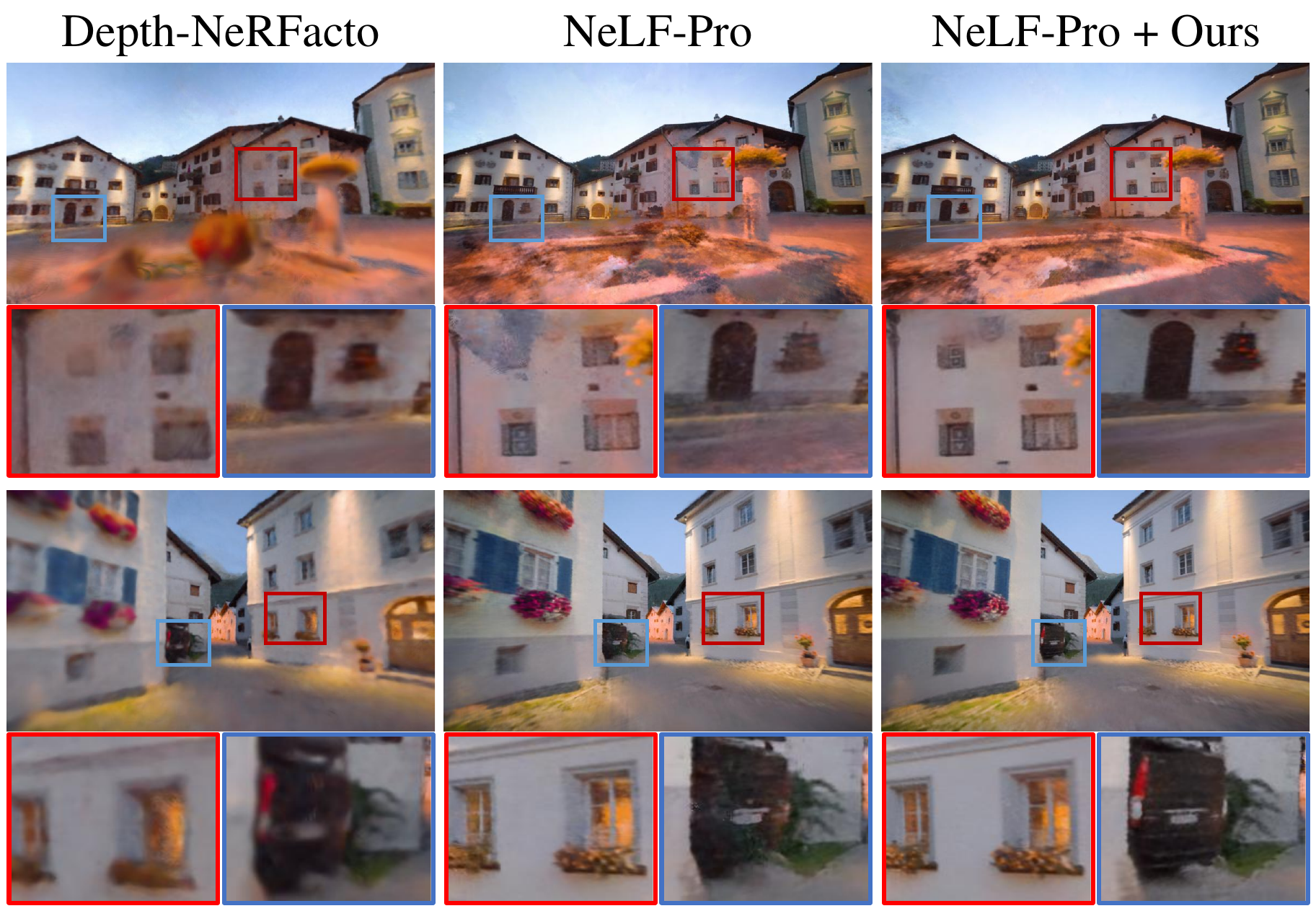}
    \caption{\textbf{Qualitative comparisons in Scuol~\cite{nelf-pro} dataset.} Within the valid geometric scaffold region, our method successfully reconstructs the scene while mitigating artifacts in sparsely observed areas.
    }
    \label{suppl_fig:outdoor_comparison}
\end{figure}

While the proposed geometric adaptation demonstrates superior performance in complex indoor layouts, the same principle can also be applied to outdoor environments.
We demonstrate the results of the outdoor experiments using Scuol~\cite{nelf-pro} and Tanks and Temples~\cite{tanks} dataset (\textsc{Barn} scene) in Figure~\ref{suppl_fig:outdoor_comparison} and Table~\ref{suppl_table:outdoor_comparison}. 
Obtaining geometric scaffold using MonoSDF~\cite{monosdf} or MapAnything~\cite{MAPAnything} struggles with sky and background regions in outdoor environments, which could degrade the optimization of the basis and geometric regularization. 
To test our method on outdoor scenarios, we manually extracted the bounding box of the valid mesh and optimized the basis placement.
While we demonstrate improved results on selected outdoor scenes, our approach faces limitations when the underlying mesh quality is severely degraded.
For example, in datasets such as KITTI-360~\cite{kitti-360}, the geometric scaffolds struggle to faithfully capture under-constrained outdoor environments, which limits their effectiveness in guiding reconstruction.
\begin{table}[t]
\centering
\caption{\textbf{Quantitative evaluation on outdoor datasets.} Applying our framework to outdoor scenes improves reconstruction quality compared to baselines.}
\resizebox{\linewidth}{!}{
    \begin{tabular}{@{}l|ccc|ccc@{}}
        \toprule
         & \multicolumn{3}{c|}{Tanks (\textsc{Barn})} & \multicolumn{3}{c}{Scuol} \\
         & PSNR↑ & SSIM↑ & LPIPS↓ & PSNR↑ & SSIM↑ & LPIPS↓ \\
        \midrule
        Depth-NeRFacto~\cite{nerfstudio} & 17.04 & 0.634 & 0.571 & 18.21 & 0.676 & 0.471 \\
        NeLF-Pro~\cite{nelf-pro} & 21.12 & 0.652 & 0.556 & 25.44 & 0.779 & 0.313 \\
        NeLF-Pro~\cite{nelf-pro}+Ours & 21.88 & 0.654 & 0.523 & \textbf{25.73} & \textbf{0.781} & \textbf{0.236} \\
        LocalRF~\cite{localrf} & 21.33 & 0.706 & 0.566 & 22.54 & 0.776 & 0.418\\
        LocalRF~\cite{localrf}+Ours & \textbf{22.20} & \textbf{0.711} & \textbf{0.511} & 23.01 & 0.780 & 0.359\\
        \bottomrule
    \end{tabular}
}
\label{suppl_table:outdoor_comparison}
\end{table}
\begin{table}[t!]
\centering
\caption{\textbf{Evaluation for different scales with baselines.} We categorized the results in the main manuscript according to the scene scale.}
\resizebox{\linewidth}{!}{
    \begin{tabular}{@{}l|c|c|c|c}
        \toprule
        Scene scale & Single-Room & Multi-Rooms & Outdoor & Town-scale\\
        \midrule
        NeLF-Pro & 22.58 & 22.44 & 21.12 & 25.44 \\
        NeLF-Pro+Ours & \textbf{23.06} & \textbf{23.08} & 21.88 & \textbf{25.73} \\
        LocalRF & Non-sequential & 21.39 & 21.33 & 22.54 \\
        LocalRF+Ours & Non-sequential & 23.02 & \textbf{22.20} & 23.01 \\
        \bottomrule
    \end{tabular}
}
\label{suppl_table:scale_comparison}
\end{table}
\subsection{Evaluation for Different Scales}
We demonstrate scalability analysis with additional results in Table~\ref{suppl_table:scale_comparison}, where scenes are categorized as single-room, multi-room, outdoor, and town scale.
For outdoor and town scale scene, we reported PSNRs for the \textsc{BARN} scene and Scoul dataset, which were demonstrated in Figure~\ref{suppl_fig:outdoor_comparison} and Table~\ref{suppl_table:outdoor_comparison}.
Since the original LocalRF~\cite{localrf} assumes sequentially captured images, we report results on all scenes except the single-room cases from the ScanNet++~\cite{scannetpp} dataset.
The scalability analysis shows that our method can consistently handle the scene as scale increases.

\section{Results with Different Number of Basis}
\label{sec:different_basis_number}
\begin{figure}[t]
    \centering
    \includegraphics[trim={0, 2mm, 0, 0}, clip, width=\linewidth]{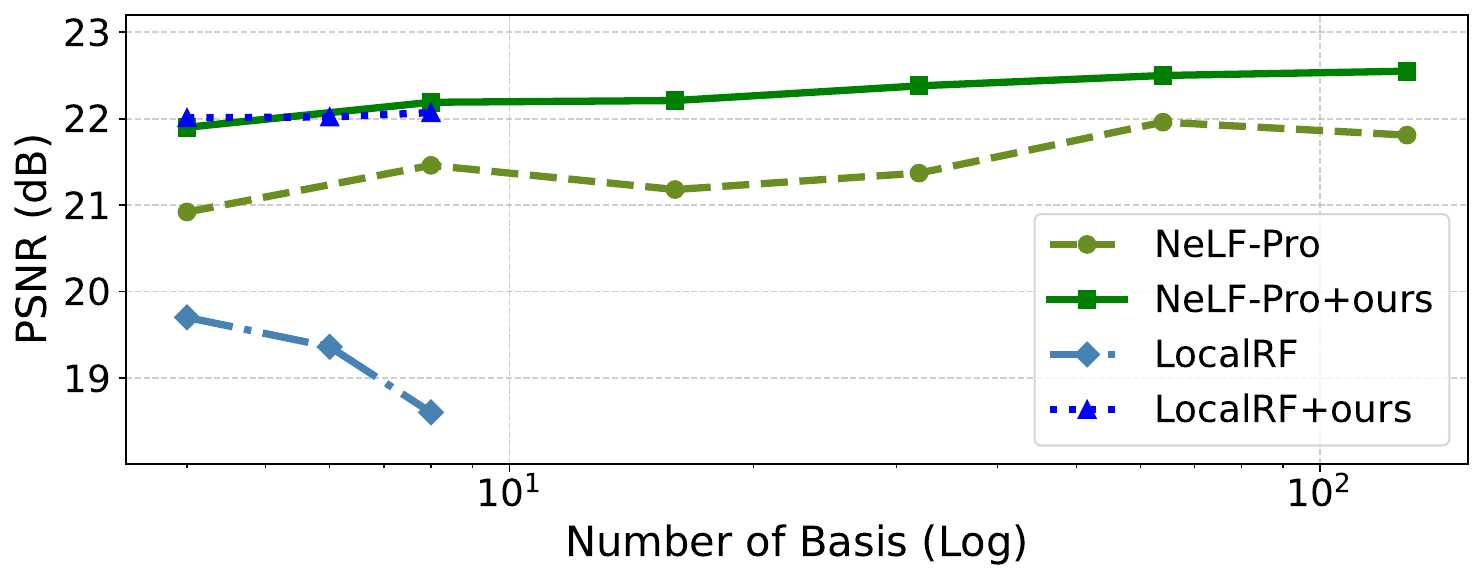}
    \caption{
    \textbf{PSNR comparison on a different number of bases.} Our scene-adaptive strategy consistently improves the
performance of NeLF-Pro~\cite{nelf-pro} and LocalRF~\cite{localrf} across all tested basis numbers
    }
    \label{suppl_fig:performance_comparison}
\end{figure}
\begin{figure}[tb!]
    \centering
    \includegraphics[width=\linewidth]{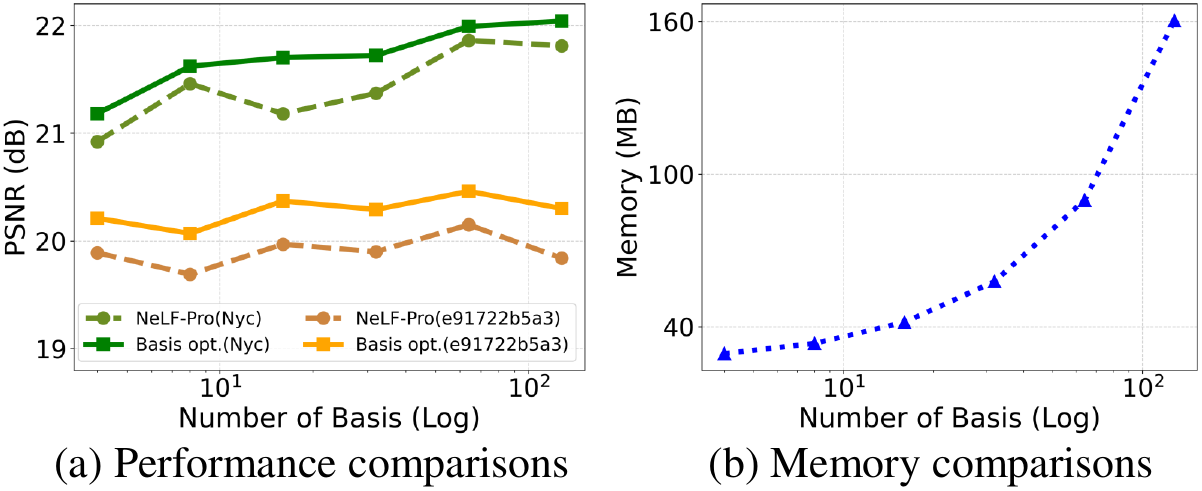}
    \caption{\textbf{Experiments for different number of bases.} (a) Performance comparisons and (b) memory comparisons are plotted.
    }
    \label{suppl_fig:differnt_number_of_basis}
\end{figure}
We analyze the performance of our framework with varying numbers of allocated bases.
For both NeLF-Pro and LocalRF variants, we tested on Zip-NeRF's \textsc{NYC} dataset.
NeLF-Pro was tested with varying basis configurations: (4,2), (8,4), (16,4), (32,8), (64,16), and (128,16), where the first value indicates the total number of bases and the second denotes the number of nearby bases. 
The number of core probes is fixed at 3.
To adjust the number of NeRF blocks in the original LocalRF, we set different values for the maximum allowable drift before adding a new block.
We examined PSNRs of the test views with the changing number of allocated bases in Figure~\ref{suppl_fig:performance_comparison}.
Our framework robustly outperforms the baselines in every basis number.

In LocalRF, NeRF blocks are added progressively without considering the layout of the input trajectory or scene geometry, and may suffer from scarce training frames within individual blocks, especially when the trajectory returns to previously visited areas.
In contrast, our framework determines basis positions and assigns a distance threshold around each basis, allowing nearby frames to be modeled together and thus avoiding this issue.
To isolate the impact of the basis placement, we further evaluated PSNR performance with and without the basis placement under a varying number of neural lightfield probes.
% Different from the experiments conducted in Figure 10 of the main manuscript, we removed the effects of our geometric regularization strategy. 
Different from the experiments conducted in Figure~\ref{suppl_fig:performance_comparison}, we removed the effects of our geometric regularization strategy. 
We set the number of bases within the range of 4 to 128, identical to those used in the previous experiment.
For evaluation, we selected \textsc{e91722b5a3} scene from ScanNet++~\cite{scannetpp} and \textsc{NYC} scene from Zip-NeRF~\cite{zip-nerf} dataset.
Additionally, we report the required memory for different numbers of bases to show the trade-off between performance and computational cost.
As shown in Figure~\ref{suppl_fig:differnt_number_of_basis}(a), our basis placement strategy consistently improves the performance of NeLF-Pro~\cite{nelf-pro} across all tested basis numbers.
While our framework uses the same memory as the NeLF-Pro, the memory consumption scales with the number of bases.
These results confirm that the optimal basis placement enhances reconstruction quality for different basis settings while remaining the number of parameter and memory usage unchanged.
\begin{table}[!t]
\centering
\caption{\textbf{Baseline configurations.} The table is separated based on the number of bases, where methods in lower rows leverage multiple bases. For cases where depth regularization is not used, we filled those cells with \xmark. Similarly, for single-basis representations, we filled those cells with \xmark.}
\resizebox{\linewidth}{!}{
    \begin{tabular}{l|c|c|c}
        \toprule
        Method & Iterations & Depth input & Basis type \\
        \midrule
        MonoSDF~\cite{monosdf} & 200K & Omnidata~\cite{omnidata} & \xmark \\
        NeRFacto~\cite{nerfstudio} & 30K & \xmark & \xmark \\
        NeRFacto~\cite{nerfstudio}-Big & 60K & \xmark & \xmark \\
        NeRFacto~\cite{nerfstudio}-Huge & 60K & \xmark & \xmark \\
        Depth-NeRFacto~\cite{nerfstudio} & 30K & Geo. scaffold & \xmark \\
        Zip-NeRF~\cite{zip-nerf} & 30K & \xmark & \xmark \\
        3DGS~\cite{3dgs} & 30K & \xmark & \xmark \\
        Mip-Splatting~\cite{Mip-Splatting} &  30K &  \xmark &  \xmark \\
        Scaffold-GS~\cite{Scaffold-GS} &  30K &  \xmark &  \xmark \\
        PlanarGS~\cite{PlanarGS} &  30K &  DUSt3R~\cite{dust3r} &  \xmark \\
        FSGS~\cite{FSGS} & 30K & MiDaS~\cite{midas} & \xmark \\
        DNGaussian~\cite{dngaussian} & 30K & DPT~\cite{midas} & \xmark \\
        \midrule
        Mega-NeRF~\cite{mega-nerf} & 200K / basis & \xmark & NeRF block \\
        Hierarchical-3DGS~\cite{hierarchical_3dgs} & 30K & DPT~\cite{dpt} & Chunk \\
        LongSplat~\cite{LongSplat} &  600 / image &  MASt3R~\cite{mast3r} &  Octree Anchor \\
        NeLF-Pro~\cite{nelf-pro} & 30K & \xmark & LF probe \\
        NeLF-Pro~\cite{nelf-pro} + Ours & 30K & Geo. scaffold & LF probe \\
        LocalRF~\cite{localrf} & 1K / image & DPT~\cite{dpt} & NeRF block \\
        LocalRF~\cite{localrf} + Ours & 1K / image & Geo. scaffold & NeRF block \\
    \bottomrule
    \end{tabular}
}
\label{suppl_table:baseline_configurations}
\end{table}

\vspace{-1.5em}
\section{Baseline Details}
In the main manuscript, we compare our method with recent approaches for 3D scene modeling and novel view synthesis, including both NeRF-based and 3DGS variants as well as methods incorporating geometric regularization, largely following official training settings with minor adjustments for NeLF-Pro and LocalRF. While Section 4.1 provides a brief overview, we summarize detailed training setups—including iterations, depth supervision, and basis types—in Table~\ref{suppl_table:baseline_configurations}.

We organize the baselines into two categories based on scalability. 
First, single-scene representations, which model a scene with a unified representation:
(1) MonoSDF~\cite{monosdf} is a neural implicit surface reconstruction model trained using RGB, depth, and normal.
(2) NeRFacto~\cite{nerfstudio} combines hash encoding from Instant-NGP~\cite{instant-ngp} and proposal sampling from MipNeRF-360~\cite{mip-nerf360}. For the Zip-NeRF dataset, NeRFacto-big and NeRFacto-huge variants with larger capacity were also tested.
(3) Depth-NeRFacto extends NeRFacto with depth supervision using SparseNeRF~\cite{sparsenerf}, which showed the best performance among available options.
(4) Zip-NeRF~\cite{zip-nerf} combines anti-aliasing on multi-scale hashed feature grids to enhance training speed while preserving high-frequency details.
(5) 3DGS~\cite{3dgs} is a state-of-the-art explicit scene representation with Gaussians initialized from the same feature points used in Section 3.4 of the main manuscript.
(6) FSGS~\cite{FSGS} addresses the sparse view problem via proximity-guided densification and geometric regularization, and (7) DNGaussian~\cite{dngaussian} similarly handles few-shot settings through hard and soft depth regularization.
(8) Mip-Splatting~\cite{Mip-Splatting} extends 3DGS with anti-aliasing to reduce artifacts, (9) Scaffold-GS~\cite{Scaffold-GS} introduces structured Gaussian primitives for view-adaptive rendering, and (10) PlanarGS~\cite{PlanarGS} leverages planar priors for geometric consistency in structured indoor environments.

We also compare against scalable representations employing multiple spatial units.
(11) NeLF-Pro~\cite{nelf-pro} is applied to both datasets with tuned probe numbers and positions.
(12) LocalRF~\cite{localrf} uses multiple NeRF blocks but requires sequential input, limiting evaluation to Zip-NeRF datasets.
(13) Mega-NeRF~\cite{mega-nerf} uses a regular $4\times2$ grid of NeRF blocks, evaluated only on Zip-NeRF as ScanNet++ room-scale scenes show minimal benefit.
(14) Hierarchical-3DGS~\cite{hierarchical_3dgs} extends 3DGS to large-scale scenes with depth regularization, tested on both datasets.
(15) LongSplat~\cite{LongSplat} extends 3DGS to long video sequences by jointly optimizing Gaussian primitives and camera poses over continuous trajectories.

\section{Additional Comparisons}
\subsubsection{Measurement of Standard Deviations}
\begin{figure}[!t]
    \centering
    \includegraphics[width=\linewidth]{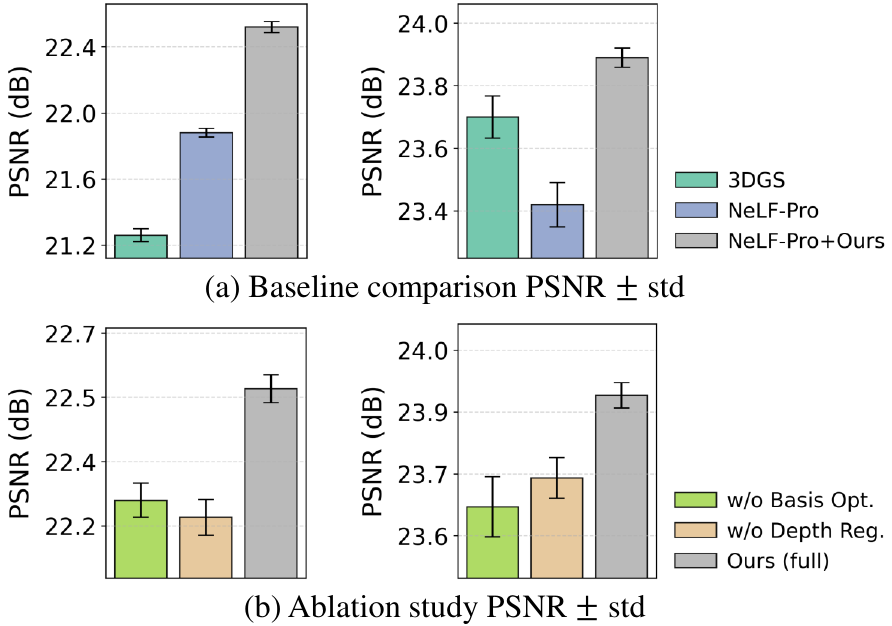}
    \caption{\textbf{Error bars for our experiments.} We compared (a) baselines and (b) ablation studies on ScanNet++ (left) and Zip-NeRF (right) dataset.}
    % \vspace{-1em}
    \label{suppl_fig:error_bar}
\end{figure}

To verify that our framework outperforms baselines, we have measured standard deviation using 5 randomly initialized seeds.
Among the baselines, we visualize the error bars of NeLF-Pro and 3DGS in Figure~\ref{suppl_fig:error_bar}. 
In addition, we have included a comparison to ablation studies, which shows the robustness of each component on performance enhancement.

\subsubsection{Effects of Each Component}
\begin{figure*}[p]
    \centering
    \includegraphics[width=\linewidth]{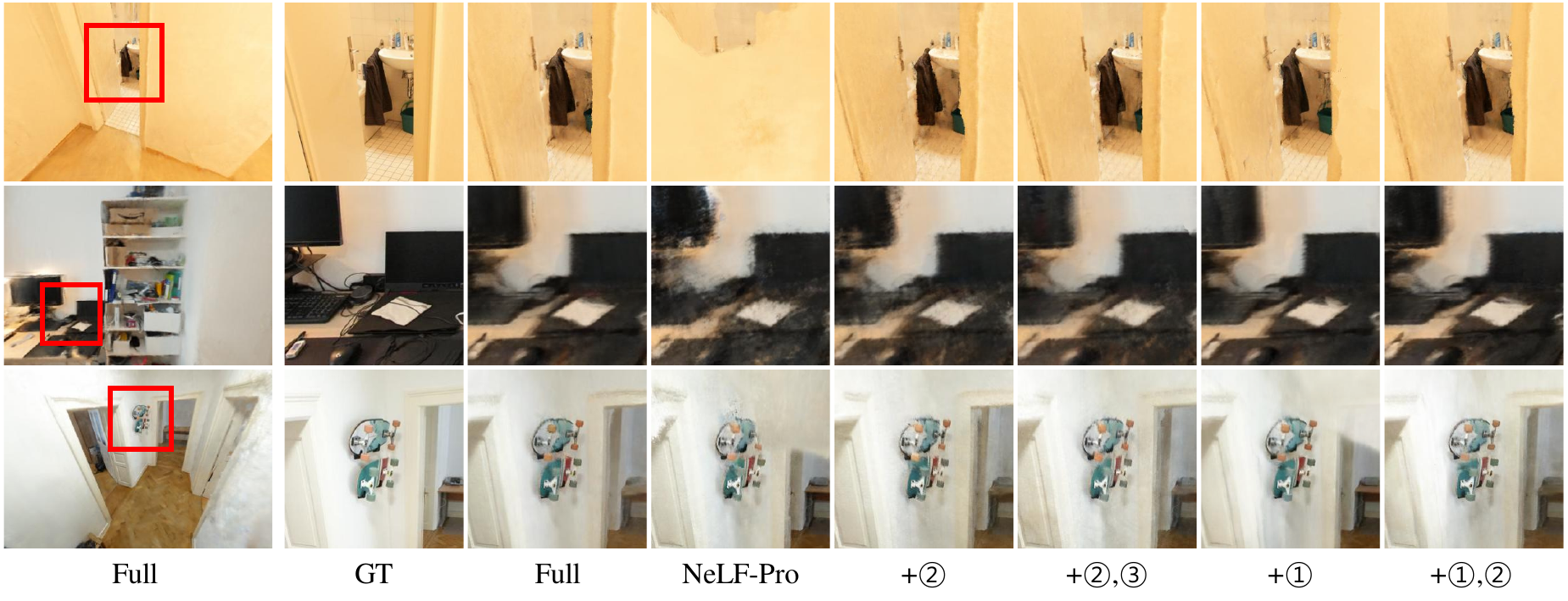}
    \caption{\textbf{Qualitative results for ablation study on ScanNet++~\cite{scannetpp} dataset.}}
    \label{suppl_fig:ablation_scannetpp}
    \vspace{1.5em}
    \includegraphics[width=\linewidth]{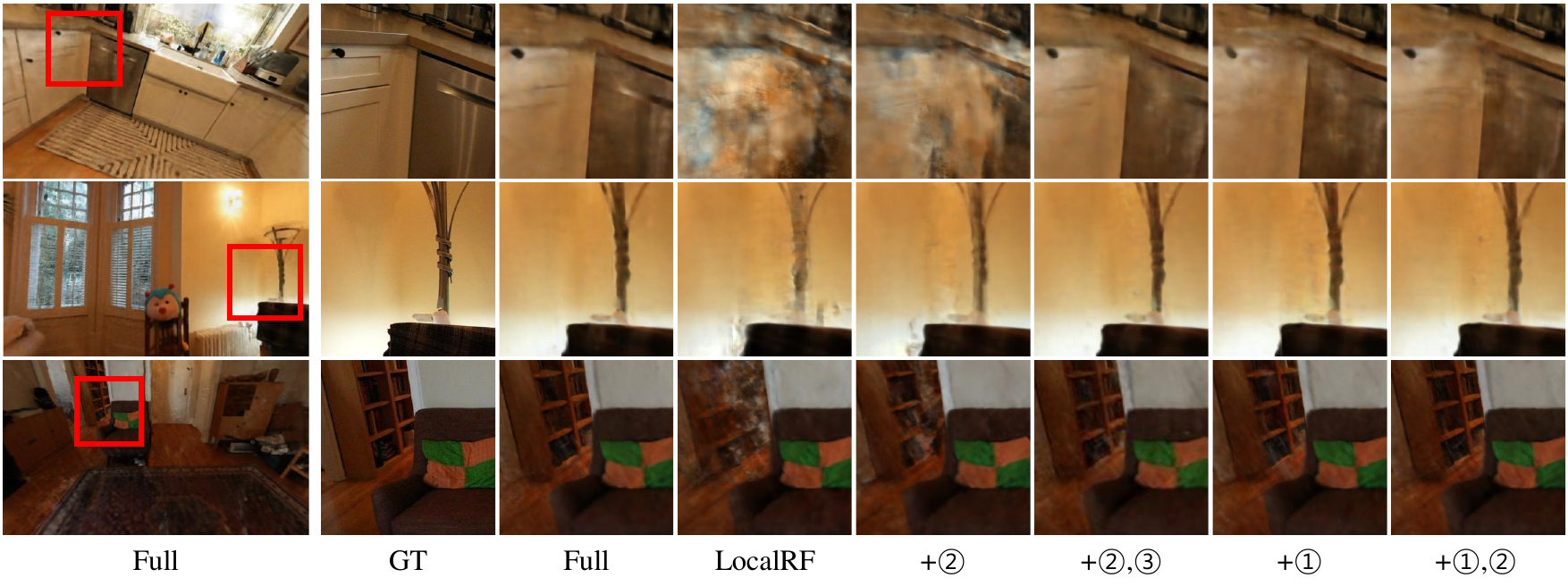}
    \caption{\textbf{Qualitative results for ablation study on Zip-NeRF~\cite{zip-nerf} dataset.}}
    \label{suppl_fig:ablation_zipnerf}
\end{figure*}

To show the effect of the components of our framework, we demonstrate visualizations for the ablation studies in Figure~\ref{suppl_fig:ablation_scannetpp} and Figure~\ref{suppl_fig:ablation_zipnerf}. % in~\cref{table:ablation}. 
We tested NeLF-Pro variants on ScanNet++ dataset and LocalRF variants on Zip-NeRF dataset, respectively.
As illustrated in both figures, while optimal basis placement helps reduce blurry artifacts, our geometric regularization with virtual viewpoints further enhances stability of the reconstruction.
In both figures, we follow the notations that were used in the main paper; $\circled{1}$: basis optimization, $\circled{2}$: robust depth loss for training views, and $\circled{3}$: novel view depth regularization.

\subsubsection{Comparison to SOTA 3DGS Algorithms}
\begin{figure*}[p]
    \centering
    \includegraphics[width=\linewidth]{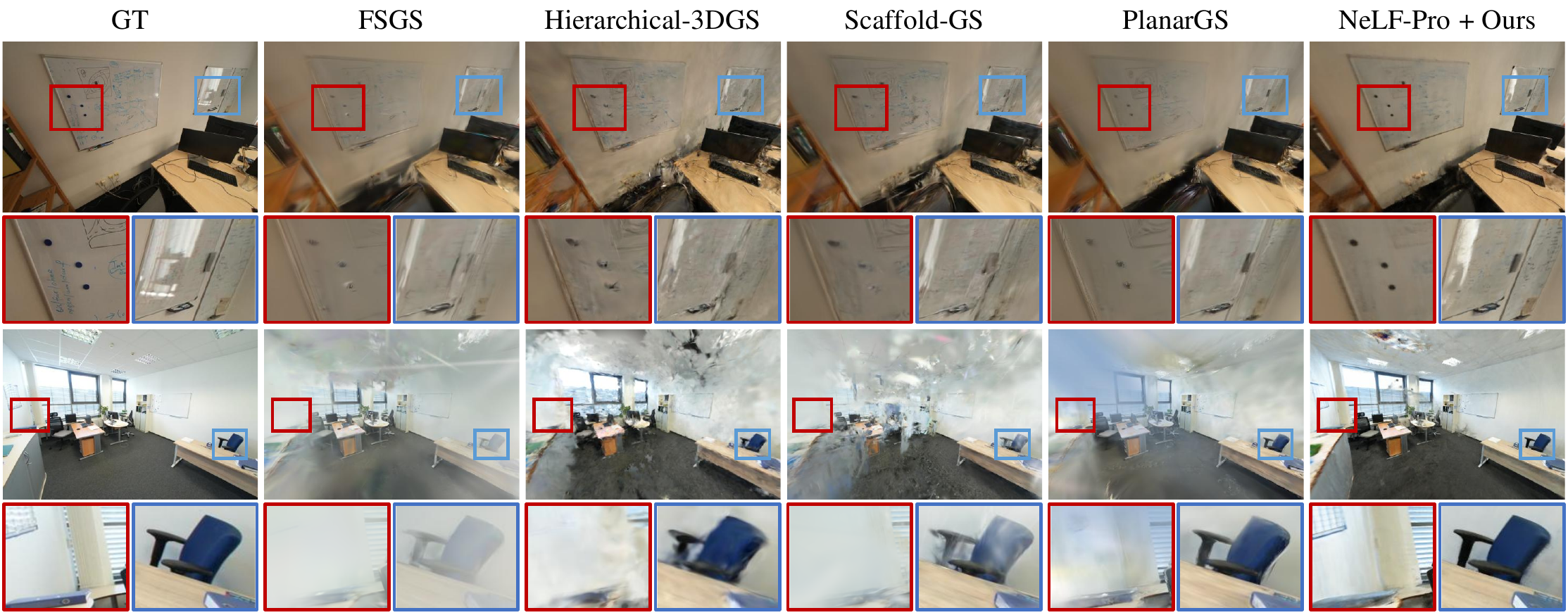}
    \caption{\textbf{Comparisons to SOTA 3DGS algorithms on ScanNet++ datasets.} We compared our frameworks using NeLF-Pro against 3DGS variants.
    }
    \label{suppl_fig:sota_3dgs_scannetpp}
    \vspace{1.5em}
    \includegraphics[width=\linewidth]{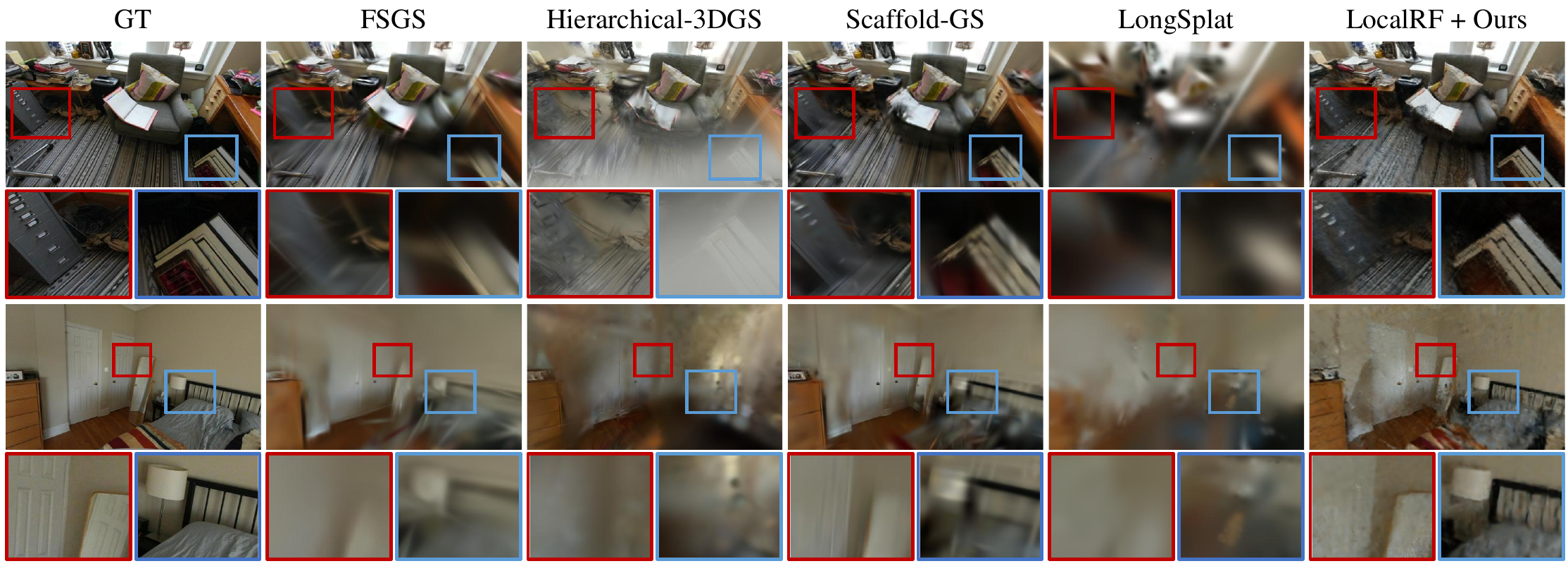}
    \caption{\textbf{Comparisons to SOTA 3DGS algorithms on Zip-NeRF datsets.} We compared our frameworks using LocalRF against 3DGS variants.
    }
    \label{suppl_fig:sota_3dgs_zipnerf}
\end{figure*}

We additionally conducted visual comparisons of our methods against SOTA 3DGS baselines.
We compared our methods to FSGS~\cite{FSGS} and Hierarchical-3DGS~\cite{hierarchical_3dgs} as in the main manuscript.
For NeLF-Pro~\cite{nelf-pro} and LocalRF~\cite{localrf} variants, we evaluated them on ScanNet++~\cite{scannetpp} and Zip-NeRF~\cite{zip-nerf} datasets, respectively.
As shown in visual comparisons in Figure~\ref{suppl_fig:sota_3dgs_scannetpp} and Figure~\ref{suppl_fig:sota_3dgs_zipnerf}, recent SOTA 3DGS algorithms still exhibit significant overfitting to training views and fail to handle our experimental settings properly.

\subsubsection{Qualitative Results}
\begin{figure*}[p]
    \centering
    \begin{tabular}{c@{\,}c@{\,}c@{\,}c@{\,}c@{\,}c}
    \small GT& \small NeRFacto & \small Depth-NeRFacto & \small 3DGS & \small NeLF-Pro & \small NeLF-Pro + Ours\\

    % 0d2ee665be
    \includegraphics[width=0.158\linewidth]{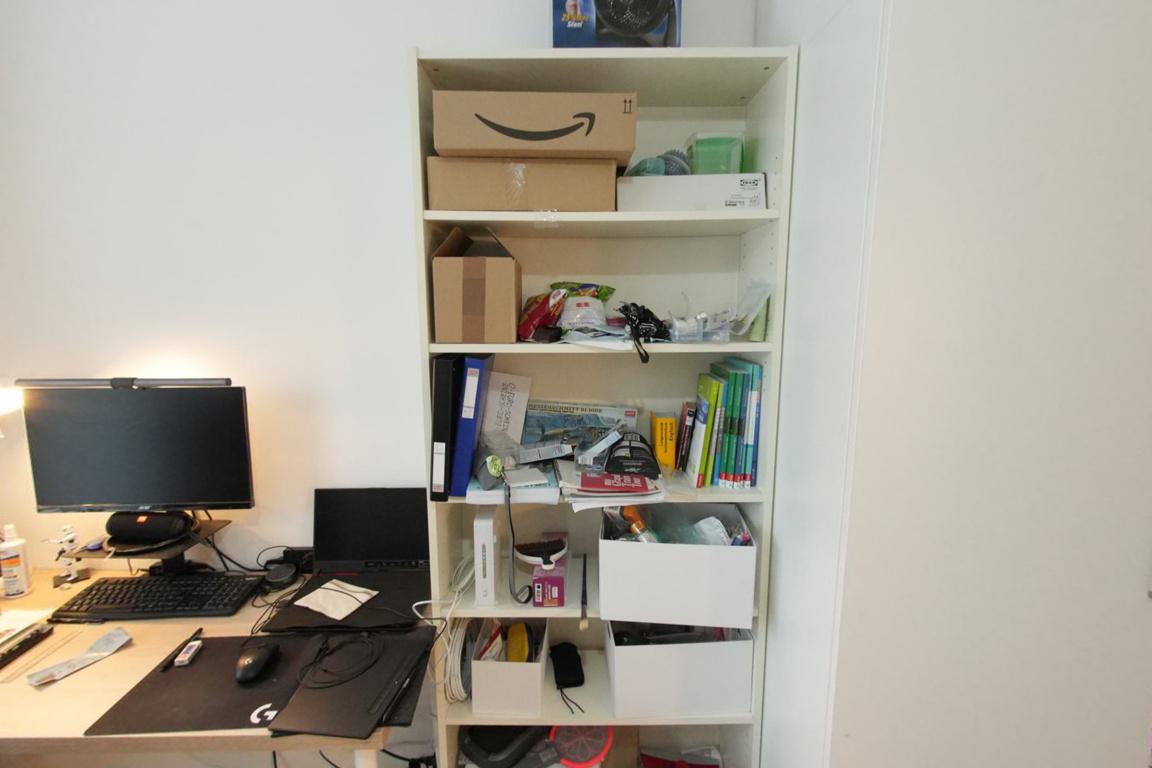}& 
    \includegraphics[width=0.158\linewidth]{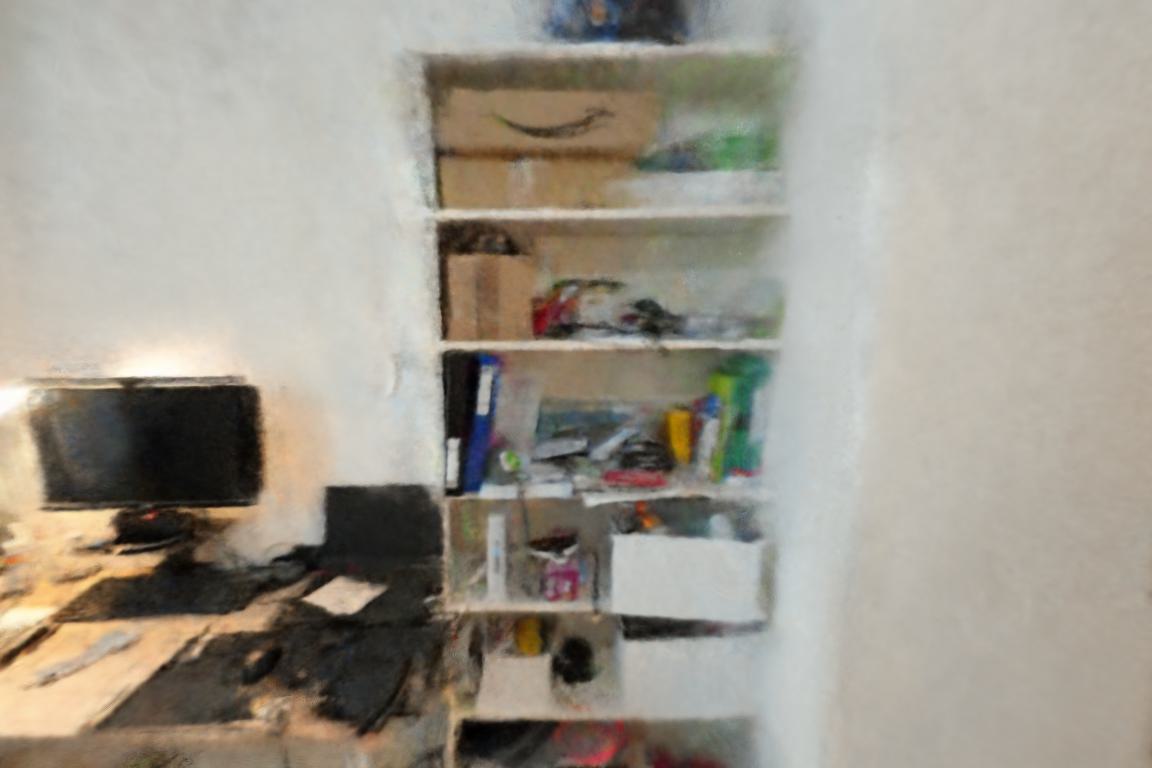}& \includegraphics[width=0.158\linewidth]{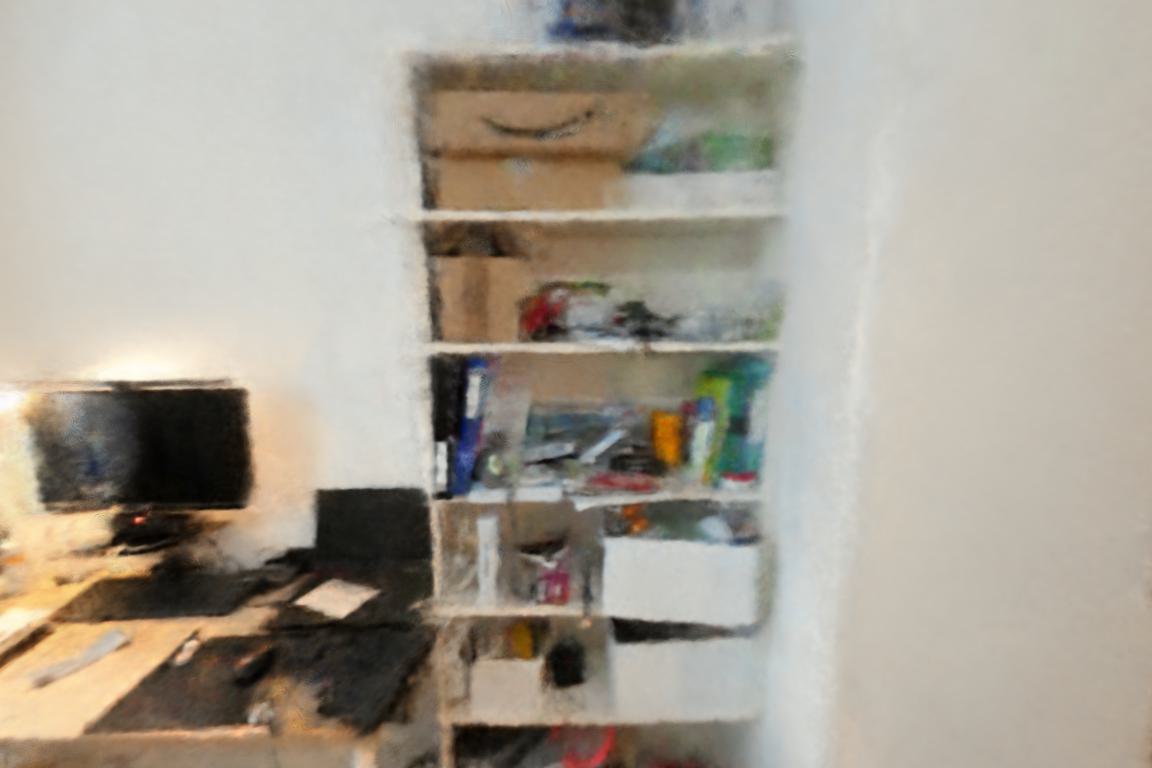}& \includegraphics[width=0.158\linewidth]{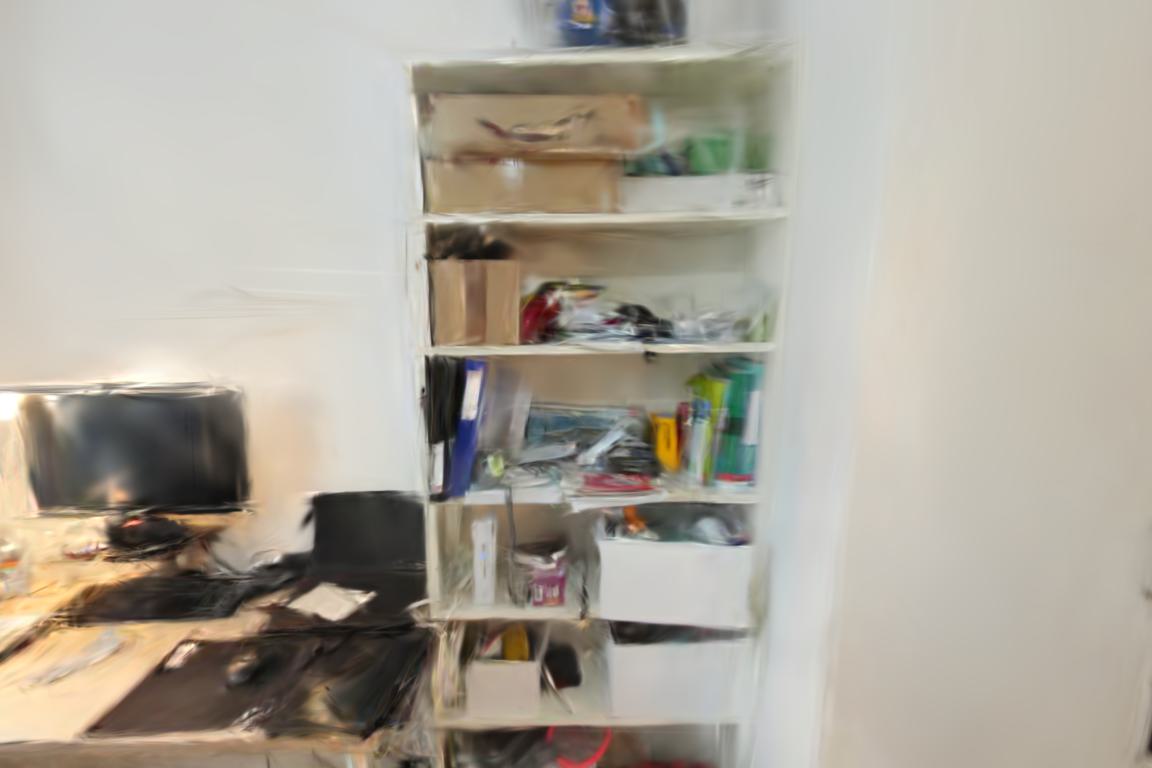}&
    \includegraphics[width=0.158\linewidth]{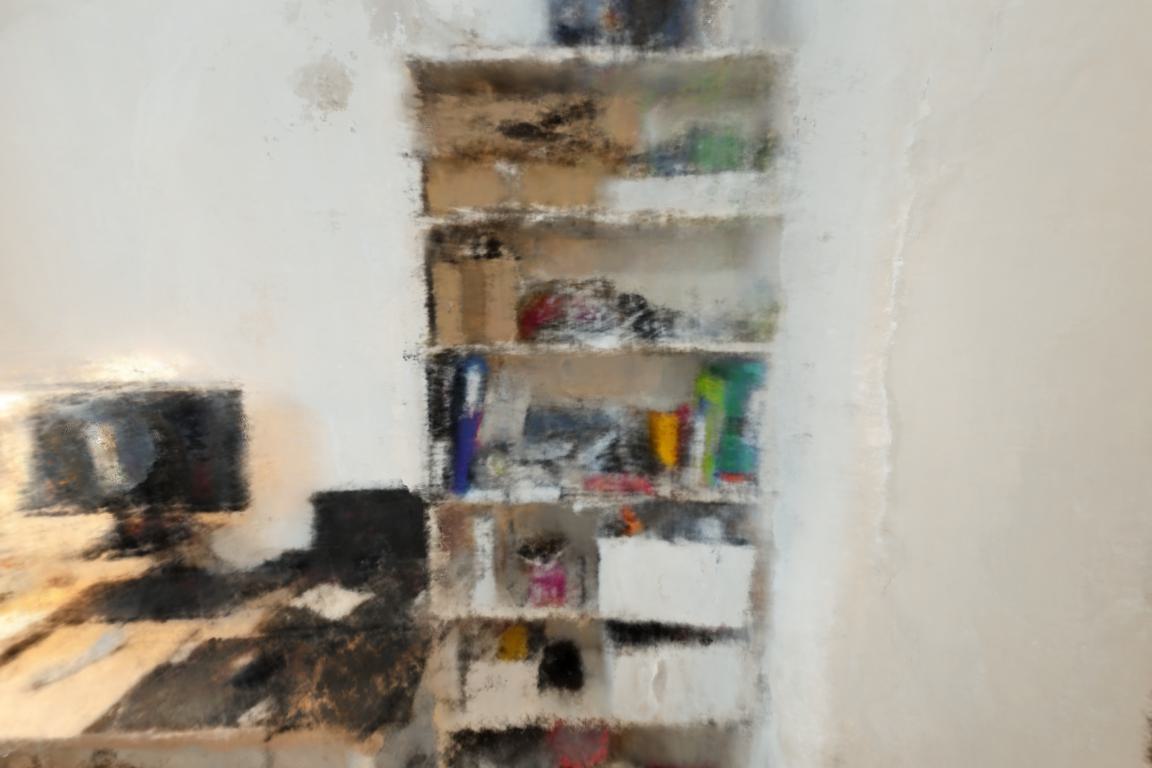}&
    \includegraphics[width=0.158\linewidth]{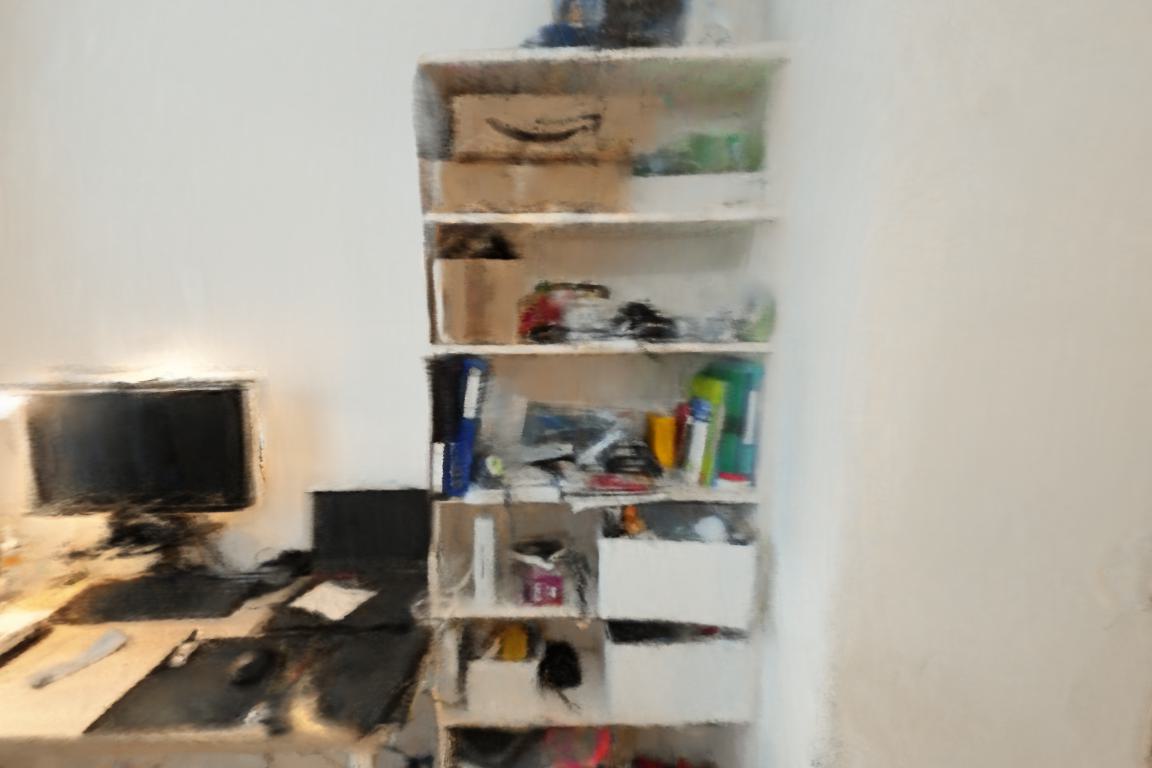}\\

    \includegraphics[width=0.158\linewidth]{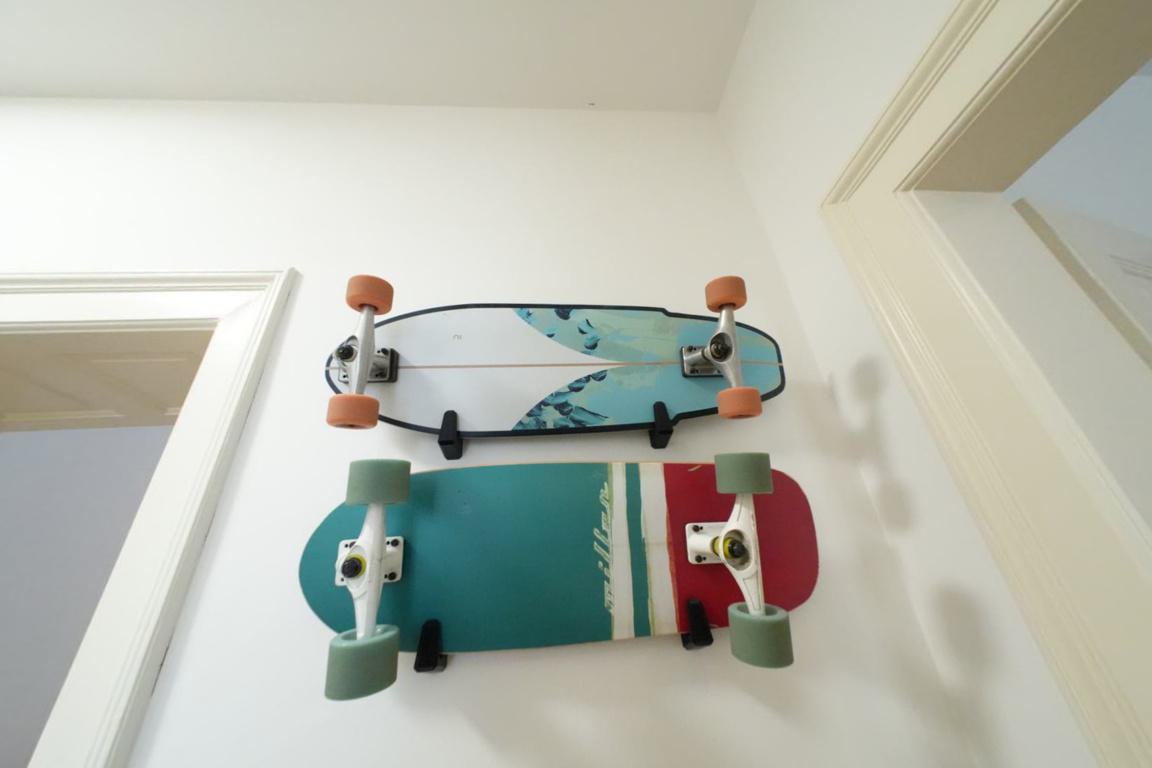}& 
    \includegraphics[width=0.158\linewidth]{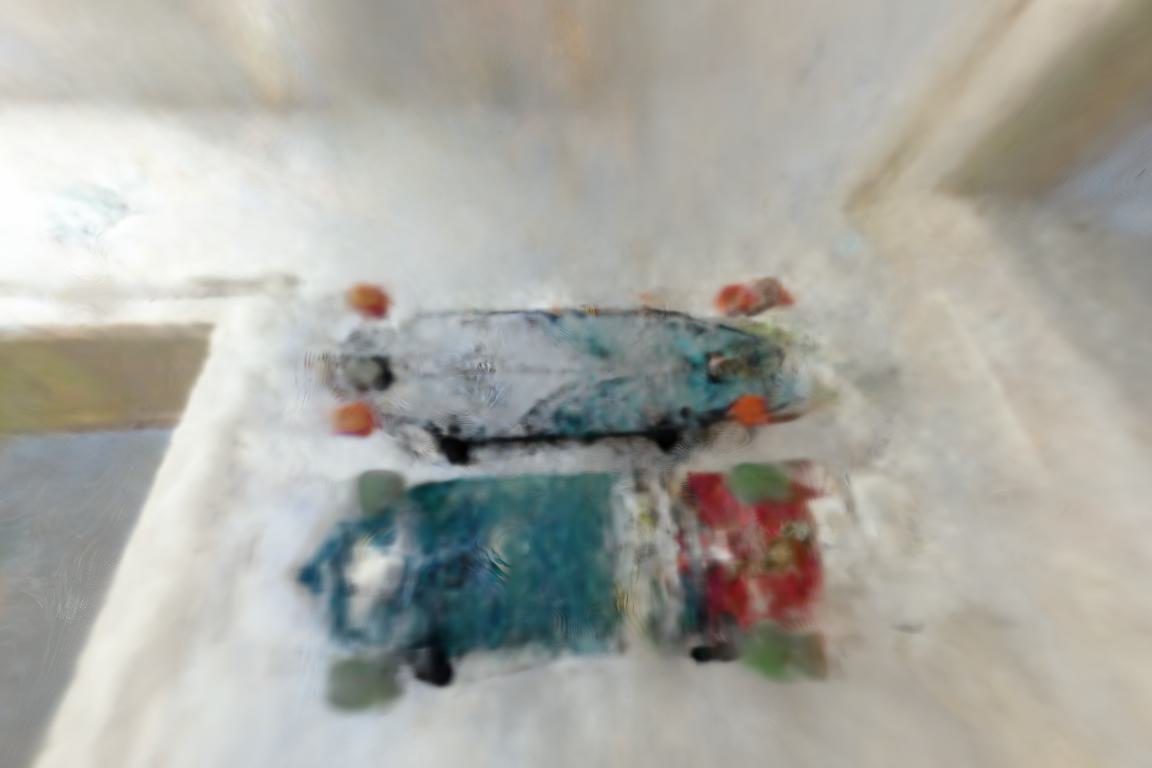}& \includegraphics[width=0.158\linewidth]{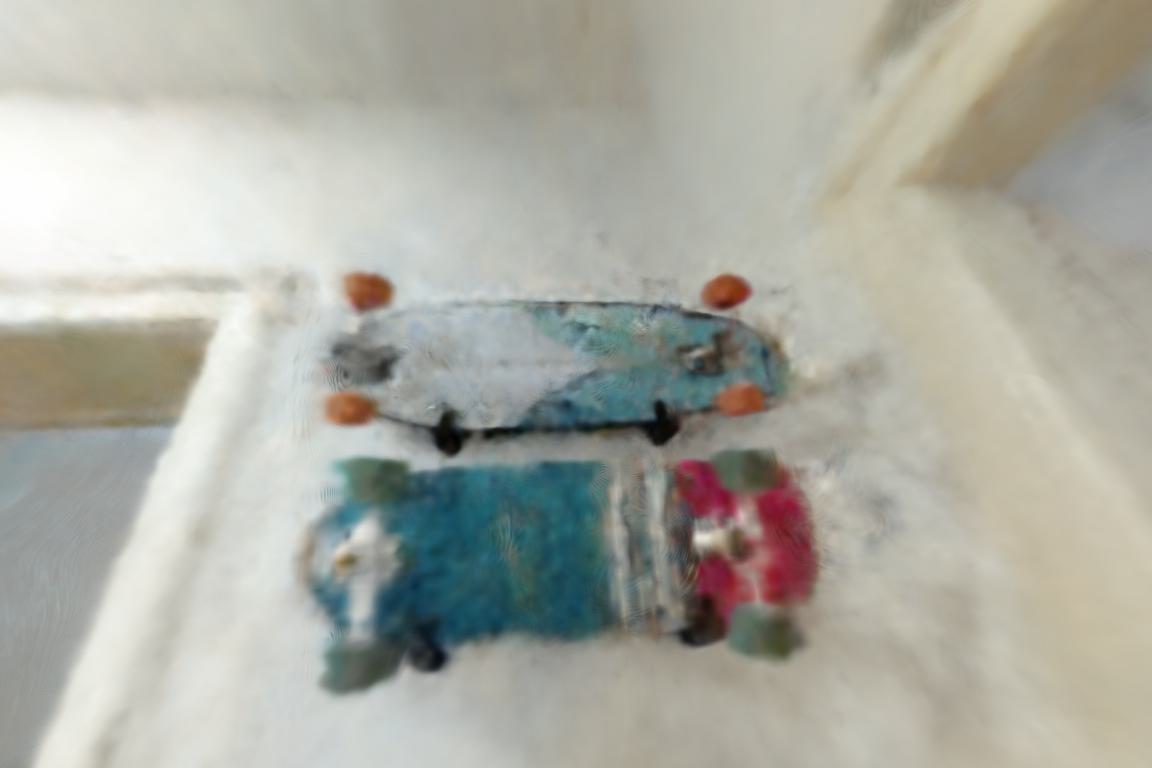}& \includegraphics[width=0.158\linewidth]{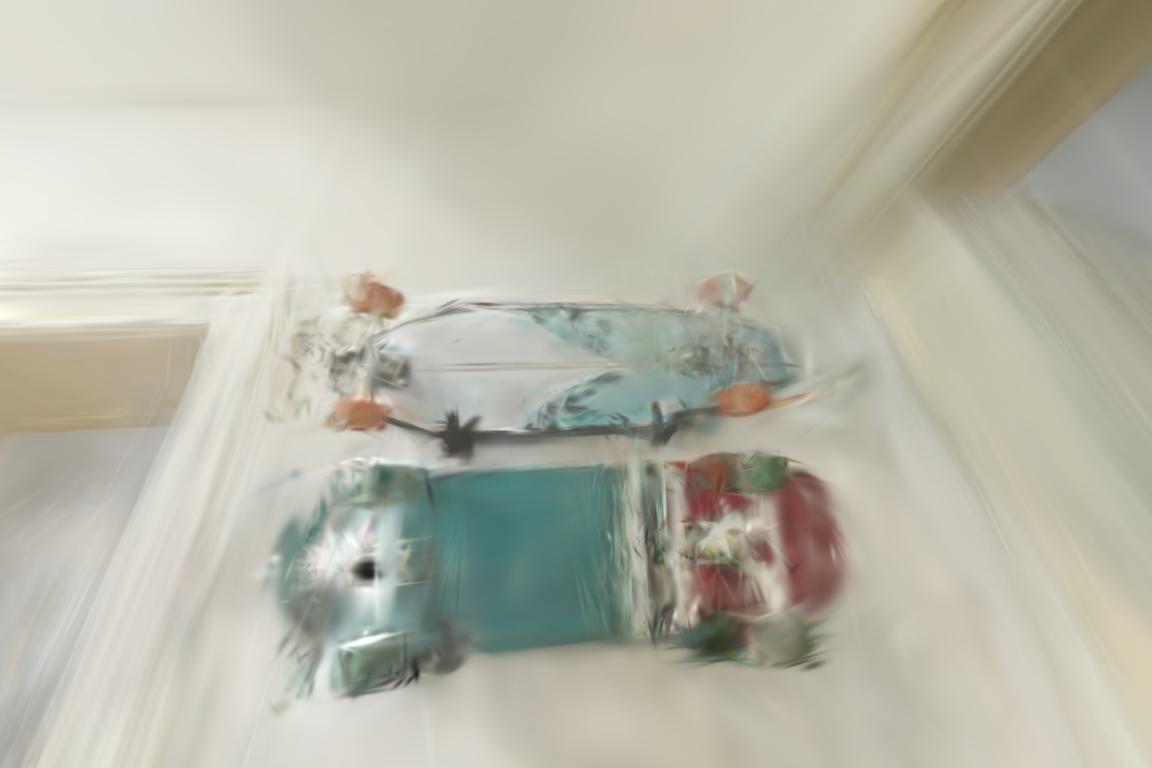}&
    \includegraphics[width=0.158\linewidth]{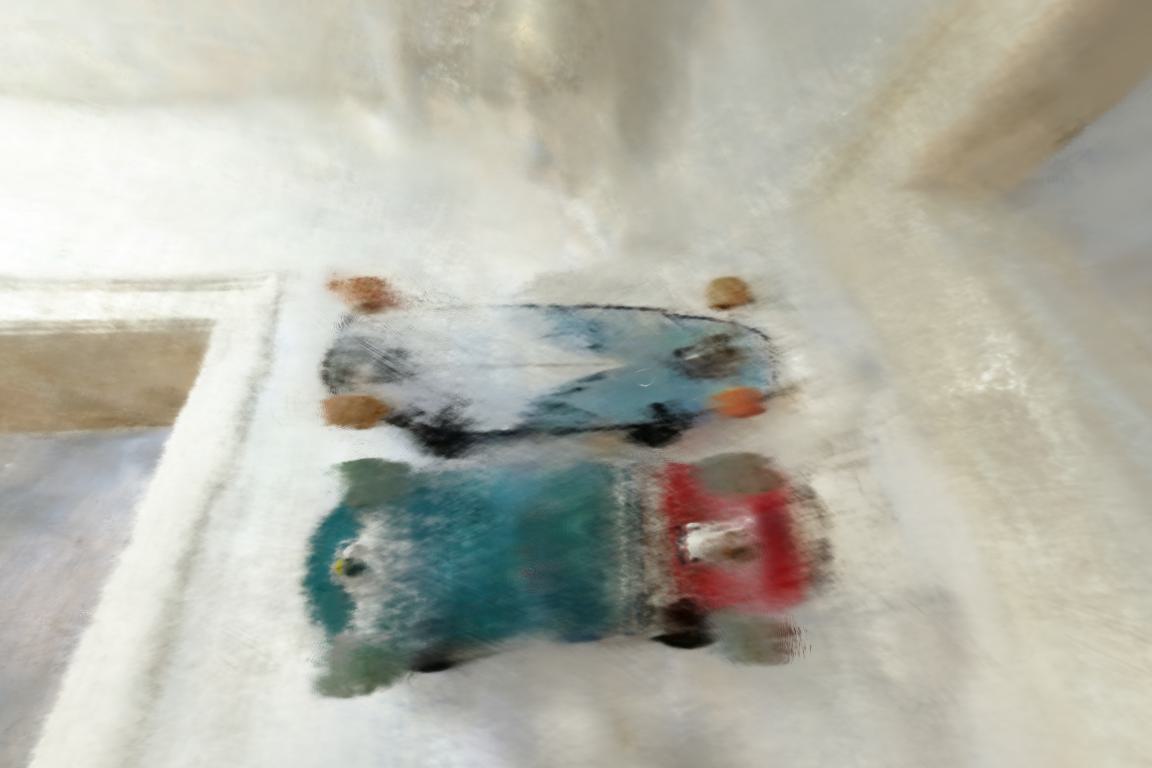}&
    \includegraphics[width=0.158\linewidth]{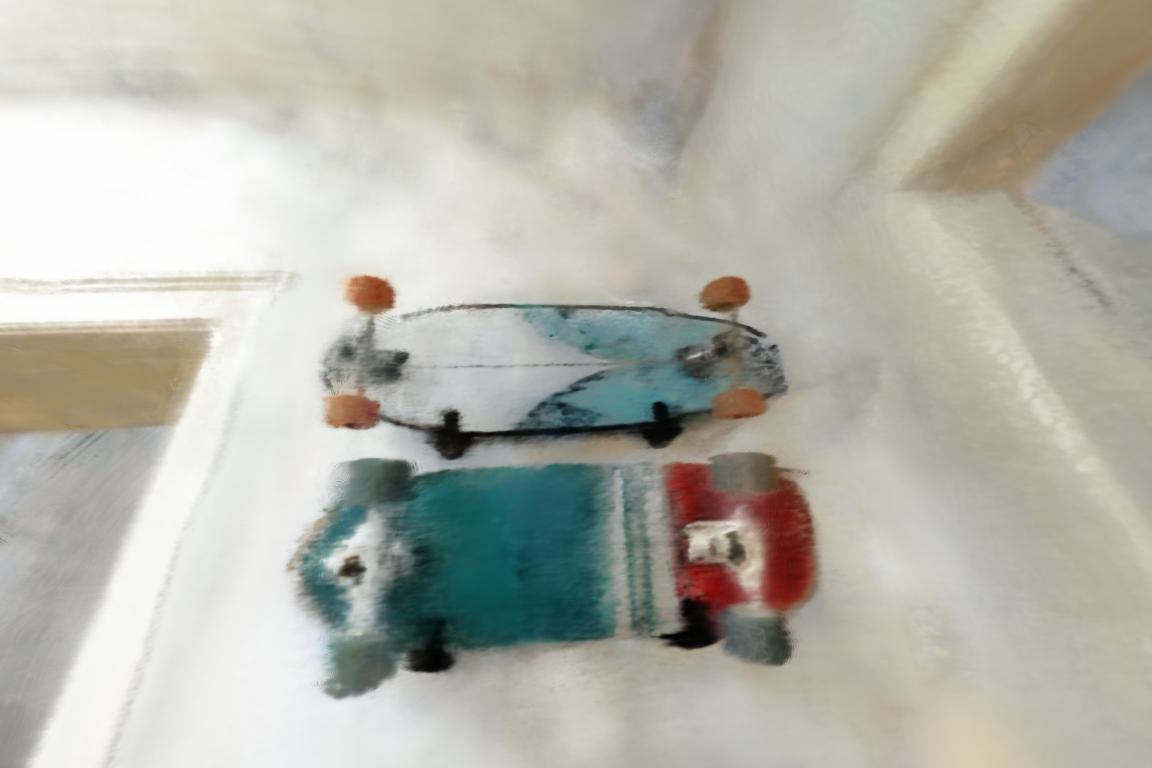}\\

    % 281ba69af1
    \includegraphics[width=0.158\linewidth]{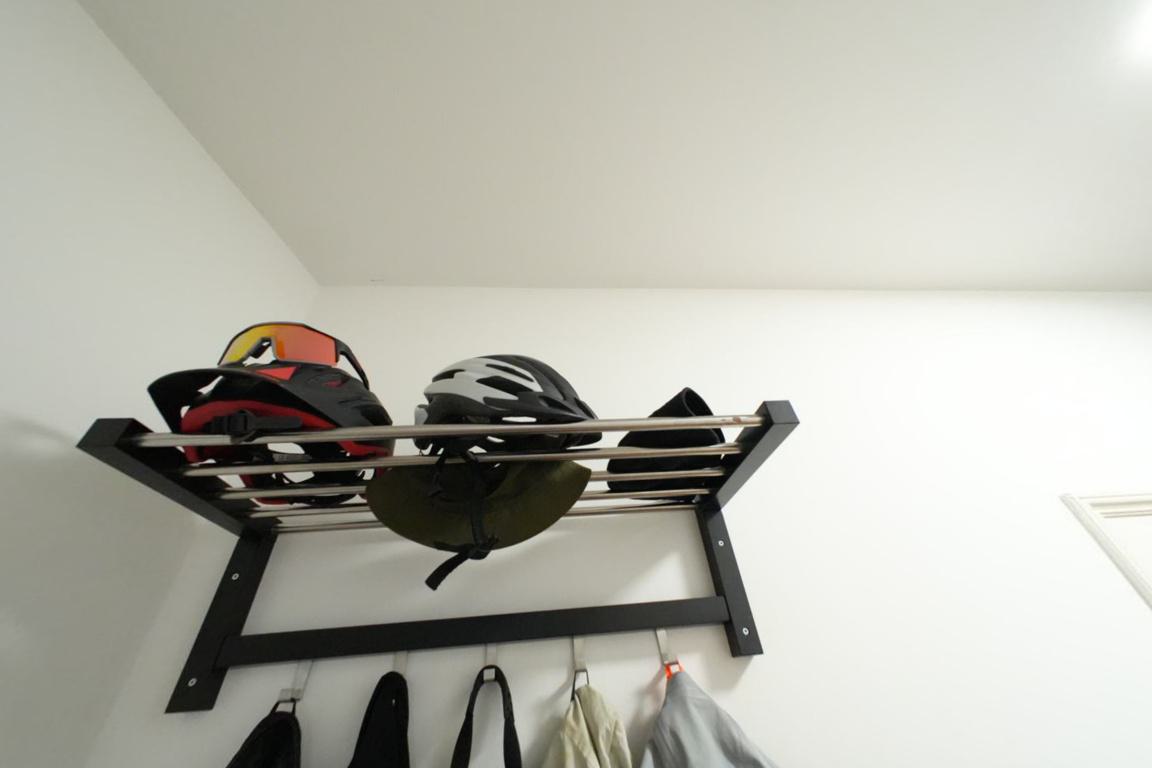}& 
    \includegraphics[width=0.158\linewidth]{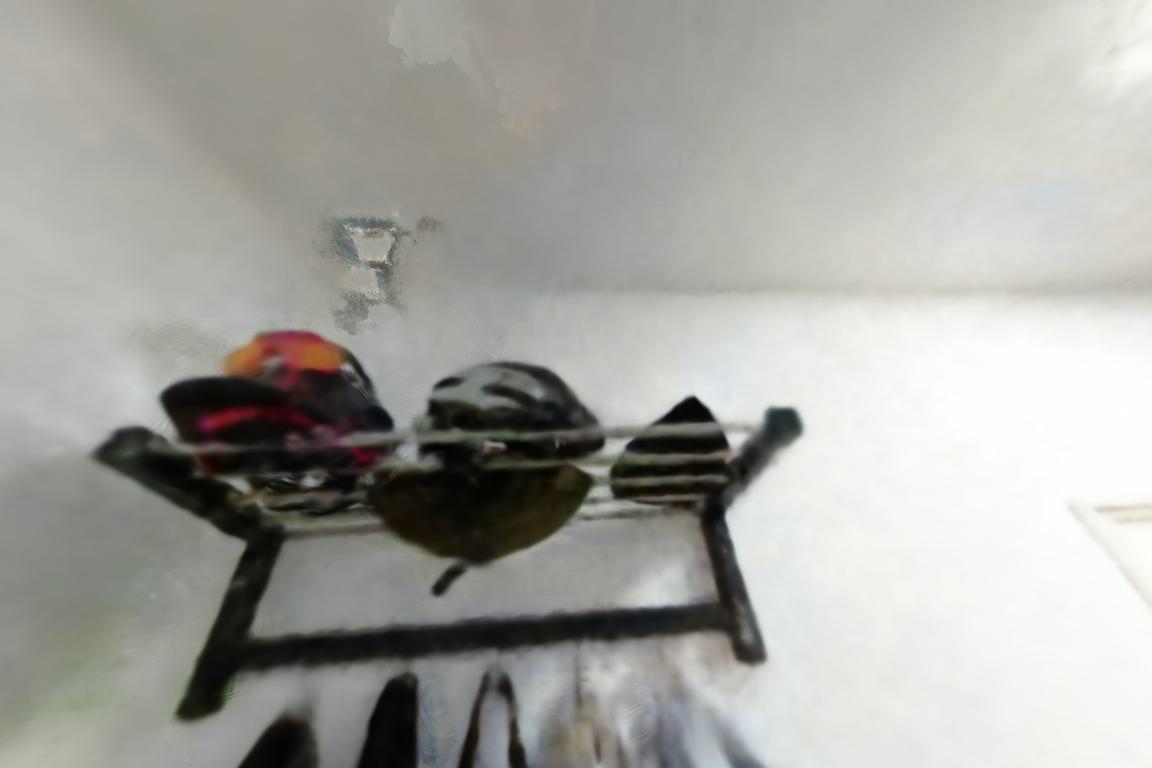}& \includegraphics[width=0.158\linewidth]{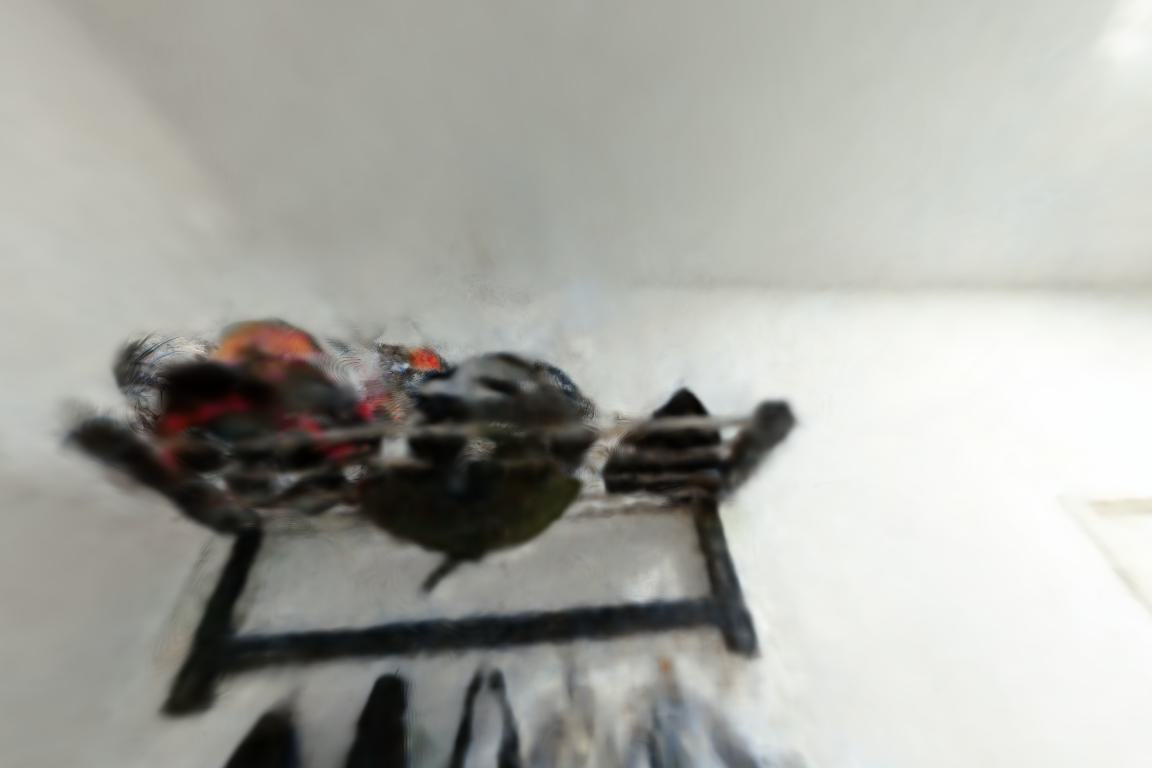}& \includegraphics[width=0.158\linewidth]{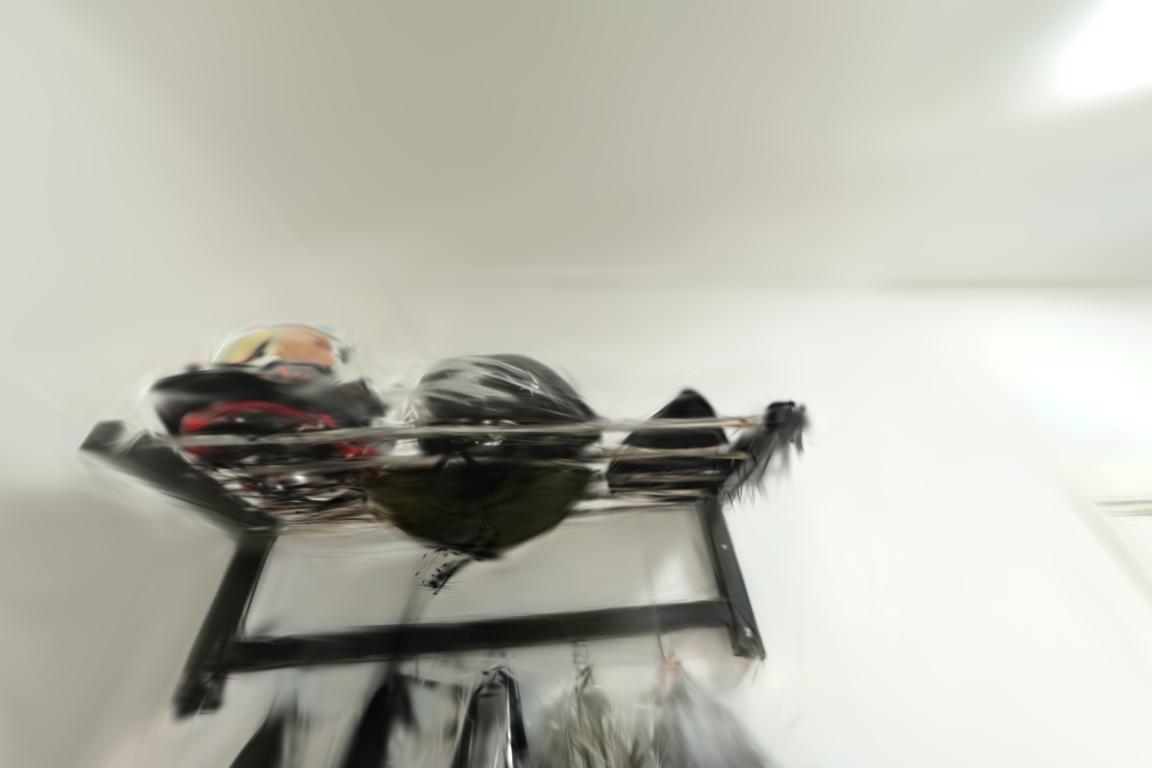}&
    \includegraphics[width=0.158\linewidth]{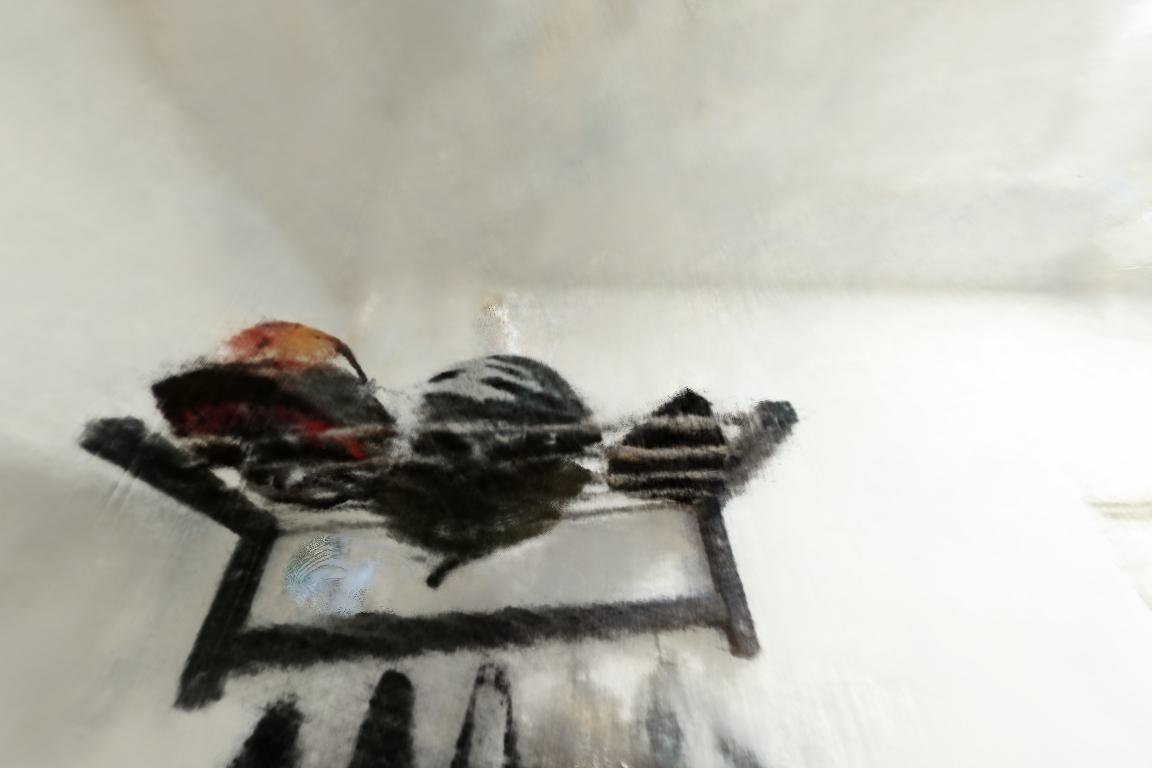}&
    \includegraphics[width=0.158\linewidth]{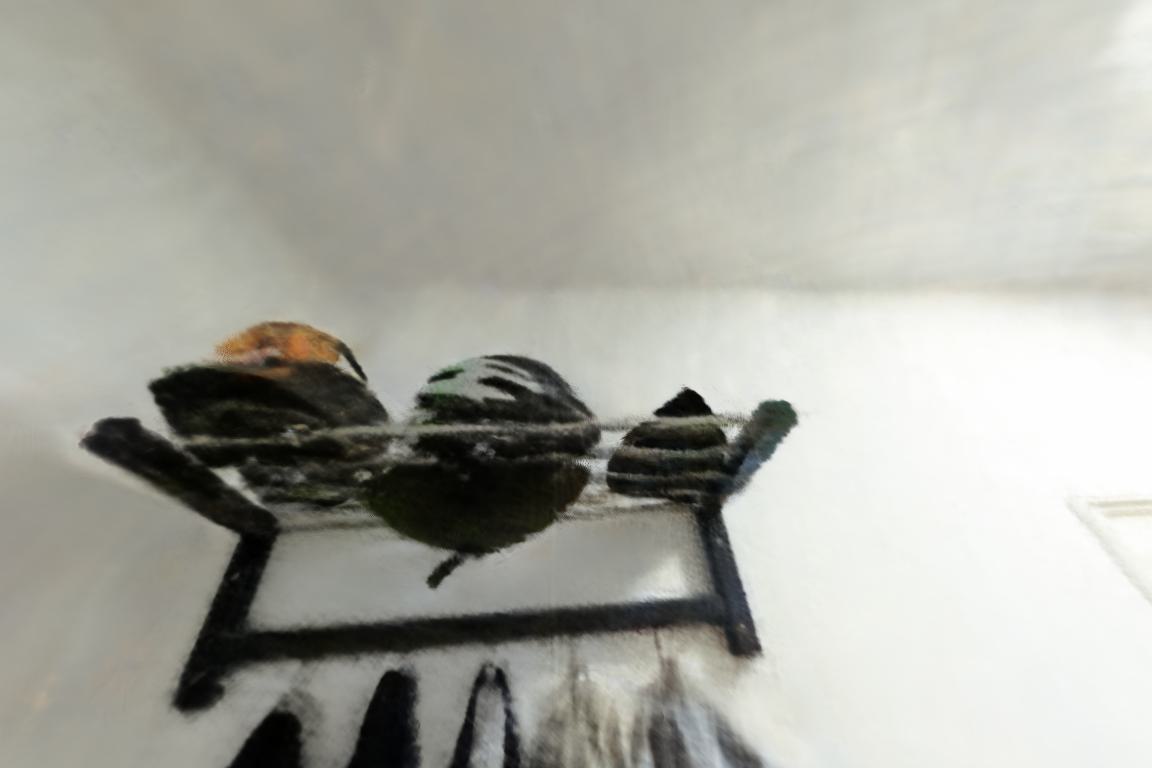}\\
    
    % \includegraphics[width=0.158\linewidth]{suppl_figures/scannetpp/281ba69af1/00006/gt.jpg}& 
    % \includegraphics[width=0.158\linewidth]{suppl_figures/scannetpp/281ba69af1/00006/nerfacto.jpg}& \includegraphics[width=0.158\linewidth]{suppl_figures/scannetpp/281ba69af1/00006/depth-nerfacto.jpg}& \includegraphics[width=0.158\linewidth]{suppl_figures/scannetpp/281ba69af1/00006/3dgs.jpg}&
    % \includegraphics[width=0.158\linewidth]{suppl_figures/scannetpp/281ba69af1/00006/nelf-pro.jpg}&
    % \includegraphics[width=0.158\linewidth]{suppl_figures/scannetpp/281ba69af1/00006/ours_appearance.jpg}\\

    % 785e7504b9
    \includegraphics[width=0.158\linewidth]{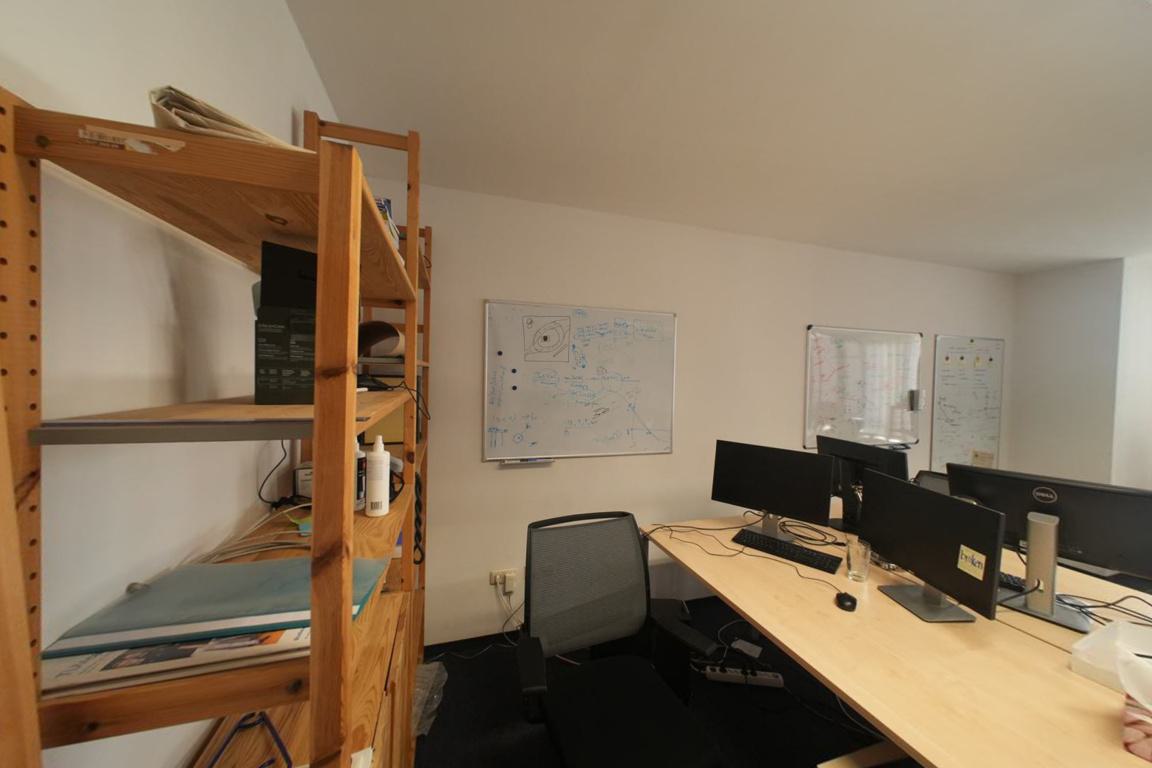}& 
    \includegraphics[width=0.158\linewidth]{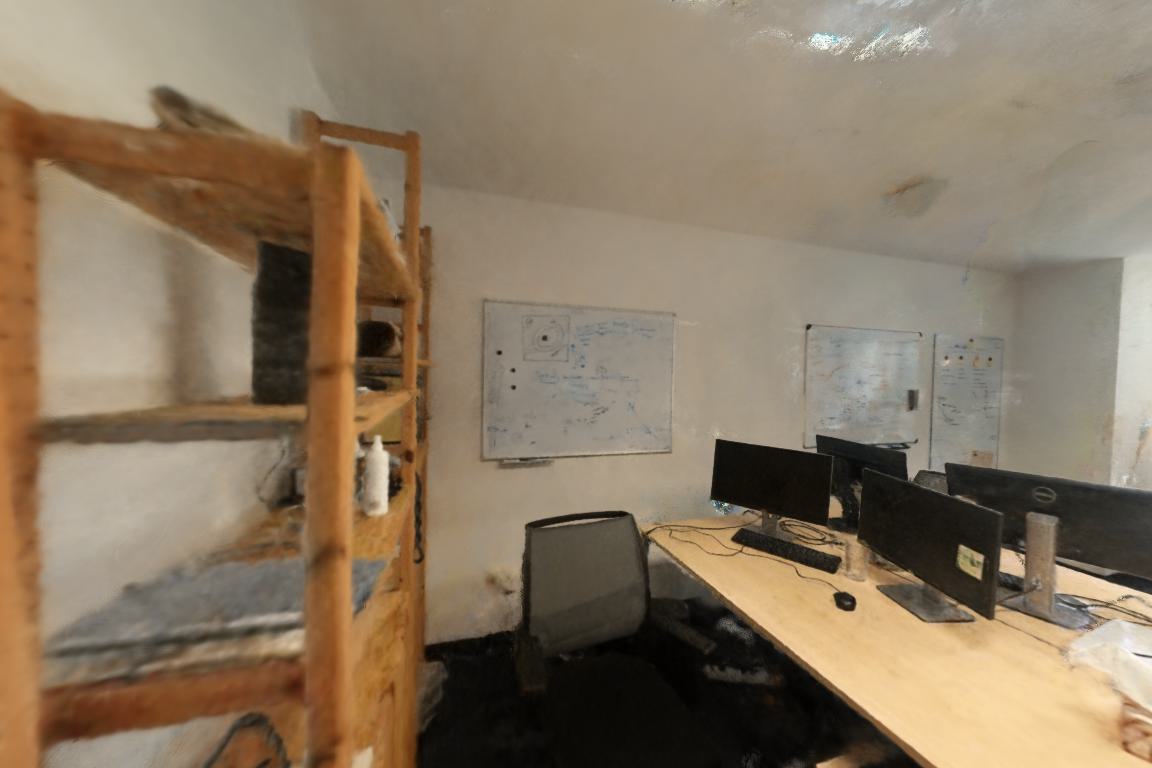}& \includegraphics[width=0.158\linewidth]{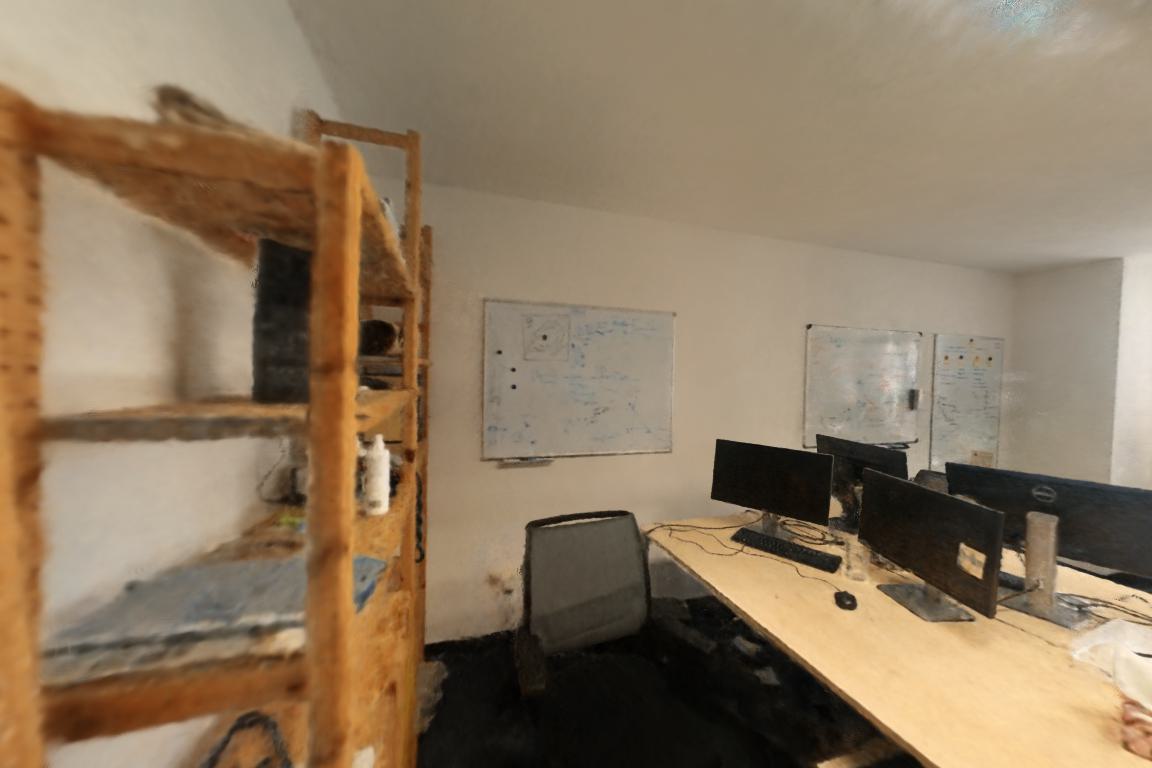}& \includegraphics[width=0.158\linewidth]{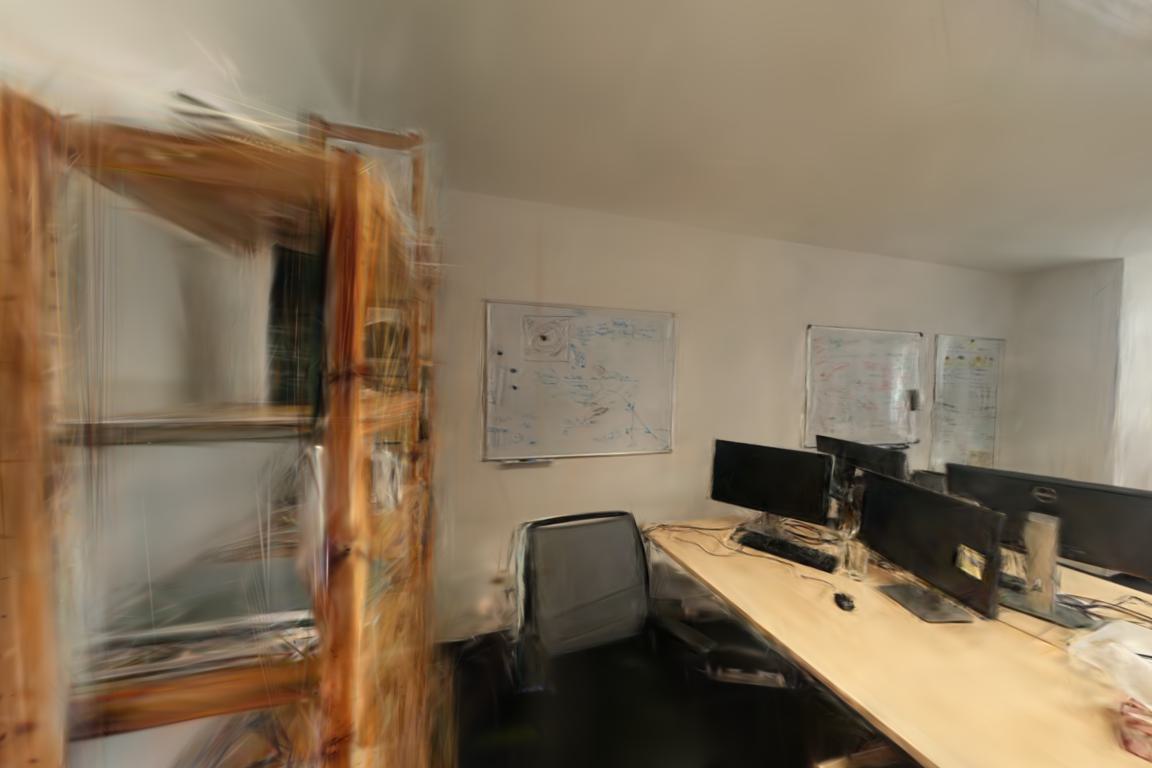}&
    \includegraphics[width=0.158\linewidth]{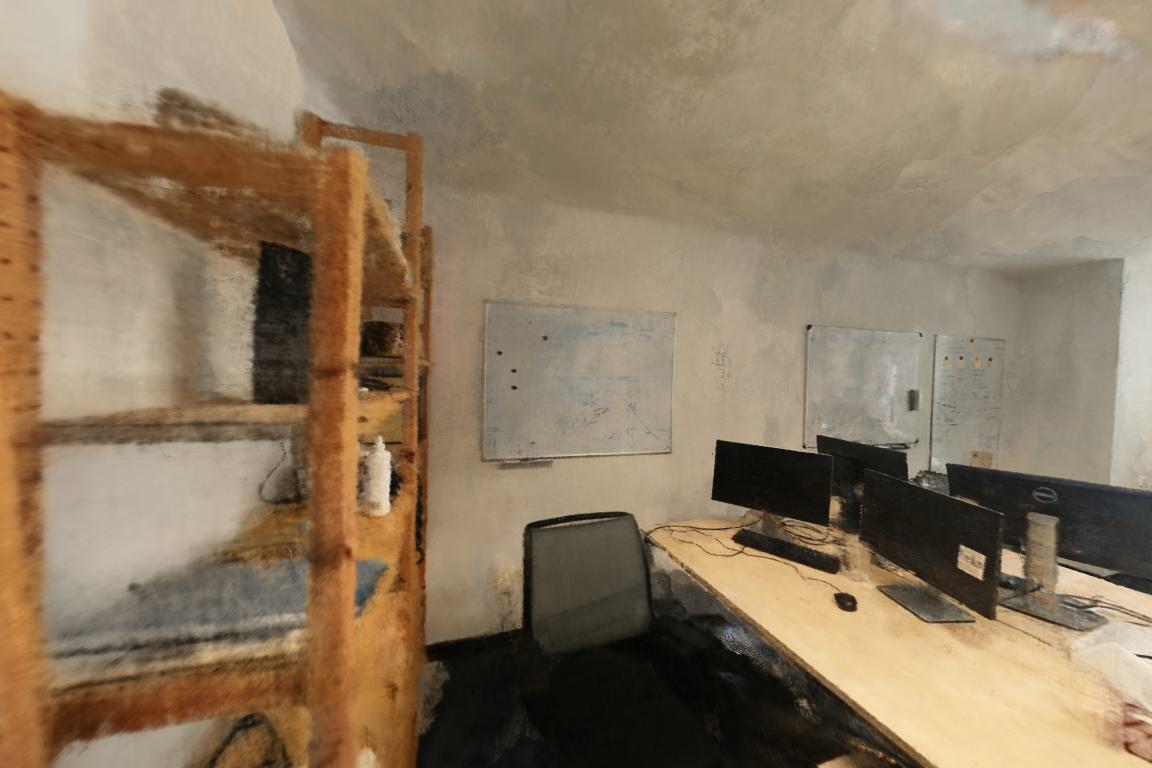}&
    \includegraphics[width=0.158\linewidth]{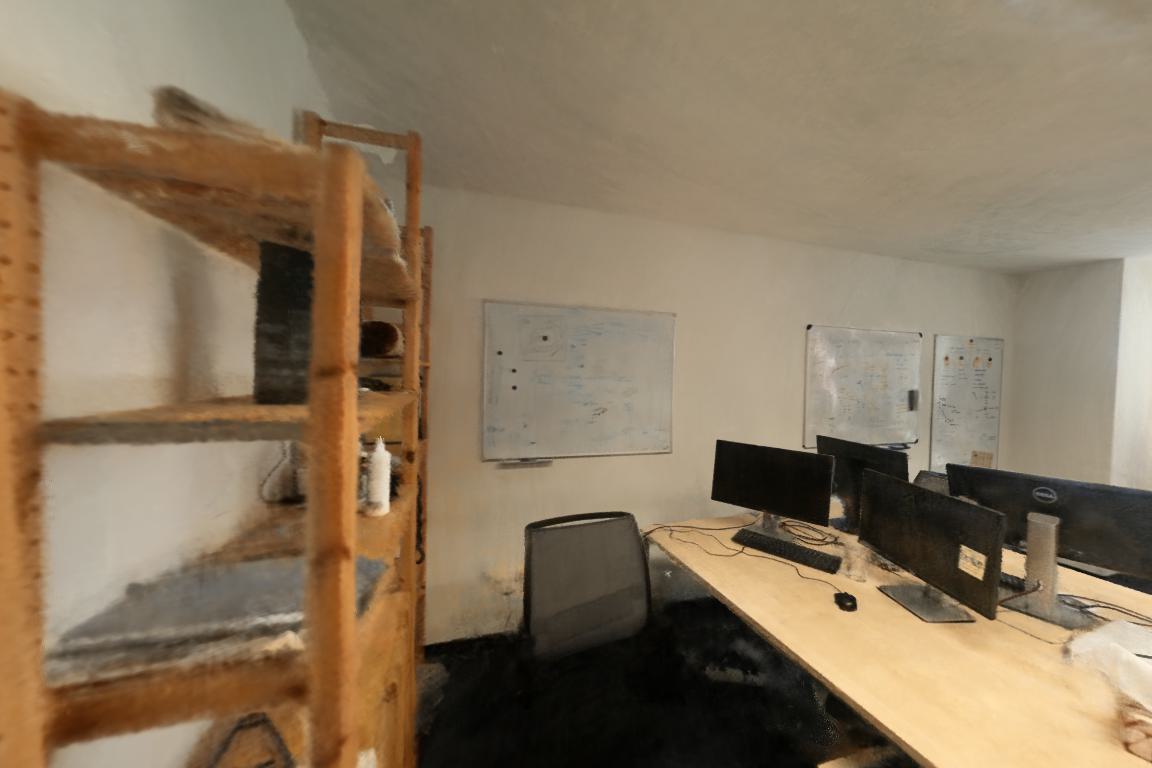}\\

    \includegraphics[width=0.158\linewidth]{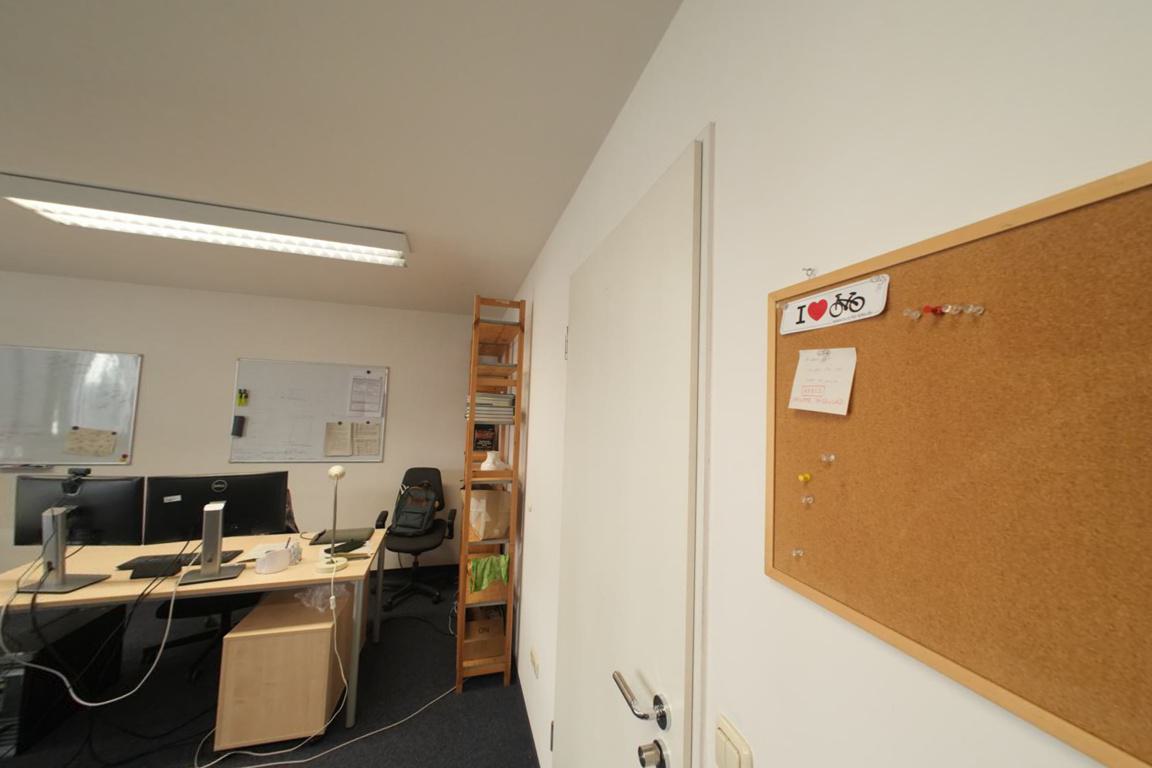}& 
    \includegraphics[width=0.158\linewidth]{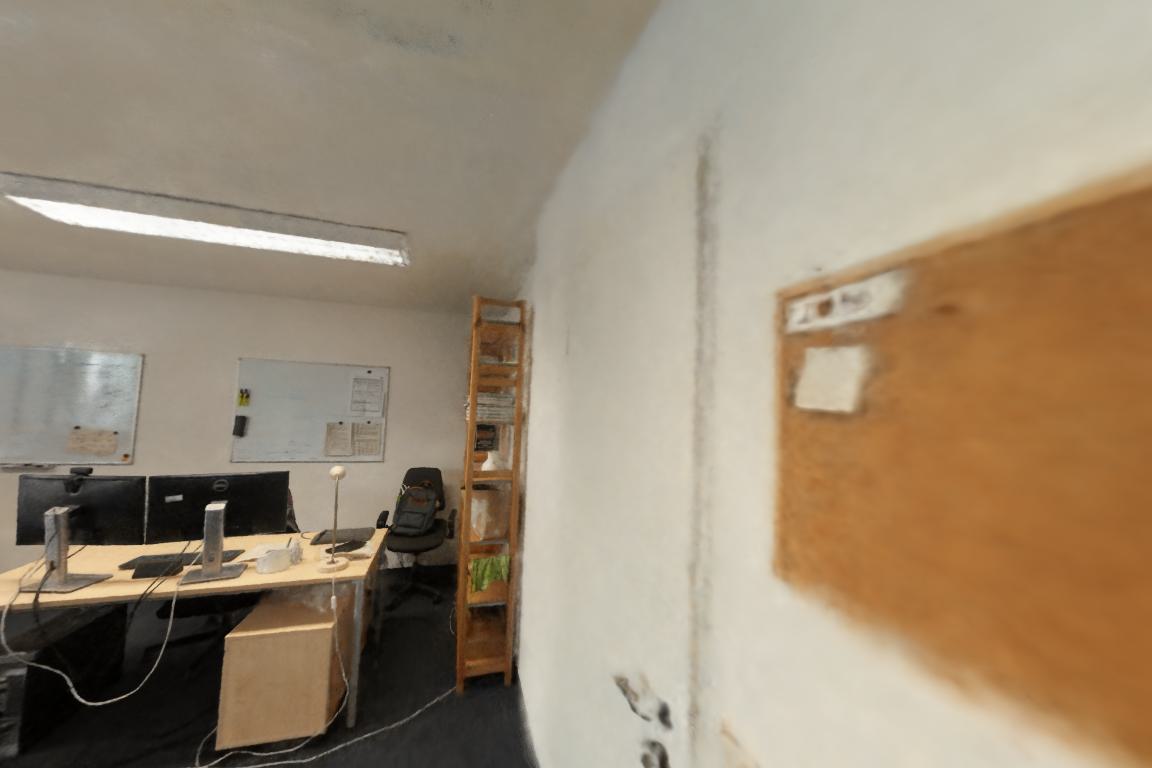}& \includegraphics[width=0.158\linewidth]{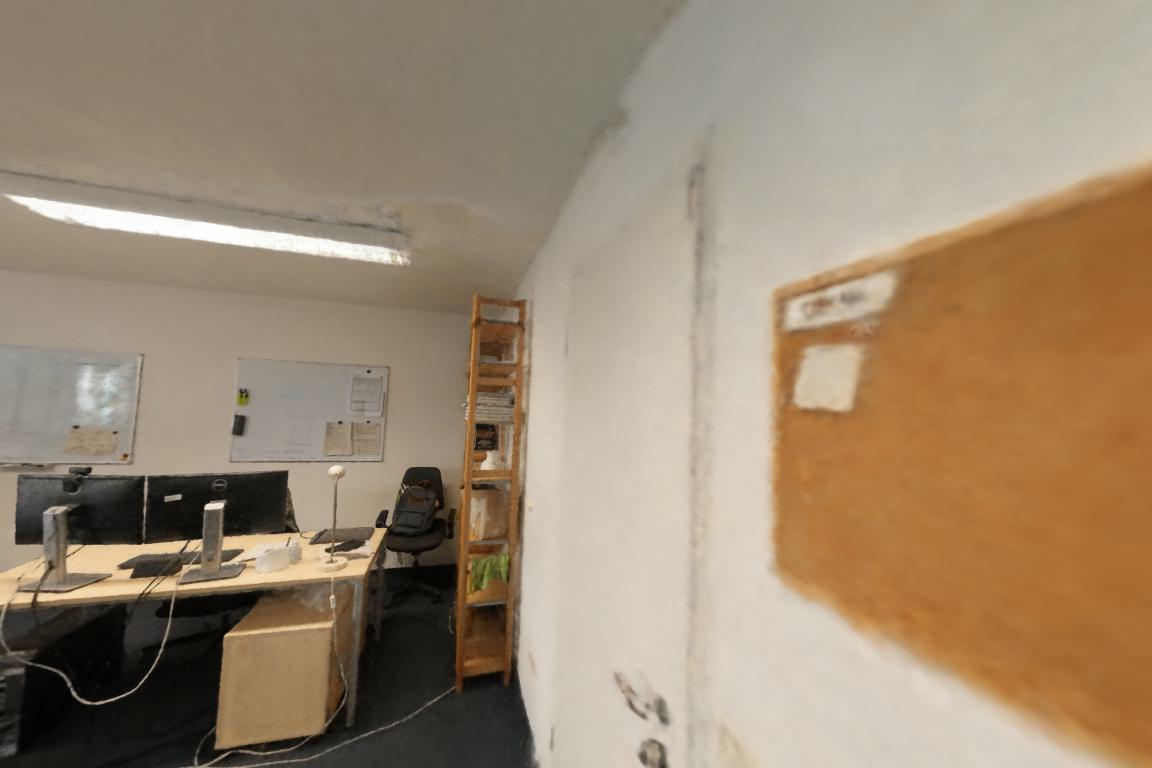}& \includegraphics[width=0.158\linewidth]{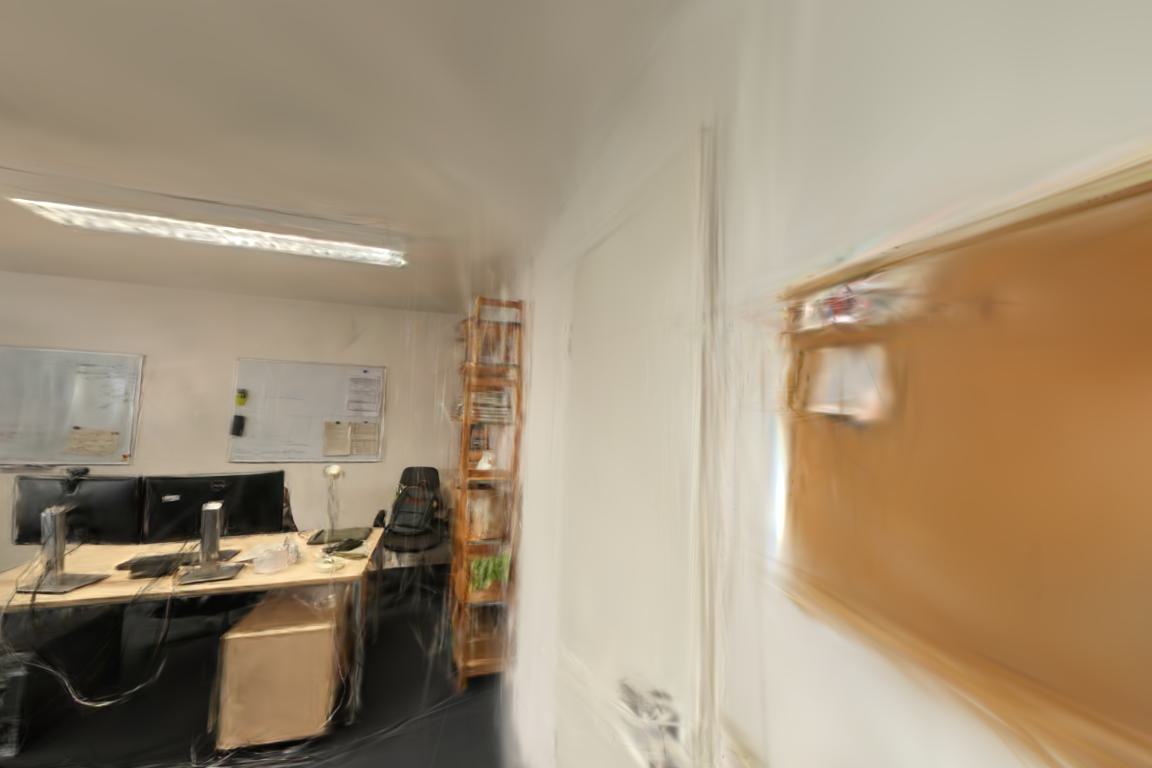}&
    \includegraphics[width=0.158\linewidth]{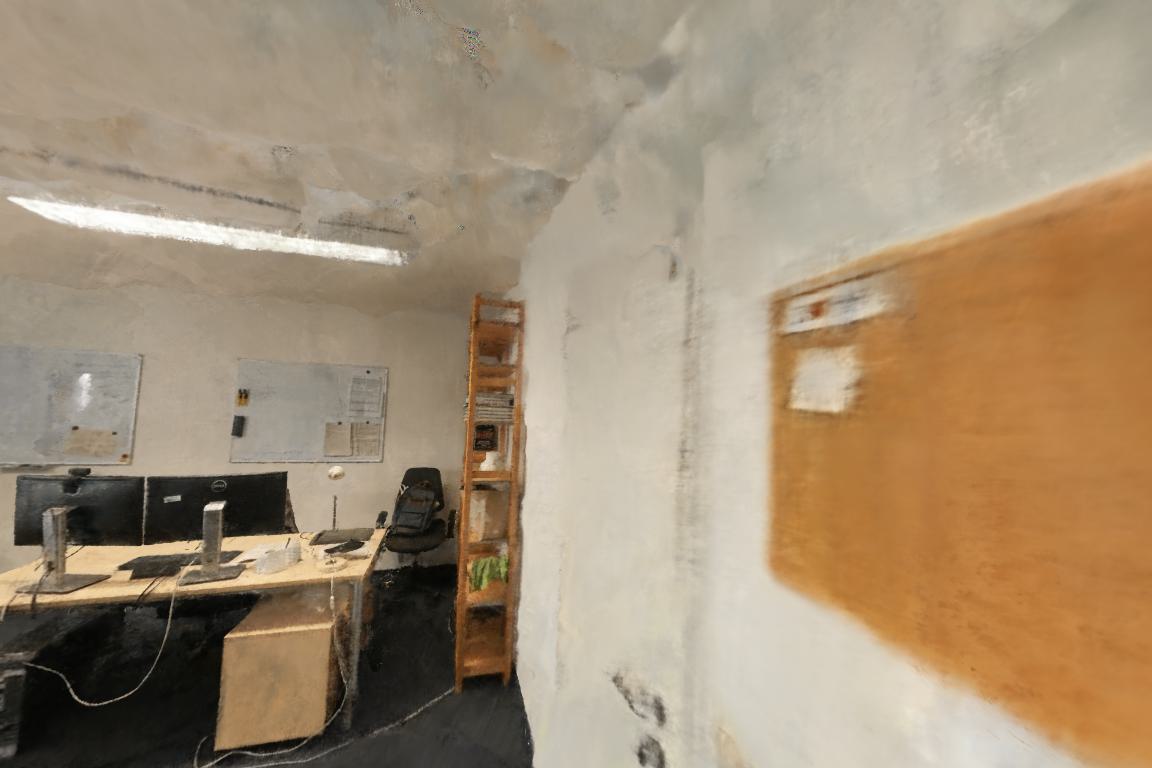}&
    \includegraphics[width=0.158\linewidth]{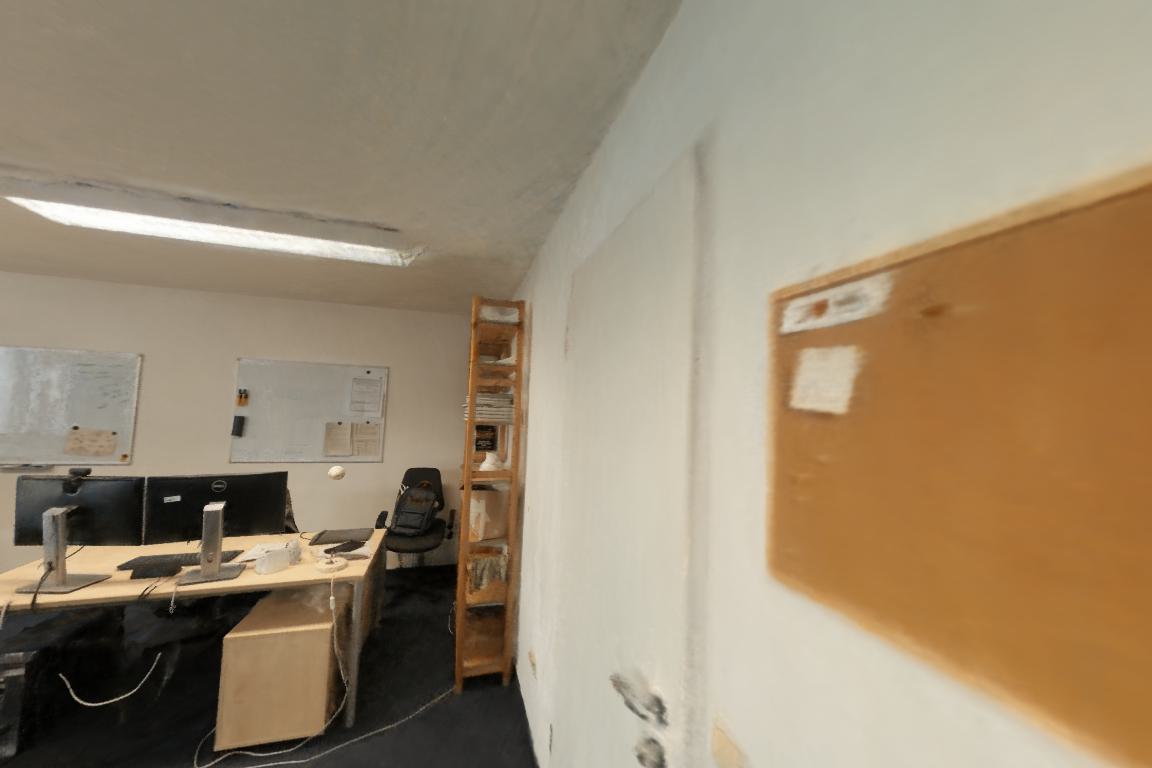}\\

    % bb87c292ad
    \includegraphics[width=0.158\linewidth]{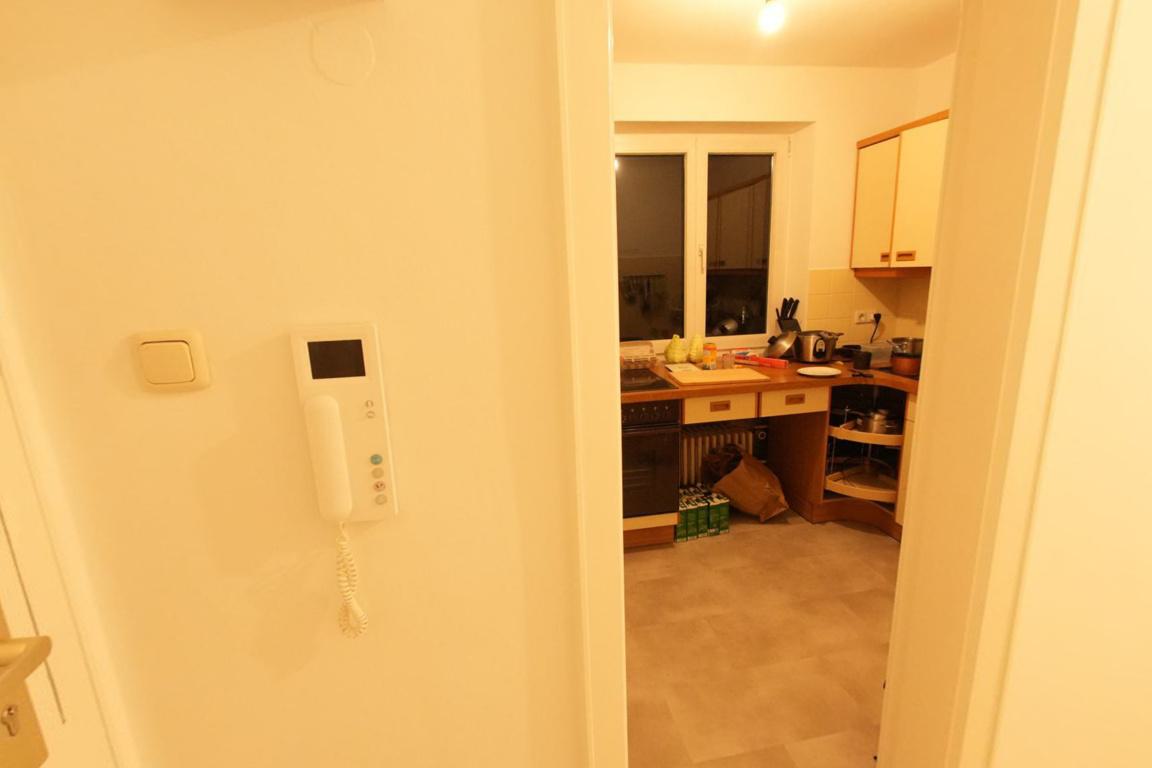}& 
    \includegraphics[width=0.158\linewidth]{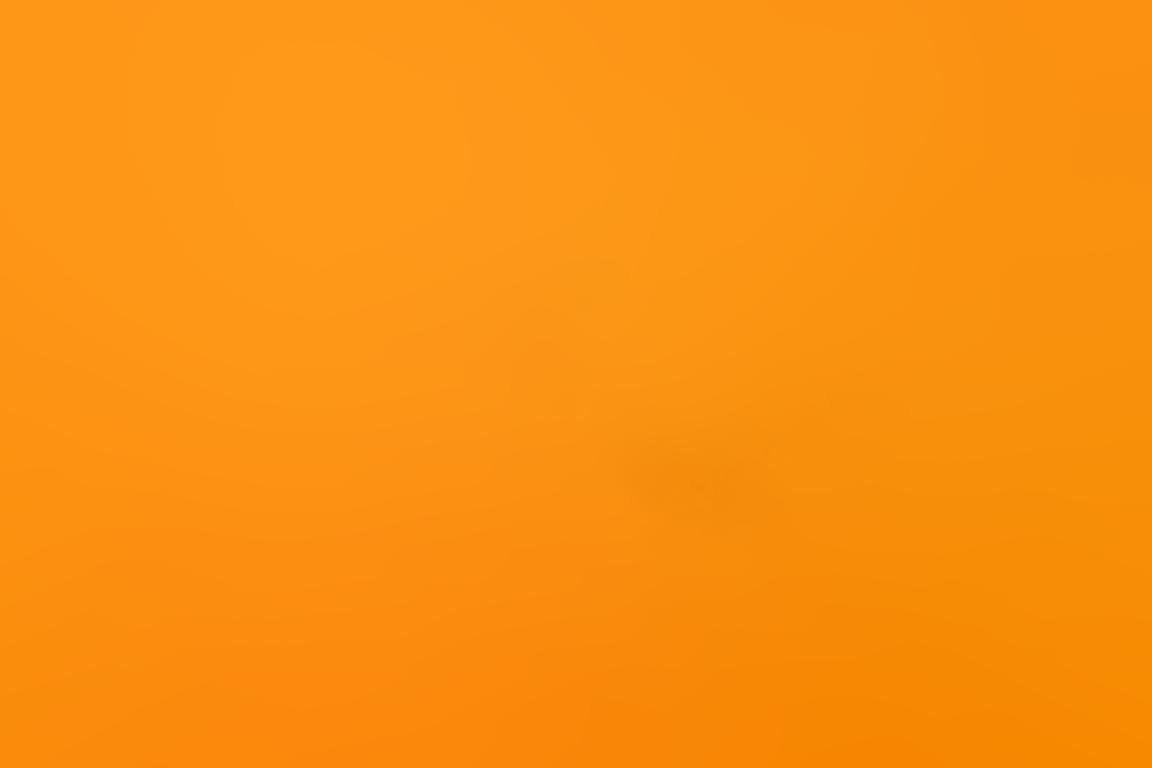}& \includegraphics[width=0.158\linewidth]{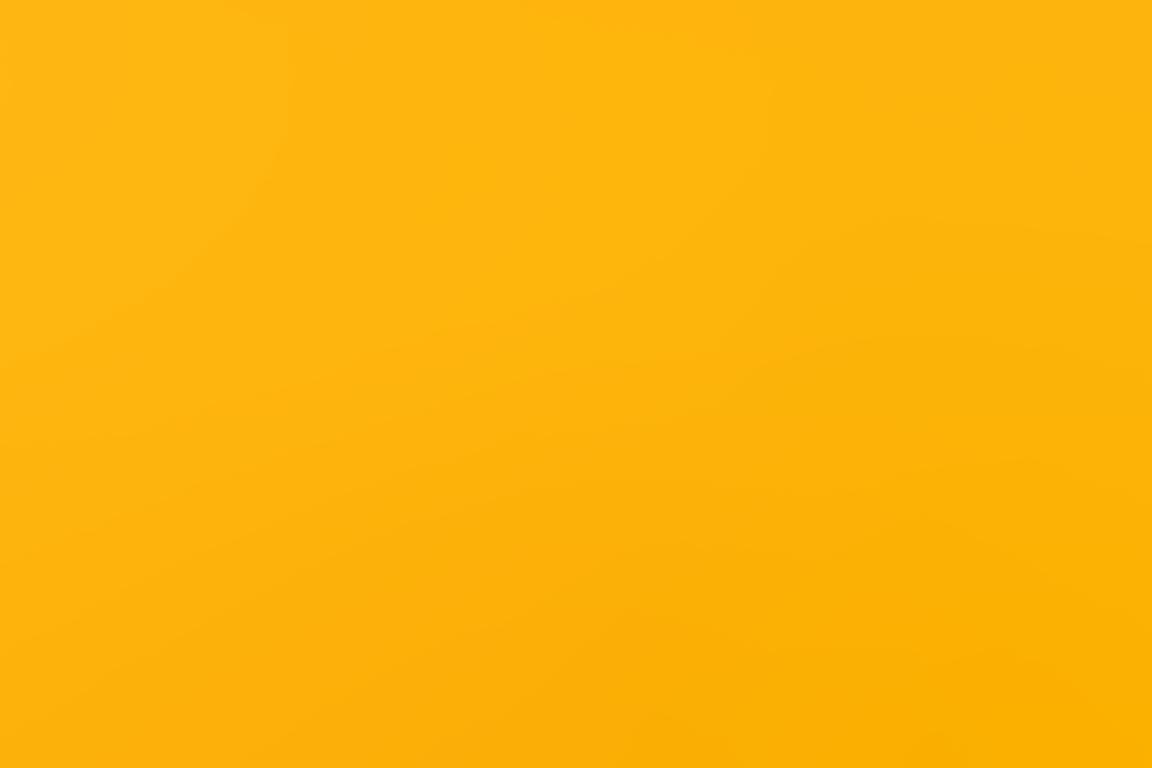}& \includegraphics[width=0.158\linewidth]{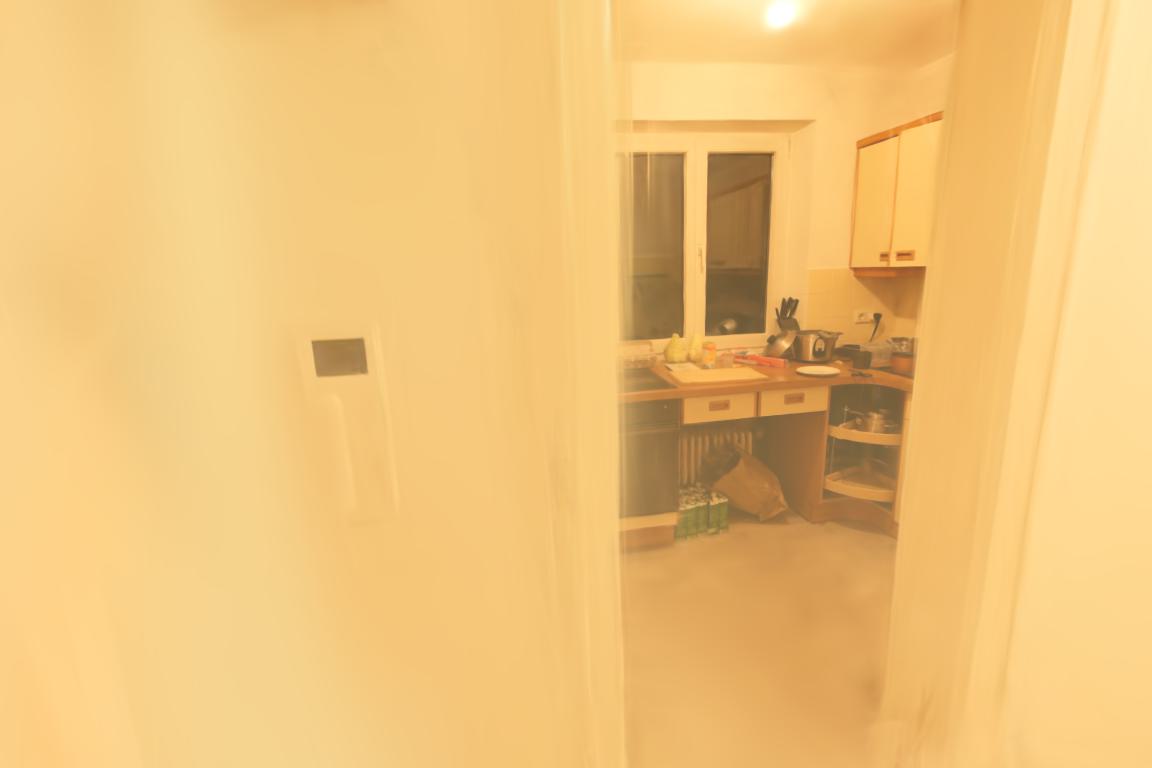}&
    \includegraphics[width=0.158\linewidth]{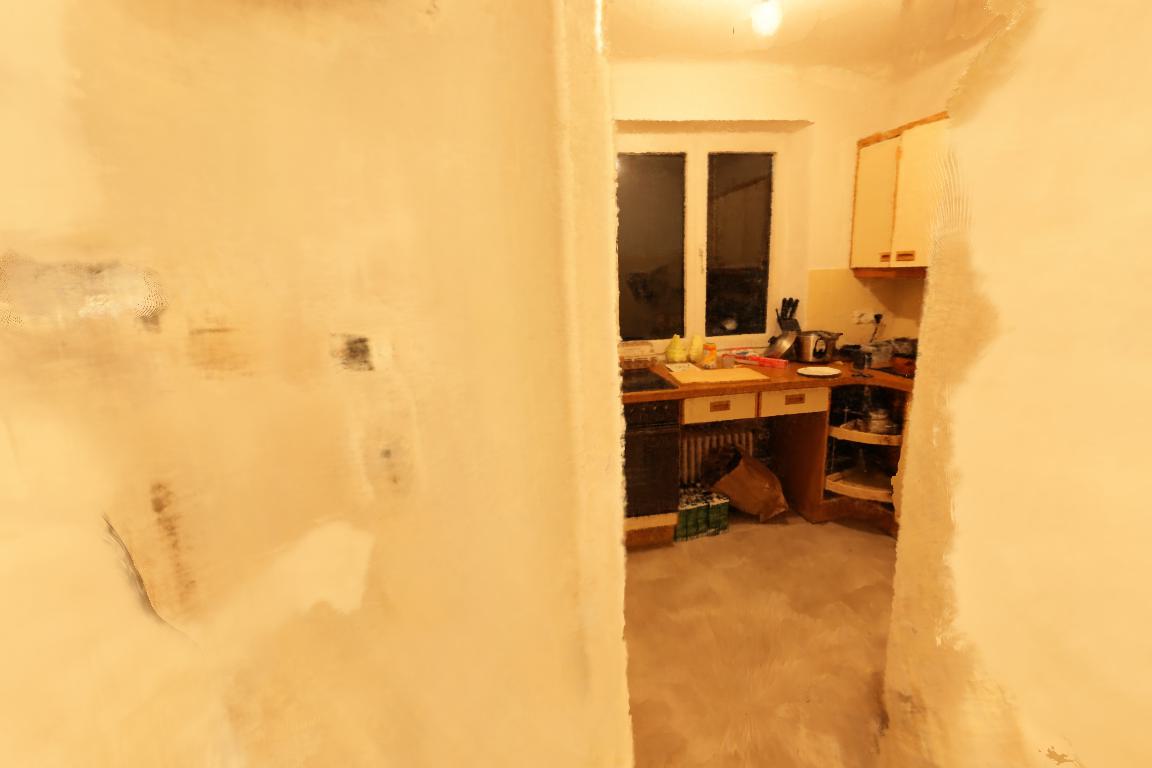}&
    \includegraphics[width=0.158\linewidth]{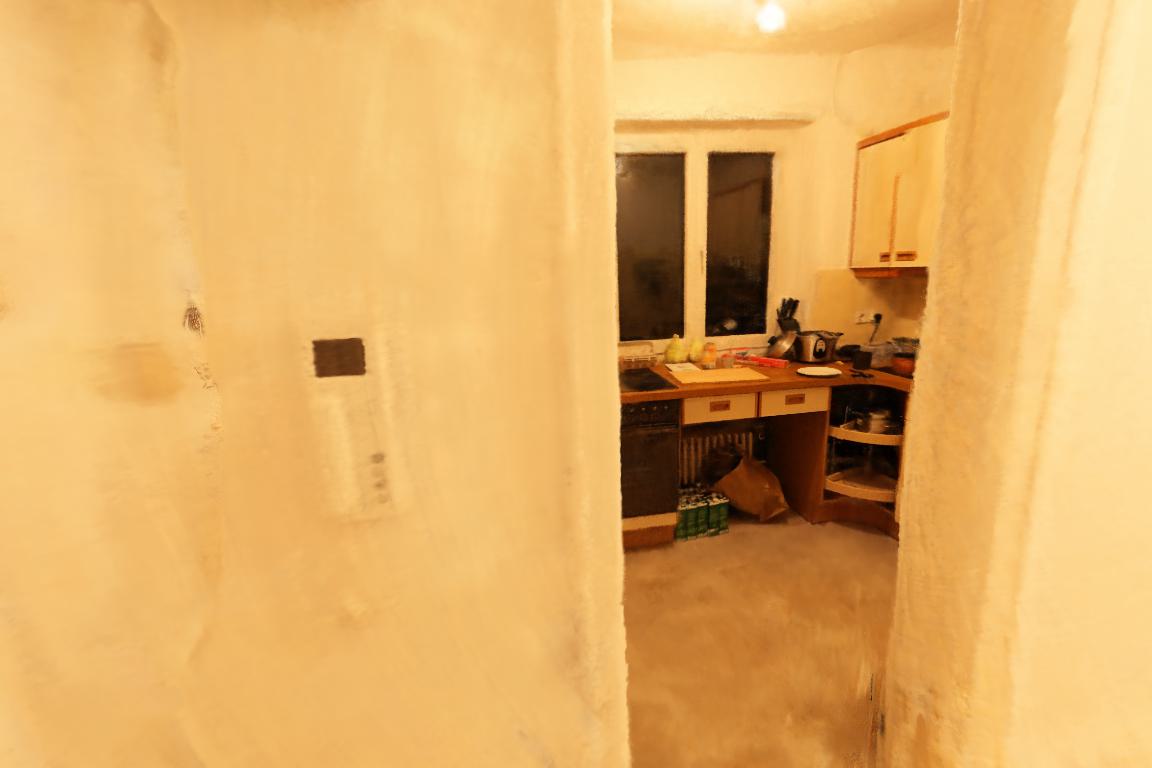}\\

    \includegraphics[width=0.158\linewidth]{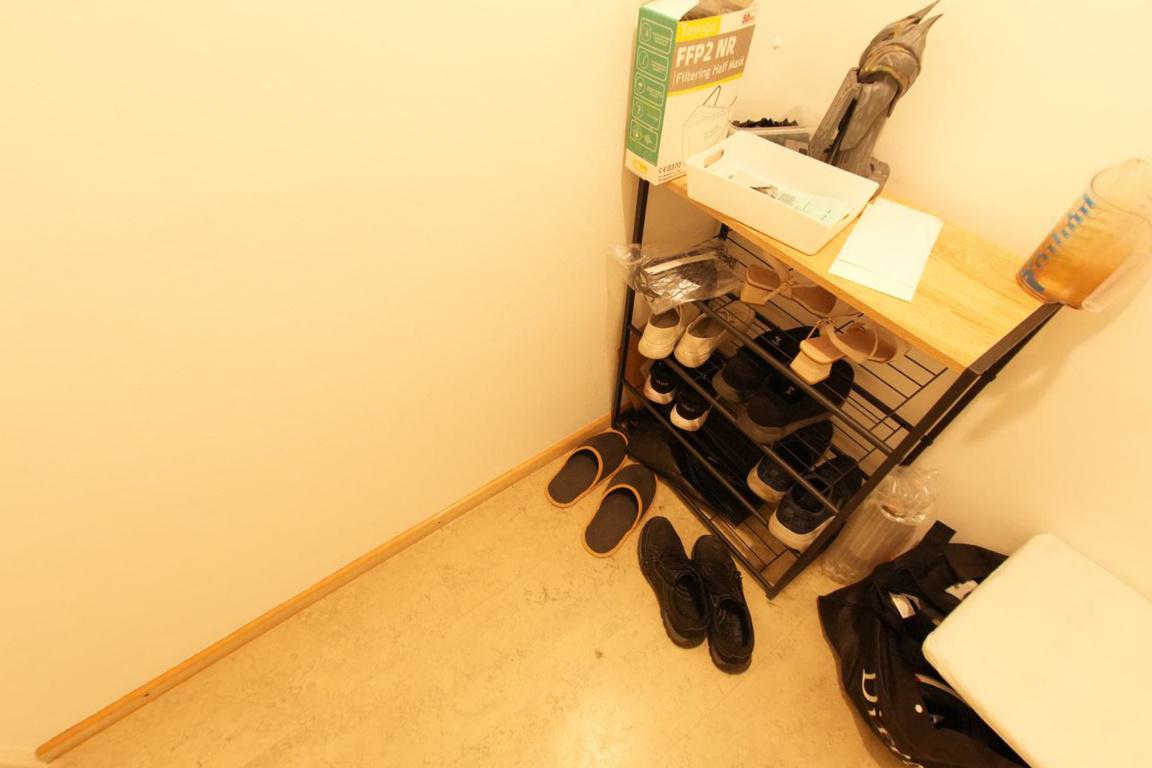}& 
    \includegraphics[width=0.158\linewidth]{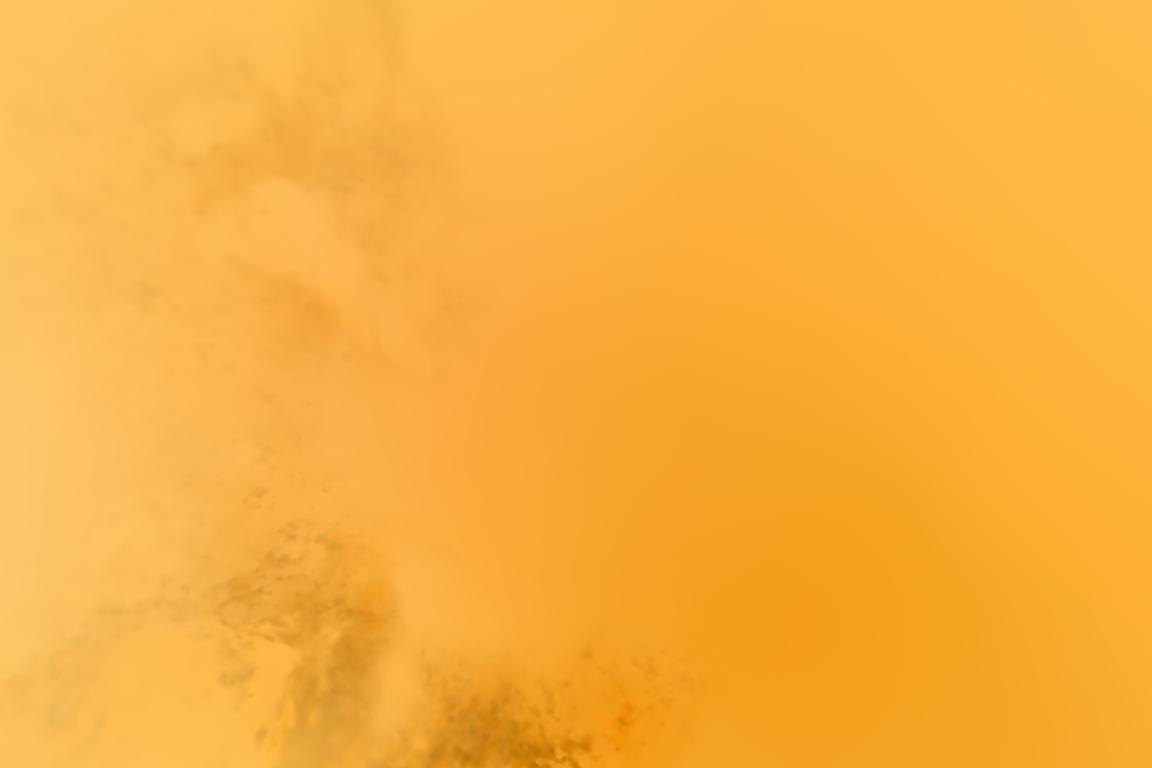}& \includegraphics[width=0.158\linewidth]{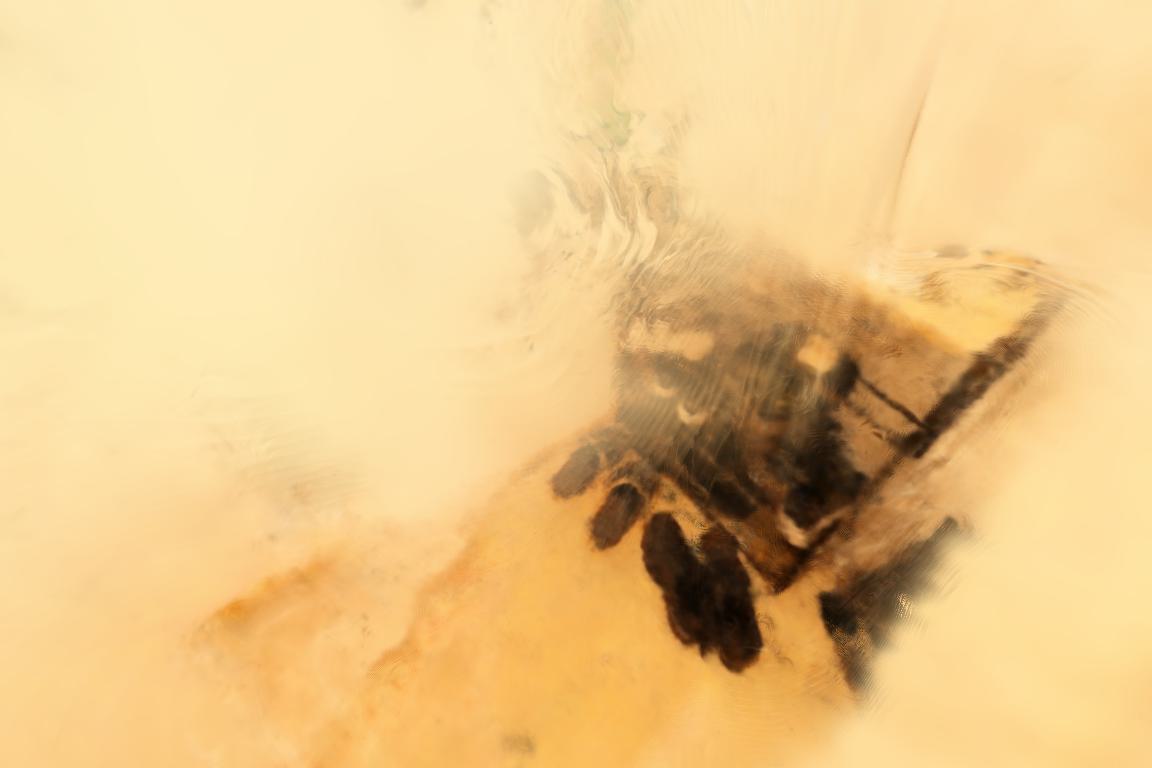}& \includegraphics[width=0.158\linewidth]{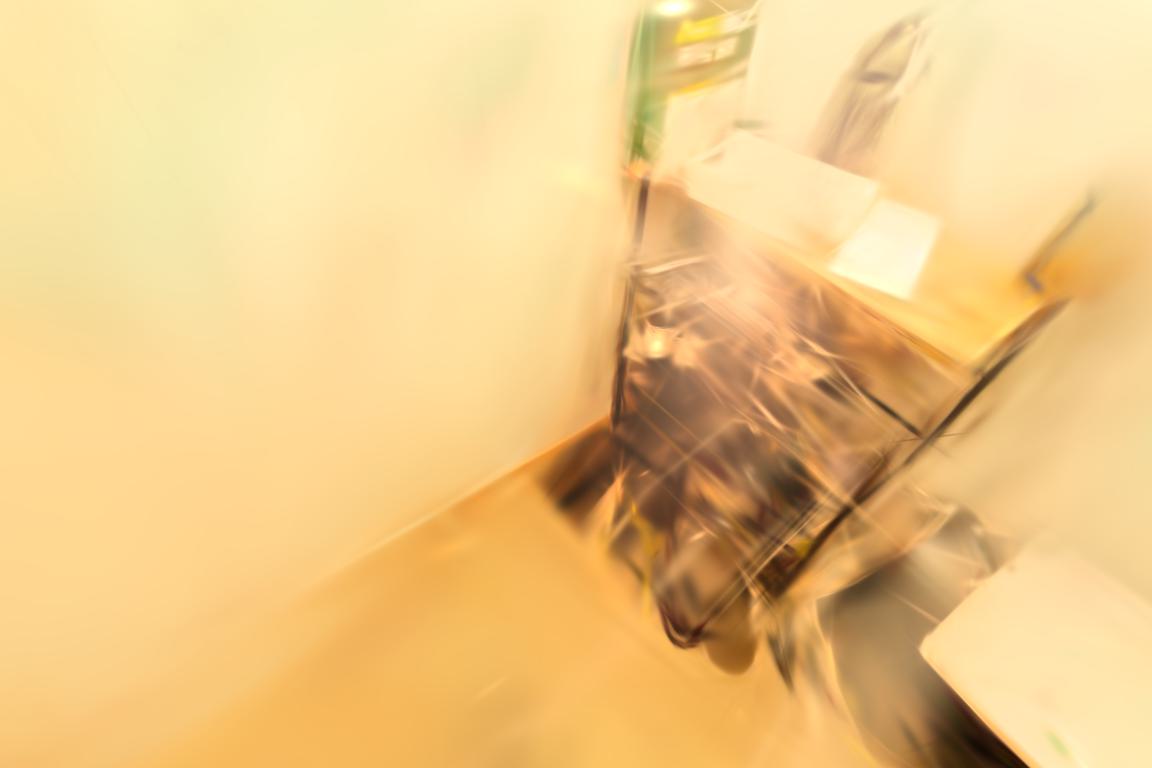}&
    \includegraphics[width=0.158\linewidth]{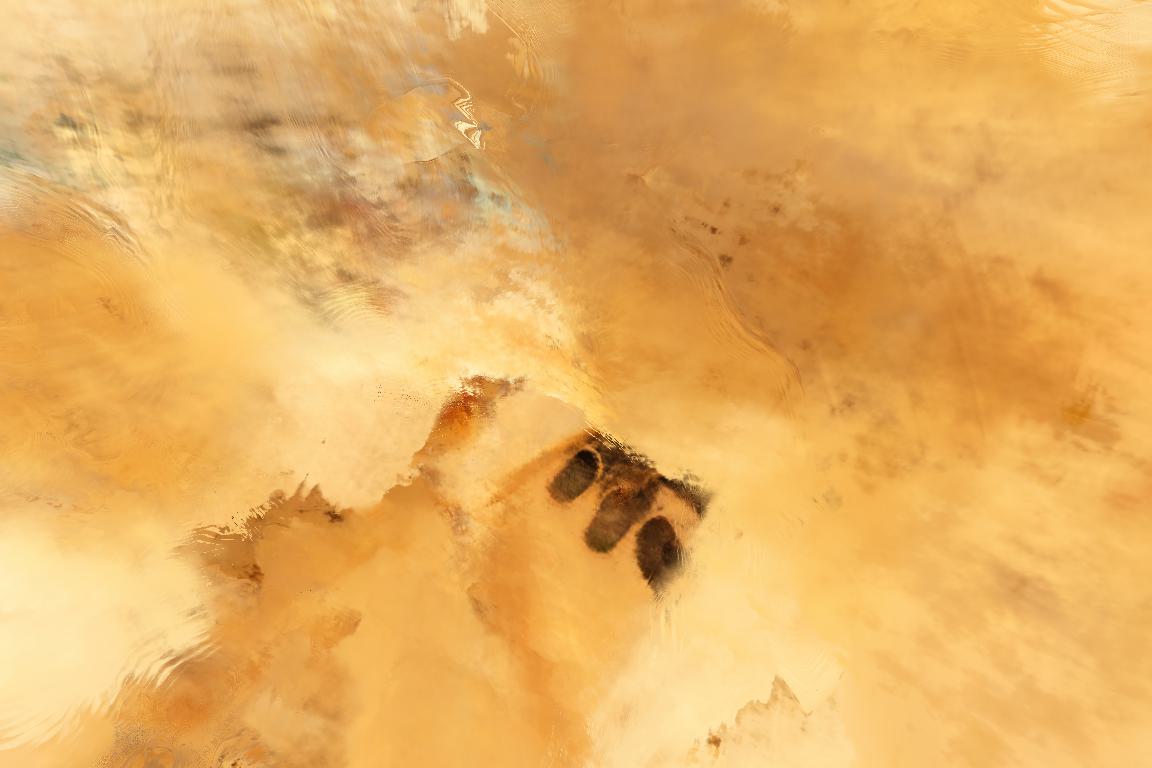}&
    \includegraphics[width=0.158\linewidth]{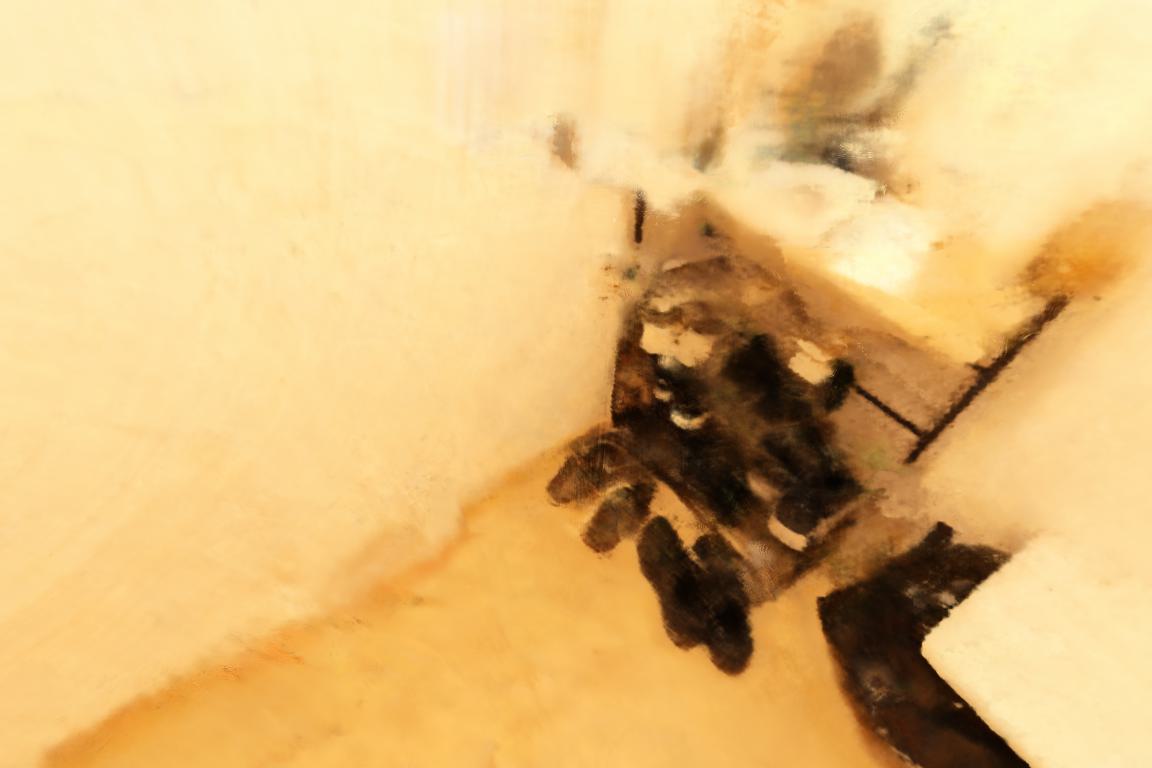}\\

    % e91722b5a3
    \includegraphics[width=0.158\linewidth]{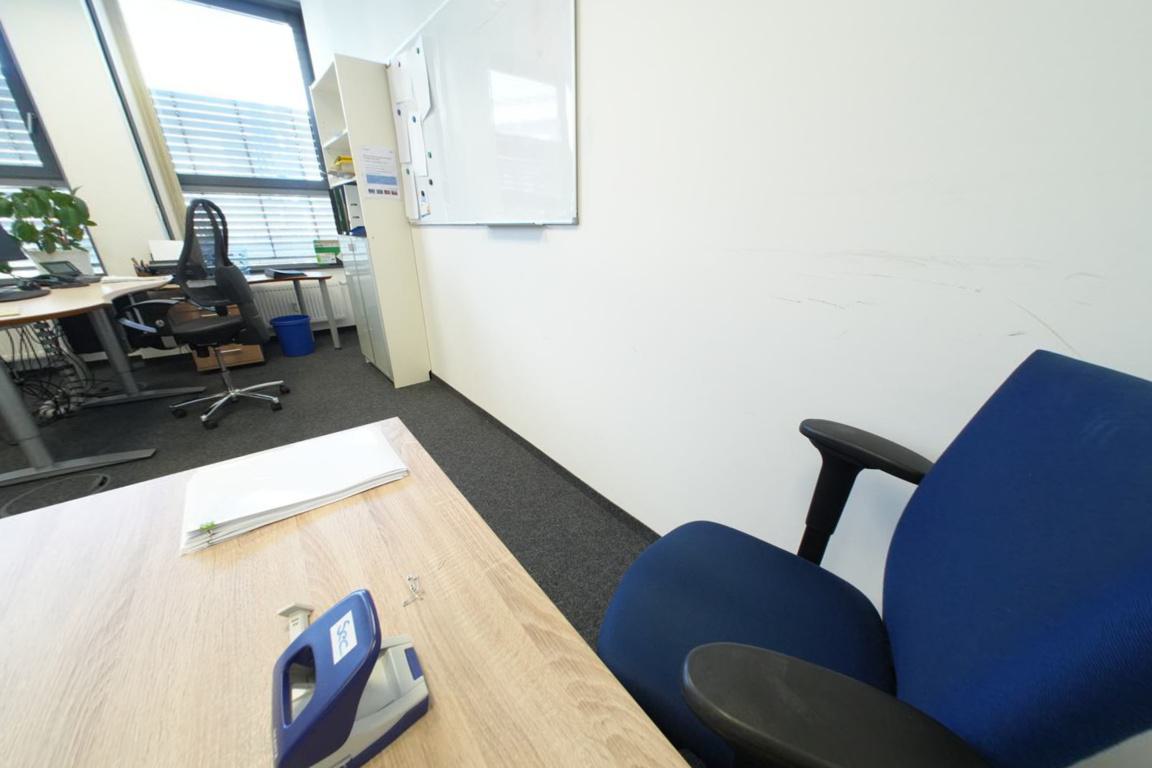}& 
    \includegraphics[width=0.158\linewidth]{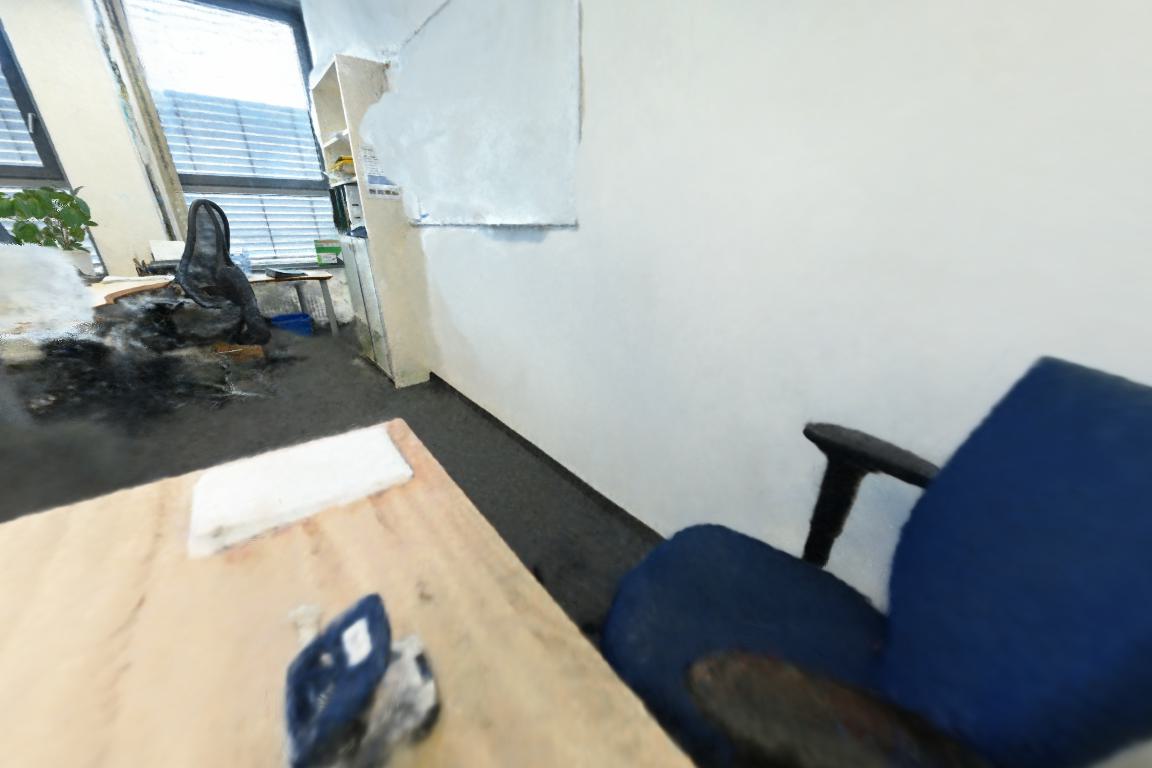}& \includegraphics[width=0.158\linewidth]{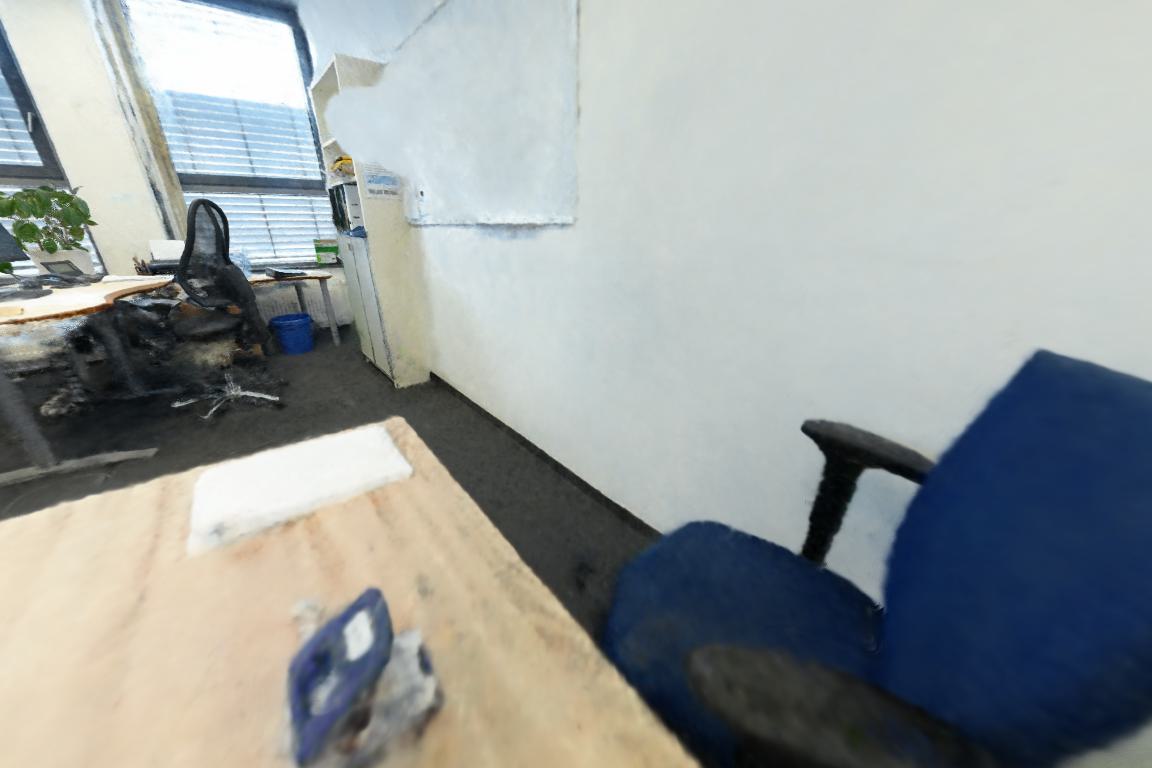}& \includegraphics[width=0.158\linewidth]{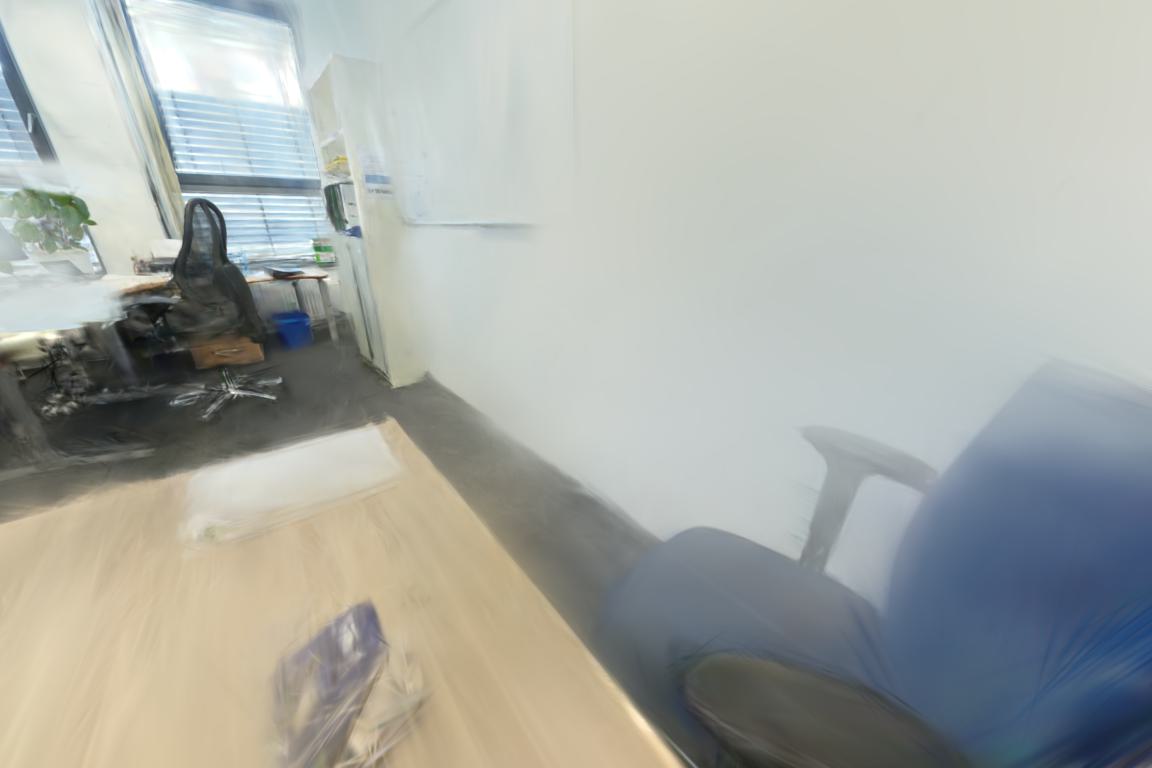}&
    \includegraphics[width=0.158\linewidth]{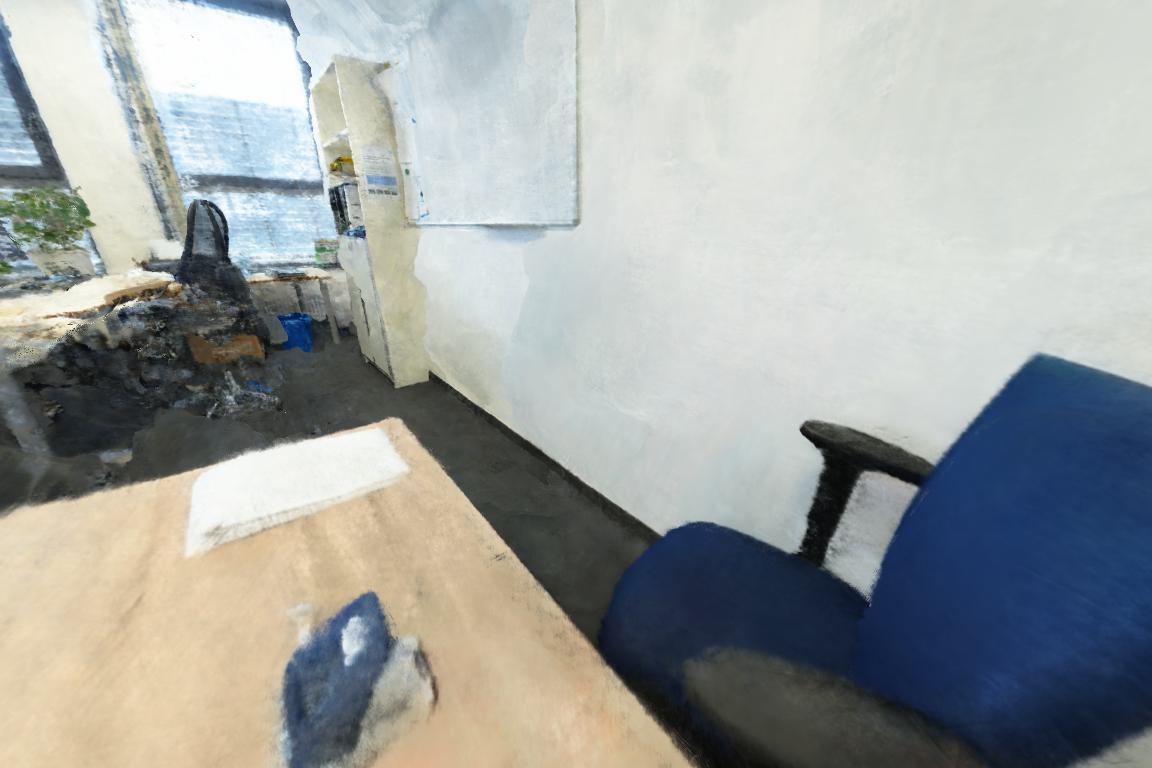}&
    \includegraphics[width=0.158\linewidth]{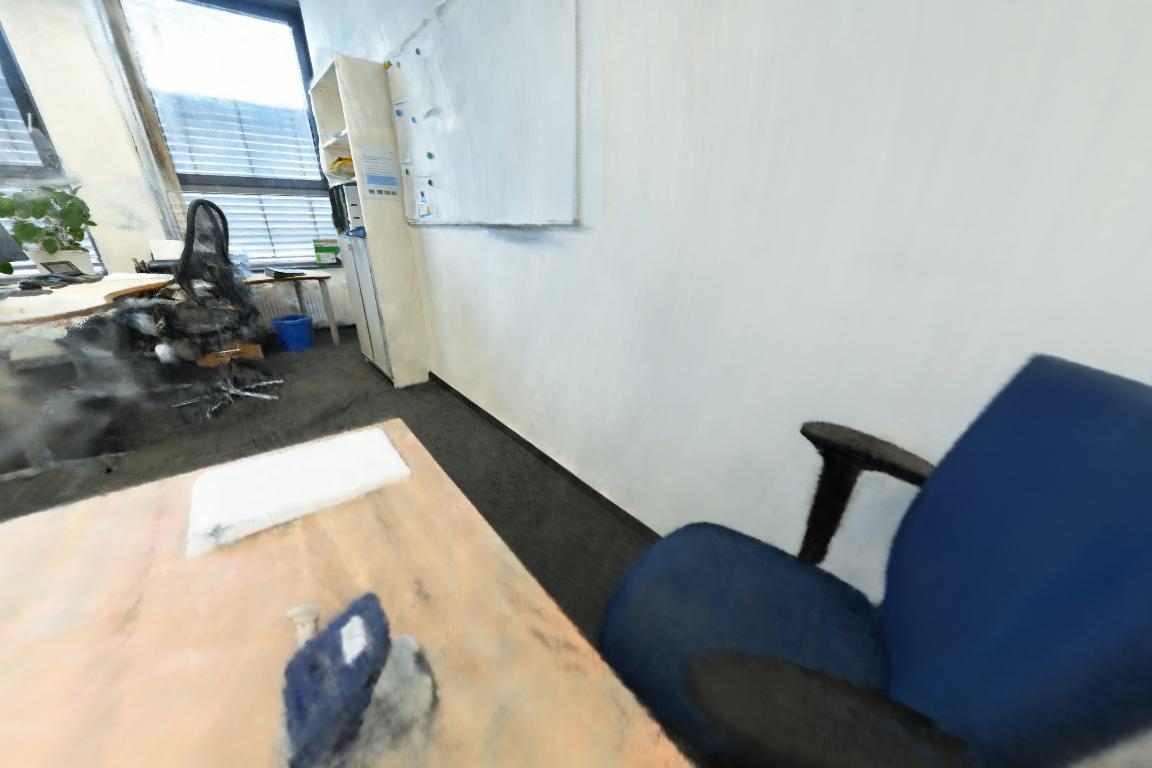}\\

    \includegraphics[width=0.158\linewidth]{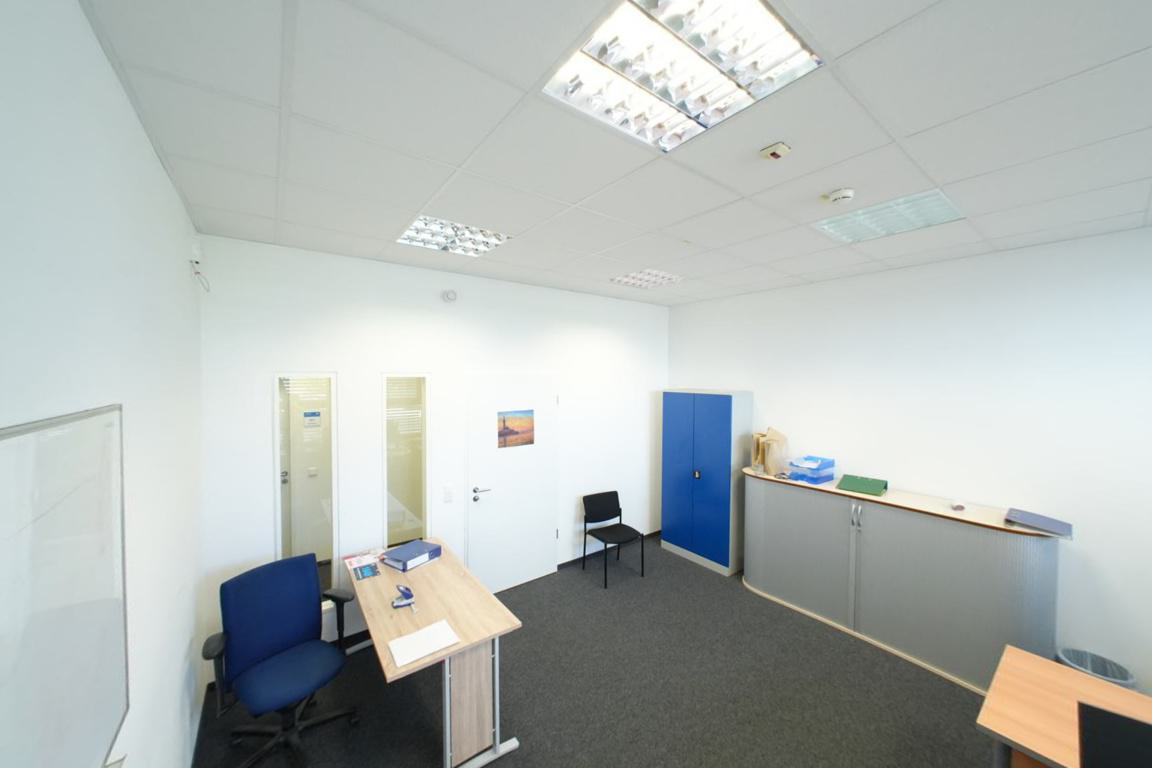}& 
    \includegraphics[width=0.158\linewidth]{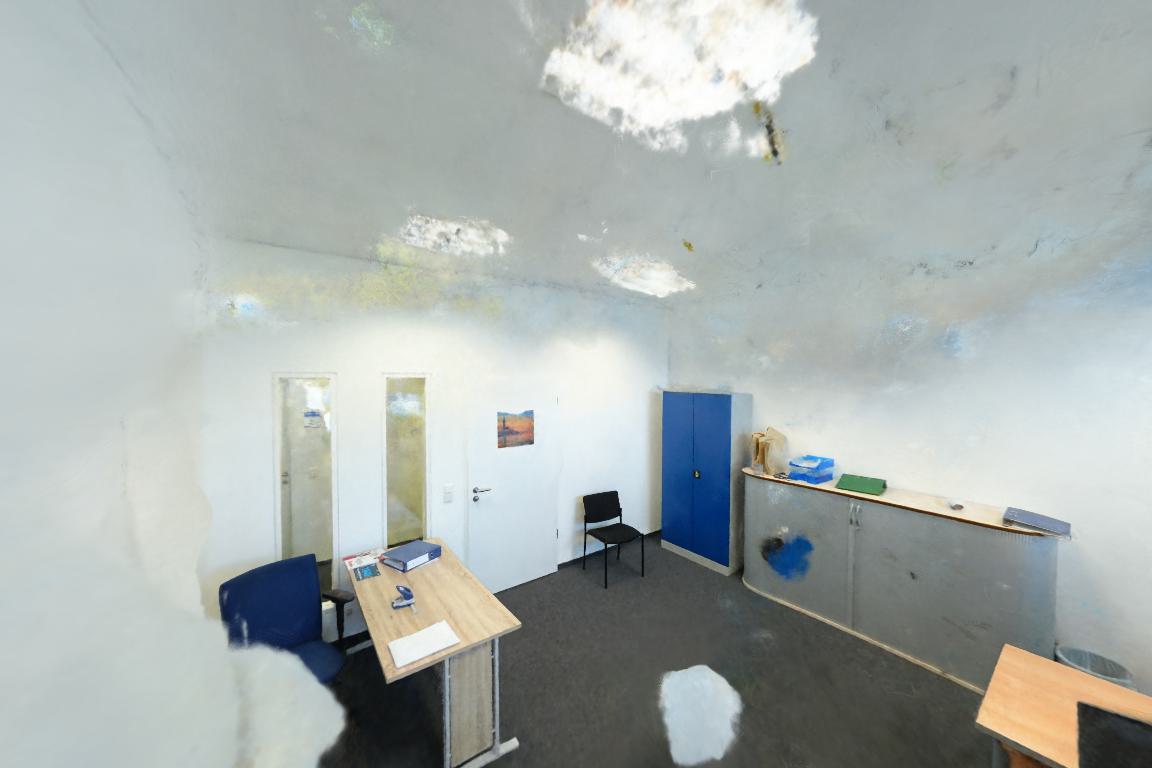}& \includegraphics[width=0.158\linewidth]{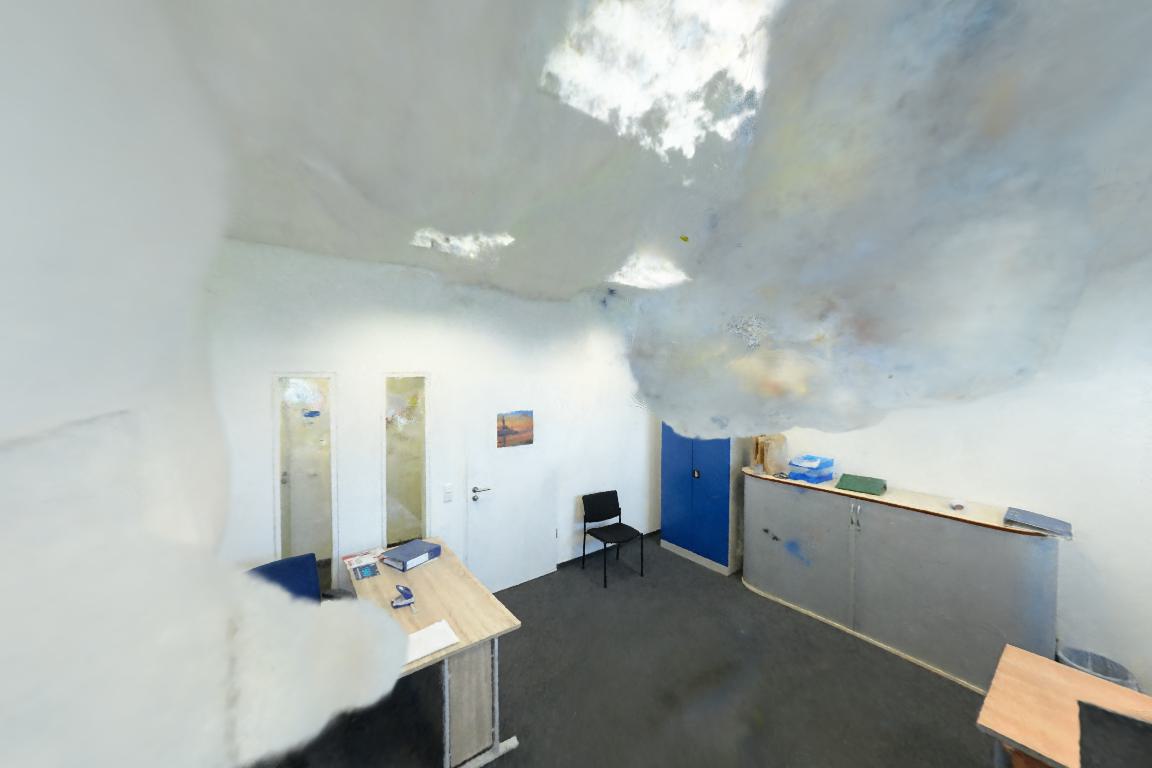}& \includegraphics[width=0.158\linewidth]{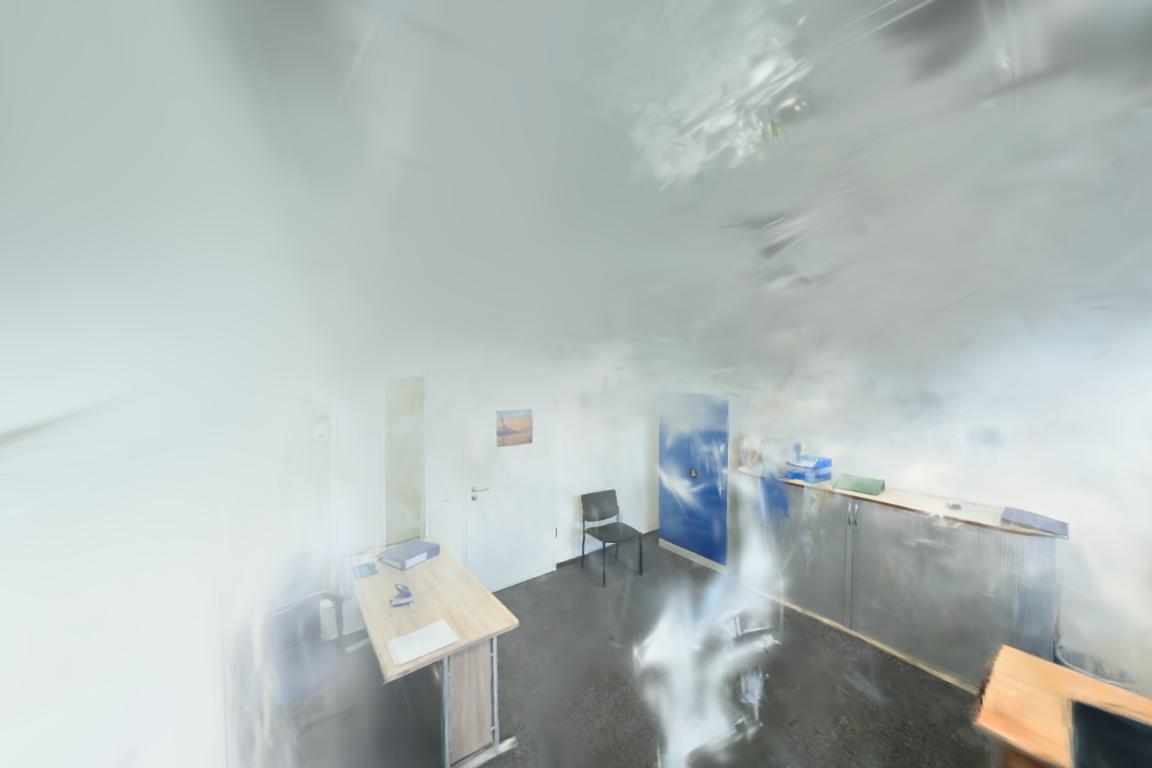}&
    \includegraphics[width=0.158\linewidth]{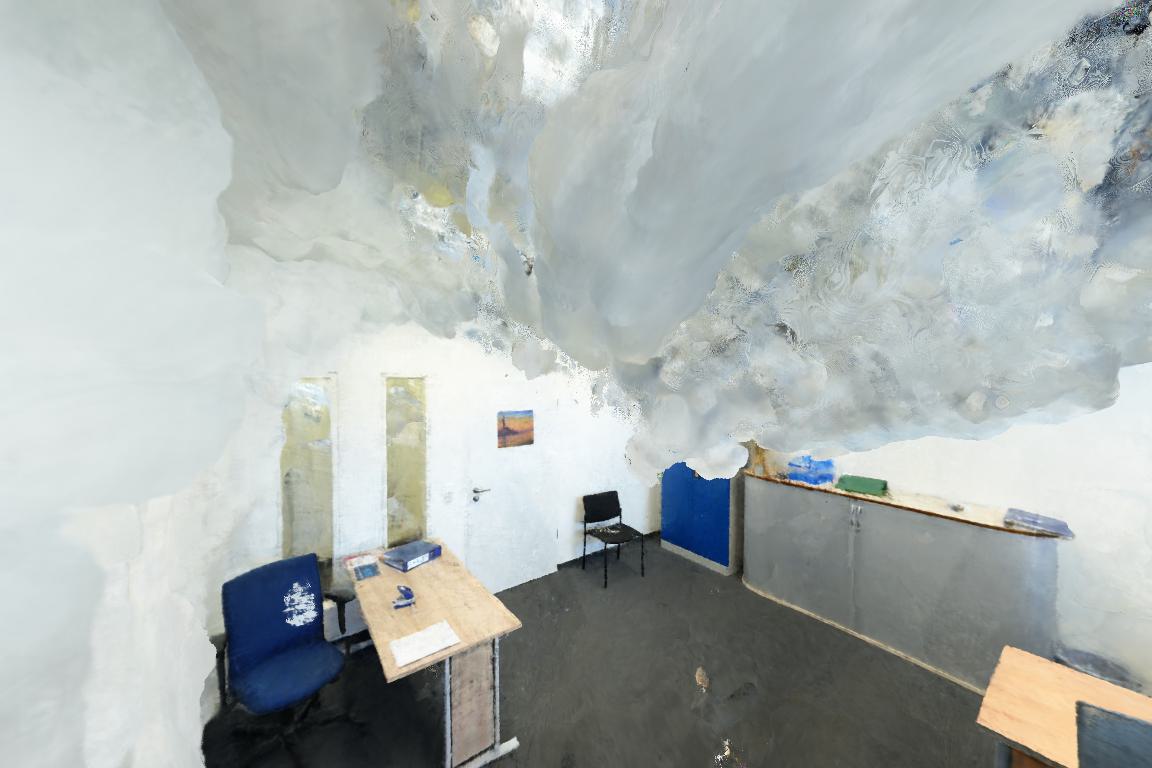}&
    \includegraphics[width=0.158\linewidth]{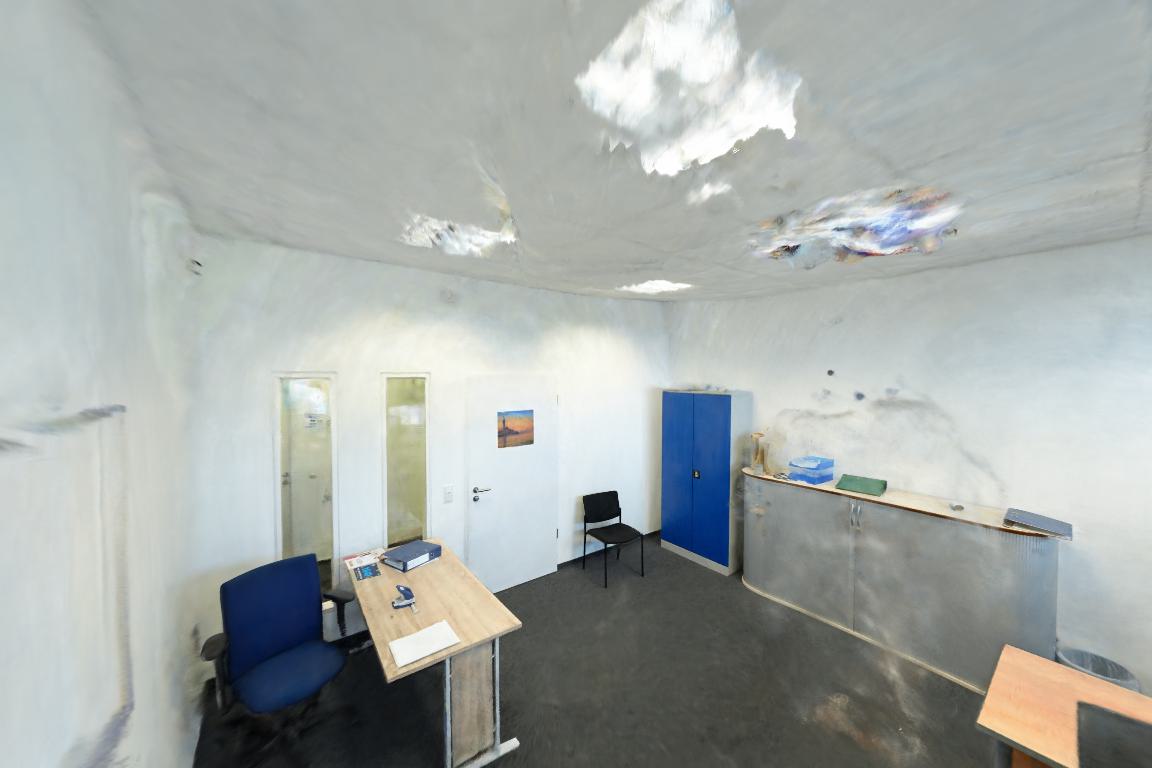}\\

    % % ef69d58016
    % \includegraphics[width=0.158\linewidth]{suppl_figures/scannetpp/ef69d58016/00009/gt.jpg}& 
    % \includegraphics[width=0.158\linewidth]{suppl_figures/scannetpp/ef69d58016/00009/nerfacto.jpg}& \includegraphics[width=0.158\linewidth]{suppl_figures/scannetpp/ef69d58016/00009/depth-nerfacto.jpg}& \includegraphics[width=0.158\linewidth]{suppl_figures/scannetpp/ef69d58016/00009/3dgs.jpg}&
    % \includegraphics[width=0.158\linewidth]{suppl_figures/scannetpp/ef69d58016/00009/nelf-pro.jpg}&
    % \includegraphics[width=0.158\linewidth]{suppl_figures/scannetpp/ef69d58016/00009/ours_appearance.jpg}\\

    \includegraphics[width=0.158\linewidth]{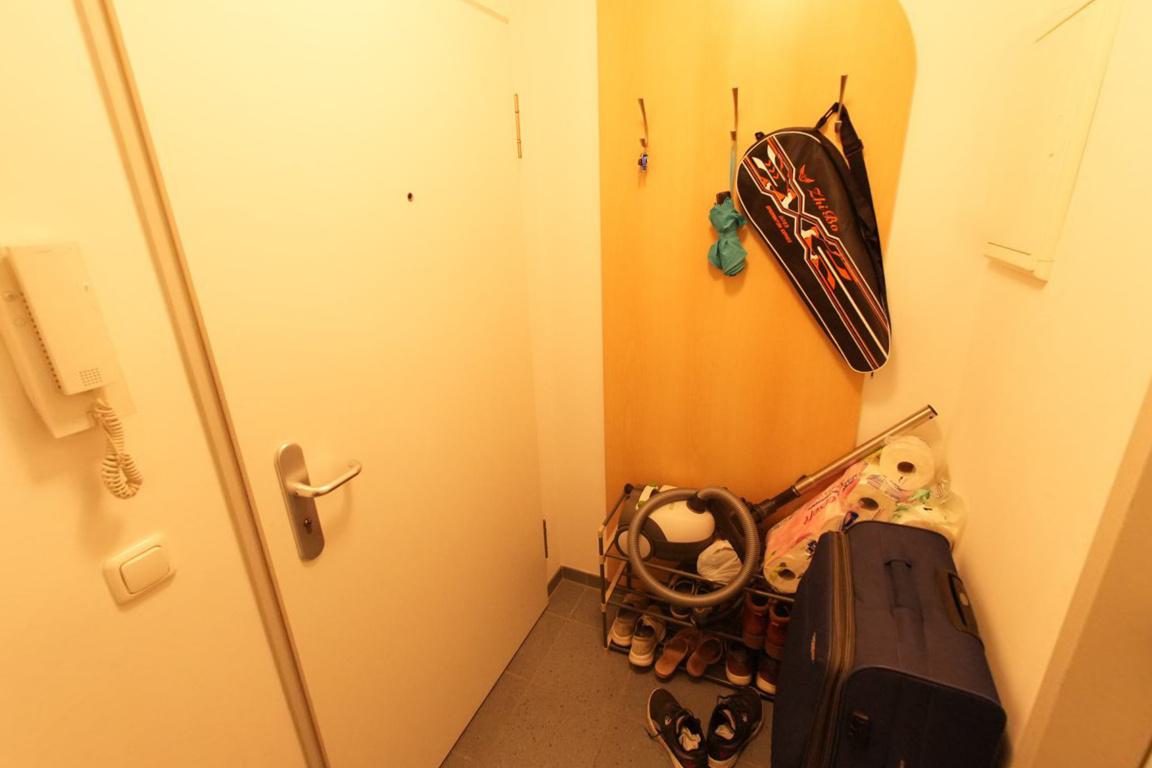}& 
    \includegraphics[width=0.158\linewidth]{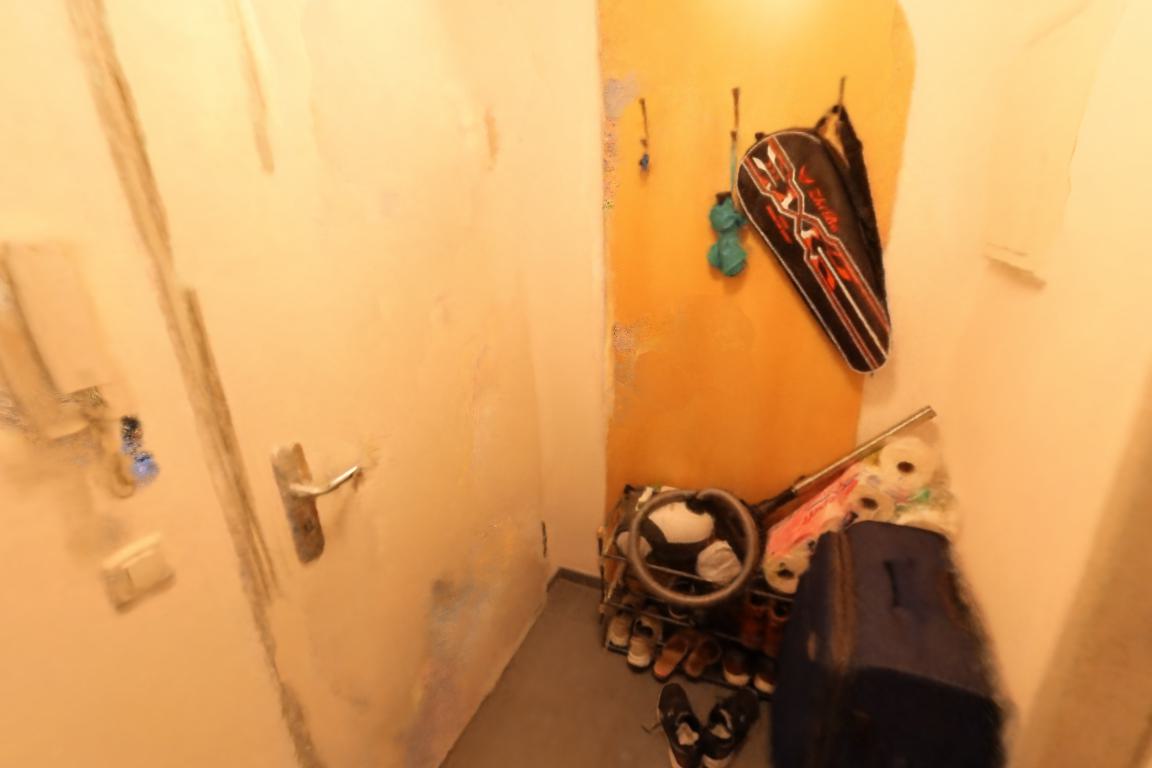}& \includegraphics[width=0.158\linewidth]{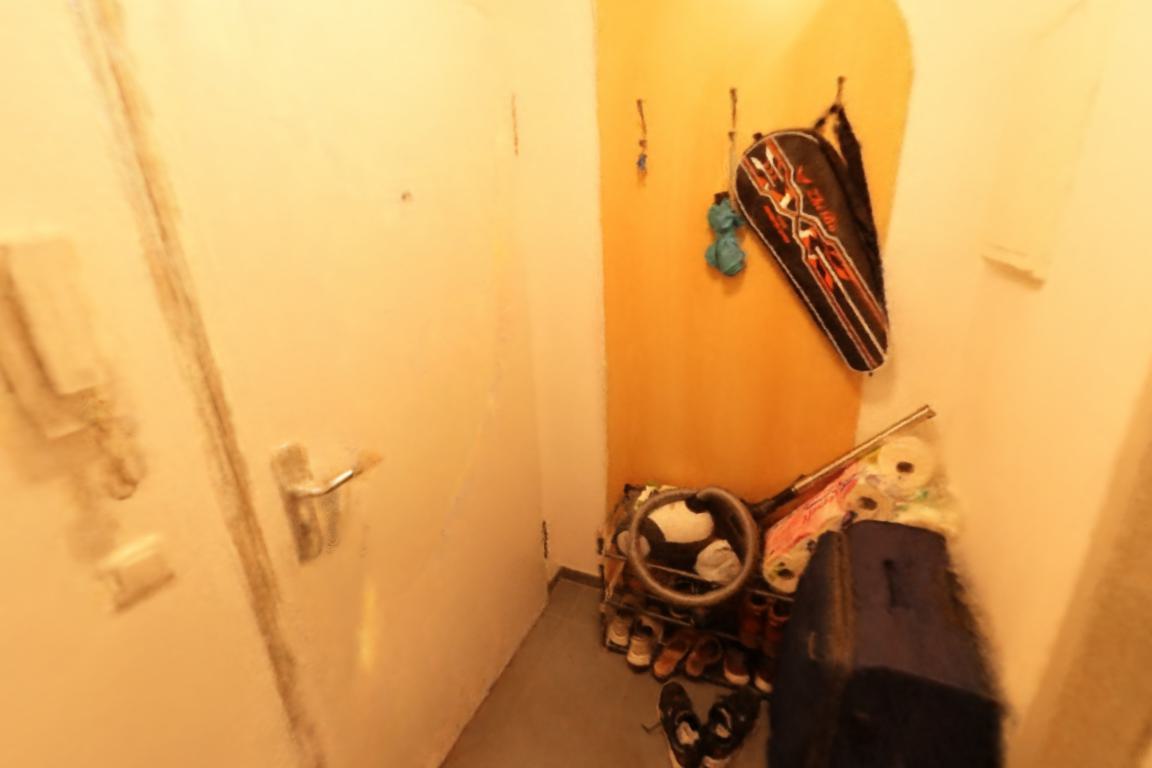}& \includegraphics[width=0.158\linewidth]{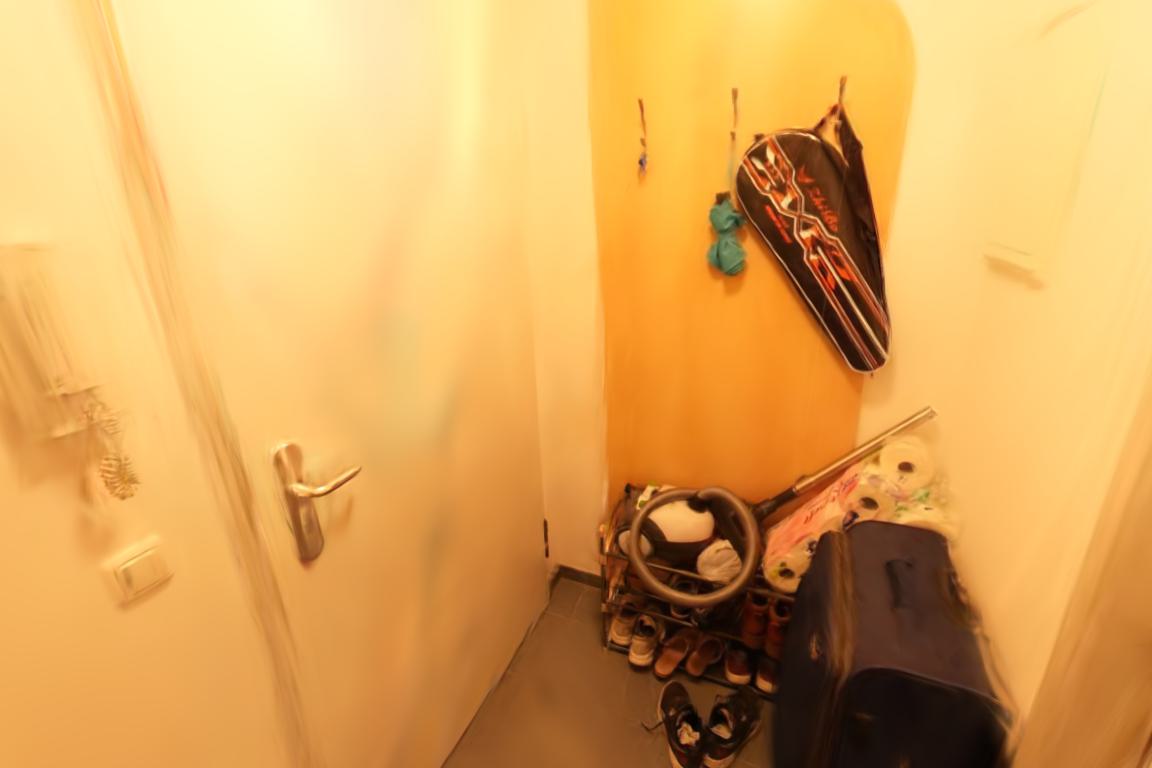}&
    \includegraphics[width=0.158\linewidth]{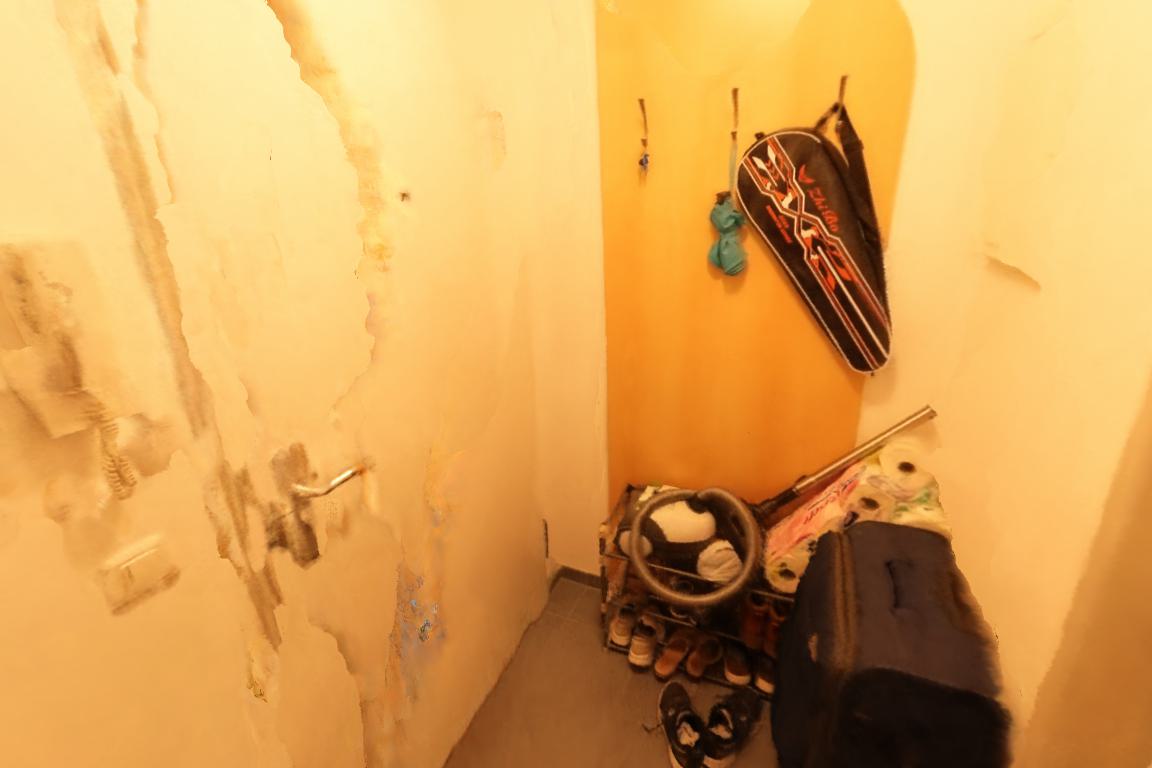}&
    \includegraphics[width=0.158\linewidth]{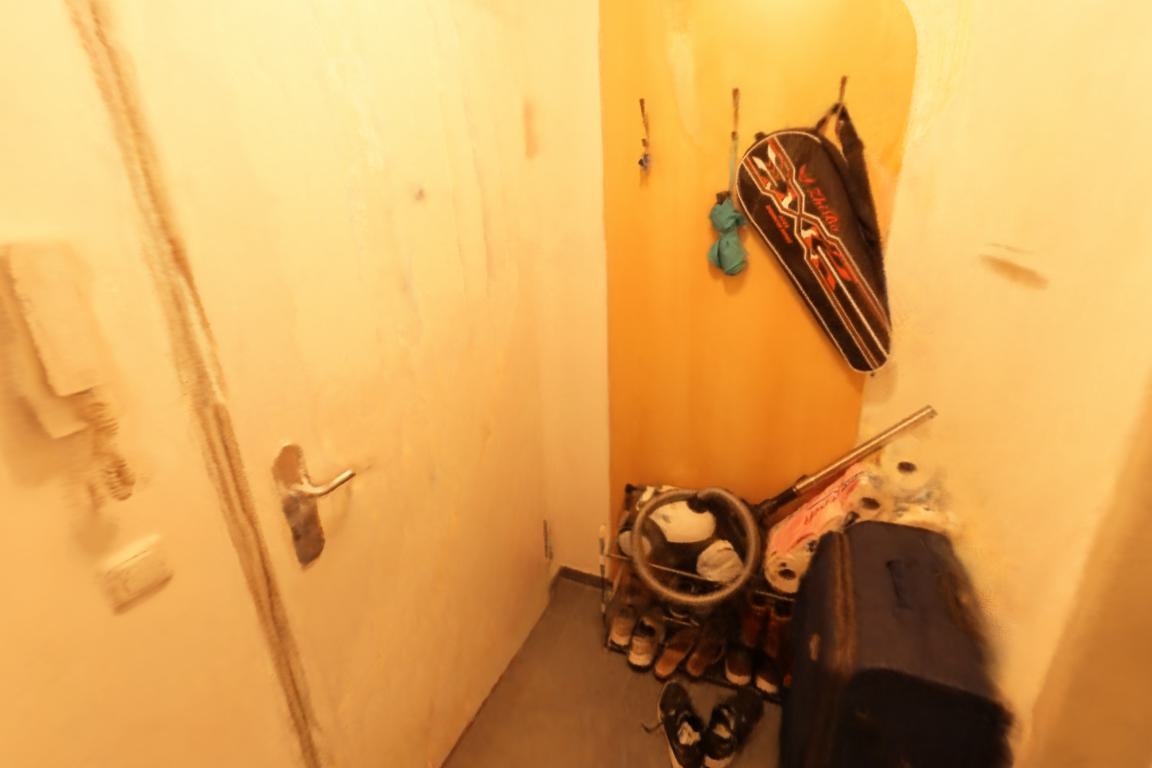}\\    
    \end{tabular}
    \caption{\textbf{Additional visualizations of novel view synthesis results on ScanNet++~\cite{scannetpp} dataset.}}
    \label{fig:suppl_qualitative_scannet++}
\end{figure*}

\begin{figure*}[p]
    \centering
    \begin{tabular}{c@{\,}c@{\,}c@{\,}c@{\,}c@{\,}c}
    \small GT& \small Mega-NeRF & \small NeLF-Pro & \small NeLF-Pro + Ours& \small LocalRF & \small LocalRF + Ours\\
    % alameda
    \includegraphics[width=0.158\linewidth]{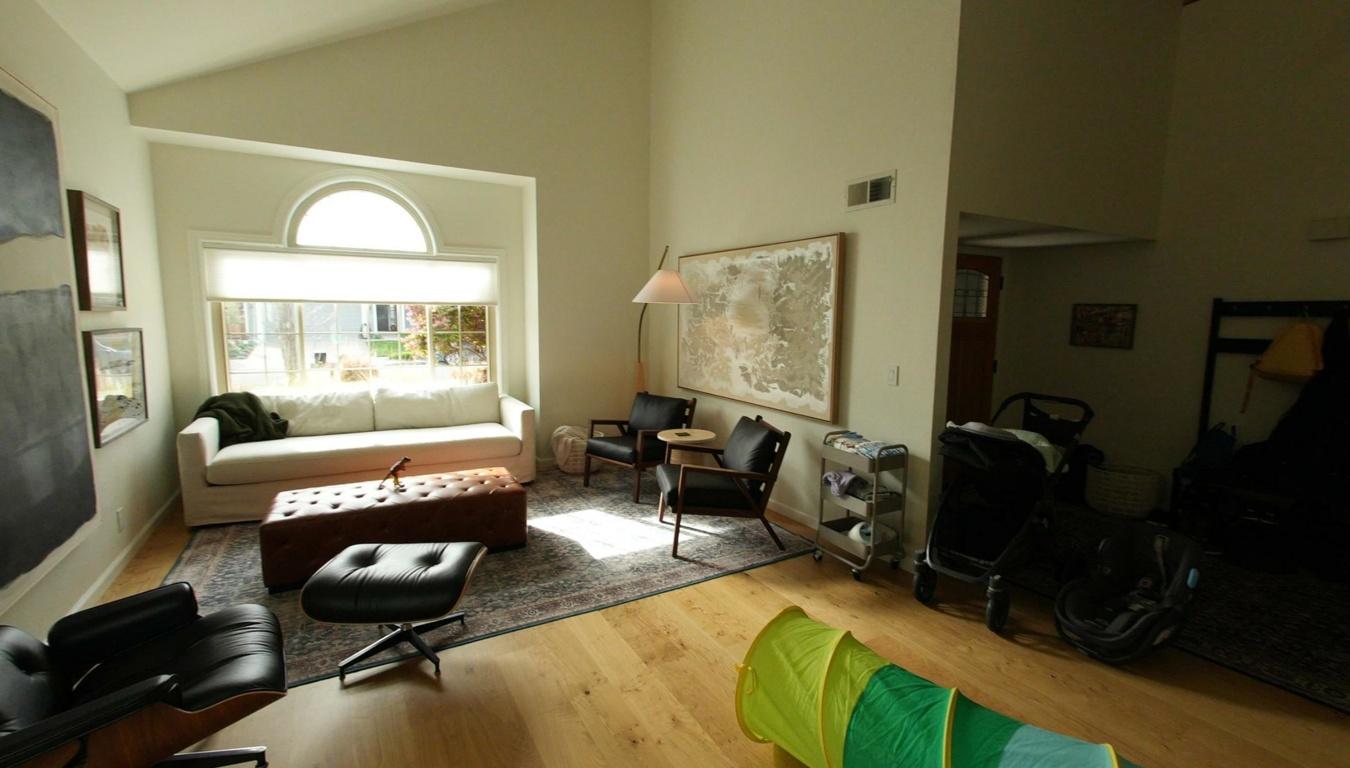}& 
    \includegraphics[width=0.158\linewidth]{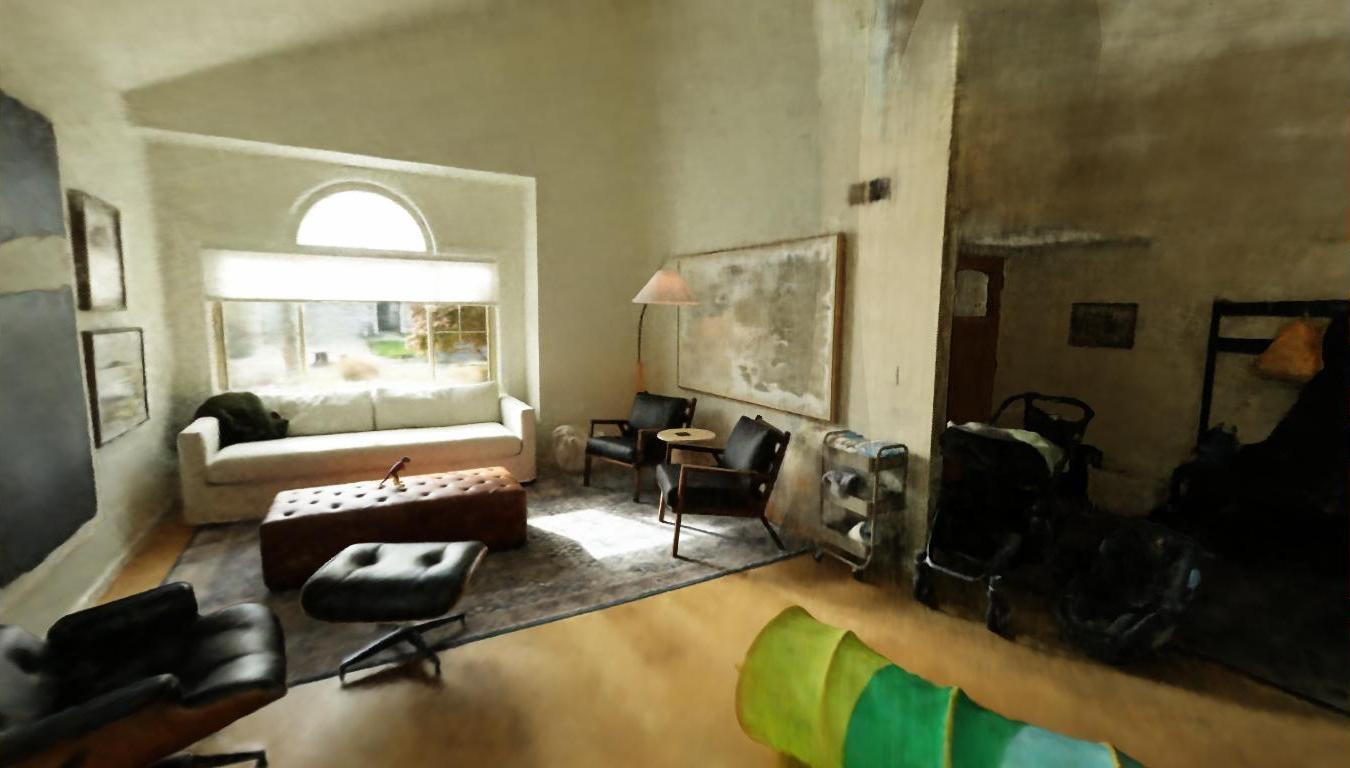}& 
    \includegraphics[width=0.158\linewidth]{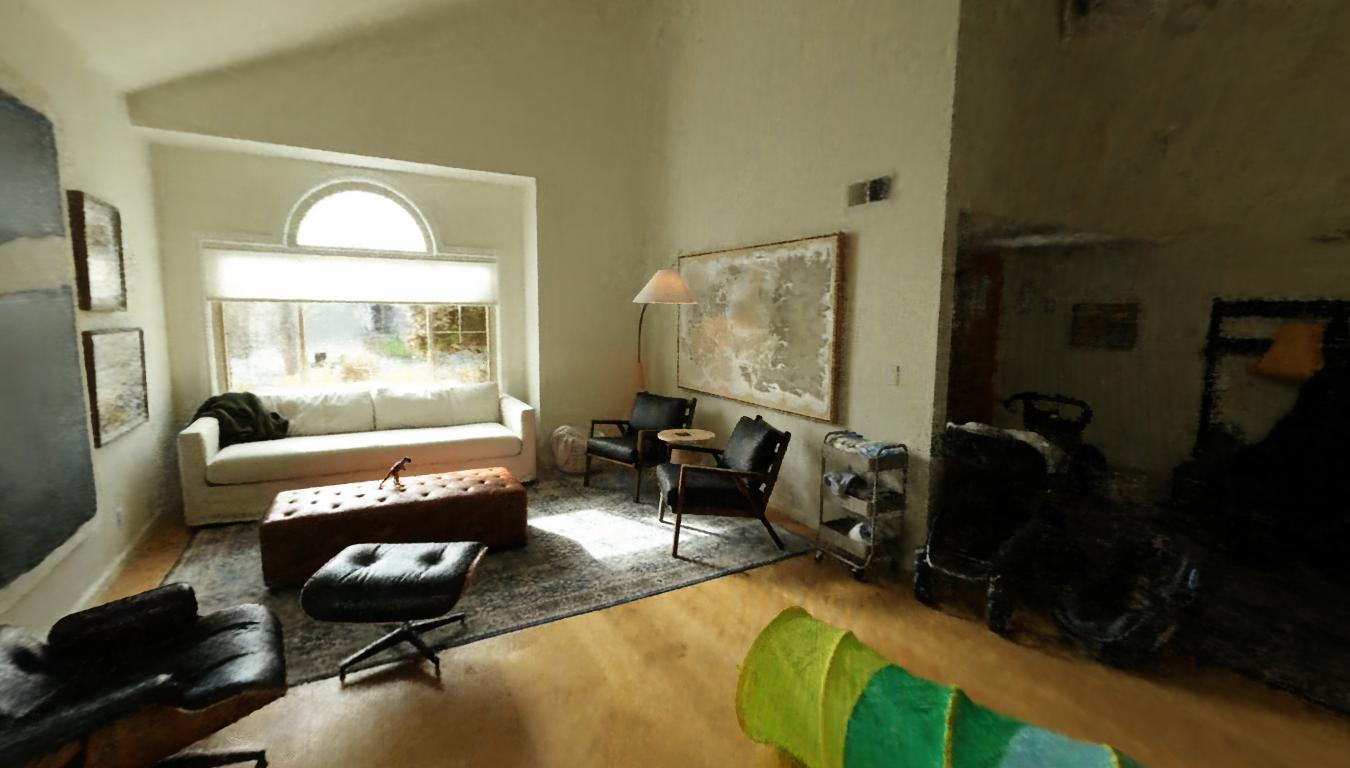}& 
    \includegraphics[width=0.158\linewidth]{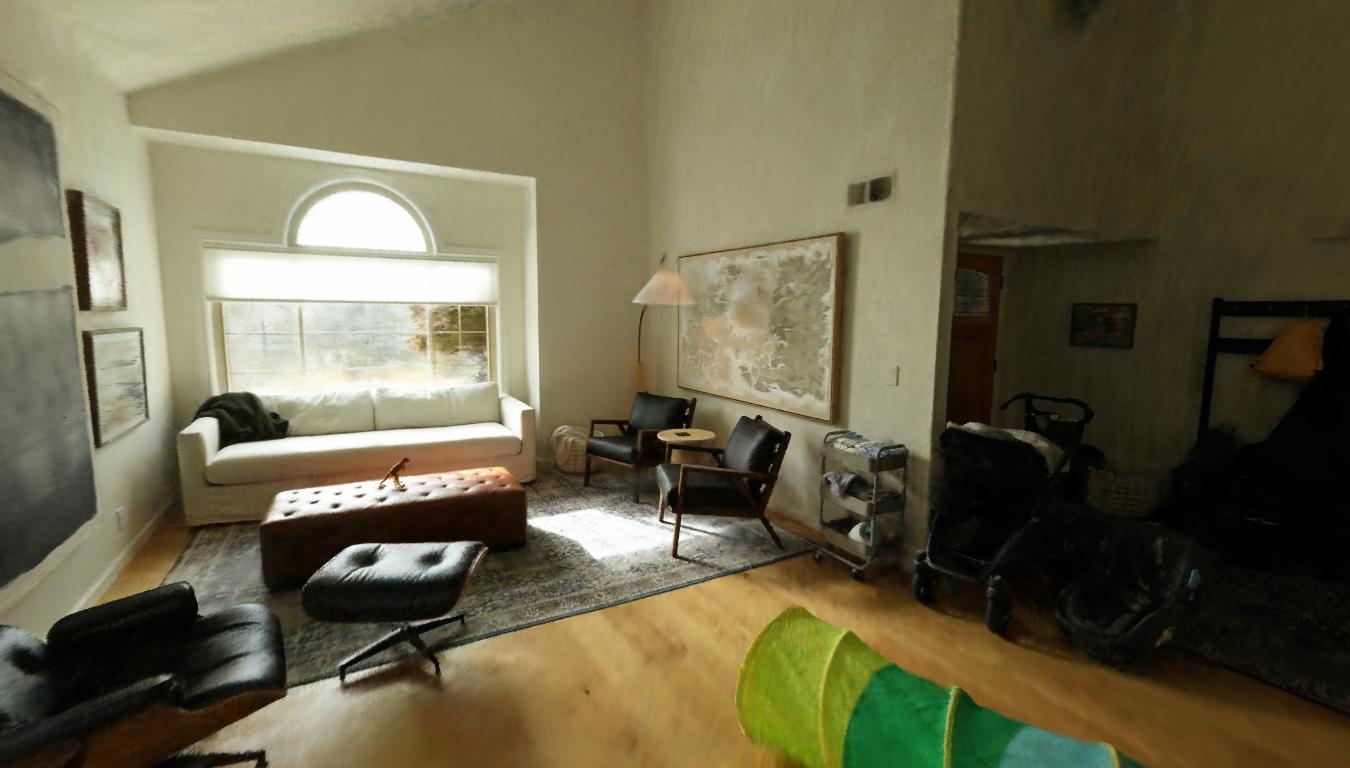}&
    \includegraphics[width=0.158\linewidth]{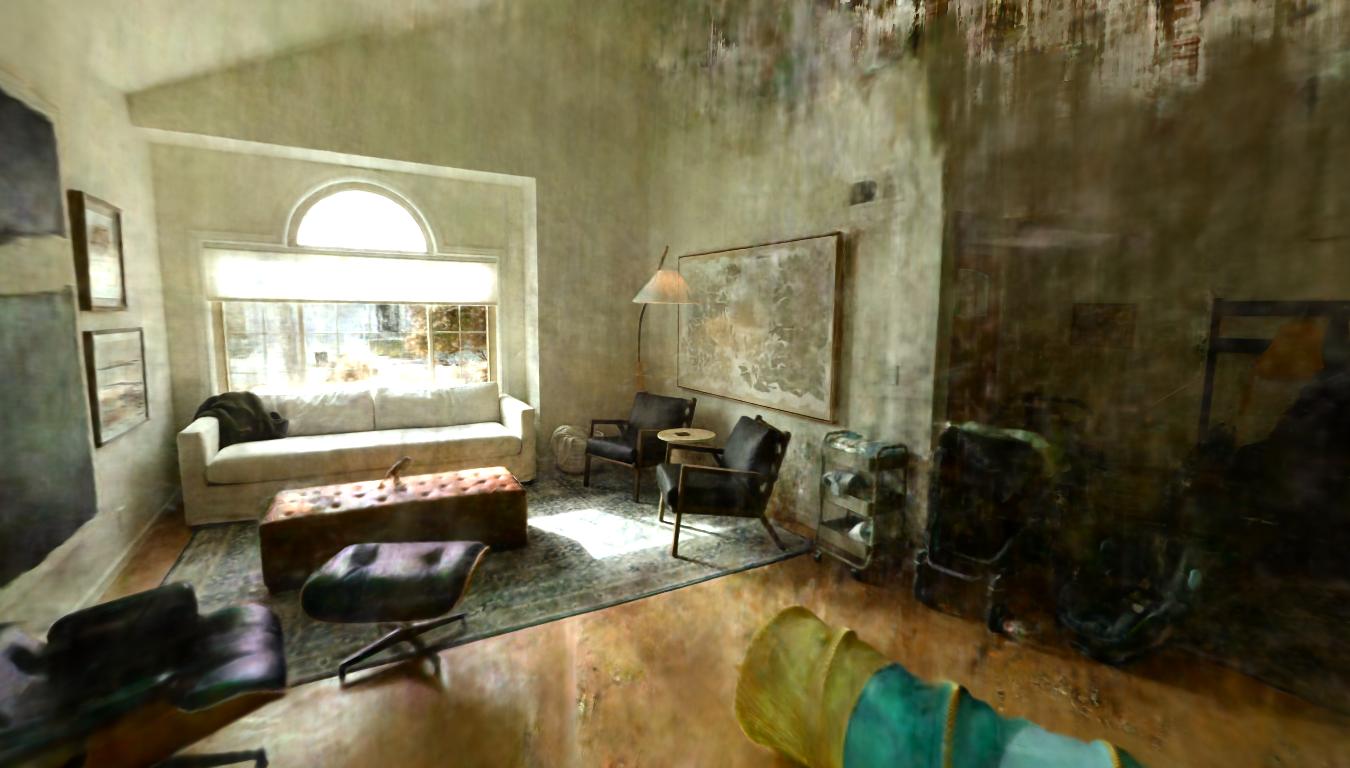}&
    \includegraphics[width=0.158\linewidth]{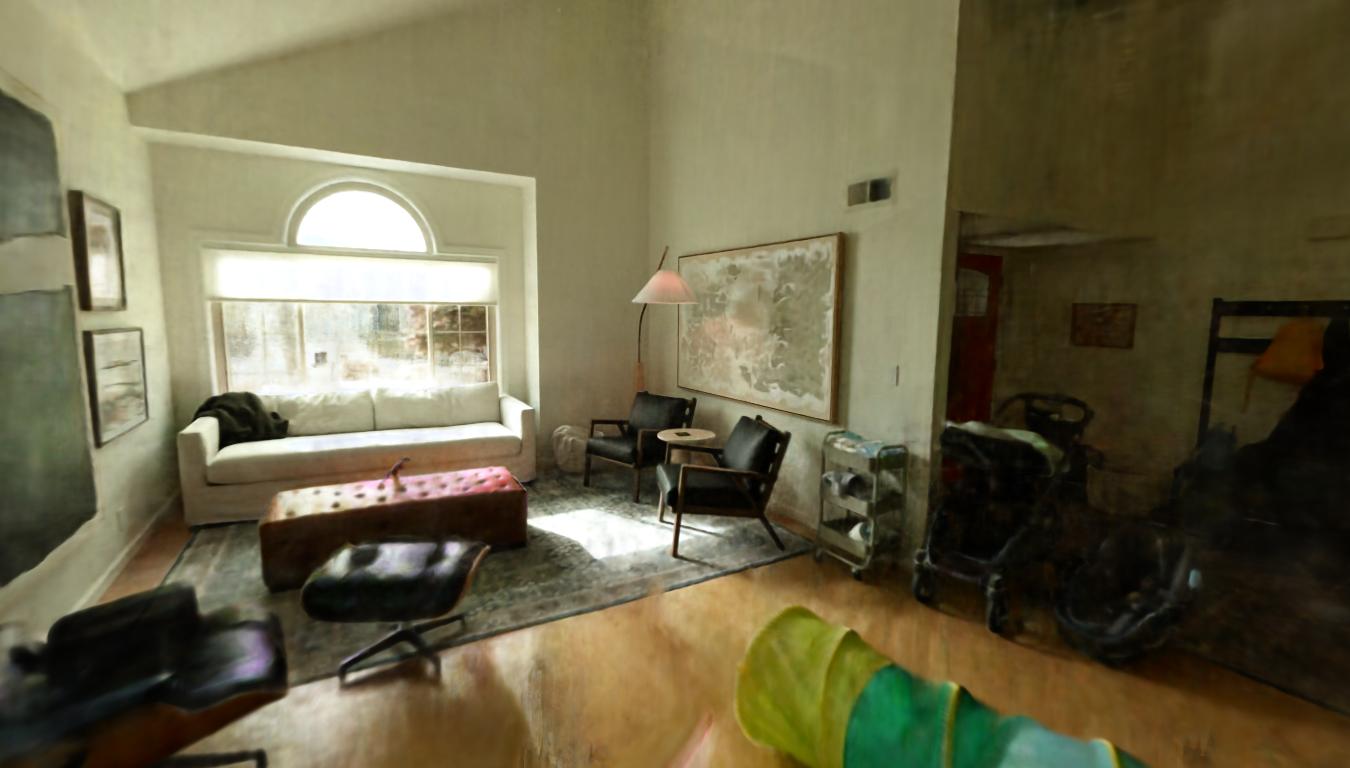}\\
    
    \includegraphics[width=0.158\linewidth]{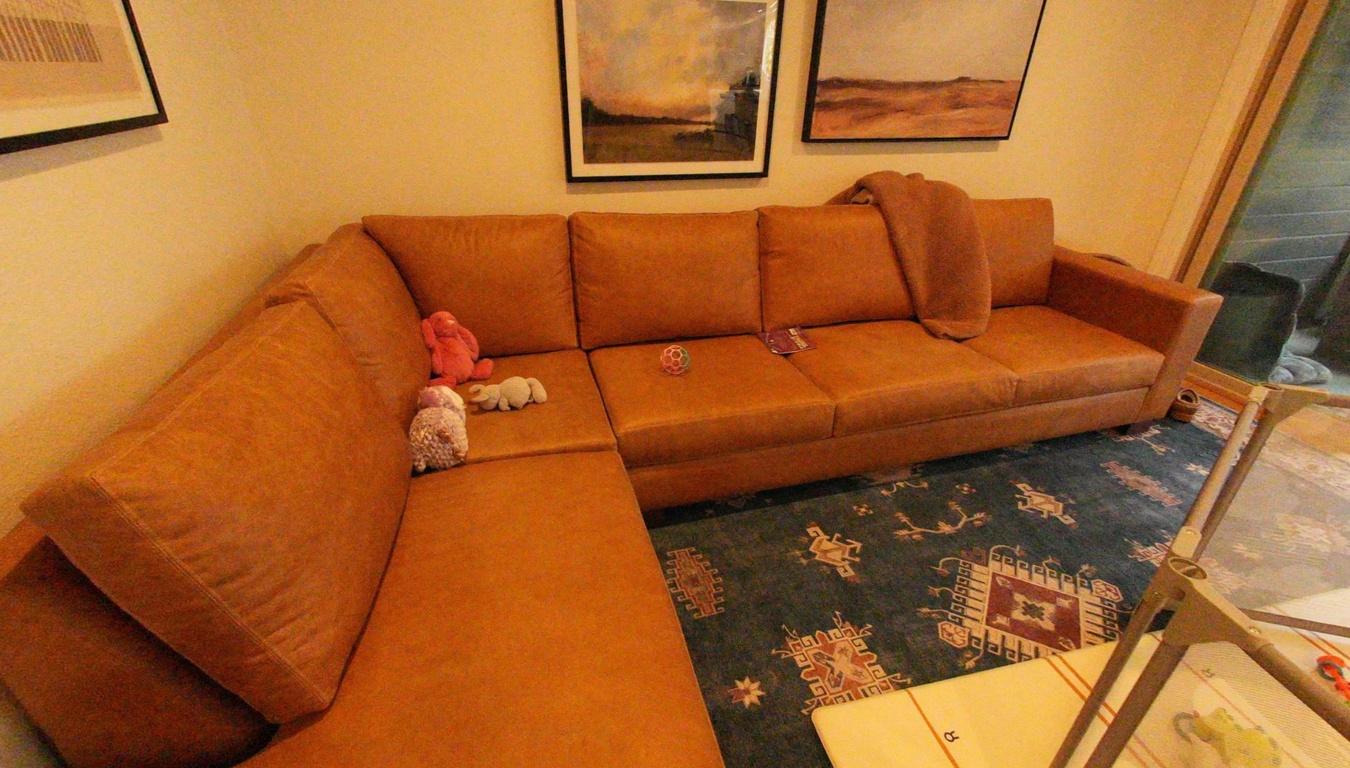}& 
    \includegraphics[width=0.158\linewidth]{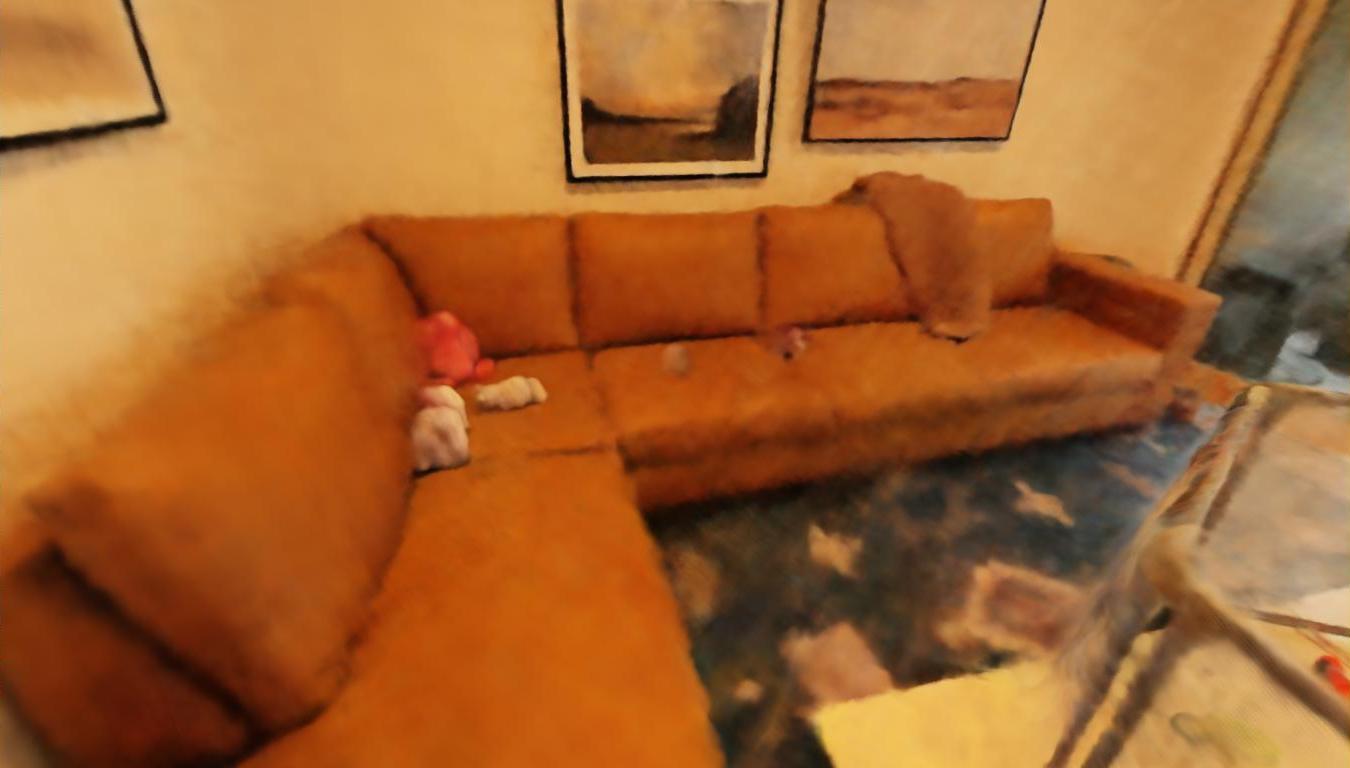}& 
    \includegraphics[width=0.158\linewidth]{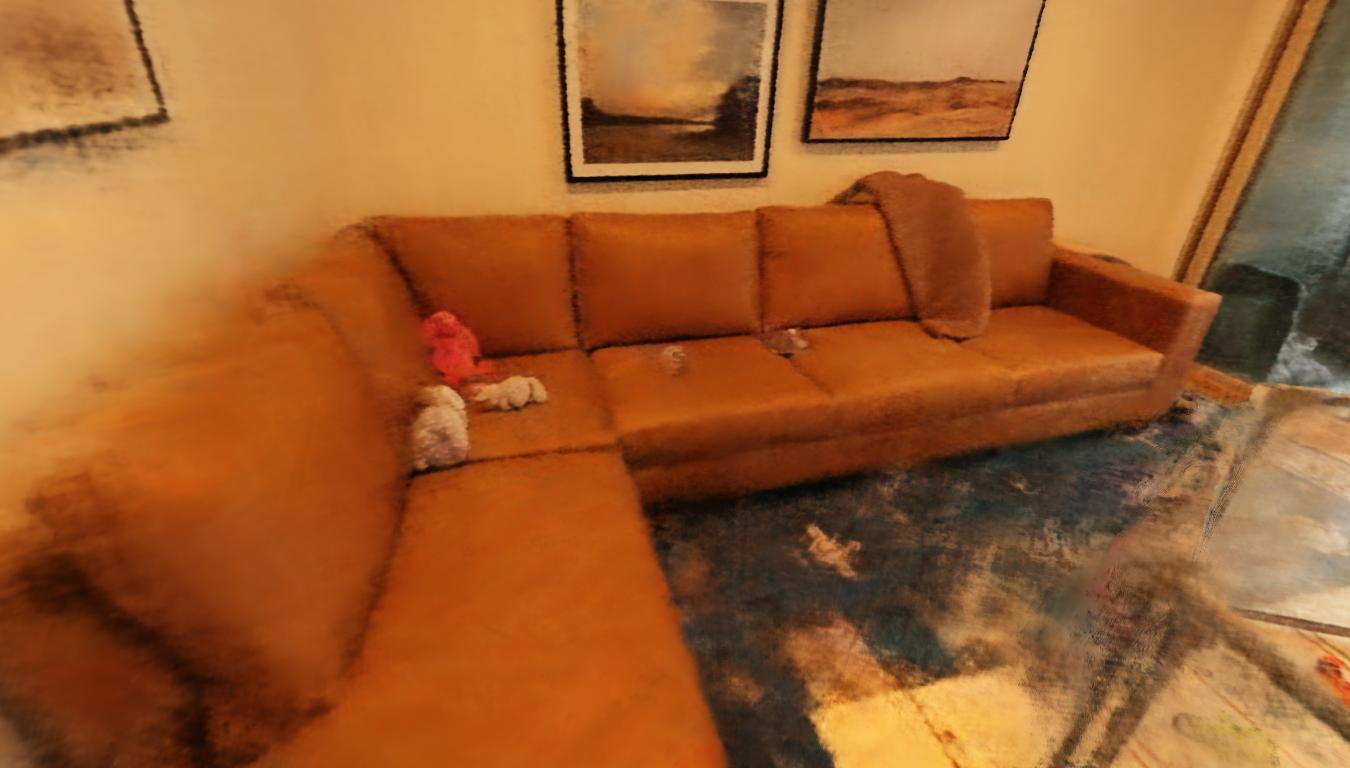}& 
    \includegraphics[width=0.158\linewidth]{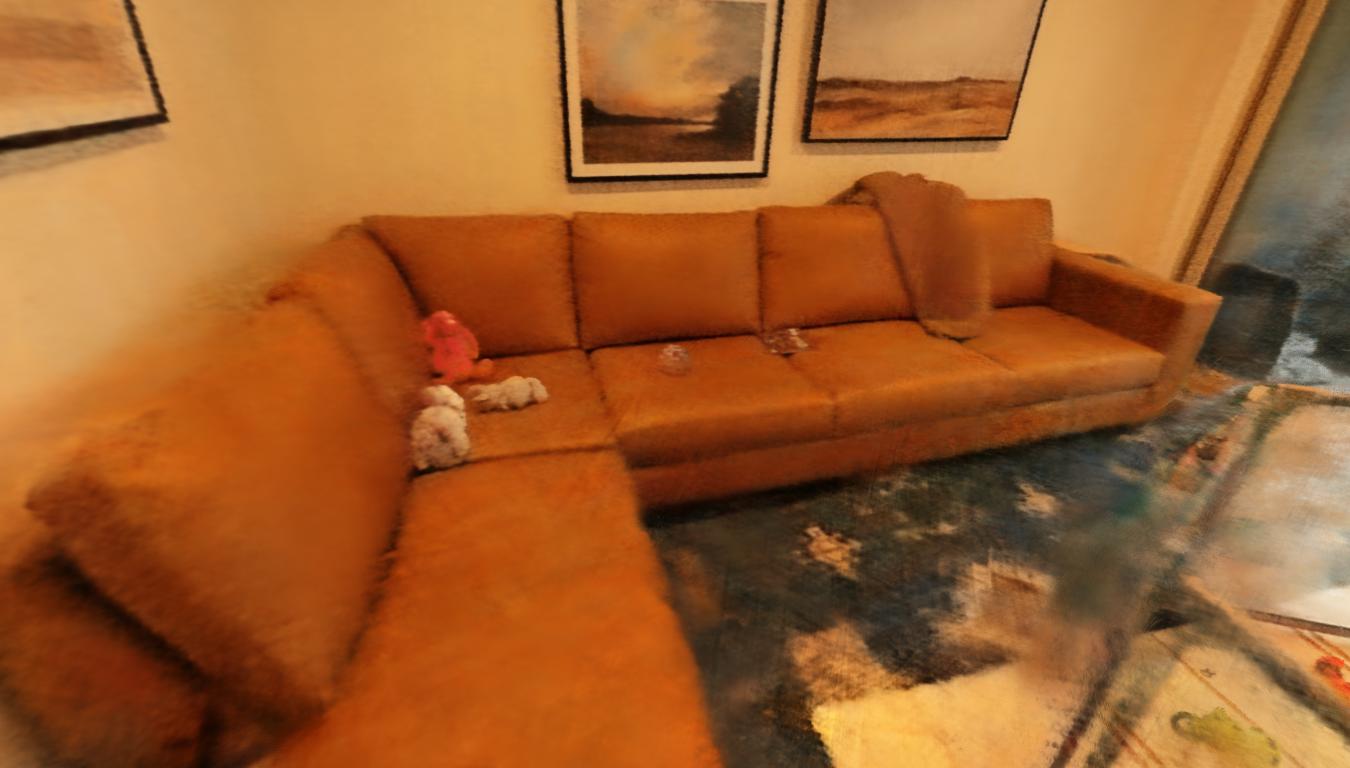}&
    \includegraphics[width=0.158\linewidth]{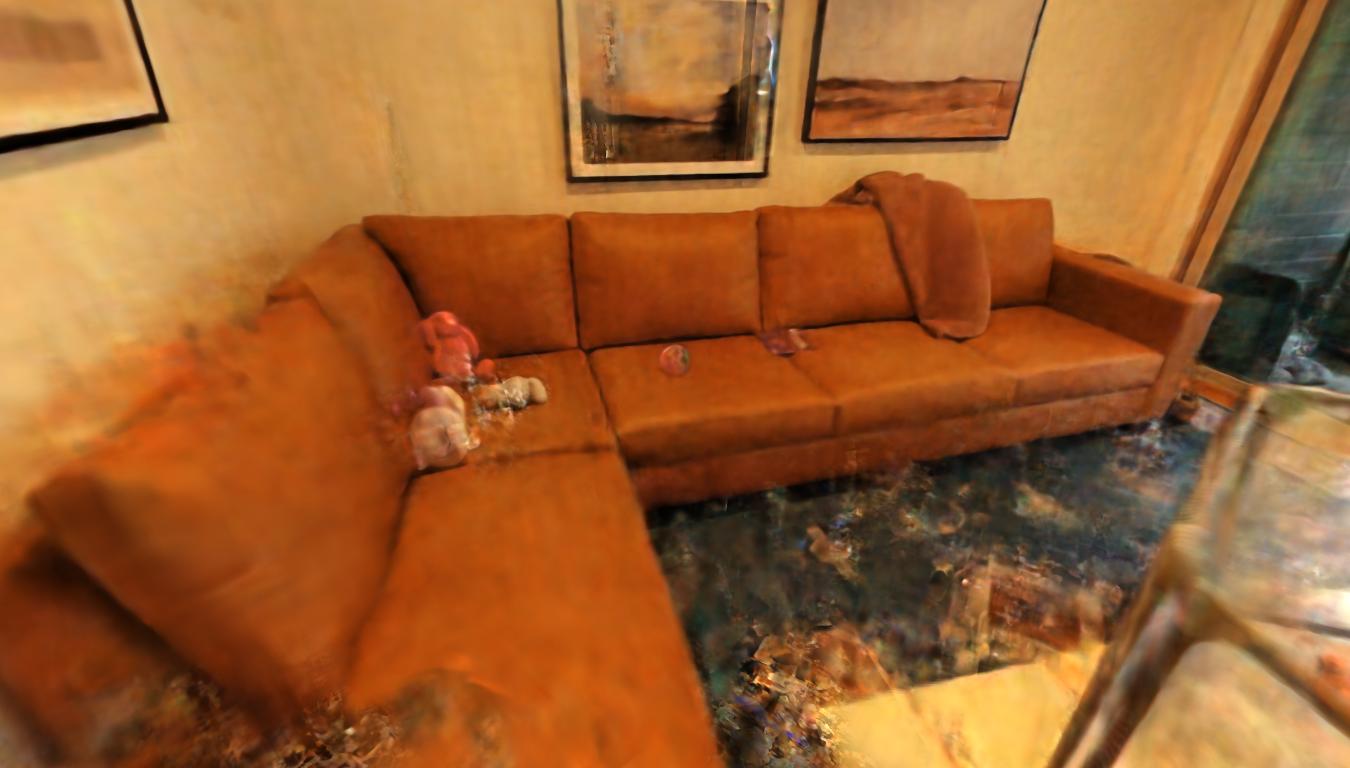}&
    \includegraphics[width=0.158\linewidth]{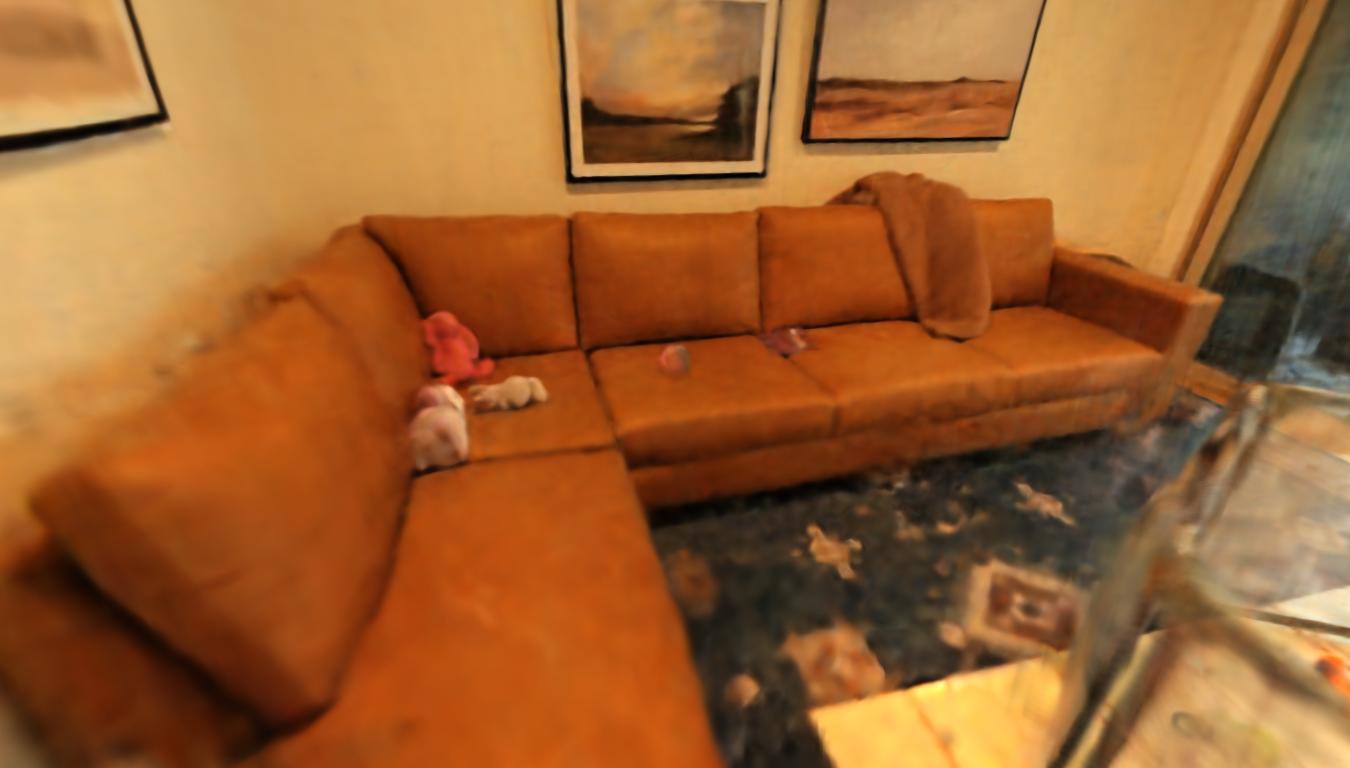}\\

    \includegraphics[width=0.158\linewidth]{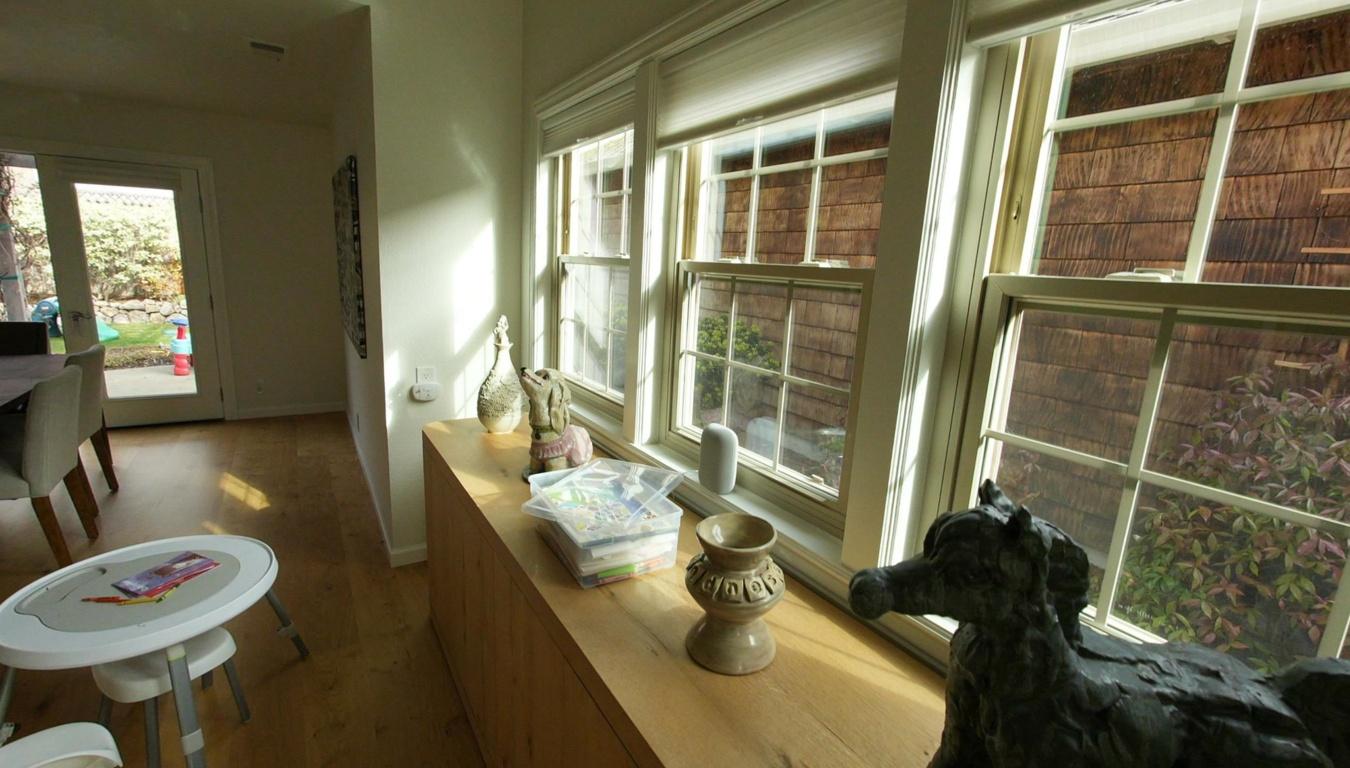}& 
    \includegraphics[width=0.158\linewidth]{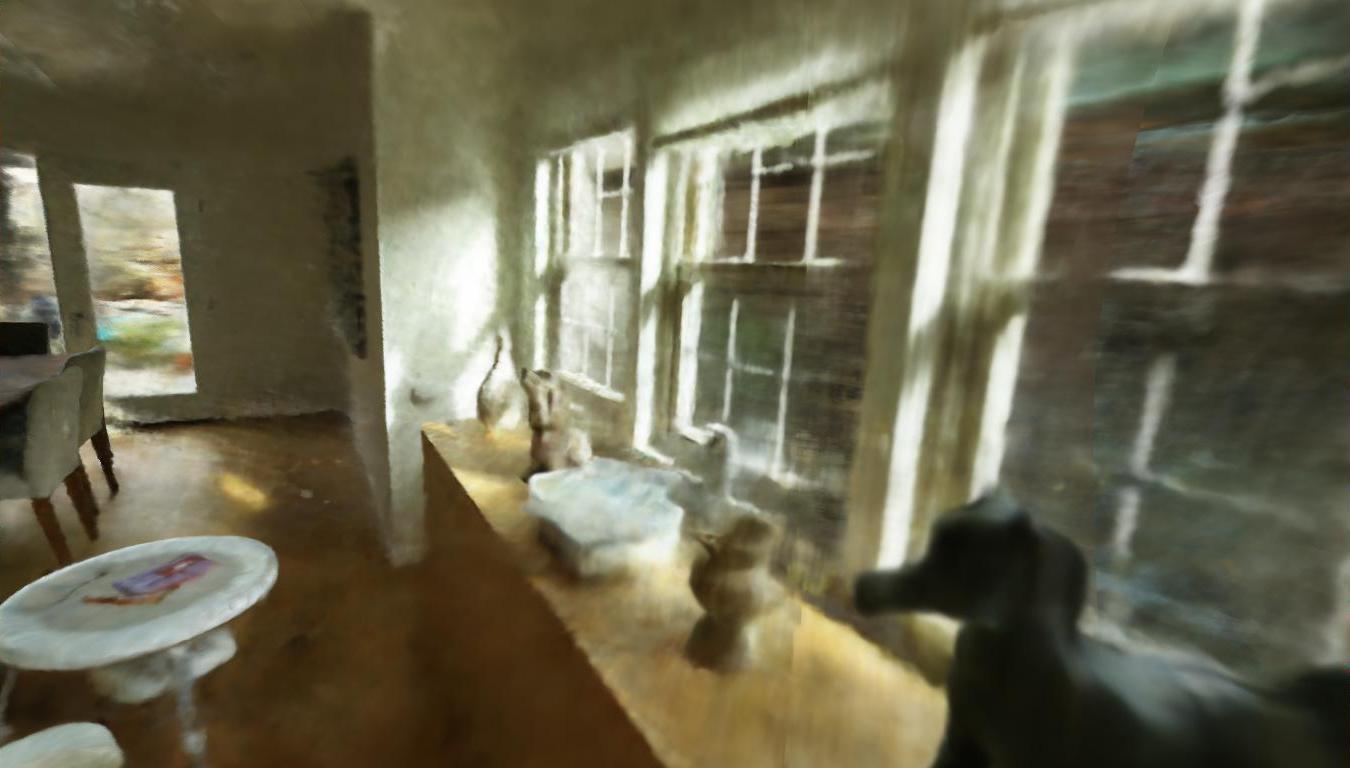}& 
    \includegraphics[width=0.158\linewidth]{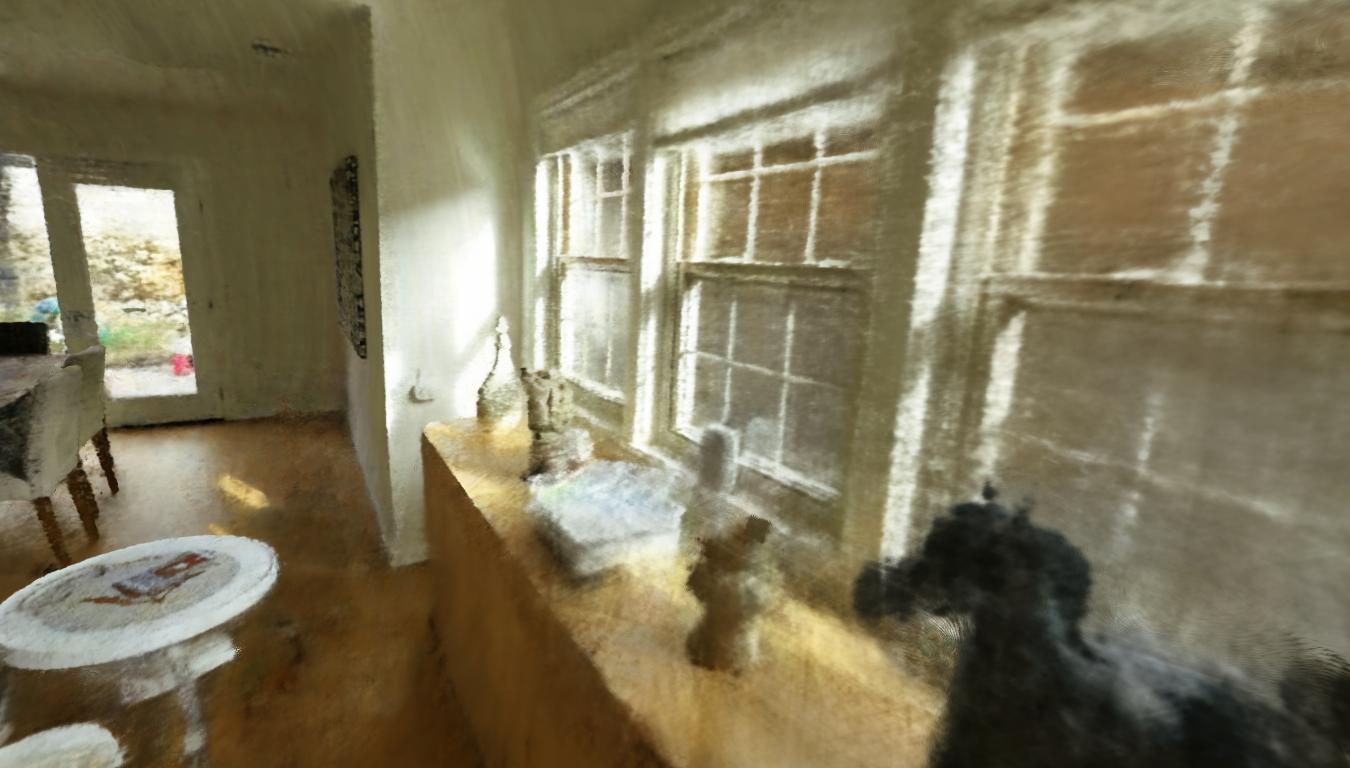}& 
    \includegraphics[width=0.158\linewidth]{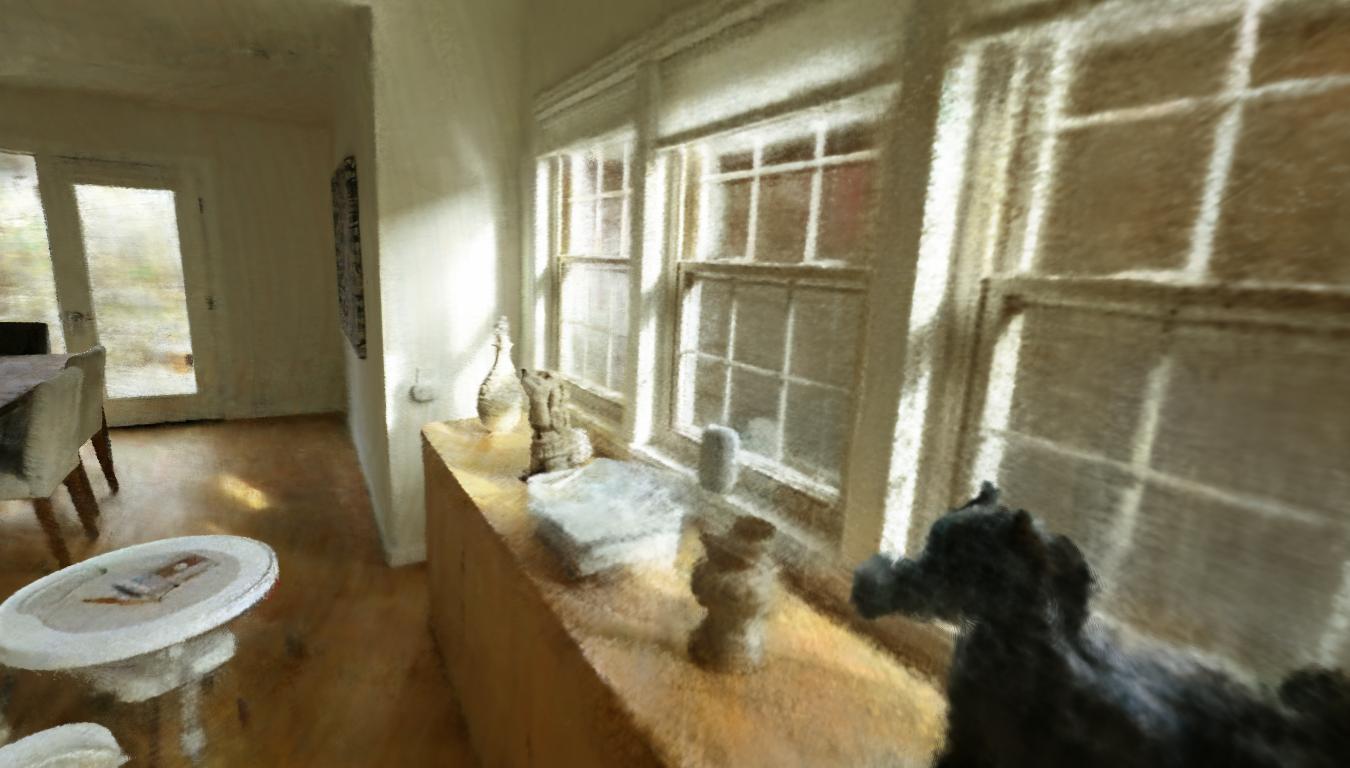}&
    \includegraphics[width=0.158\linewidth]{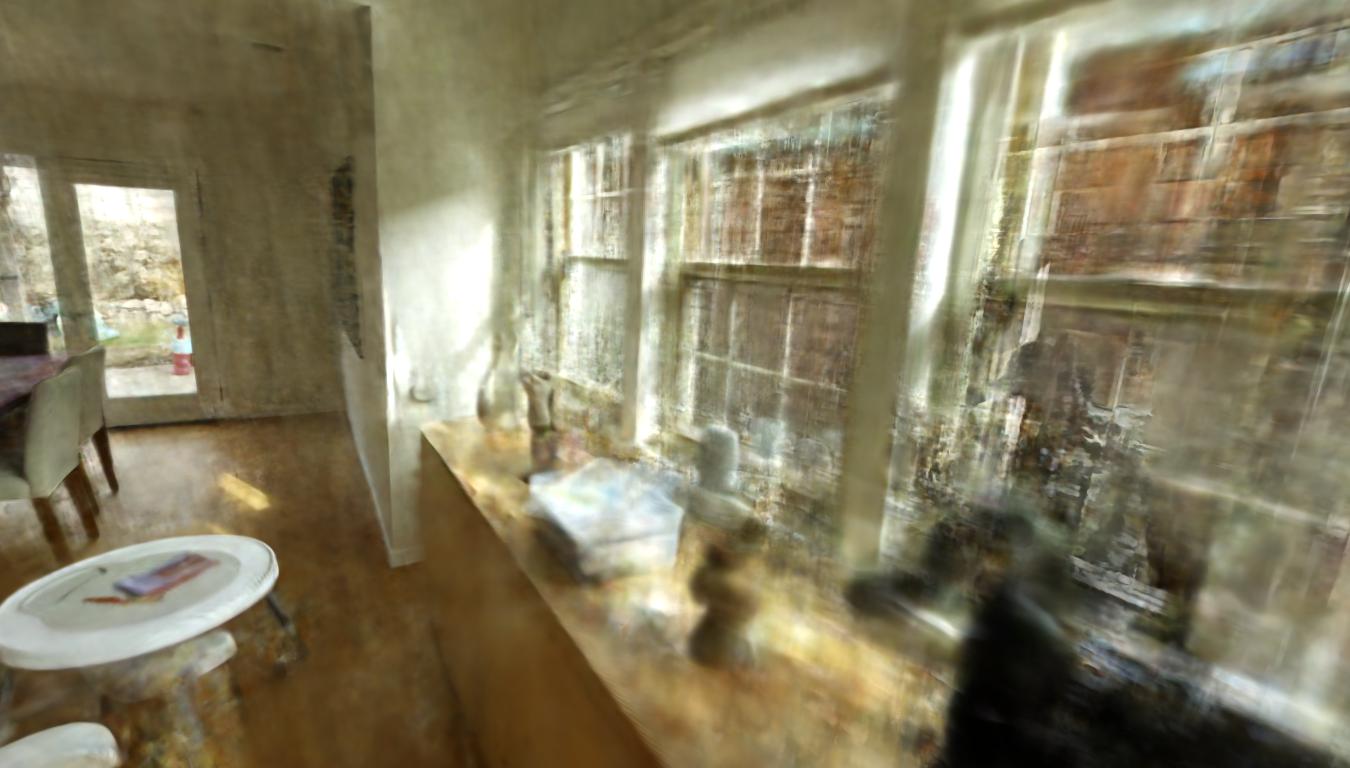}&
    \includegraphics[width=0.158\linewidth]{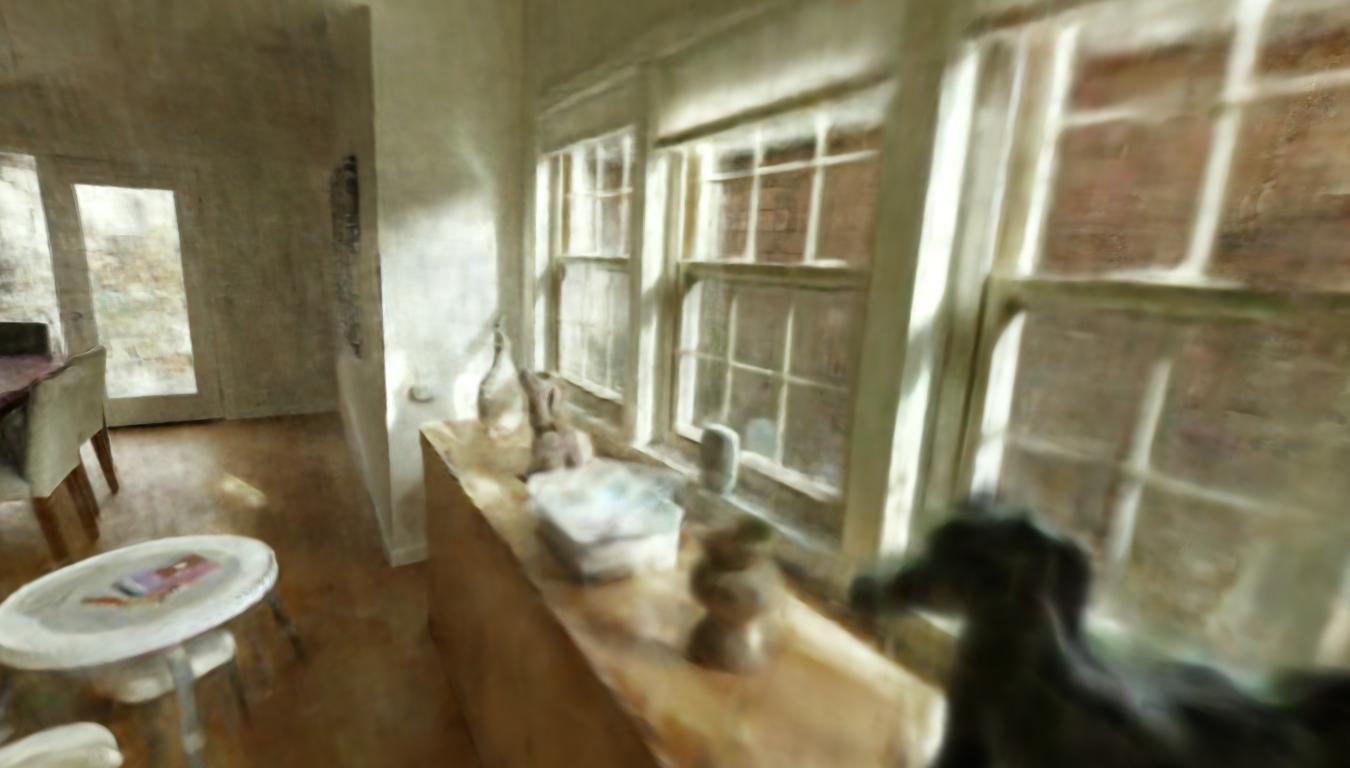}\\
    
    % berlin
    \includegraphics[width=0.158\linewidth]{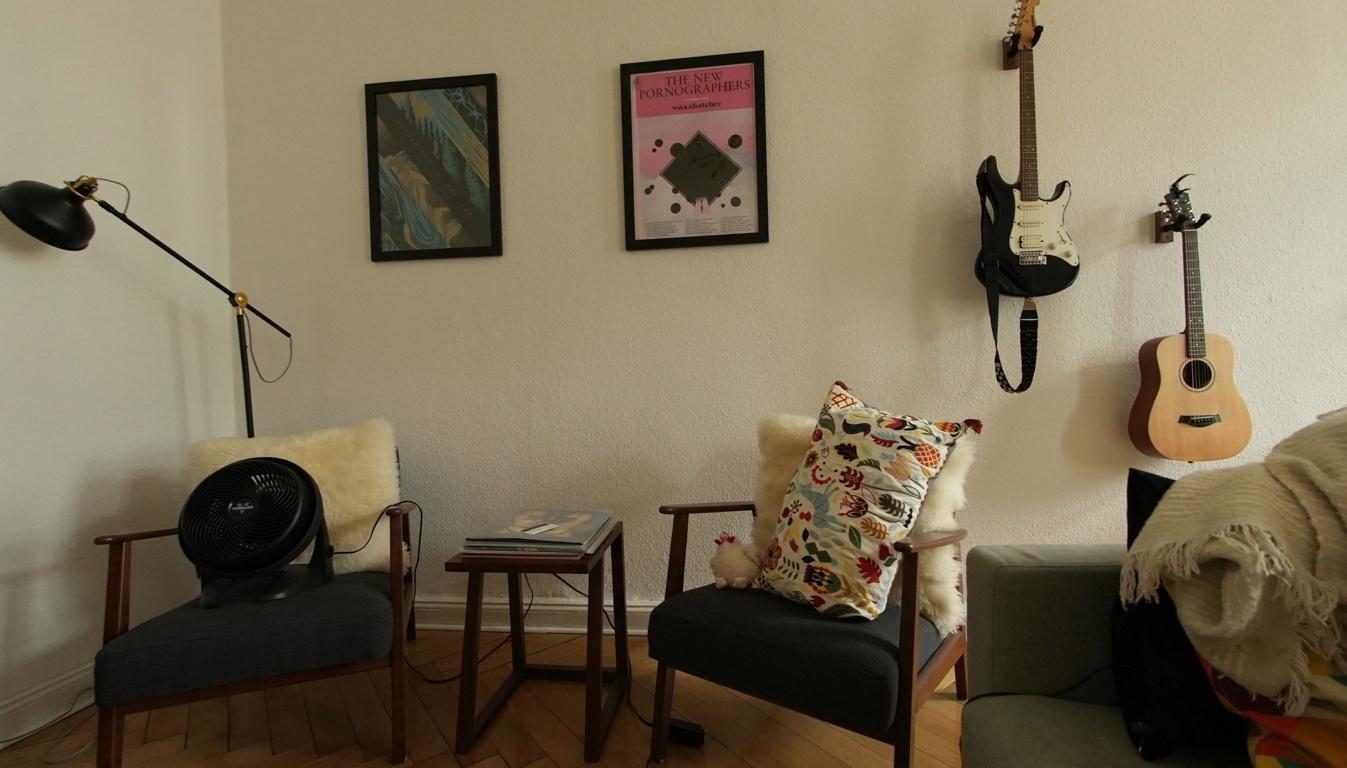}& 
    \includegraphics[width=0.158\linewidth]{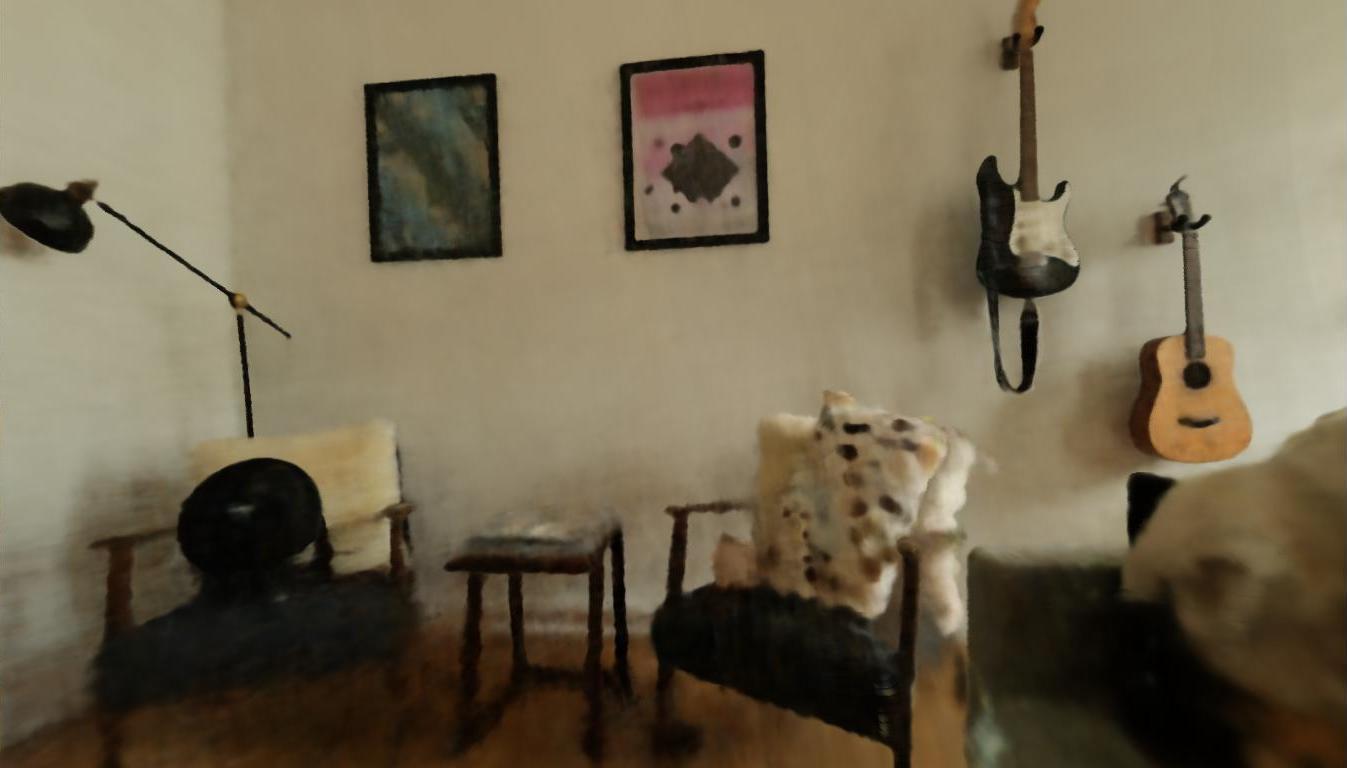}& 
    \includegraphics[width=0.158\linewidth]{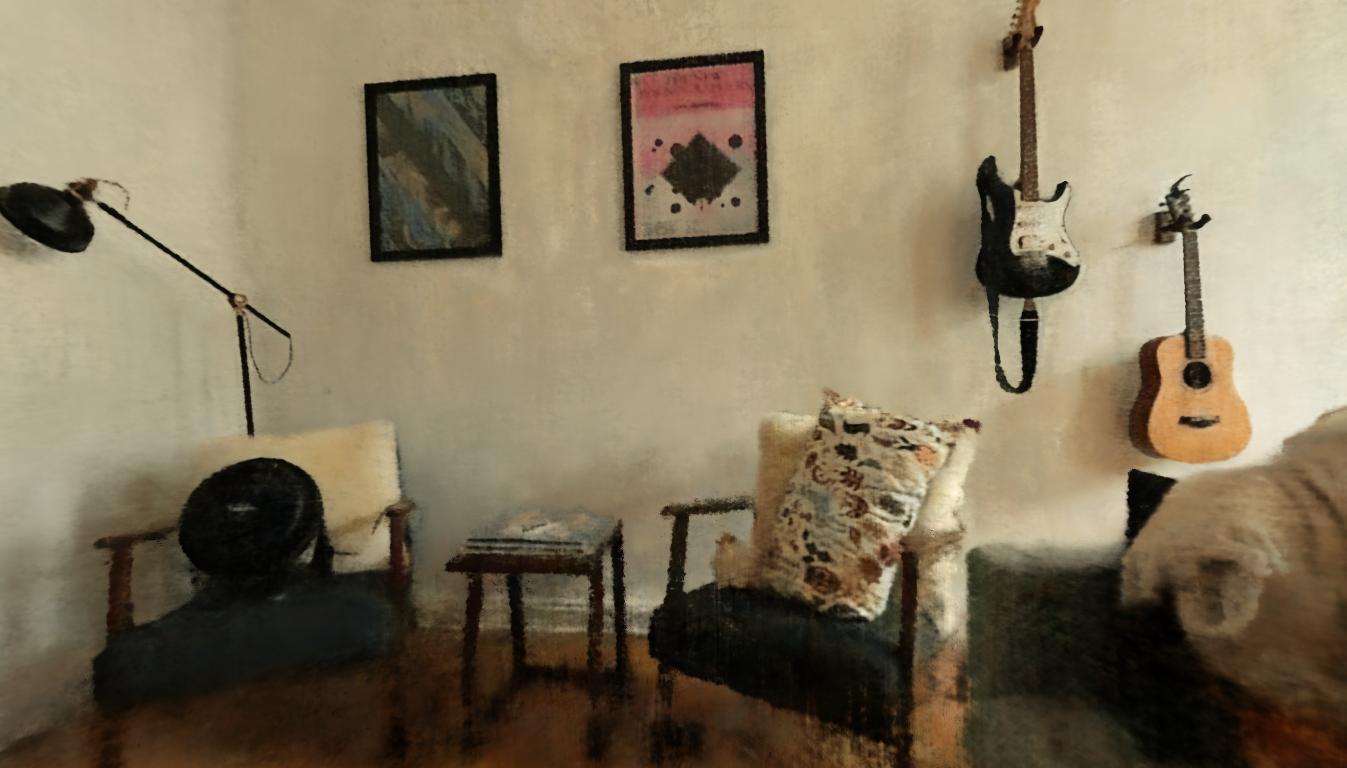}& 
    \includegraphics[width=0.158\linewidth]{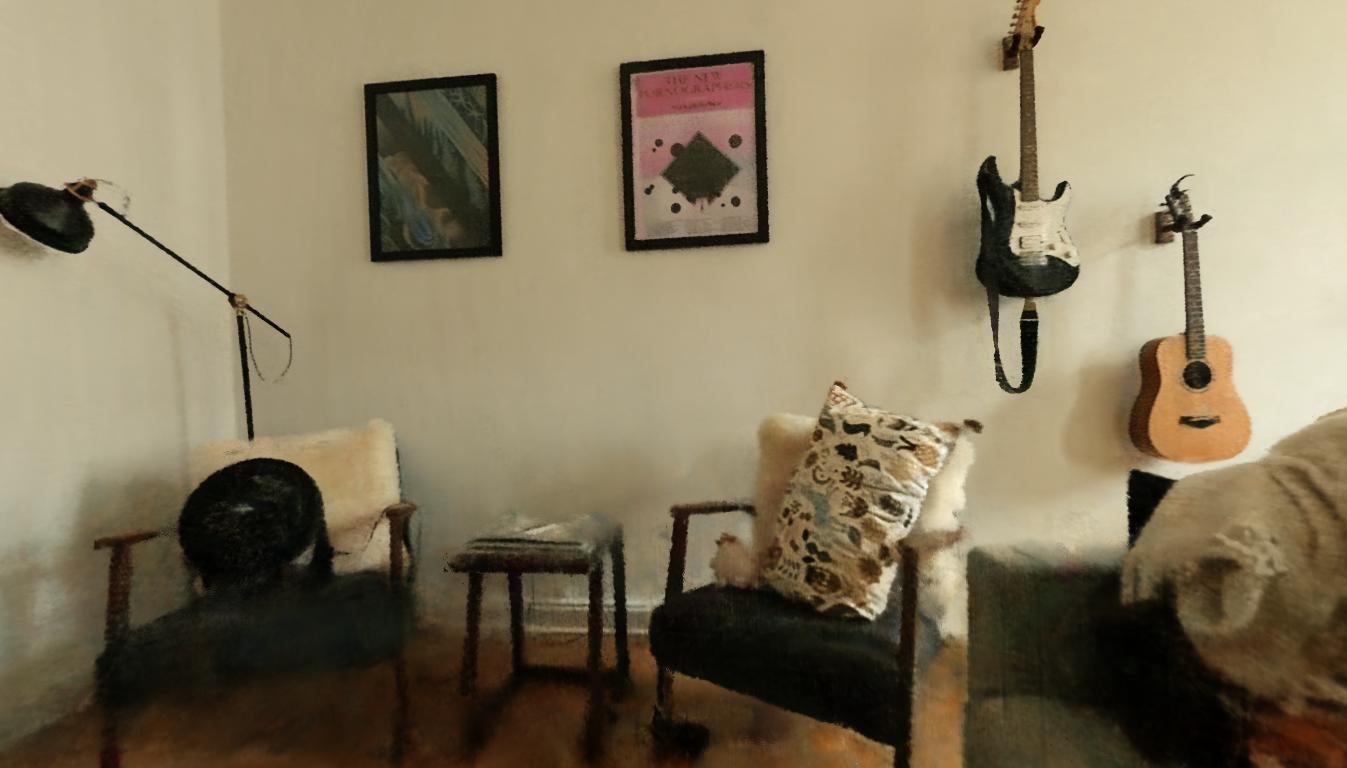}&
    \includegraphics[width=0.158\linewidth]{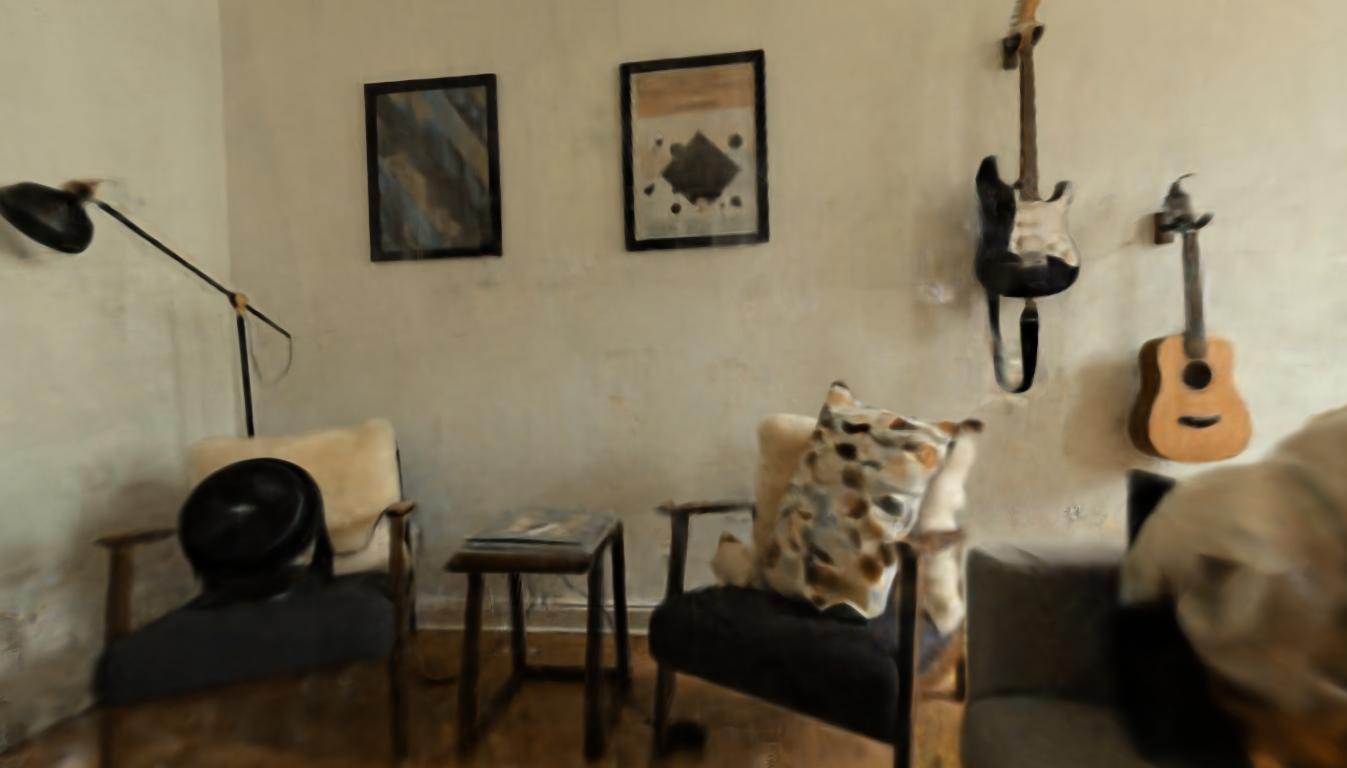}&
    \includegraphics[width=0.158\linewidth]{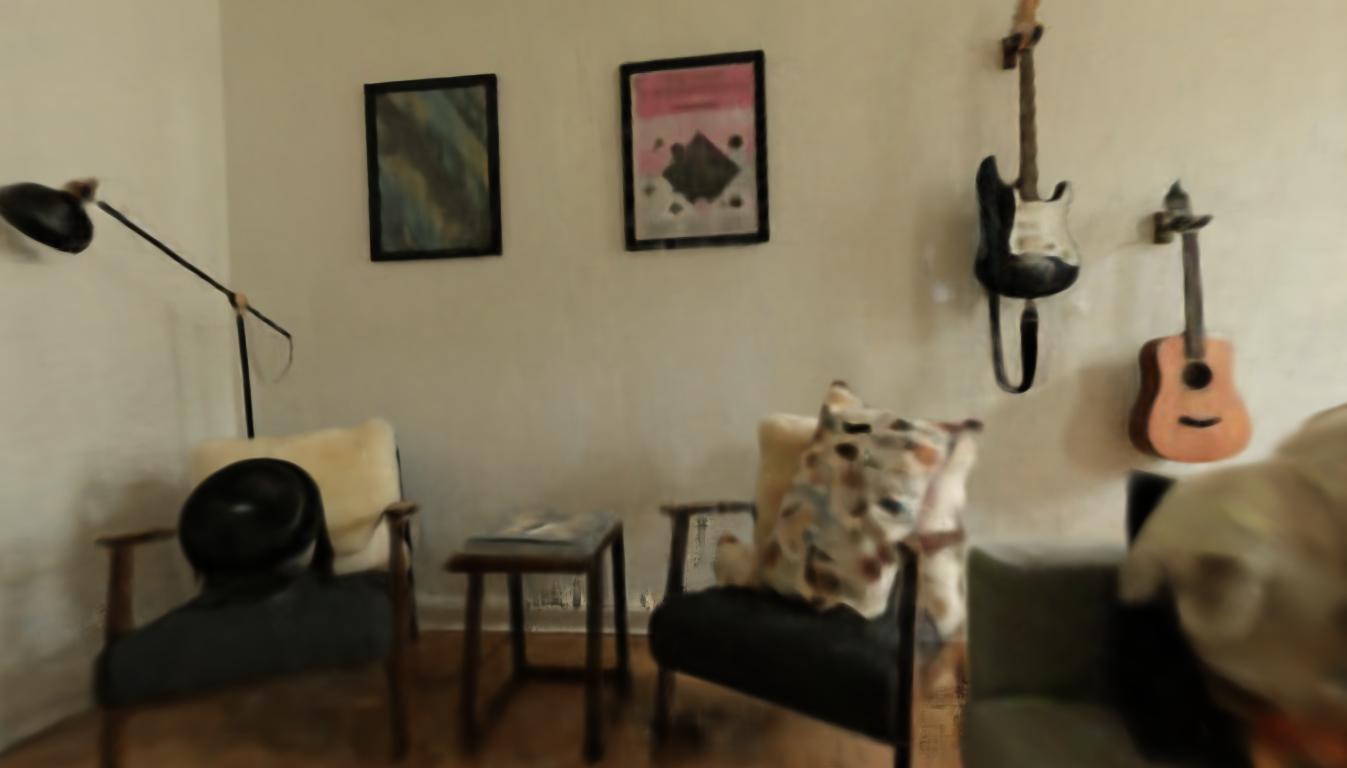}\\

    \includegraphics[width=0.158\linewidth]{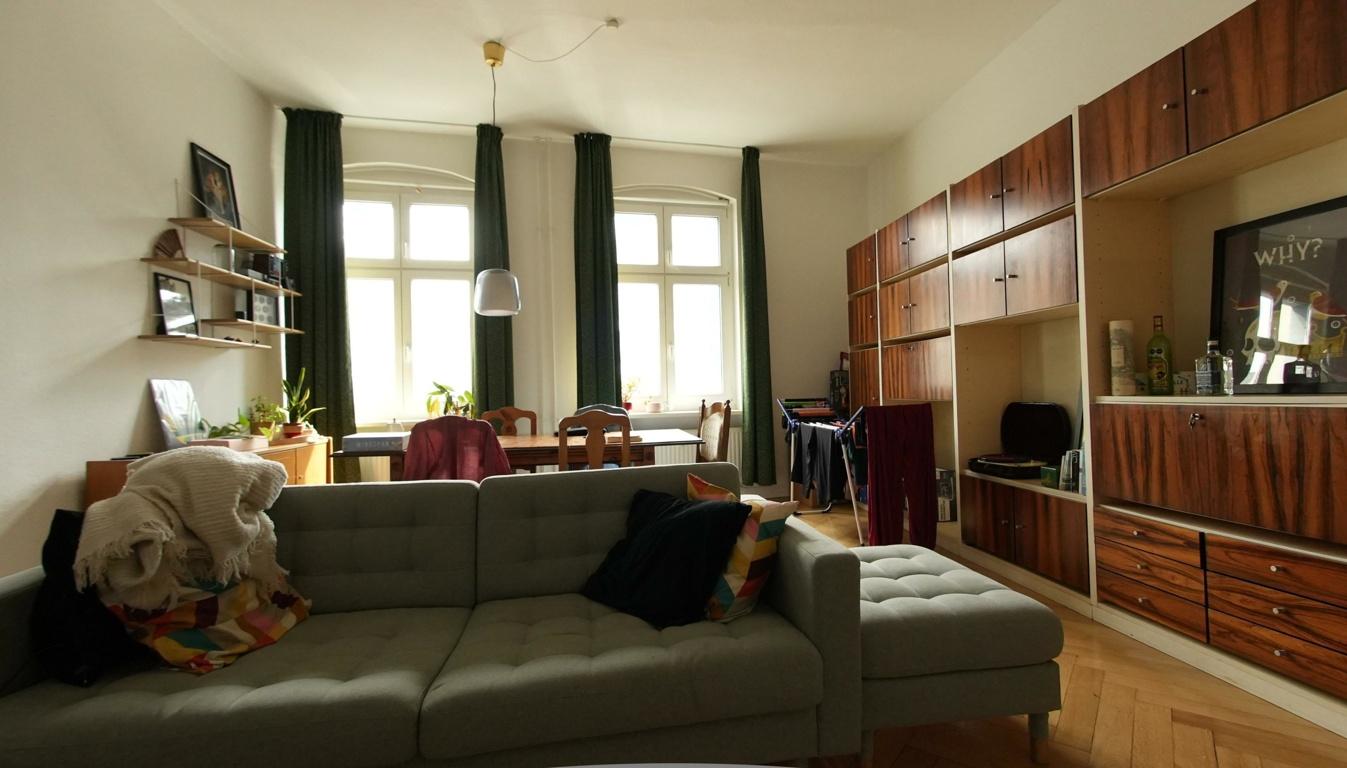}& 
    \includegraphics[width=0.158\linewidth]{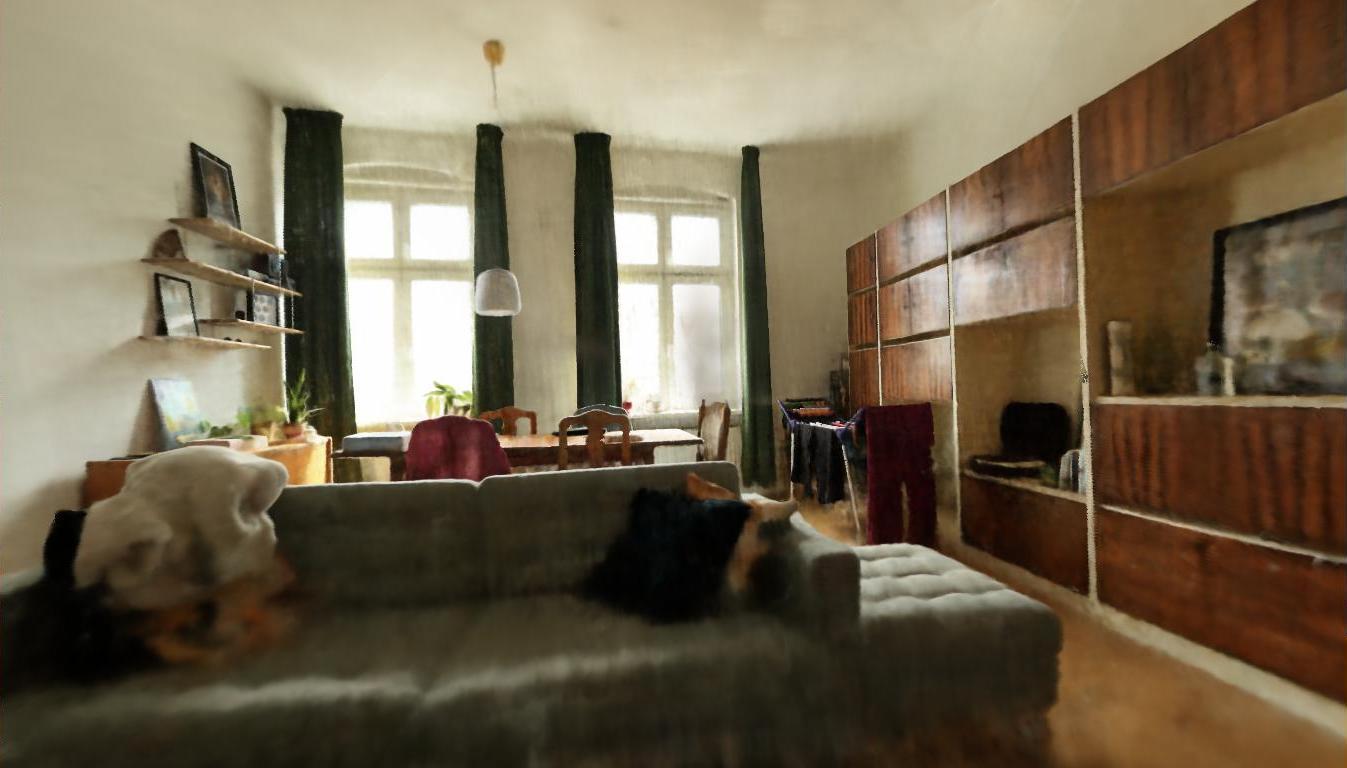}& 
    \includegraphics[width=0.158\linewidth]{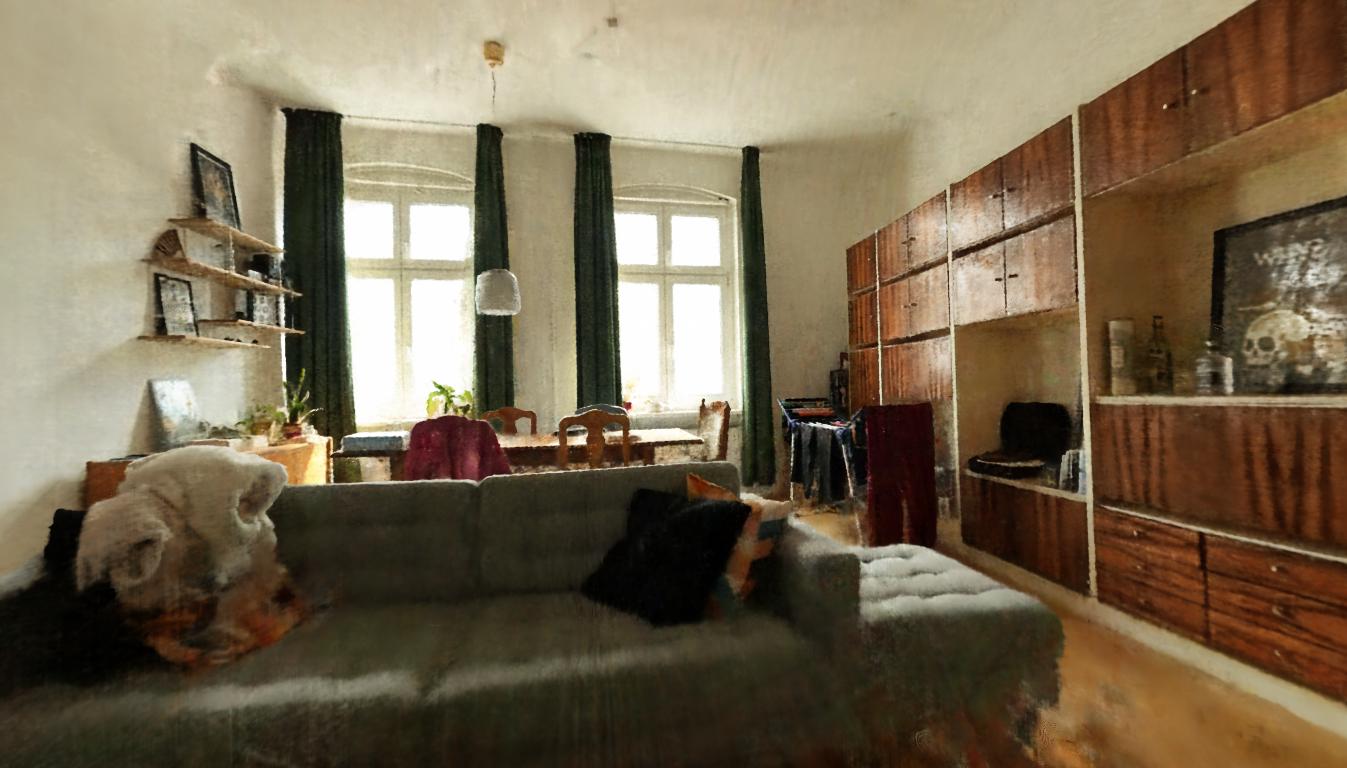}& 
    \includegraphics[width=0.158\linewidth]{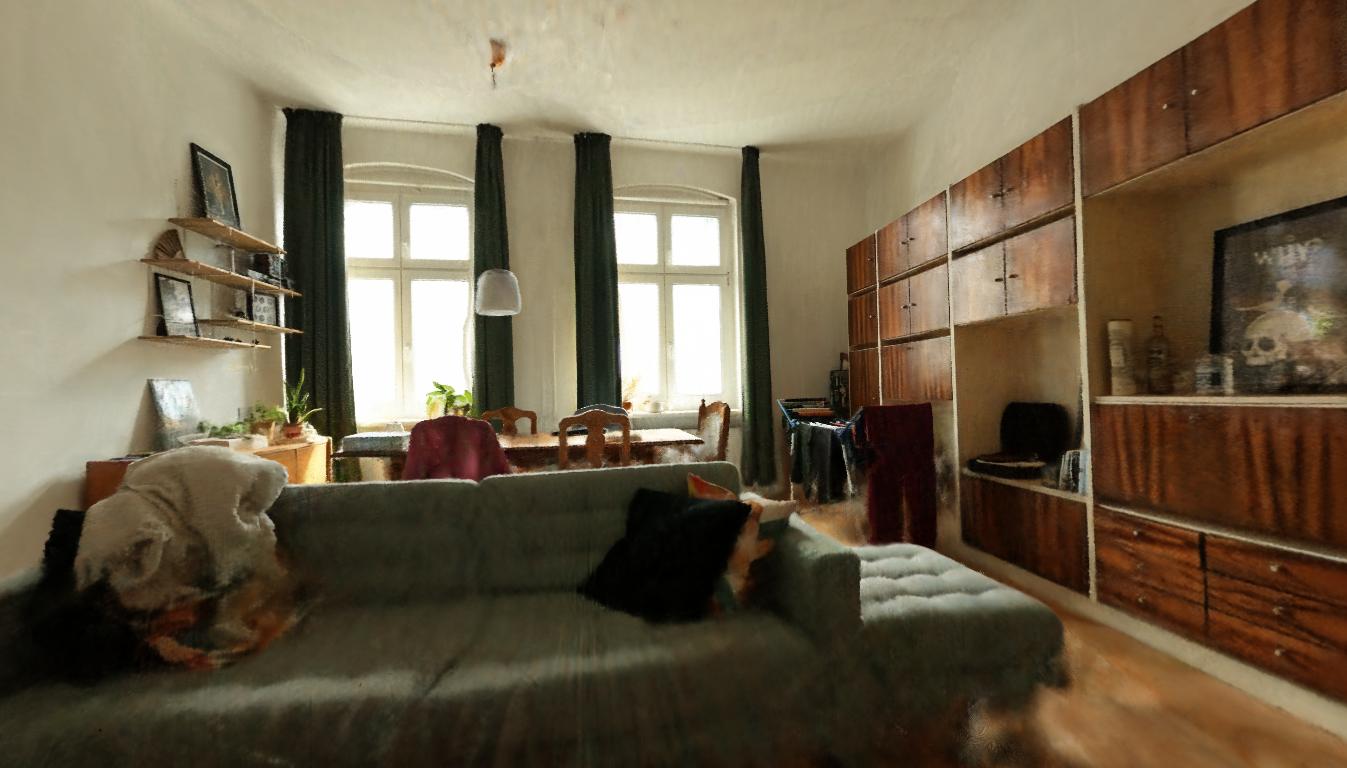}&
    \includegraphics[width=0.158\linewidth]{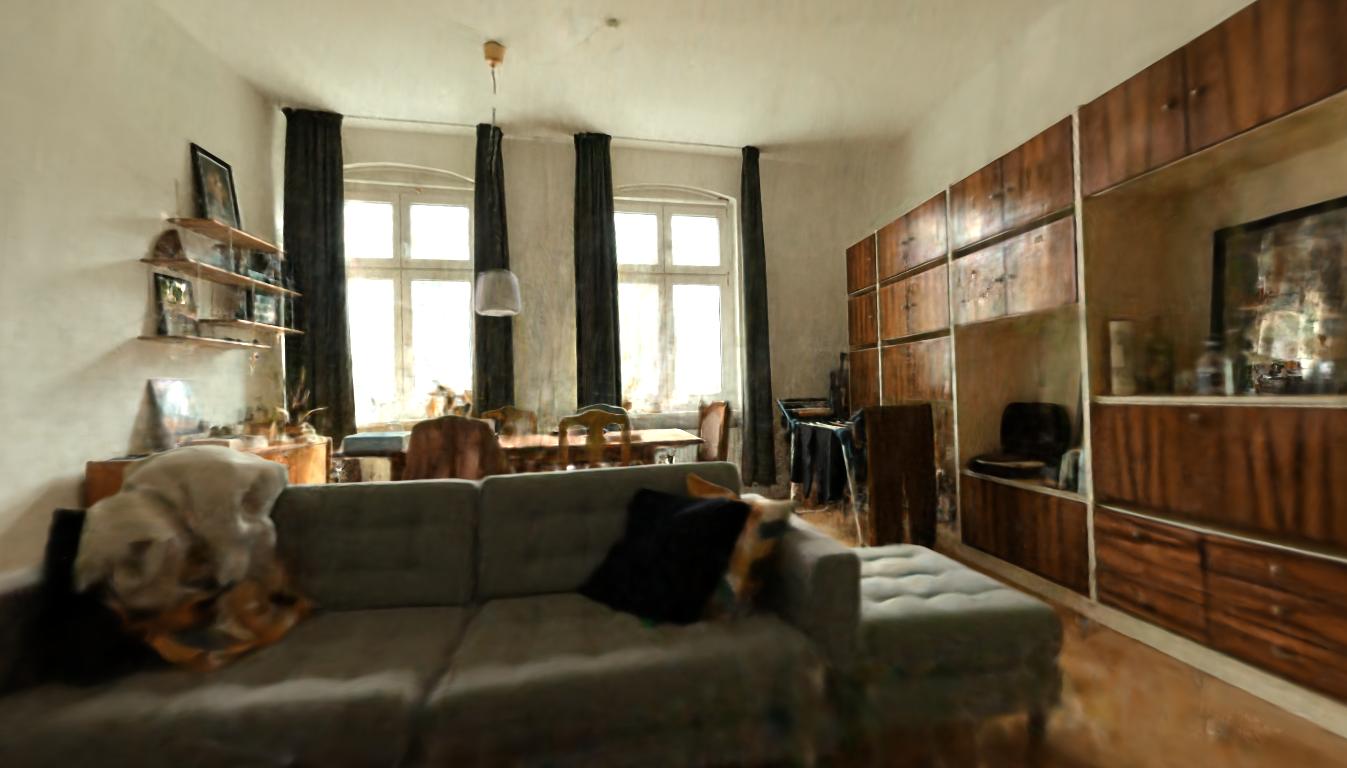}&
    \includegraphics[width=0.158\linewidth]{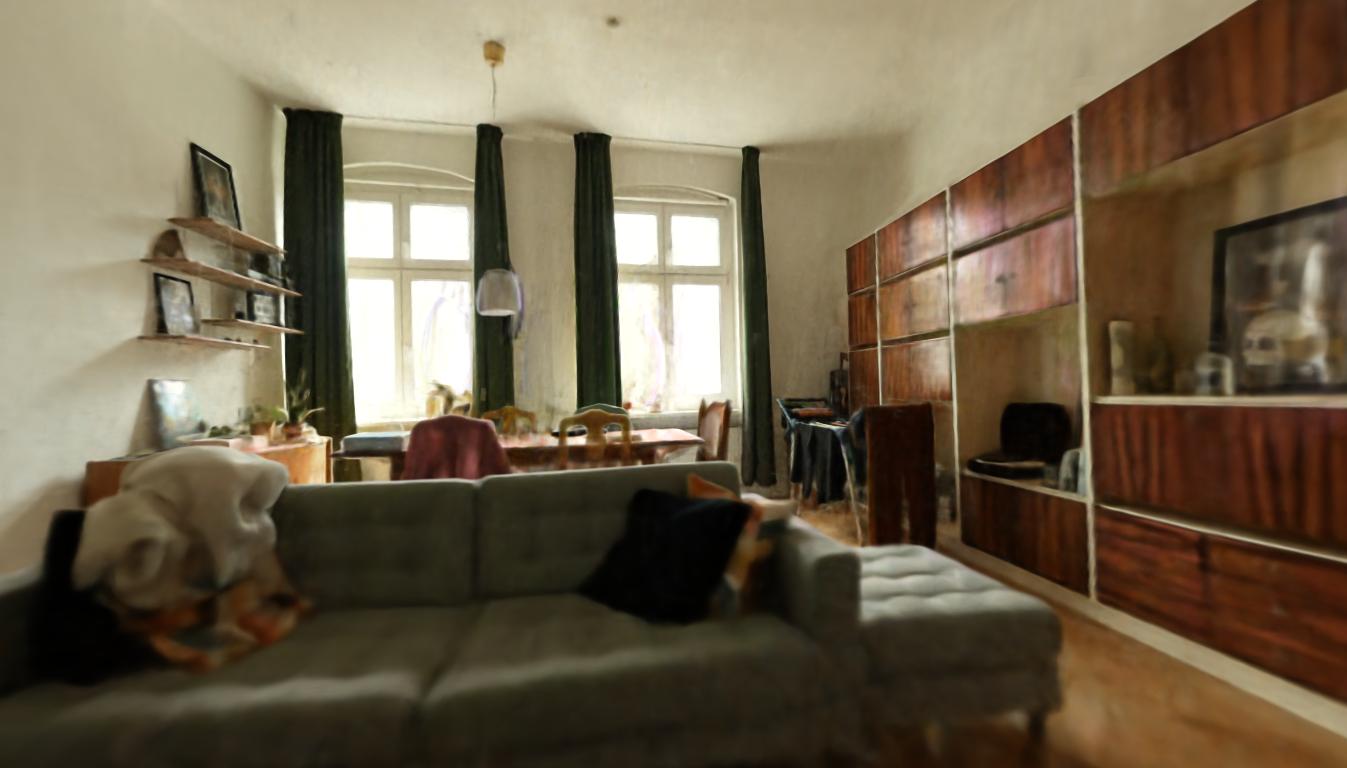}\\

    \includegraphics[width=0.158\linewidth]{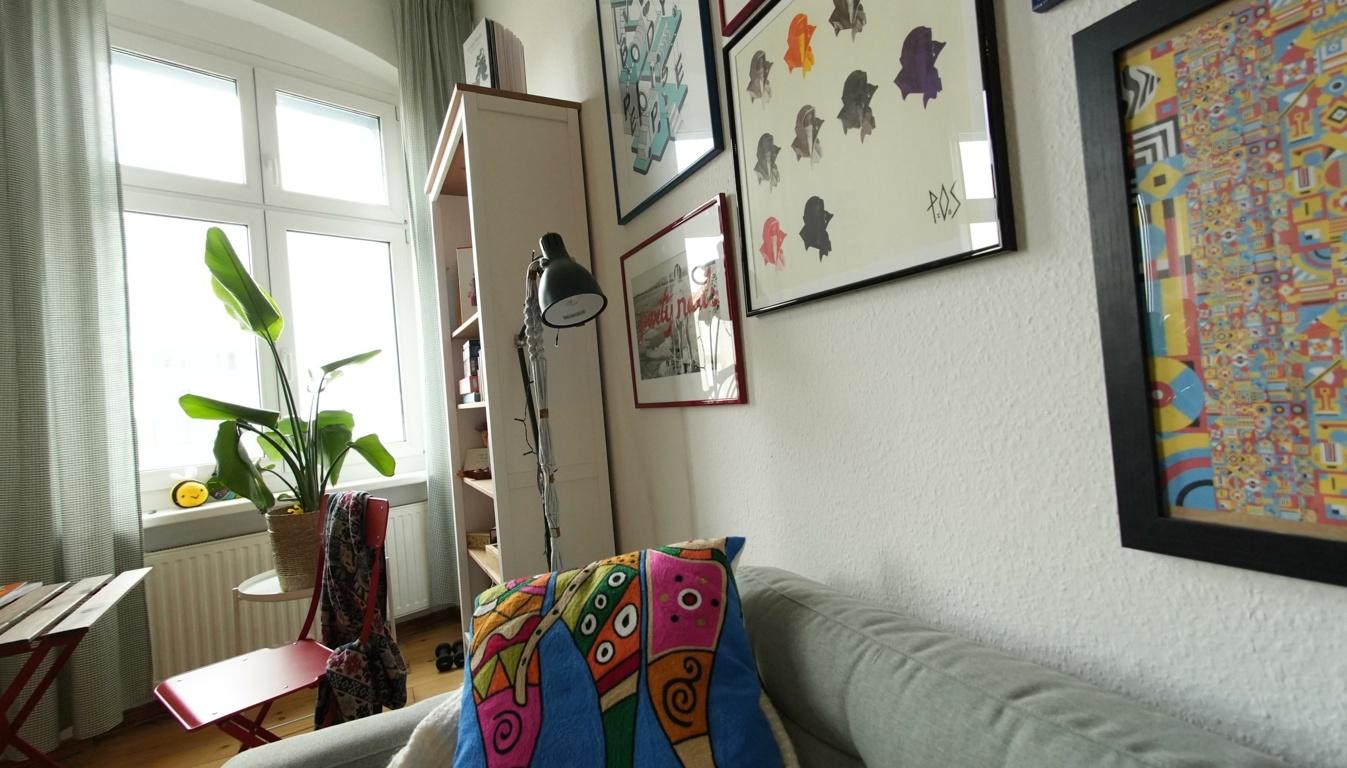}& 
    \includegraphics[width=0.158\linewidth]{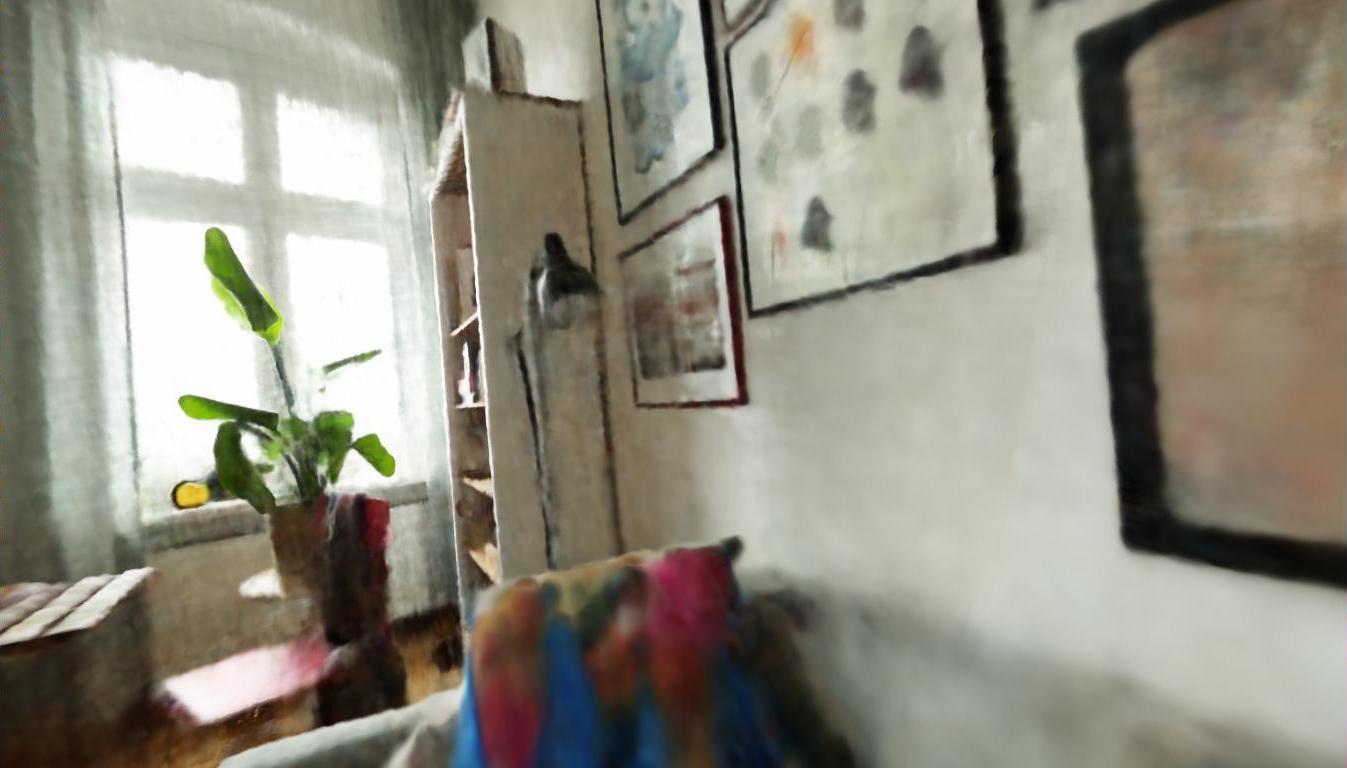}& 
    \includegraphics[width=0.158\linewidth]{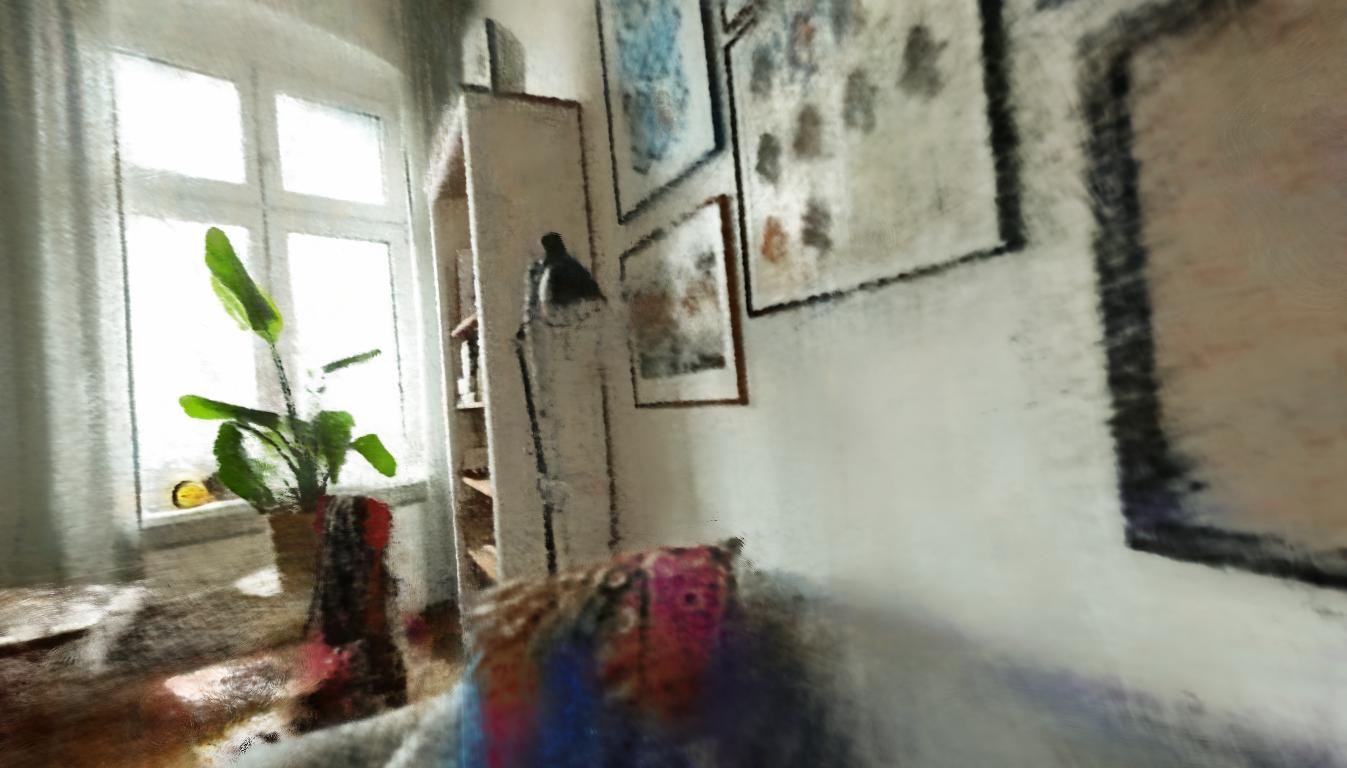}& 
    \includegraphics[width=0.158\linewidth]{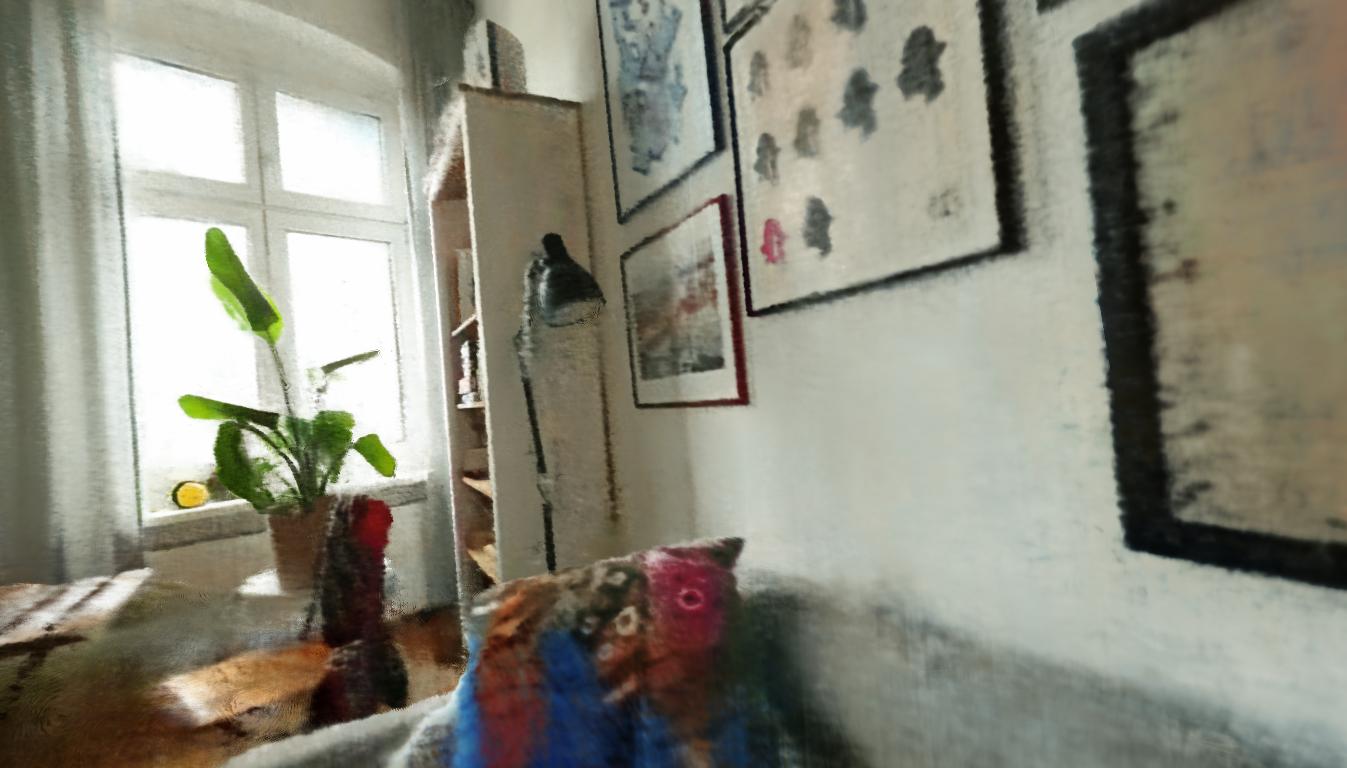}&
    \includegraphics[width=0.158\linewidth]{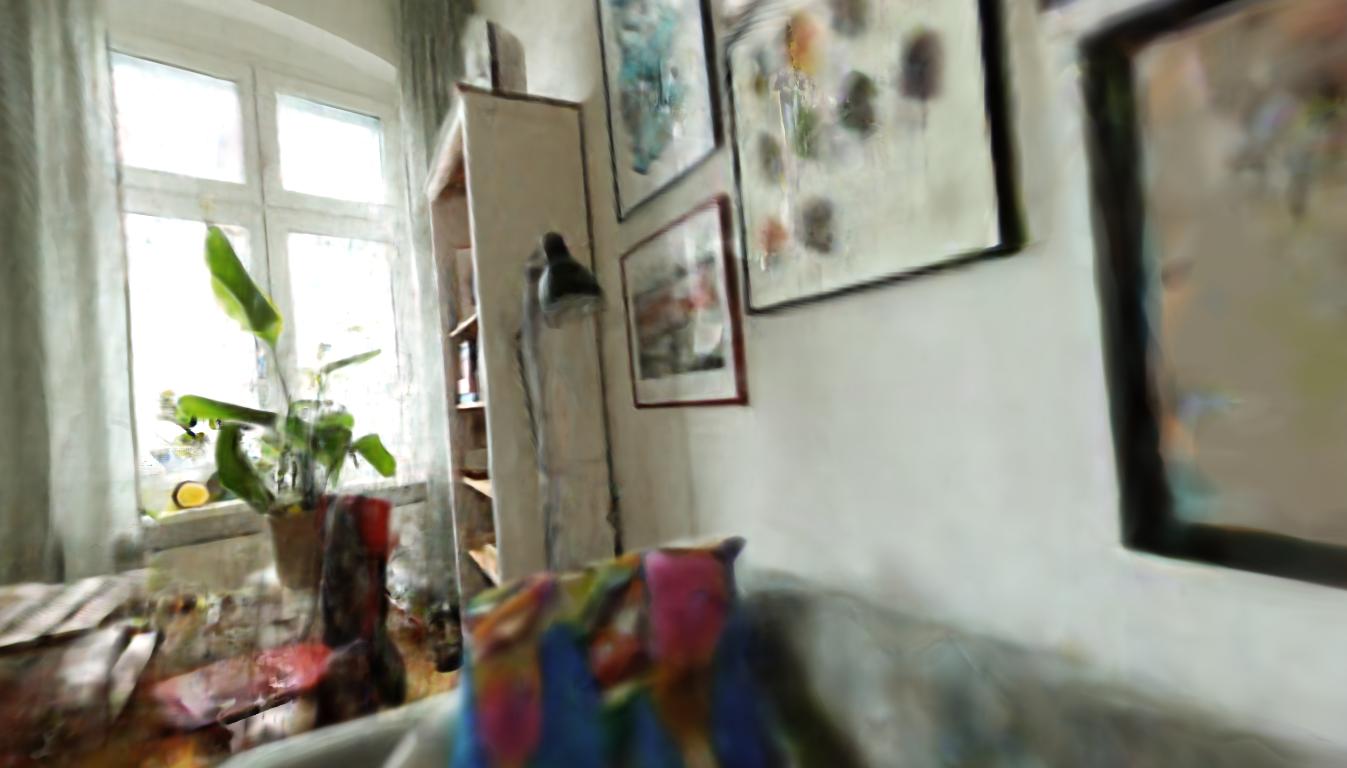}&
    \includegraphics[width=0.158\linewidth]{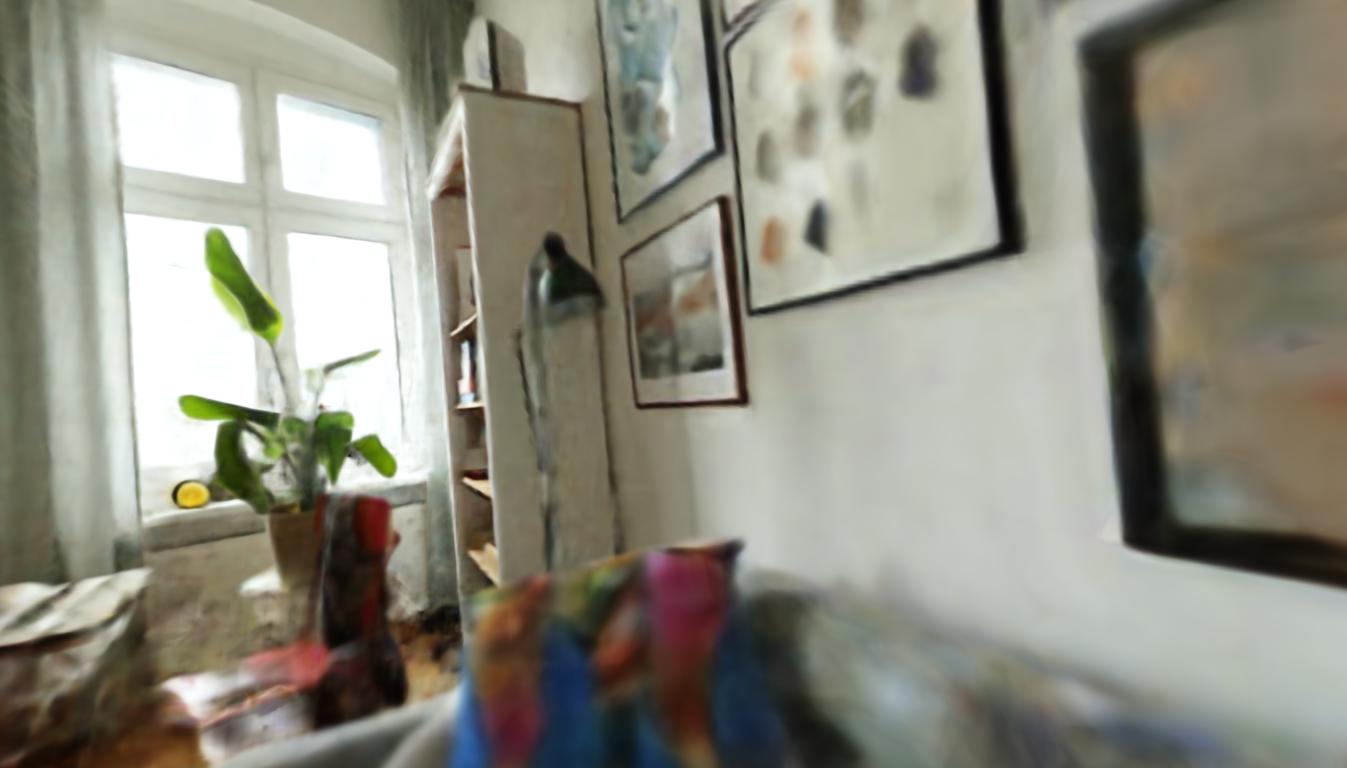}\\
    
    % london
    % \includegraphics[width=0.158\linewidth]{suppl_figures/zipnerf/london/00536/gt.jpg}& 
    % \includegraphics[width=0.158\linewidth]{suppl_figures/zipnerf/london/00536/mega-nerf.jpg}& 
    % \includegraphics[width=0.158\linewidth]{suppl_figures/zipnerf/london/00536/nelf-pro.jpg}& 
    % \includegraphics[width=0.158\linewidth]{suppl_figures/zipnerf/london/00536/nelf-pro_ours.jpg}&
    % \includegraphics[width=0.158\linewidth]{suppl_figures/zipnerf/london/00536/localrf.jpg}&
    % \includegraphics[width=0.158\linewidth]{suppl_figures/zipnerf/london/00536/localrf_ours.jpg}\\

    \includegraphics[width=0.158\linewidth]{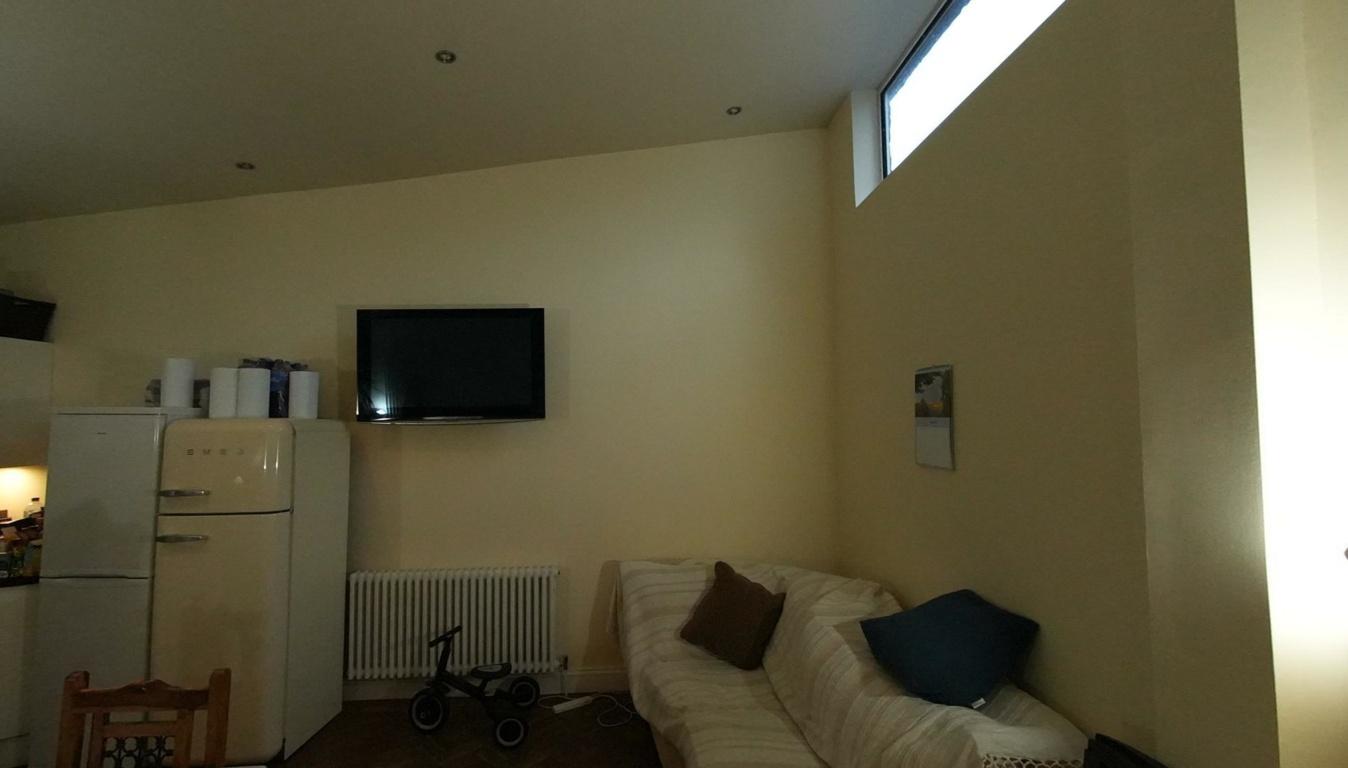}& 
    \includegraphics[width=0.158\linewidth]{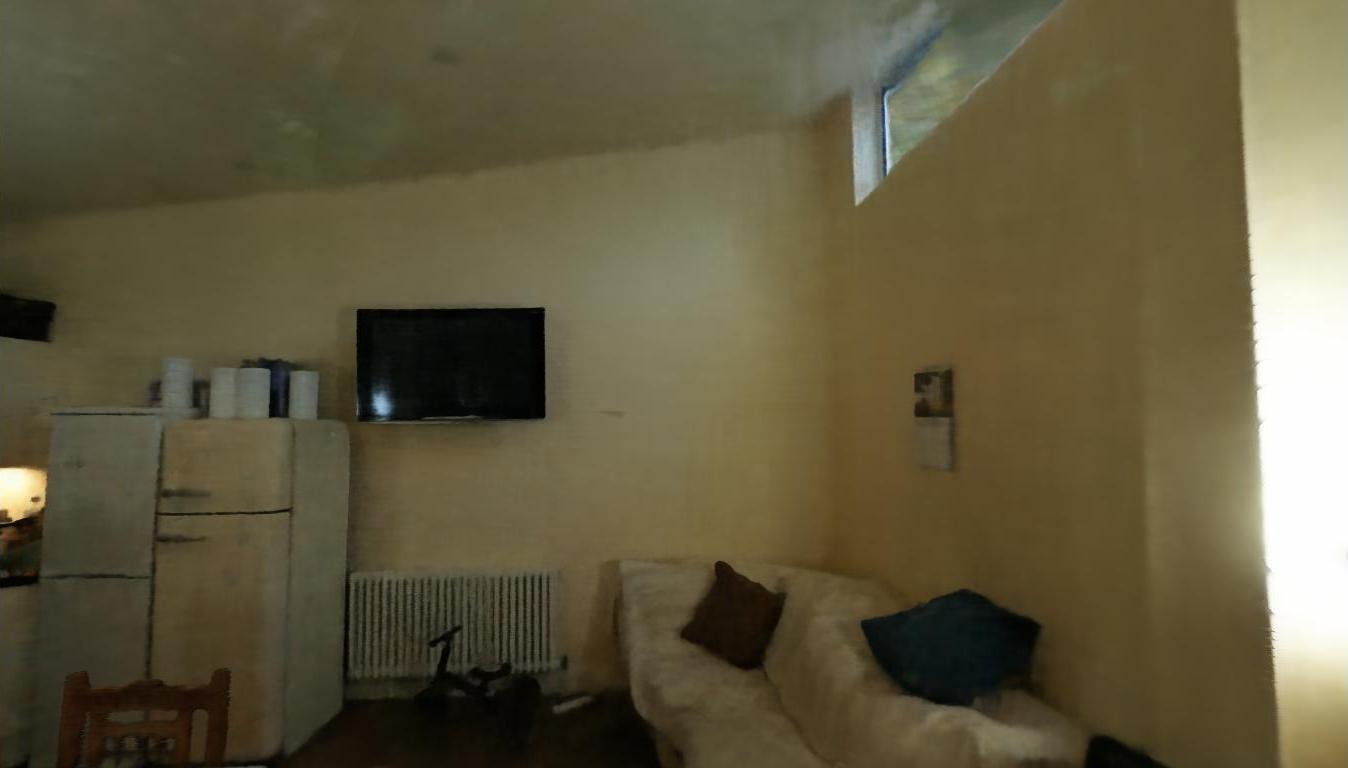}& 
    \includegraphics[width=0.158\linewidth]{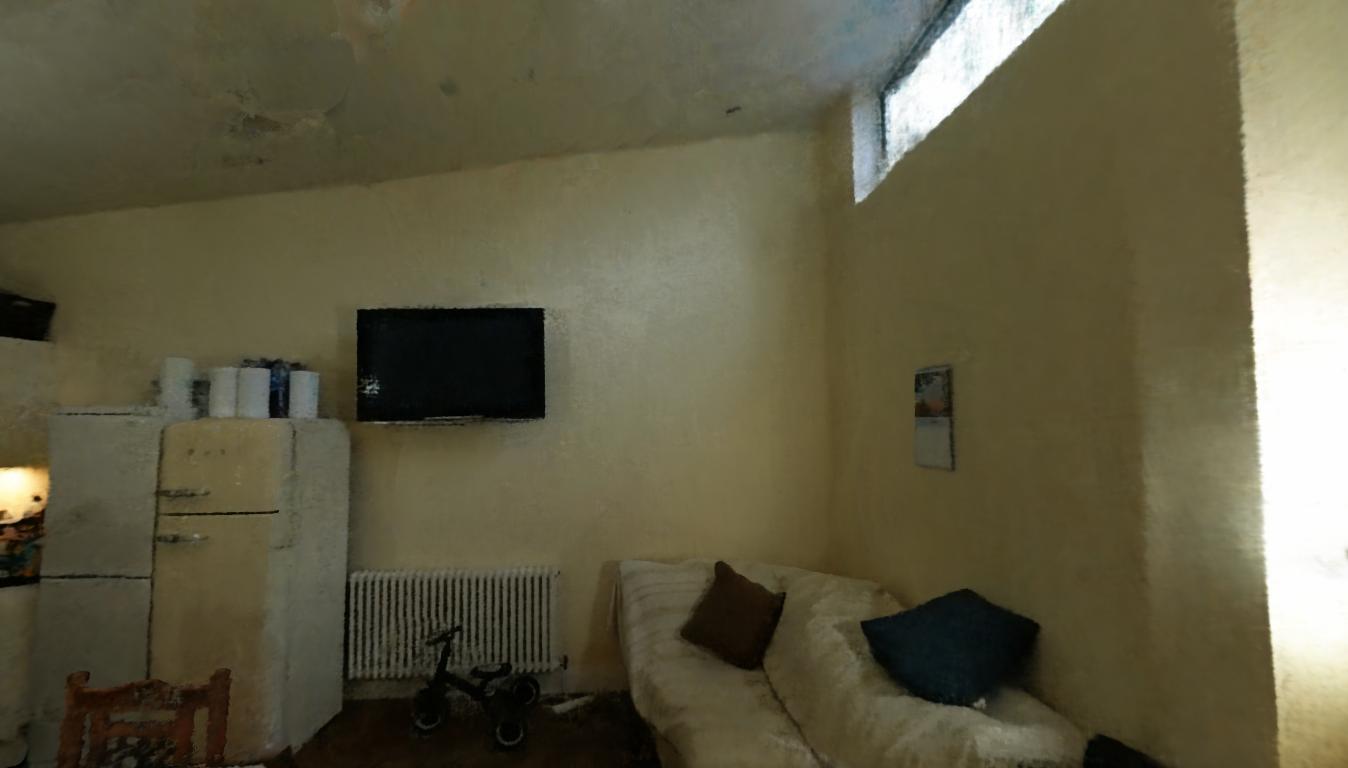}& 
    \includegraphics[width=0.158\linewidth]{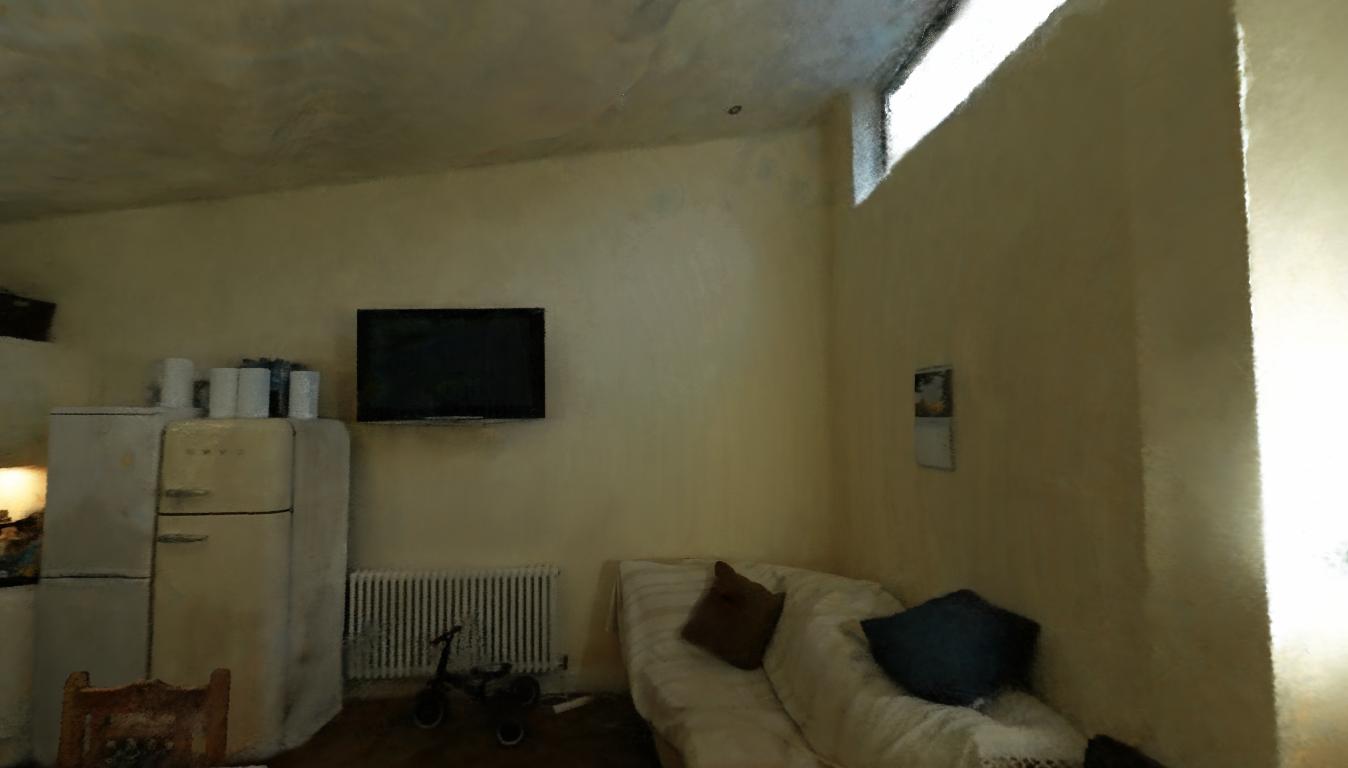}&
    \includegraphics[width=0.158\linewidth]{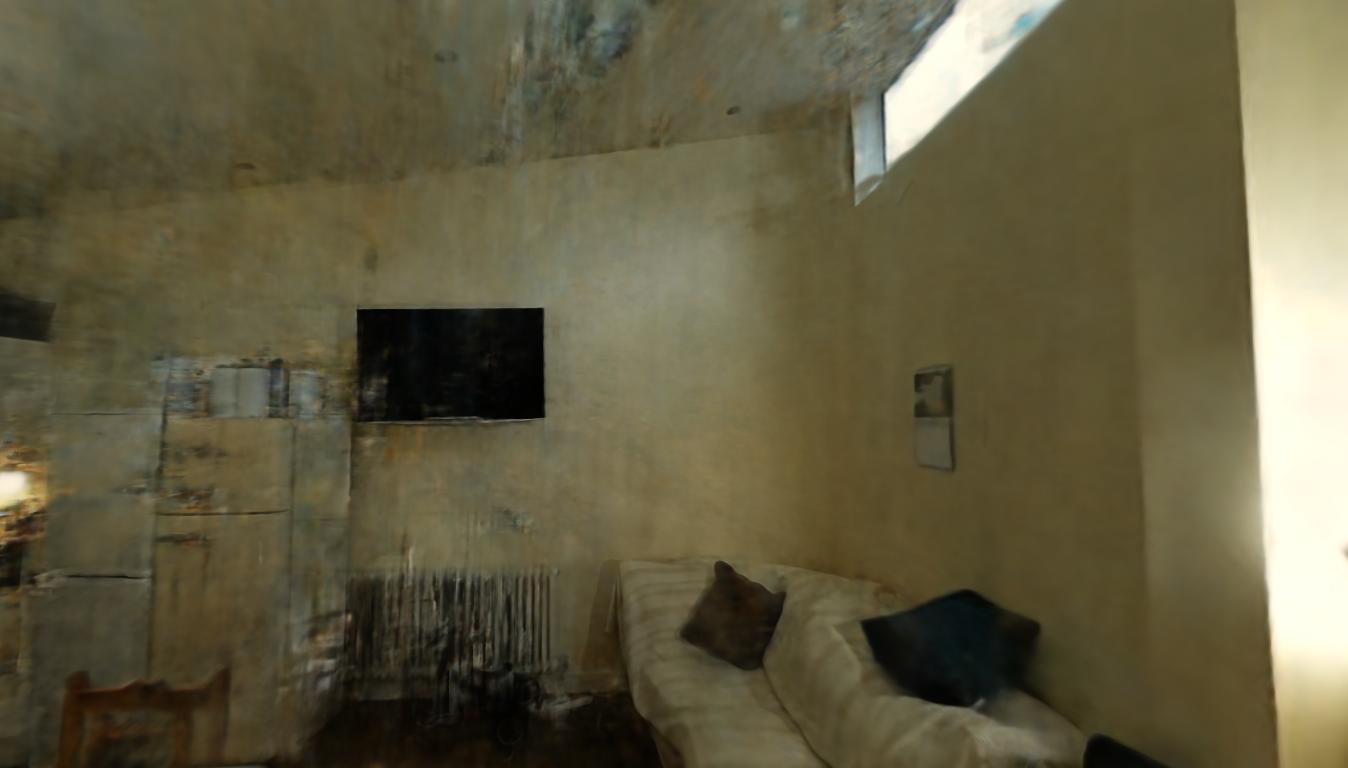}&
    \includegraphics[width=0.158\linewidth]{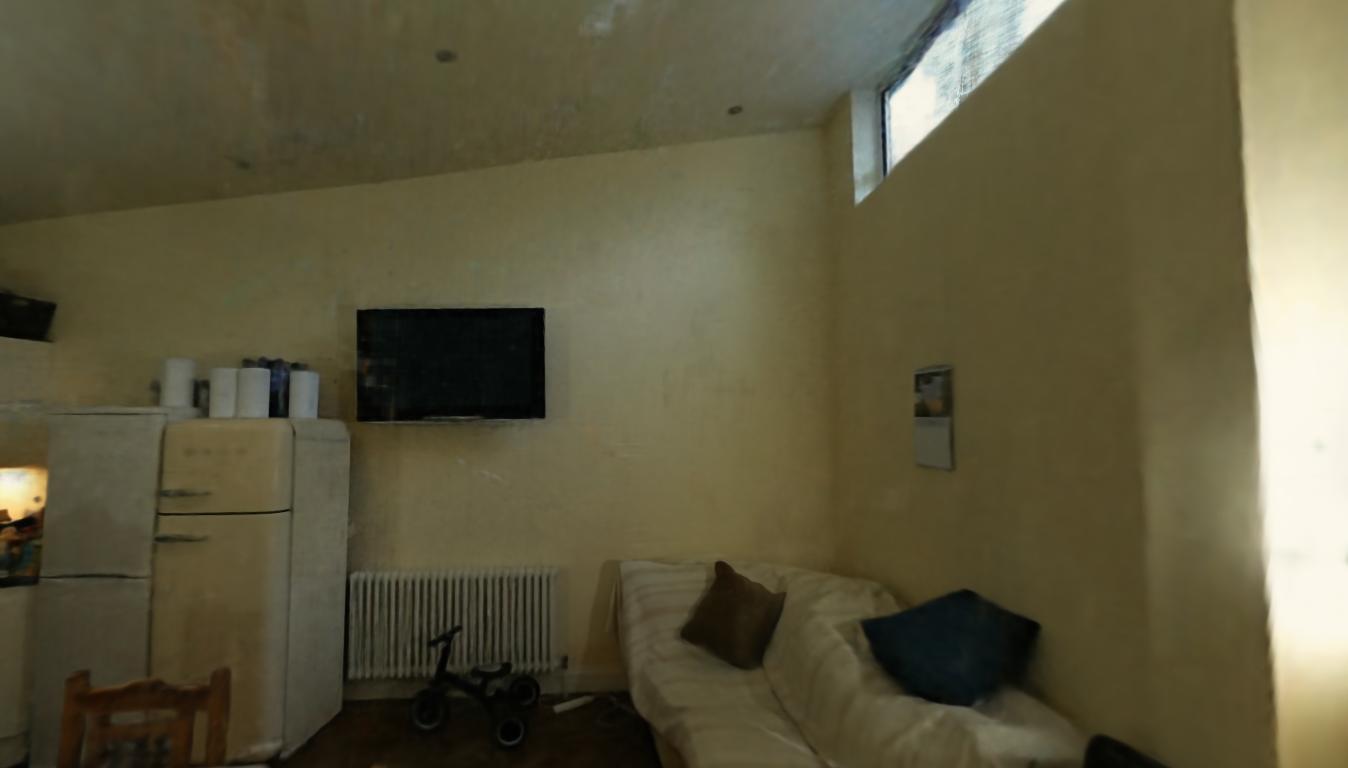}\\

    \includegraphics[width=0.158\linewidth]{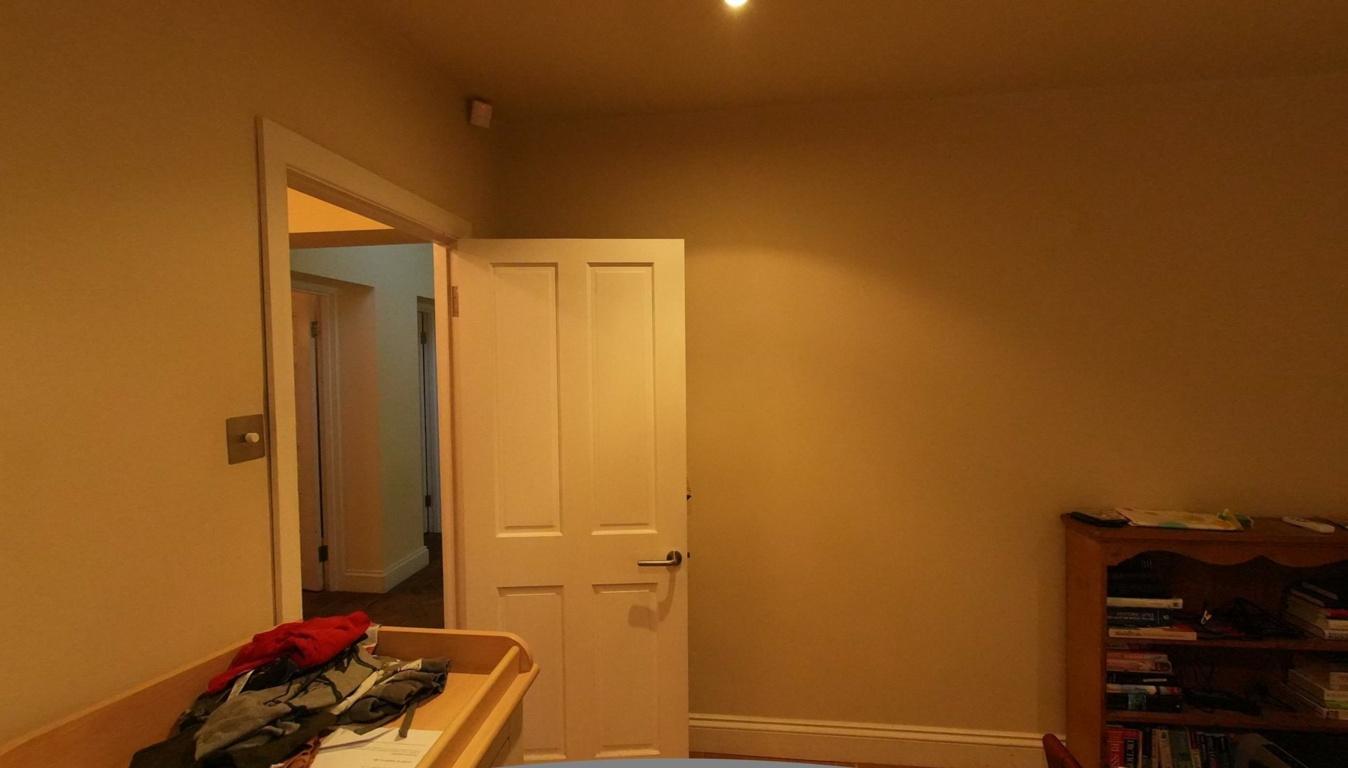}& 
    \includegraphics[width=0.158\linewidth]{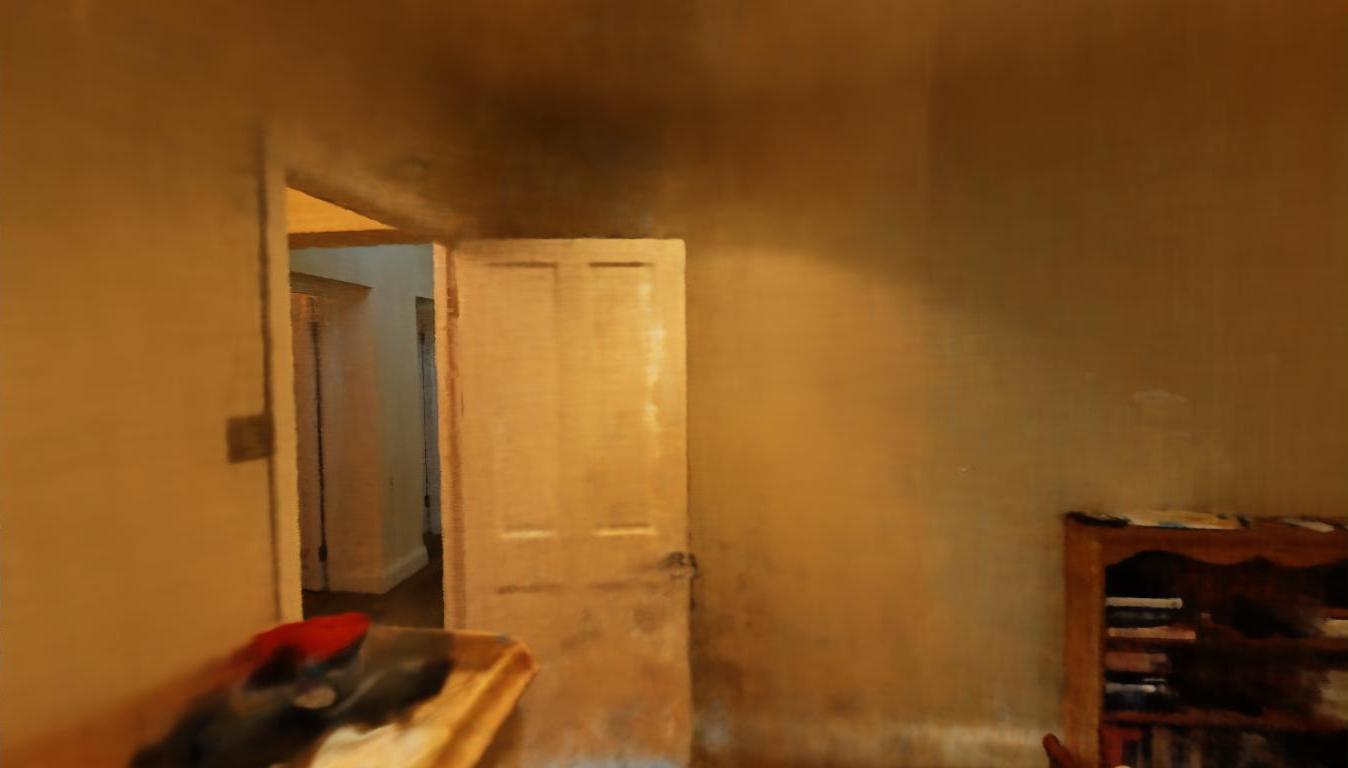}& 
    \includegraphics[width=0.158\linewidth]{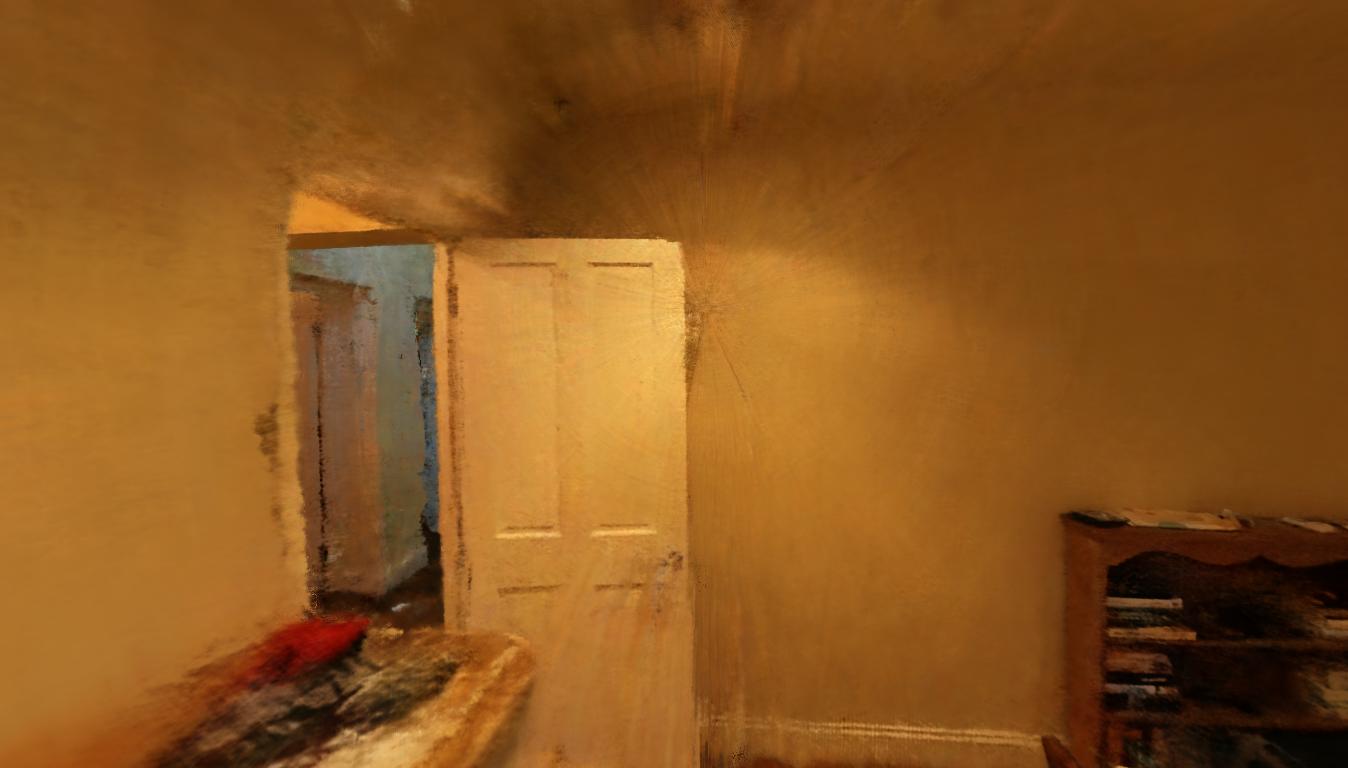}& 
    \includegraphics[width=0.158\linewidth]{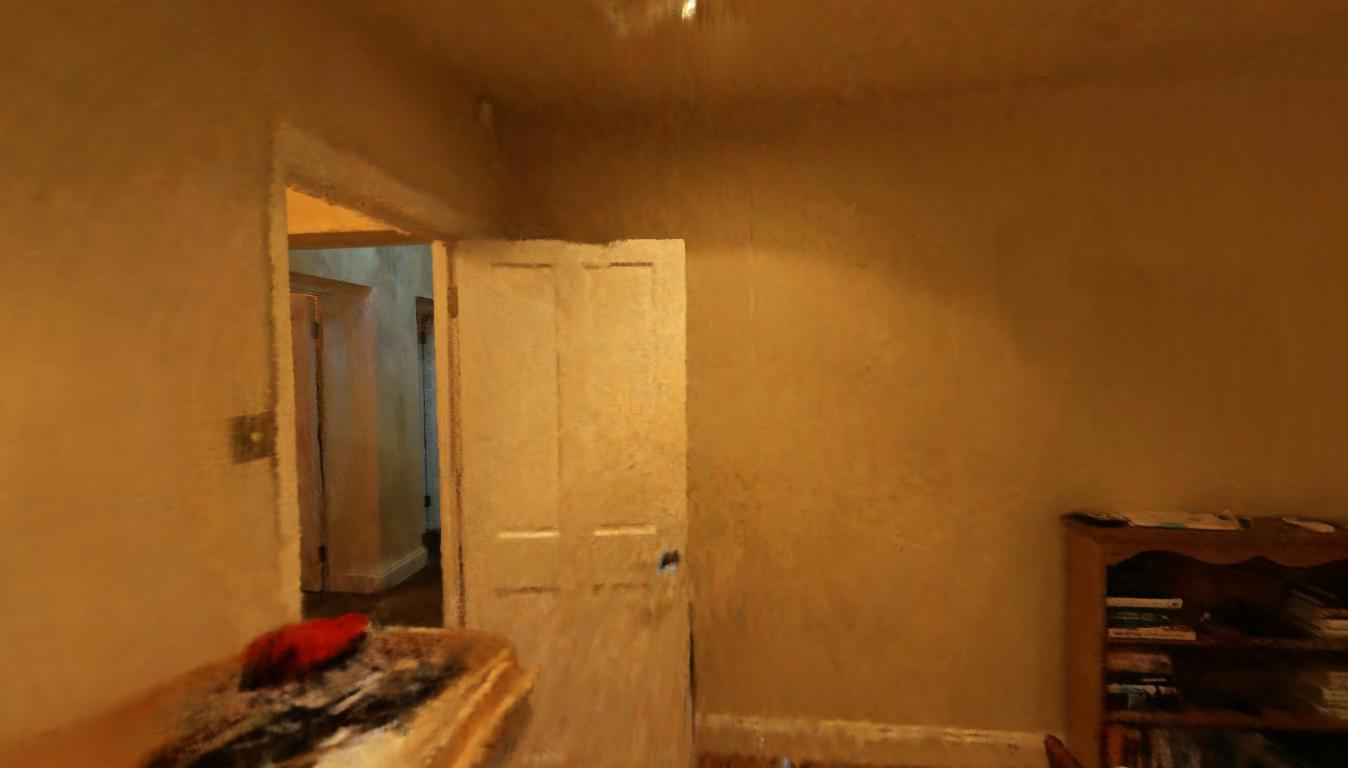}&
    \includegraphics[width=0.158\linewidth]{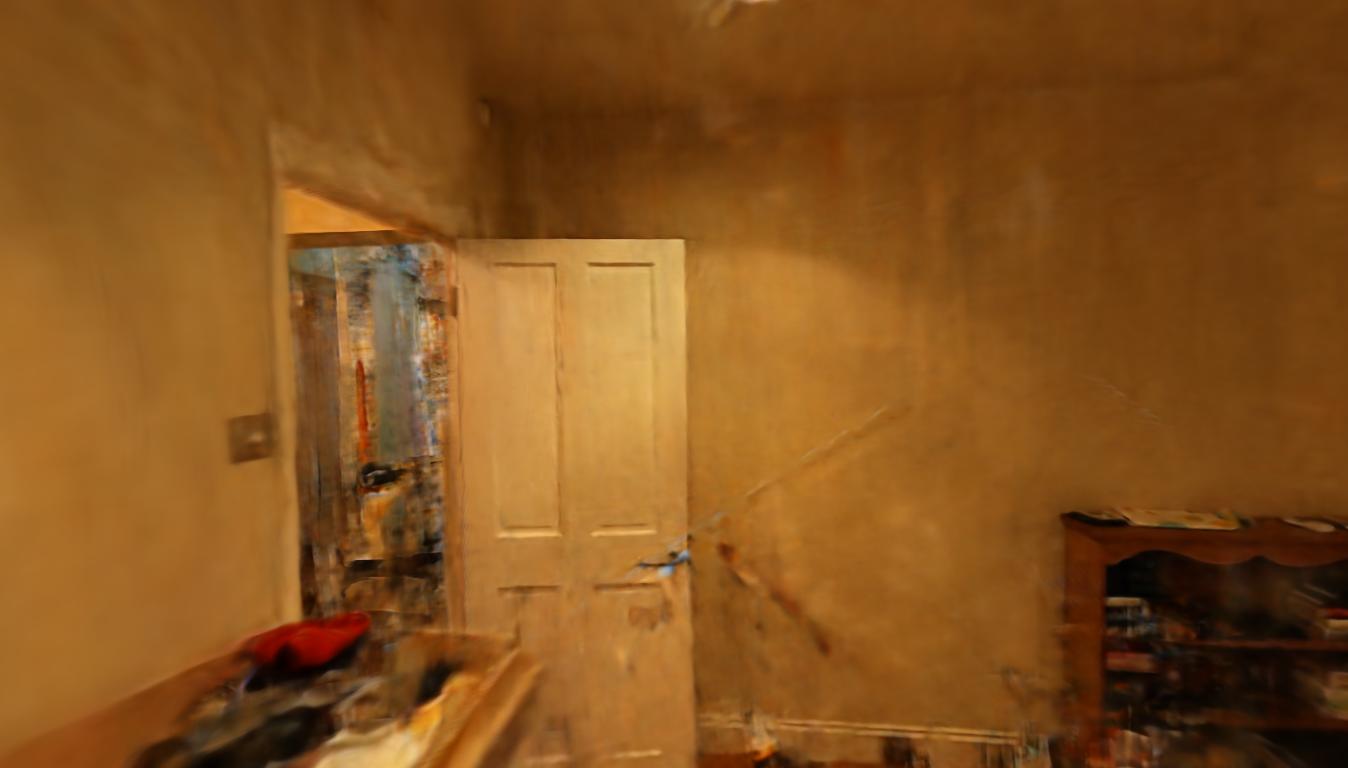}&
    \includegraphics[width=0.158\linewidth]{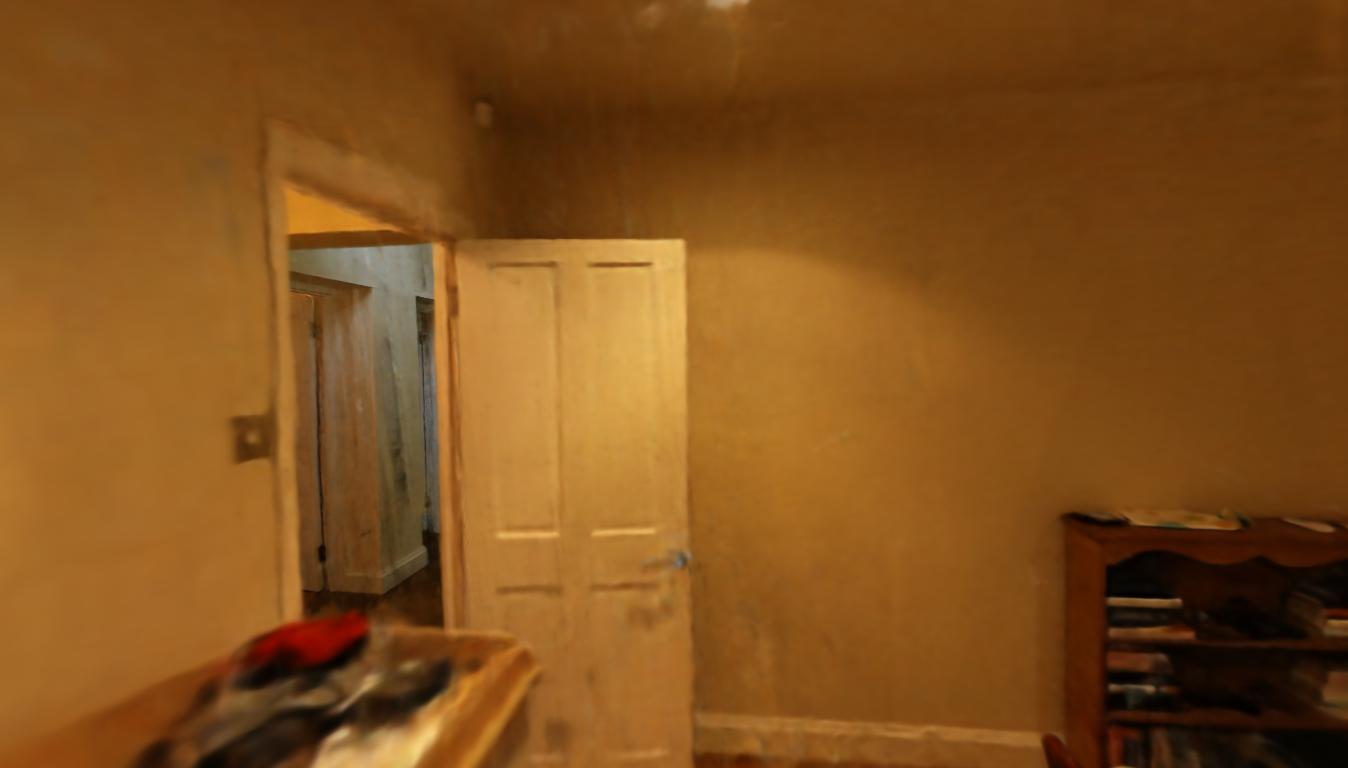}\\

    % nyc
    \includegraphics[width=0.158\linewidth]{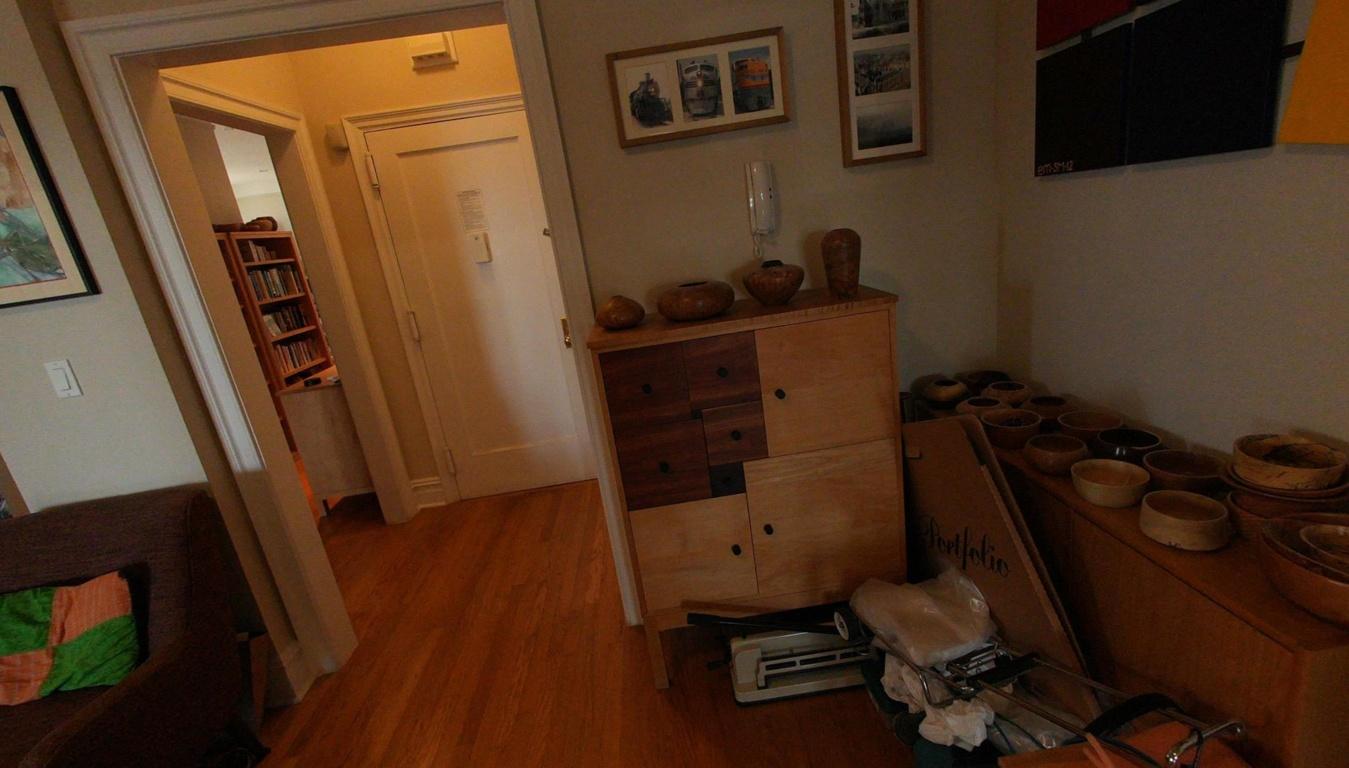}& 
    \includegraphics[width=0.158\linewidth]{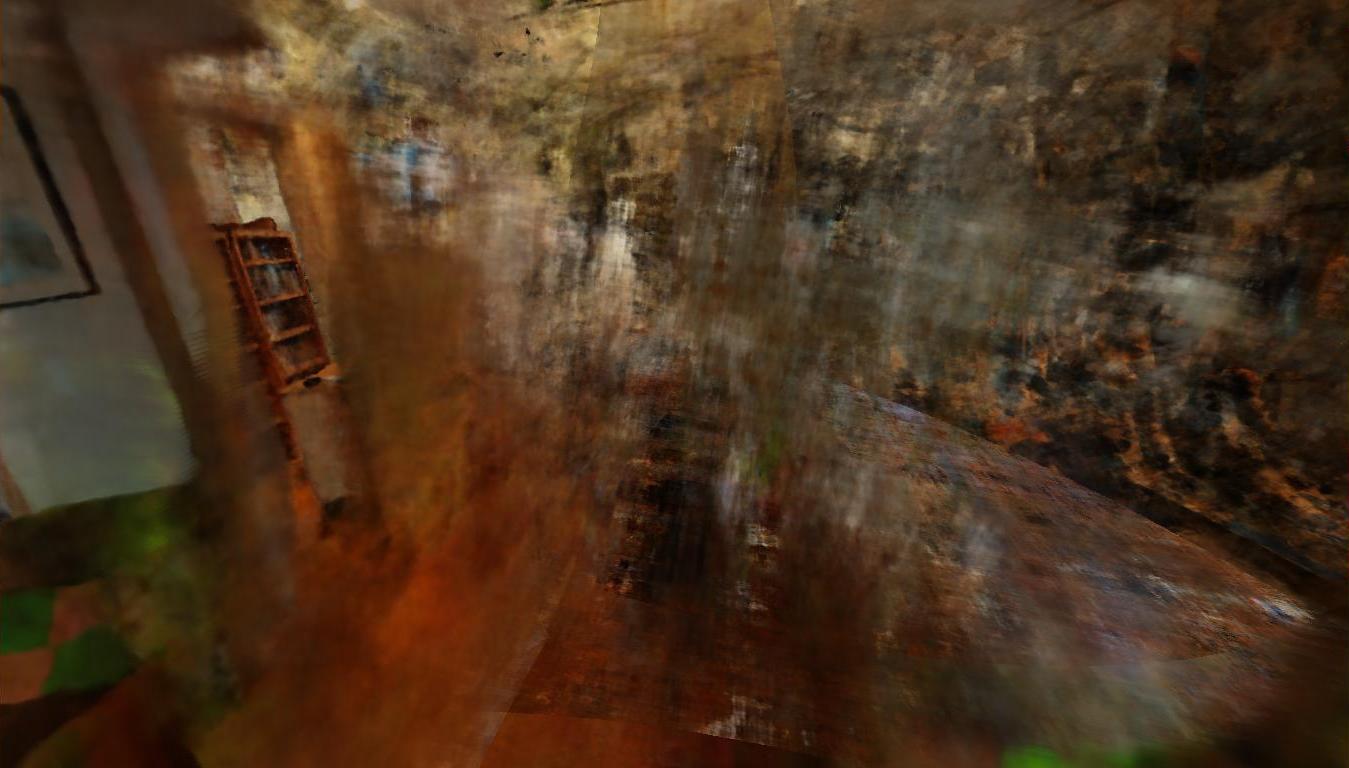}& 
    \includegraphics[width=0.158\linewidth]{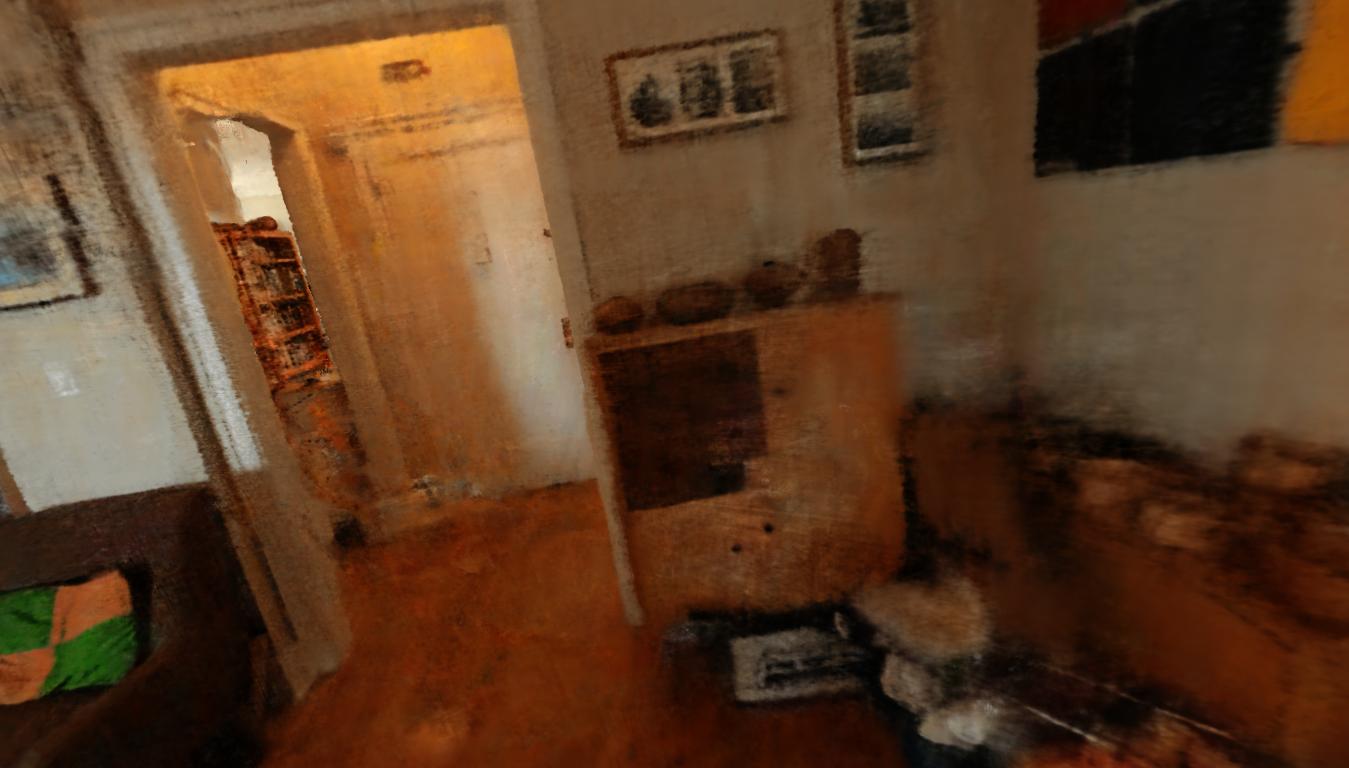}& 
    \includegraphics[width=0.158\linewidth]{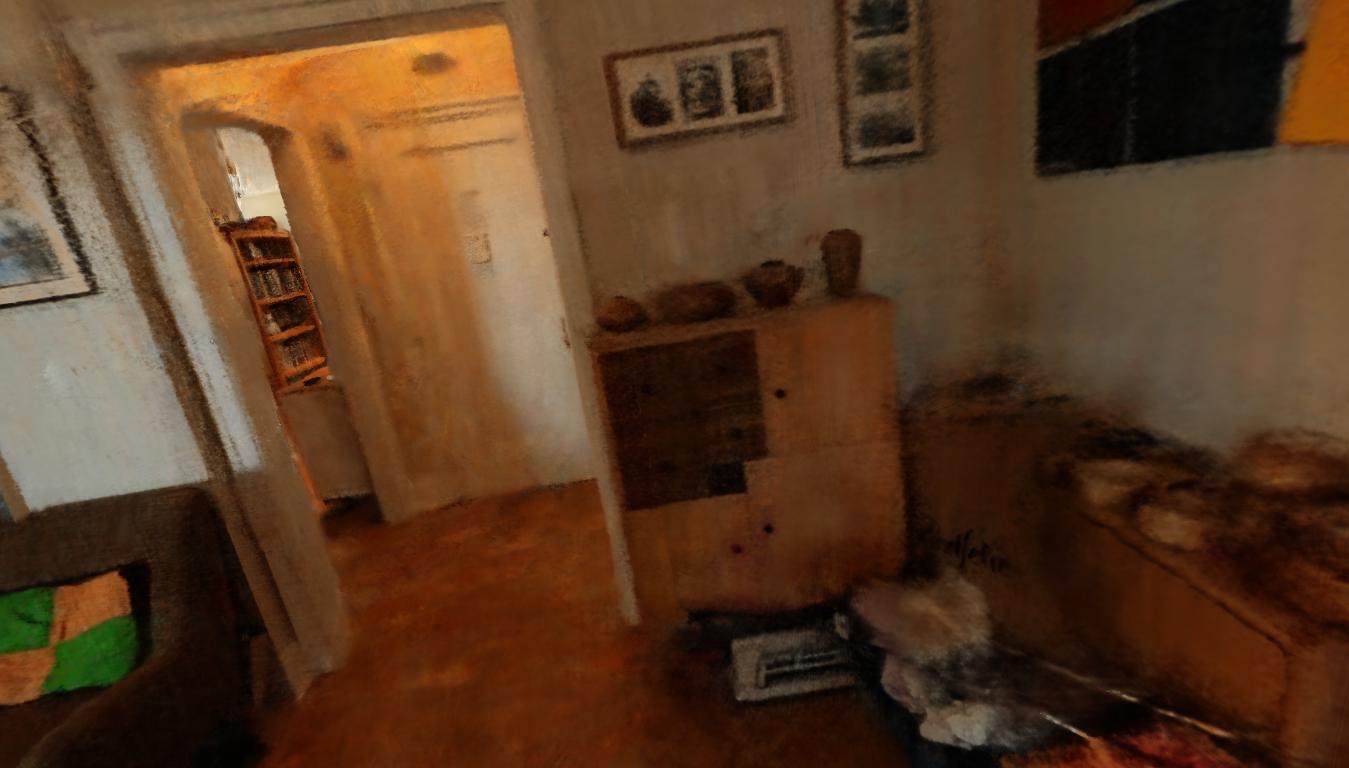}&
    \includegraphics[width=0.158\linewidth]{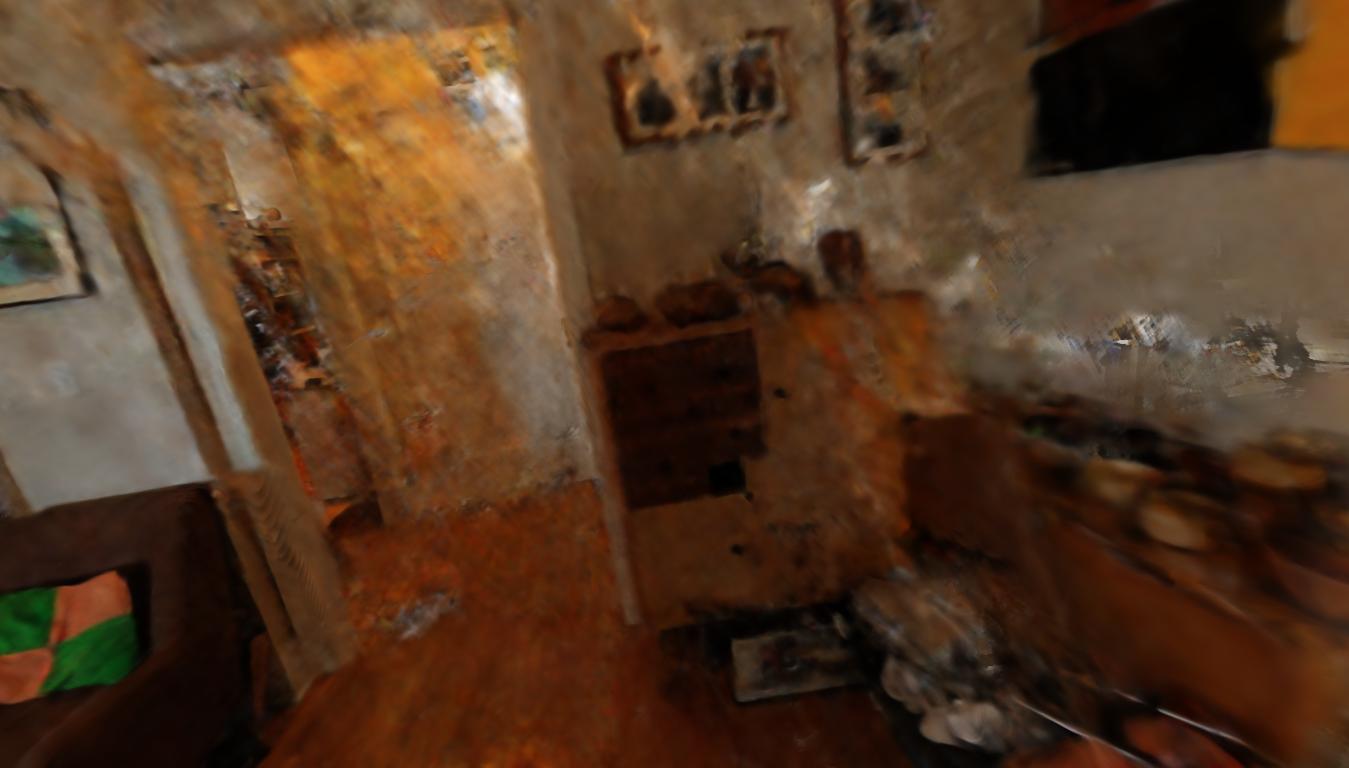}&
    \includegraphics[width=0.158\linewidth]{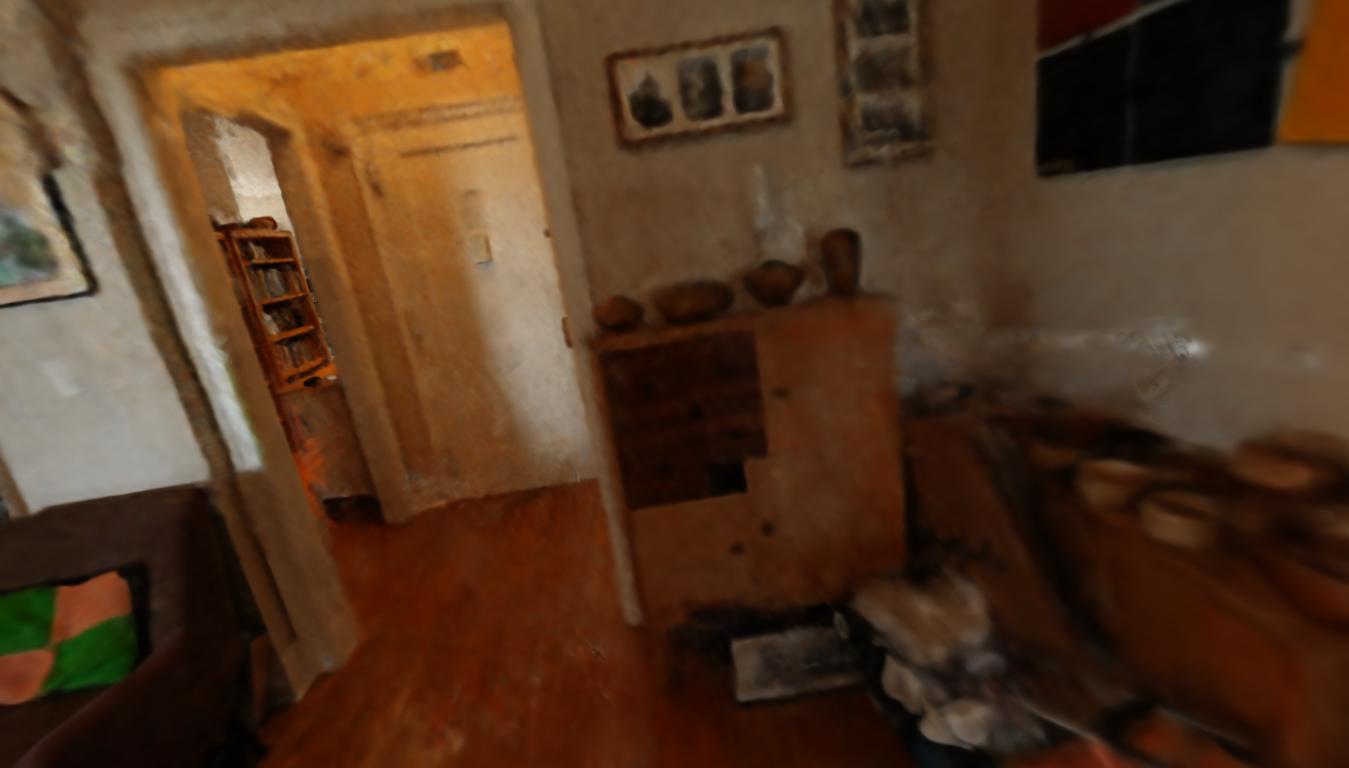}\\

    \includegraphics[width=0.158\linewidth]{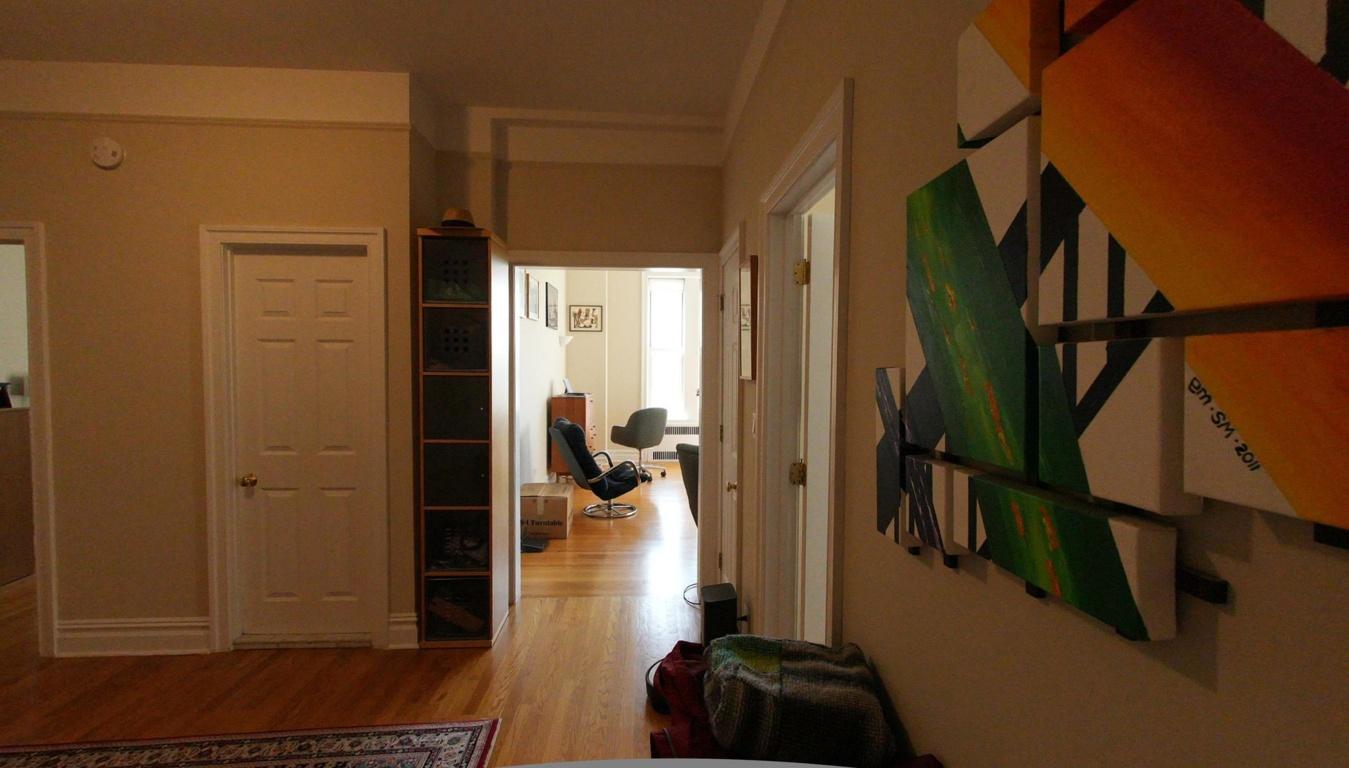}& 
    \includegraphics[width=0.158\linewidth]{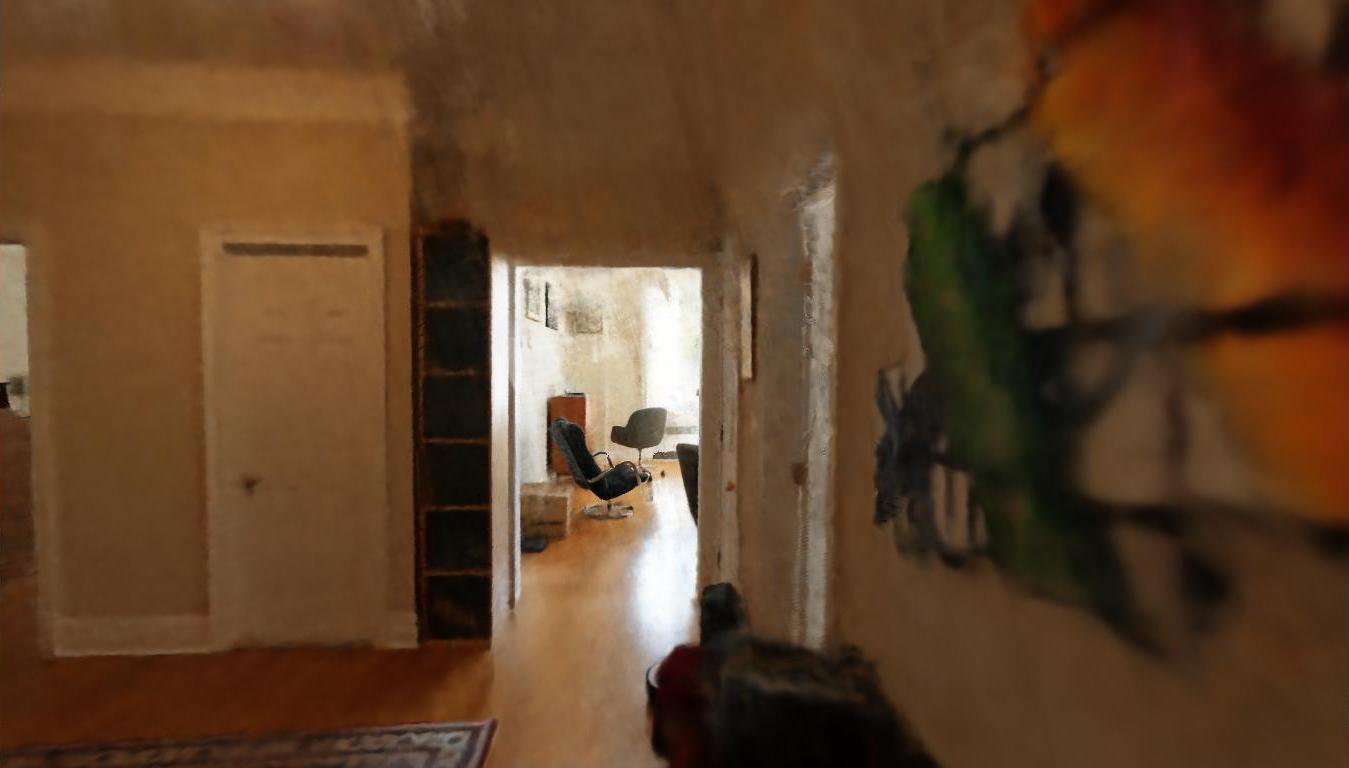}& 
    \includegraphics[width=0.158\linewidth]{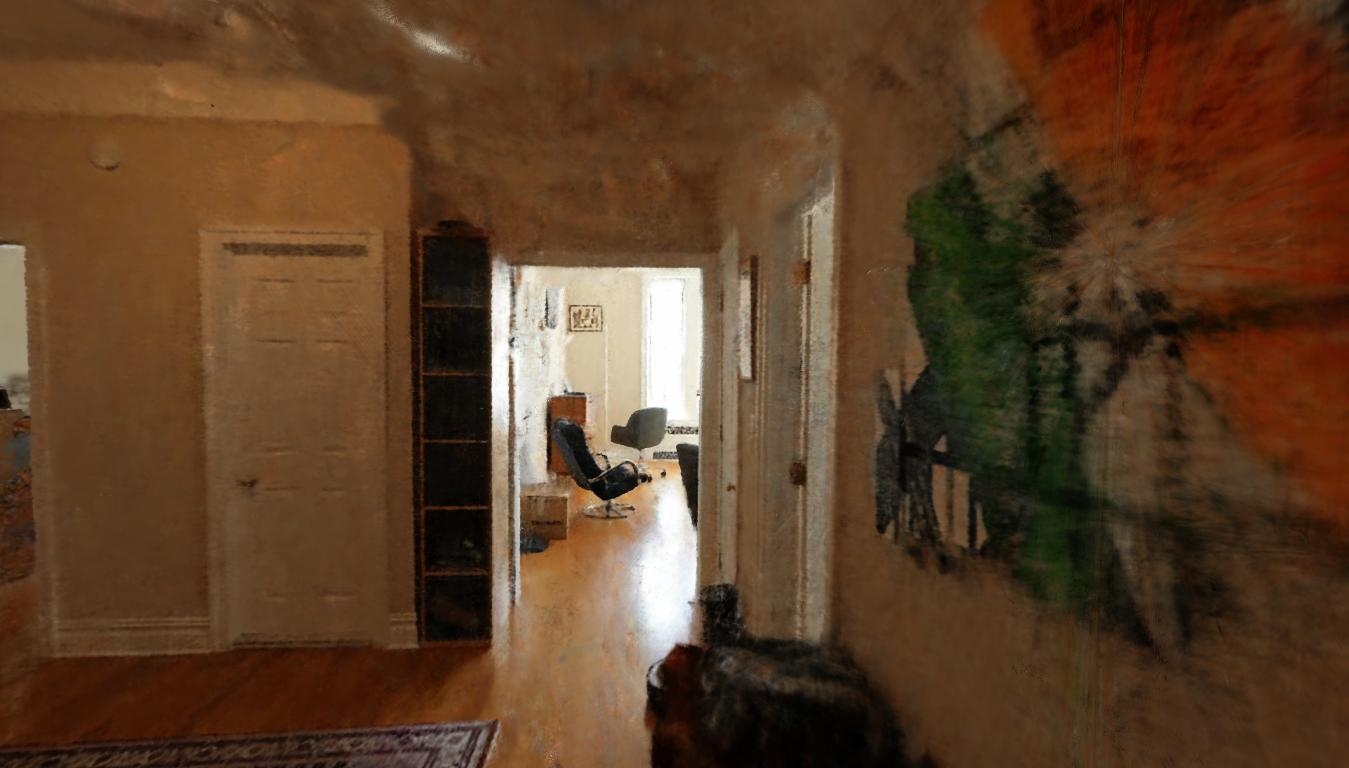}& 
    \includegraphics[width=0.158\linewidth]{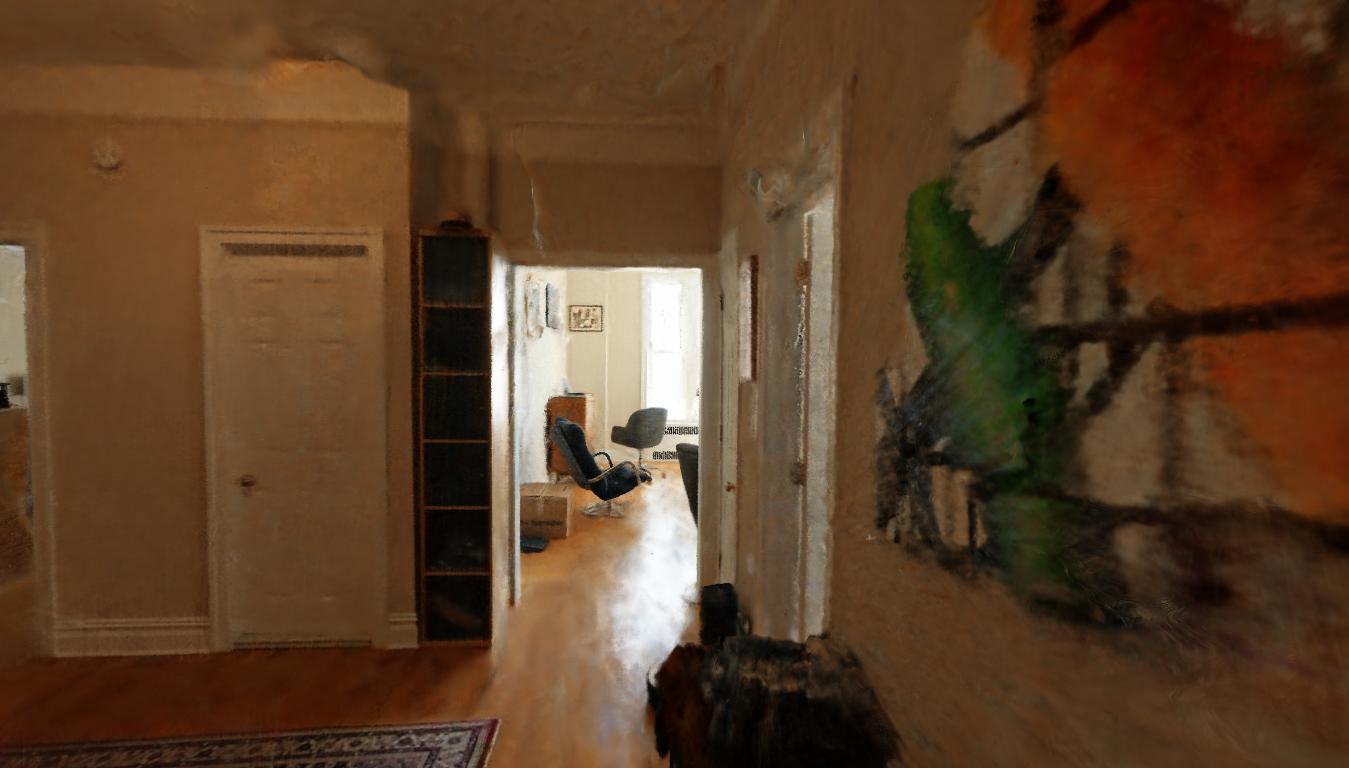}&
    \includegraphics[width=0.158\linewidth]{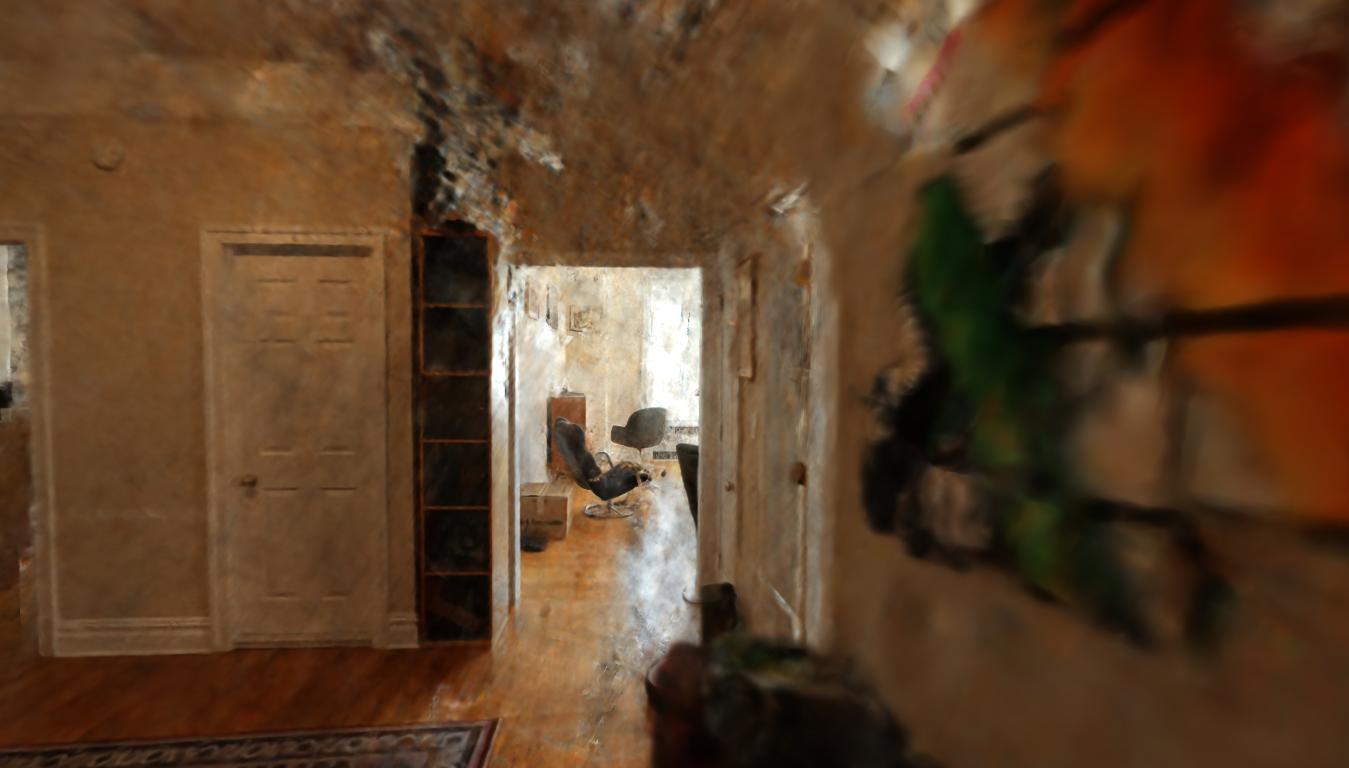}&
    \includegraphics[width=0.158\linewidth]{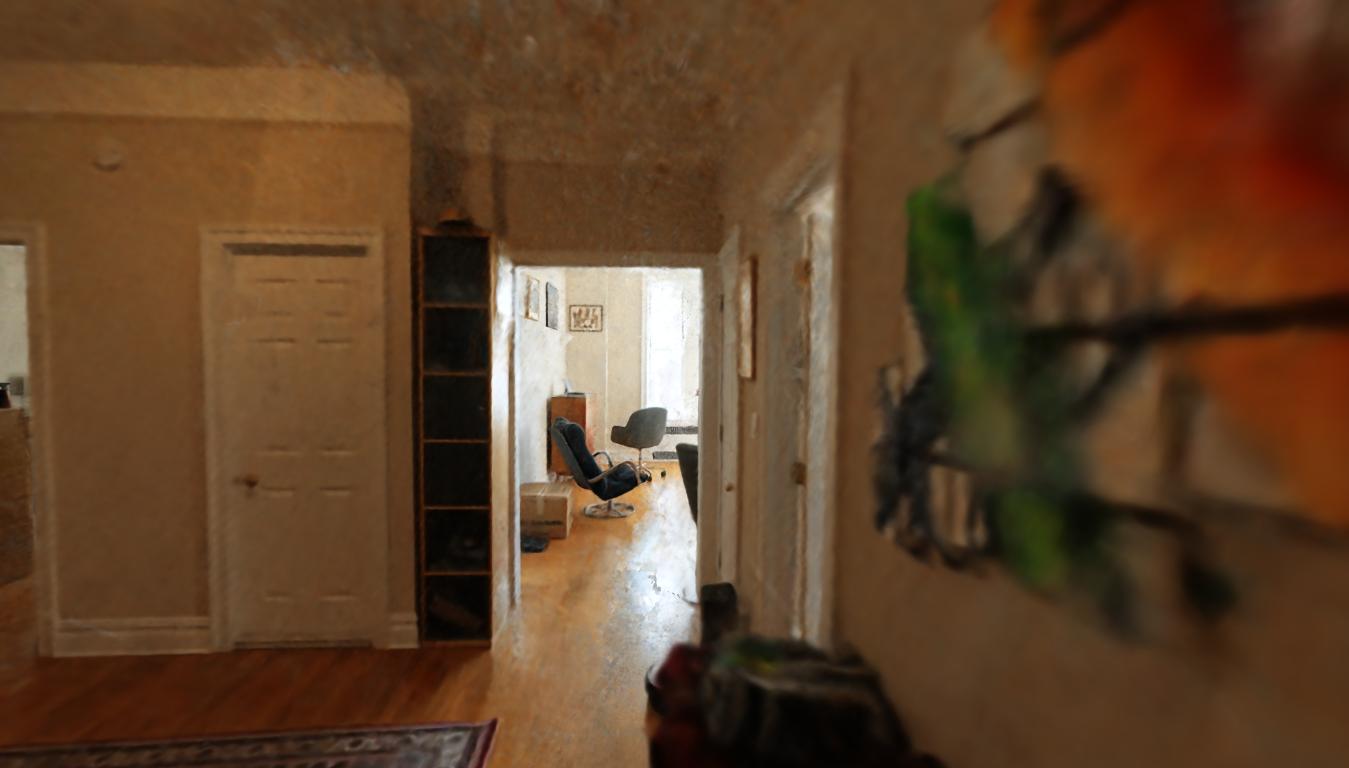}\\

    \includegraphics[width=0.158\linewidth]{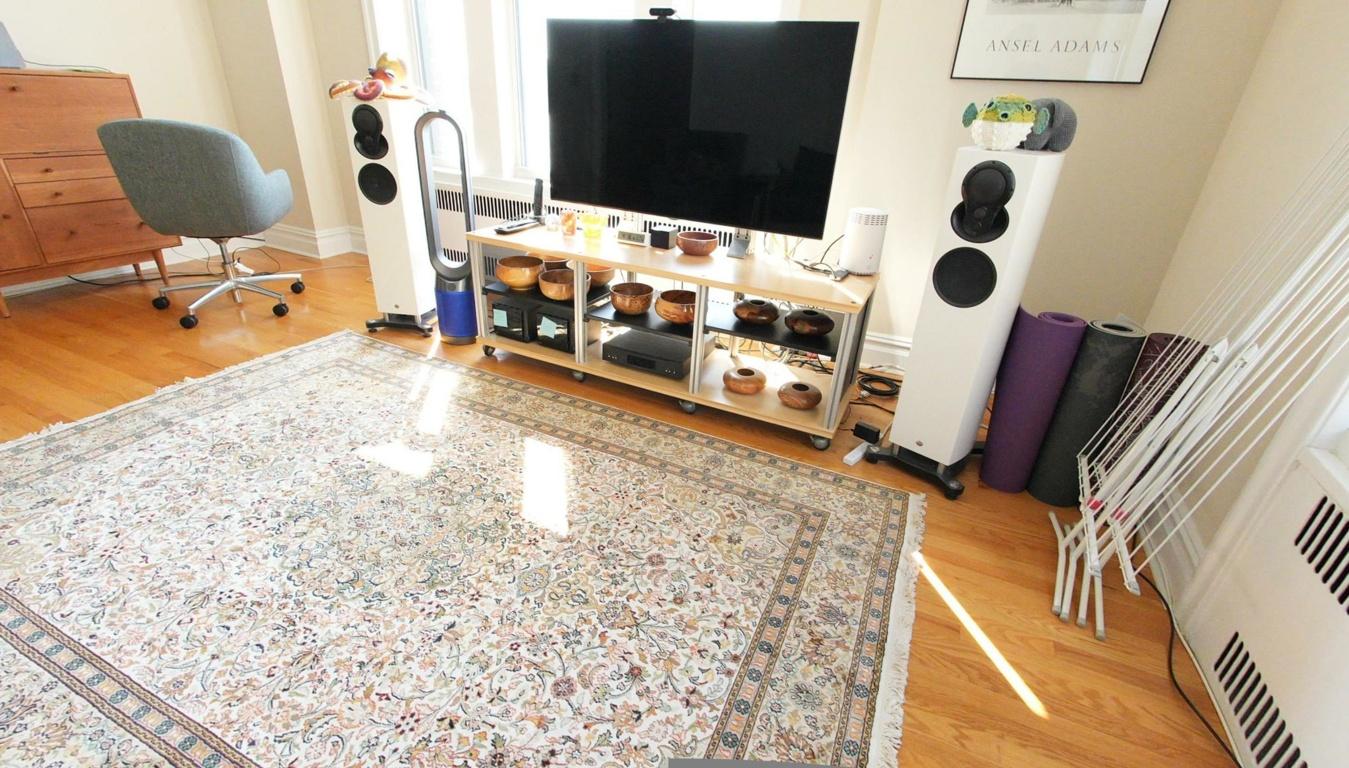}& 
    \includegraphics[width=0.158\linewidth]{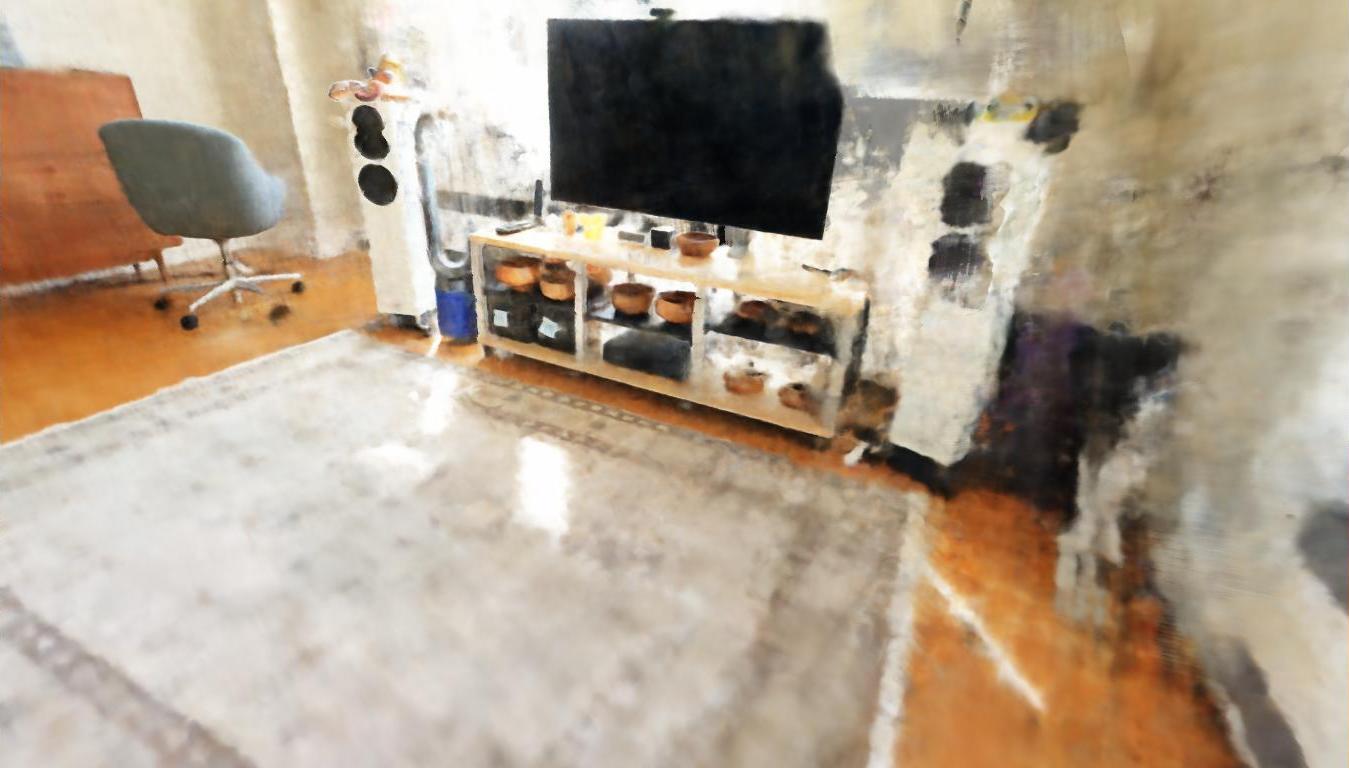}& 
    \includegraphics[width=0.158\linewidth]{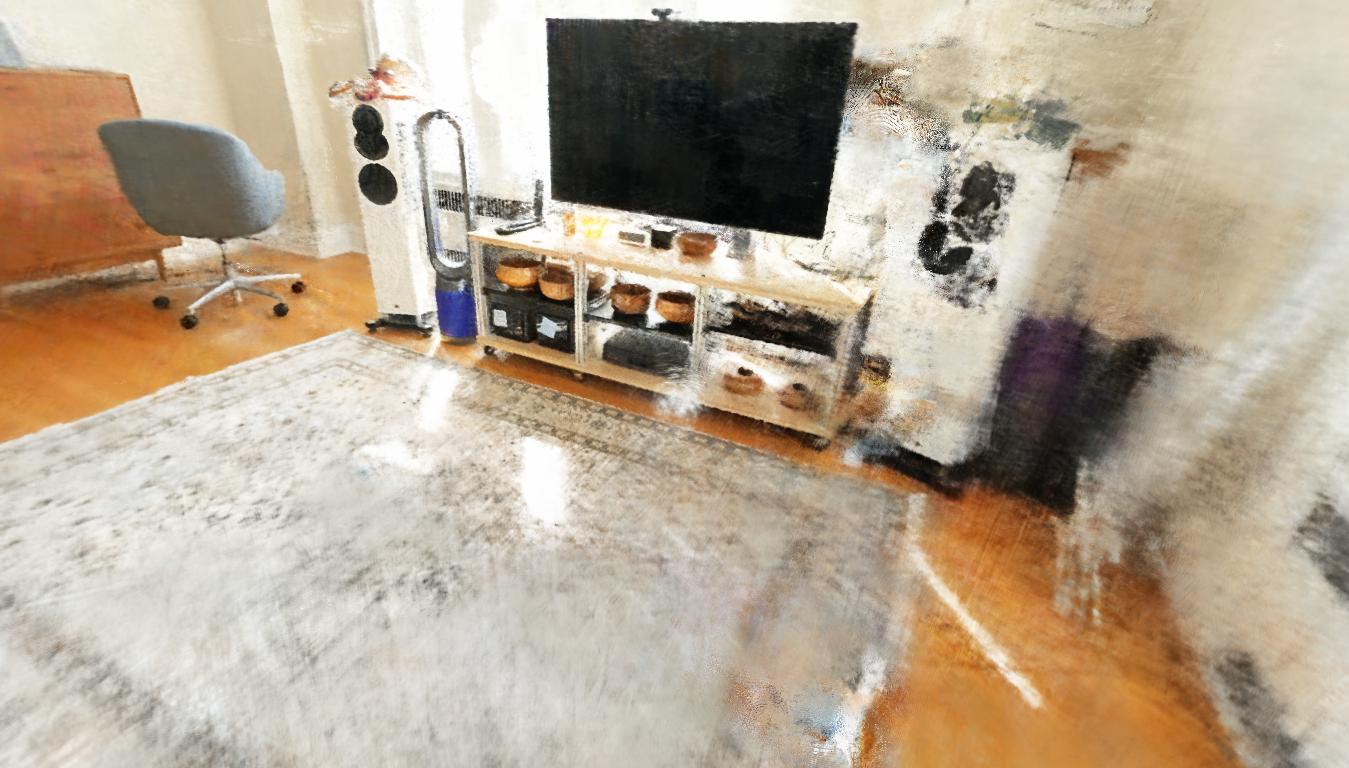}& 
    \includegraphics[width=0.158\linewidth]{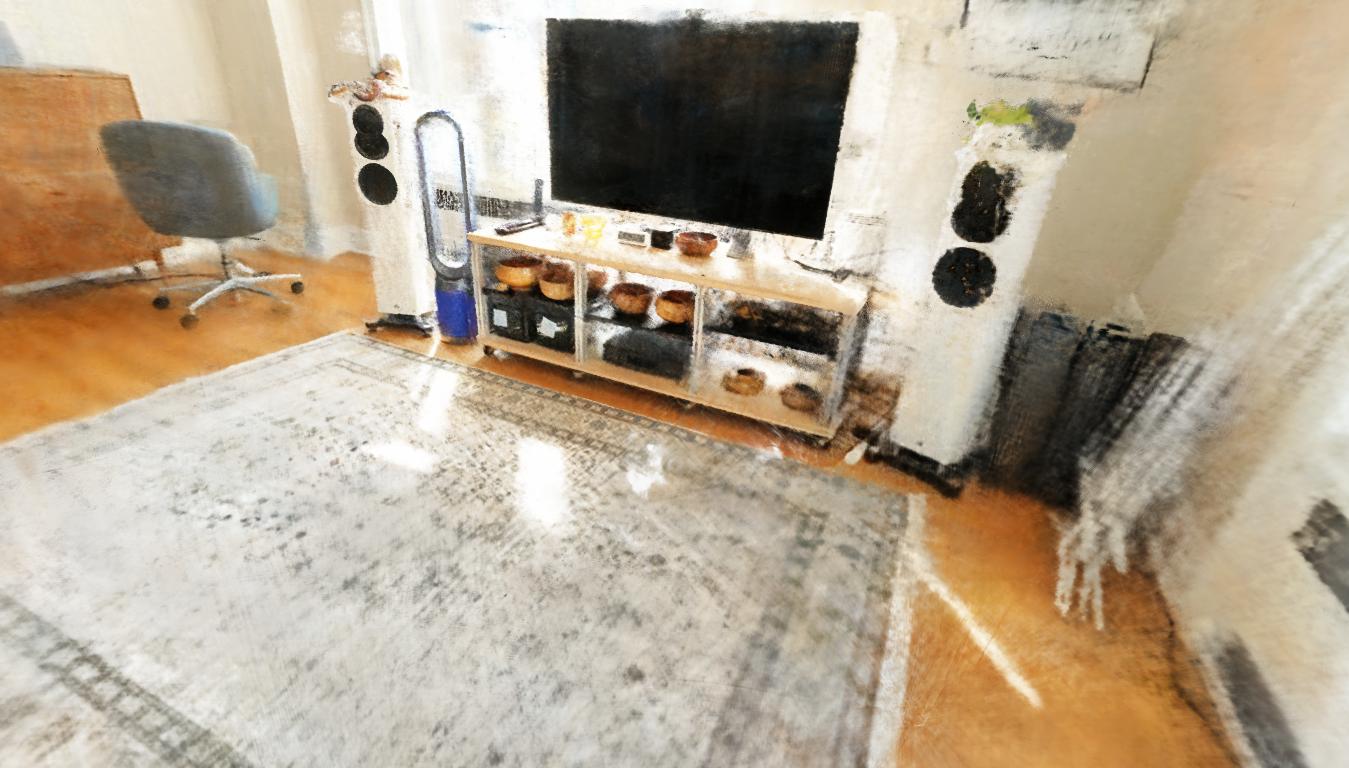}&
    \includegraphics[width=0.158\linewidth]{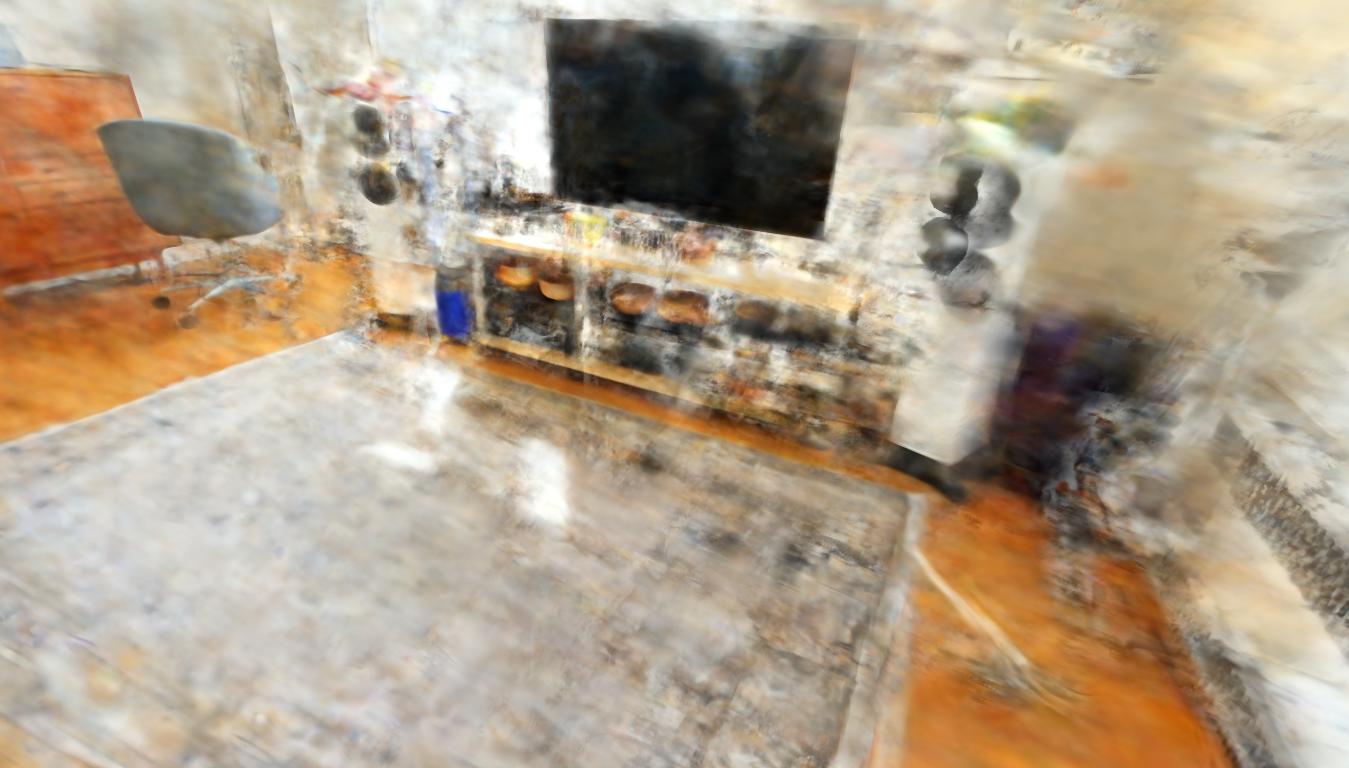}&
    \includegraphics[width=0.158\linewidth]{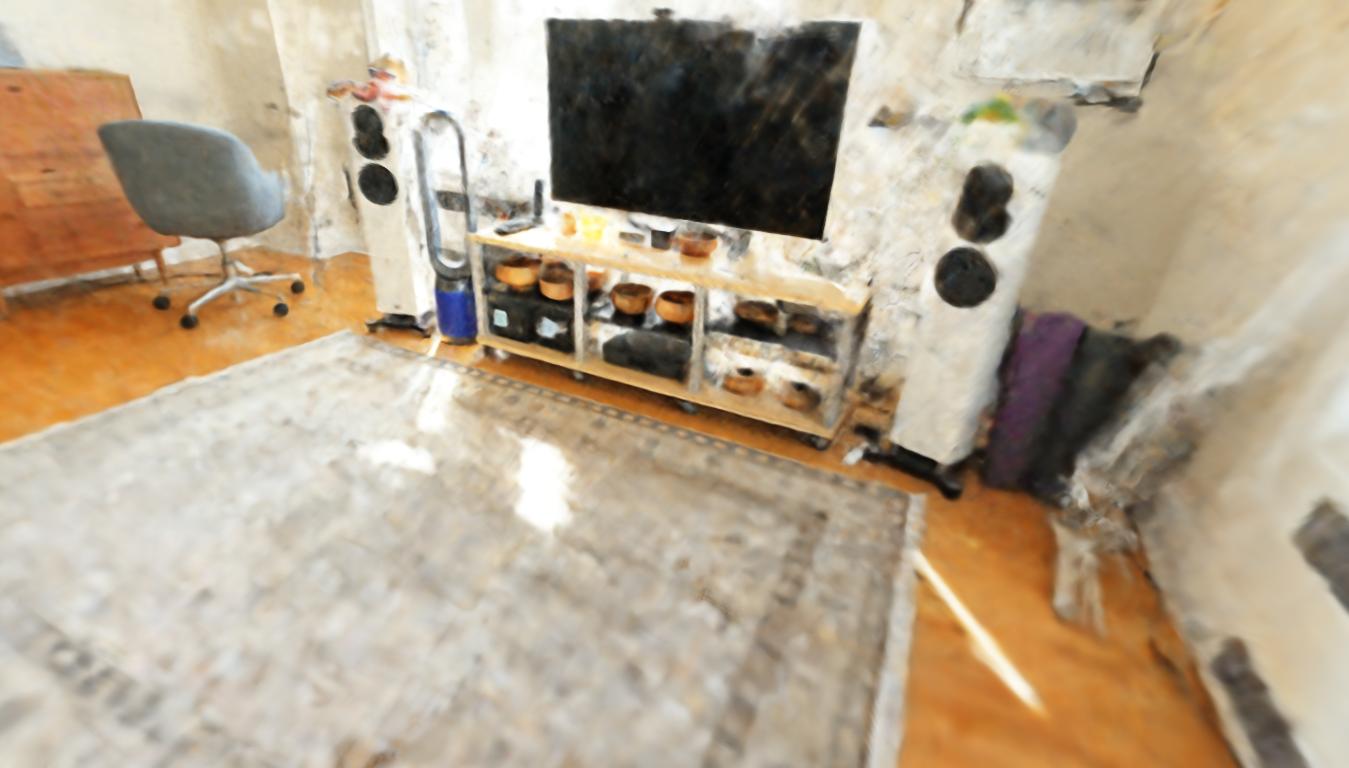}\\

    \includegraphics[width=0.158\linewidth]{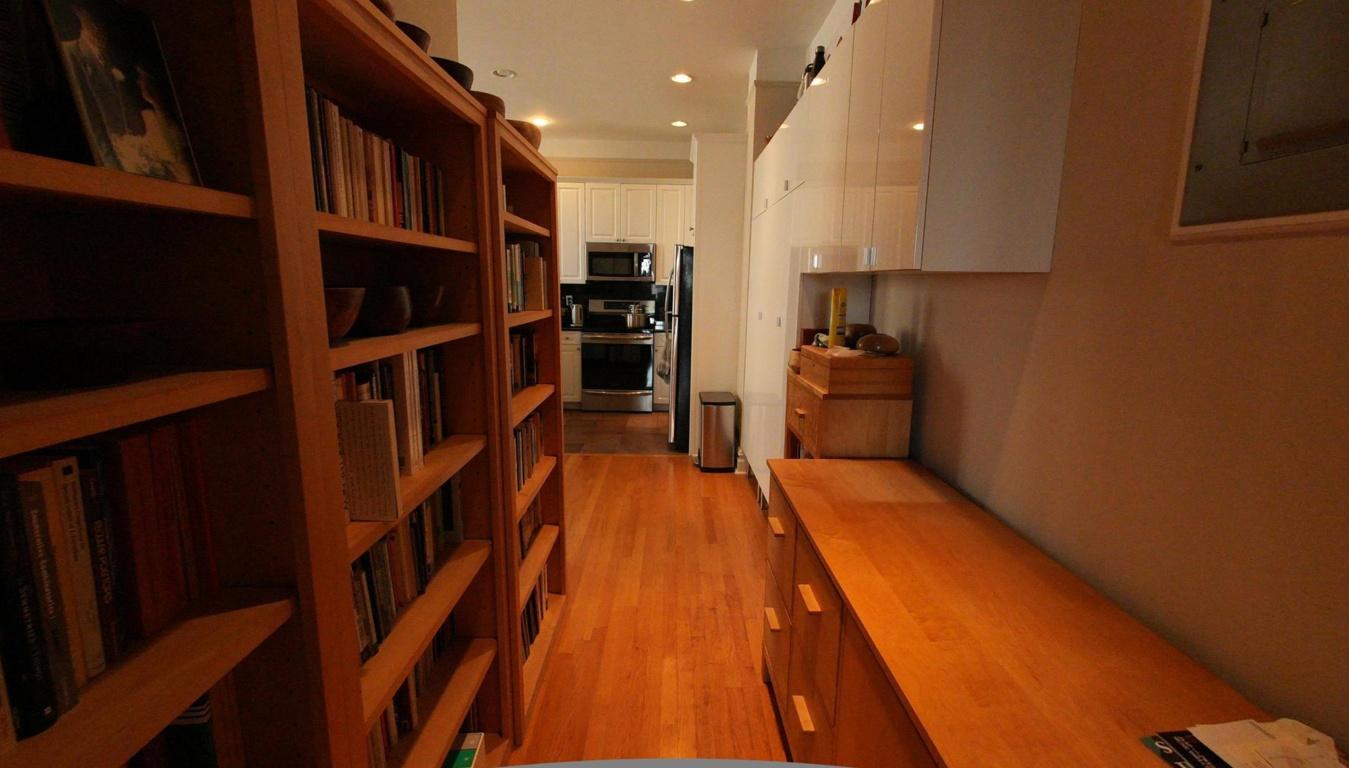}& 
    \includegraphics[width=0.158\linewidth]{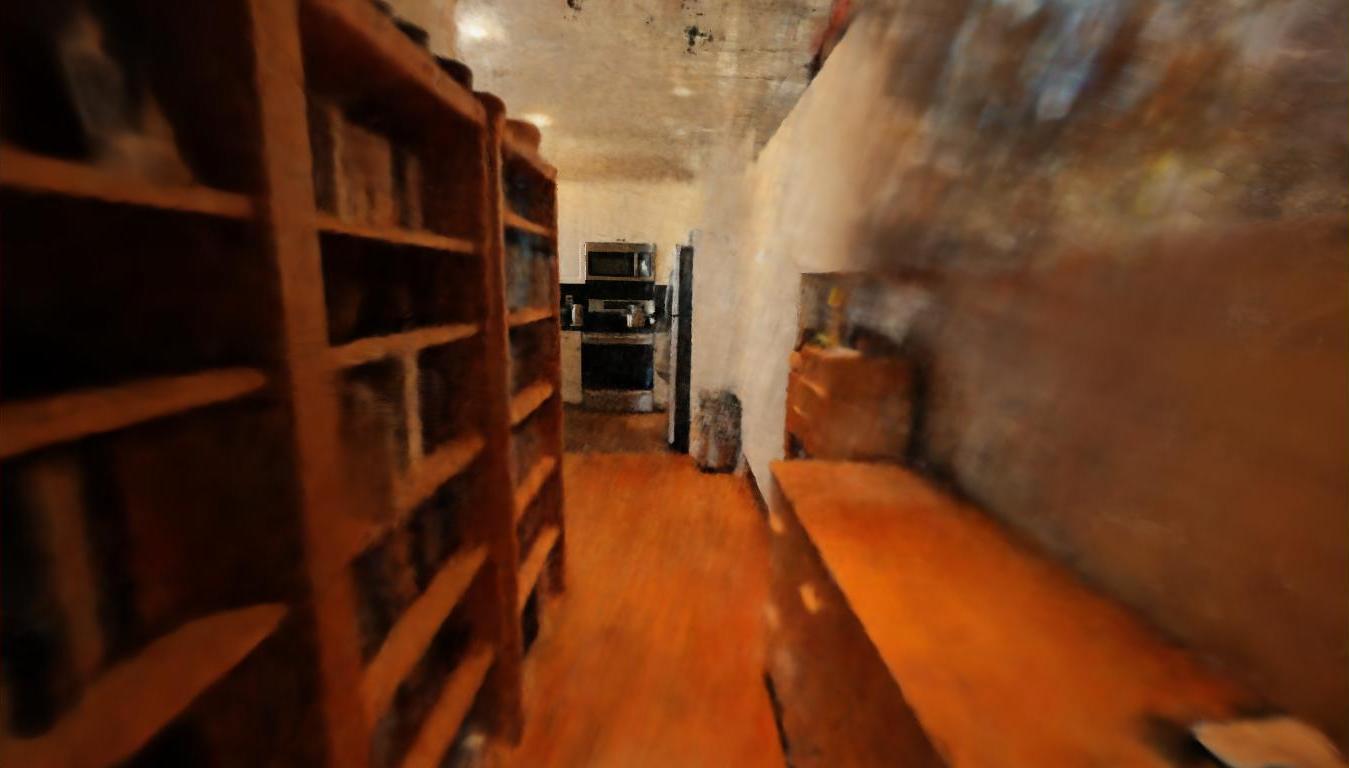}& 
    \includegraphics[width=0.158\linewidth]{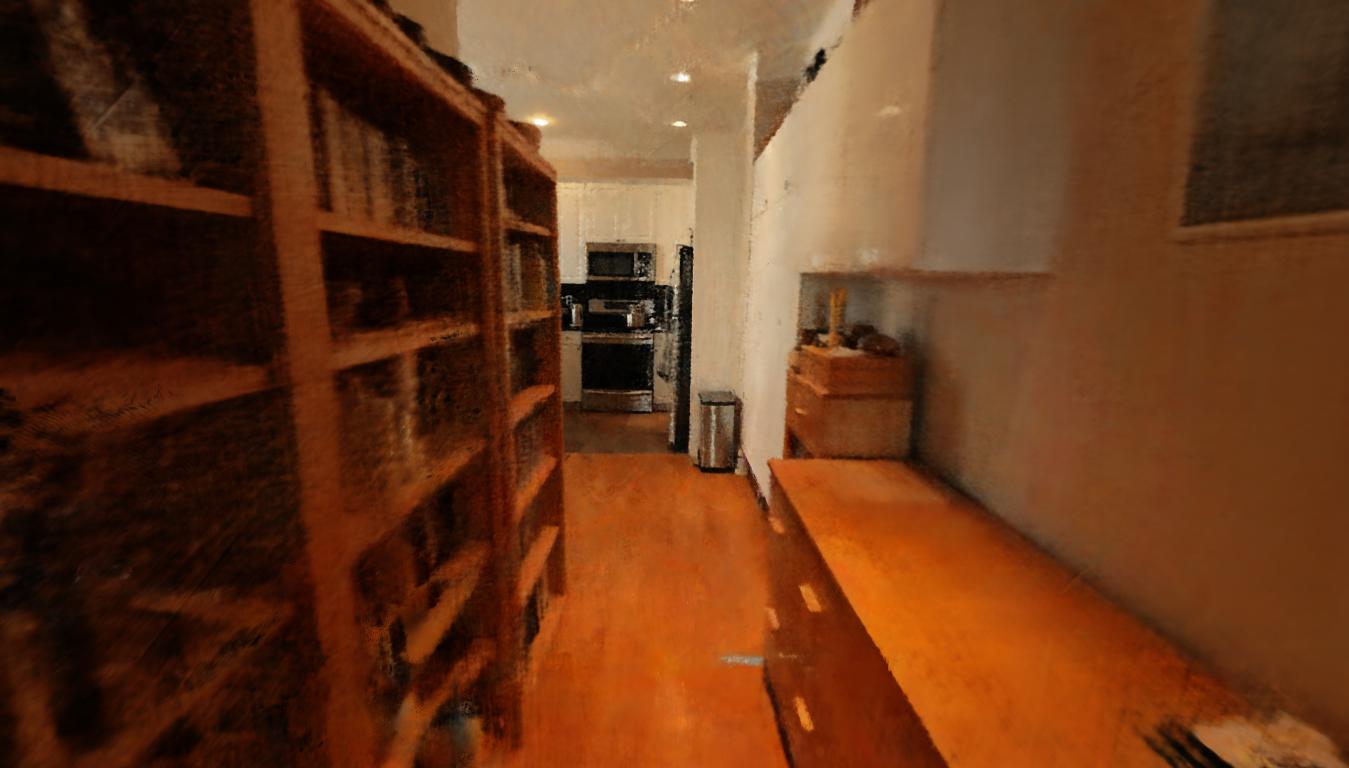}& 
    \includegraphics[width=0.158\linewidth]{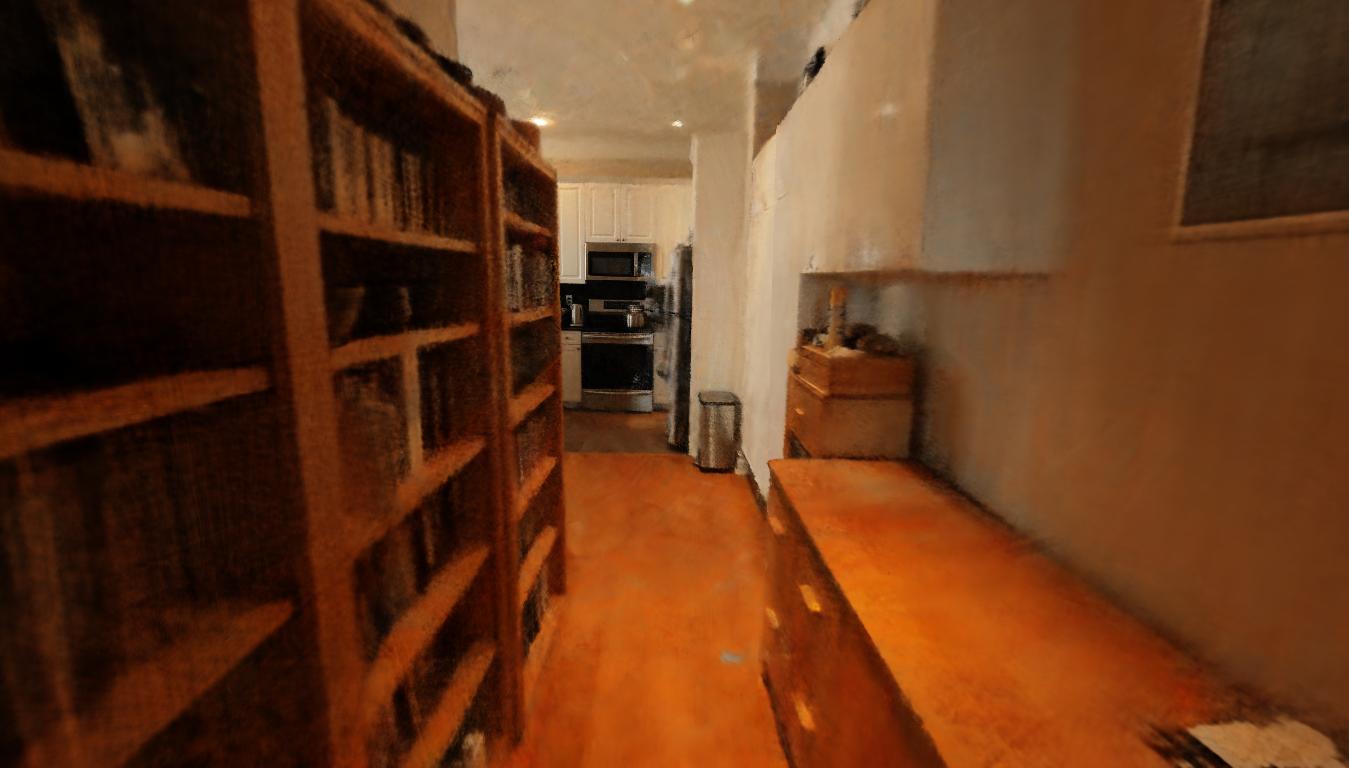}&
    \includegraphics[width=0.158\linewidth]{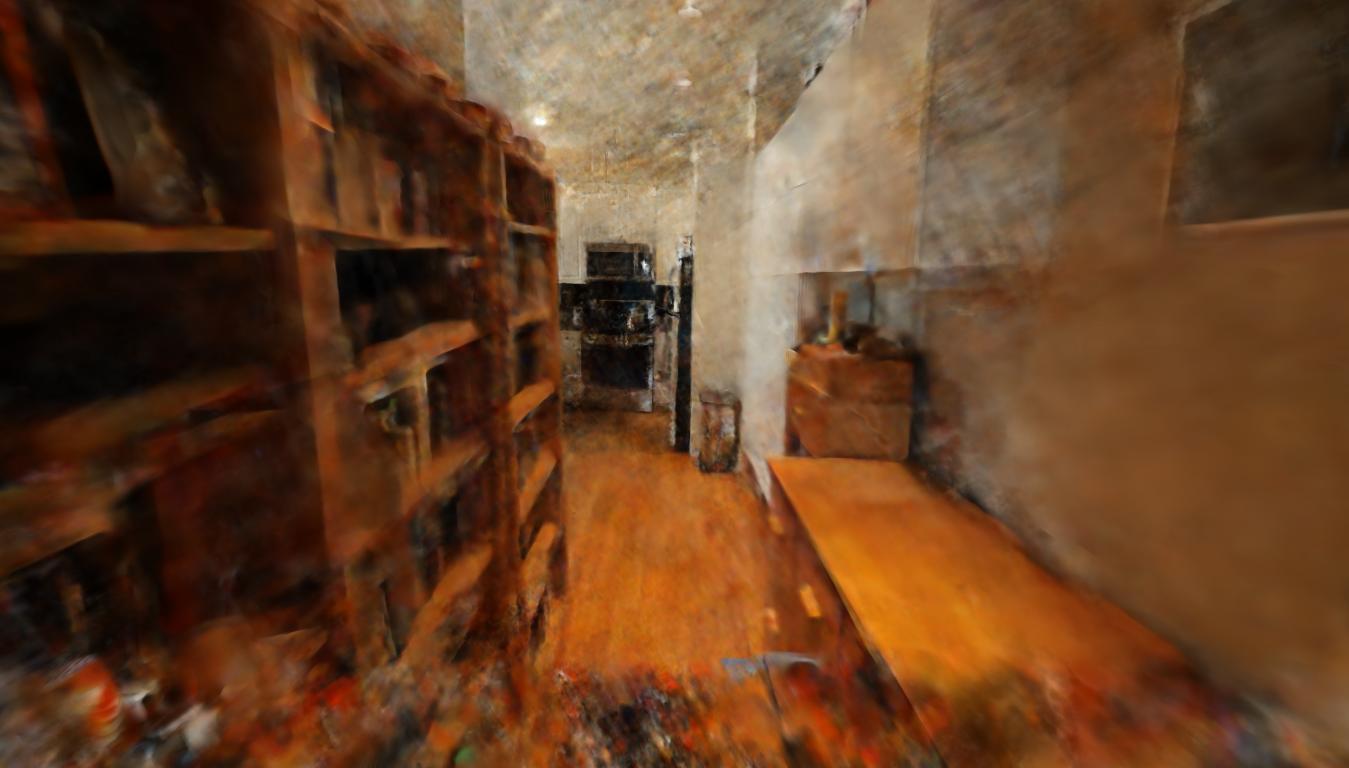}&
    \includegraphics[width=0.158\linewidth]{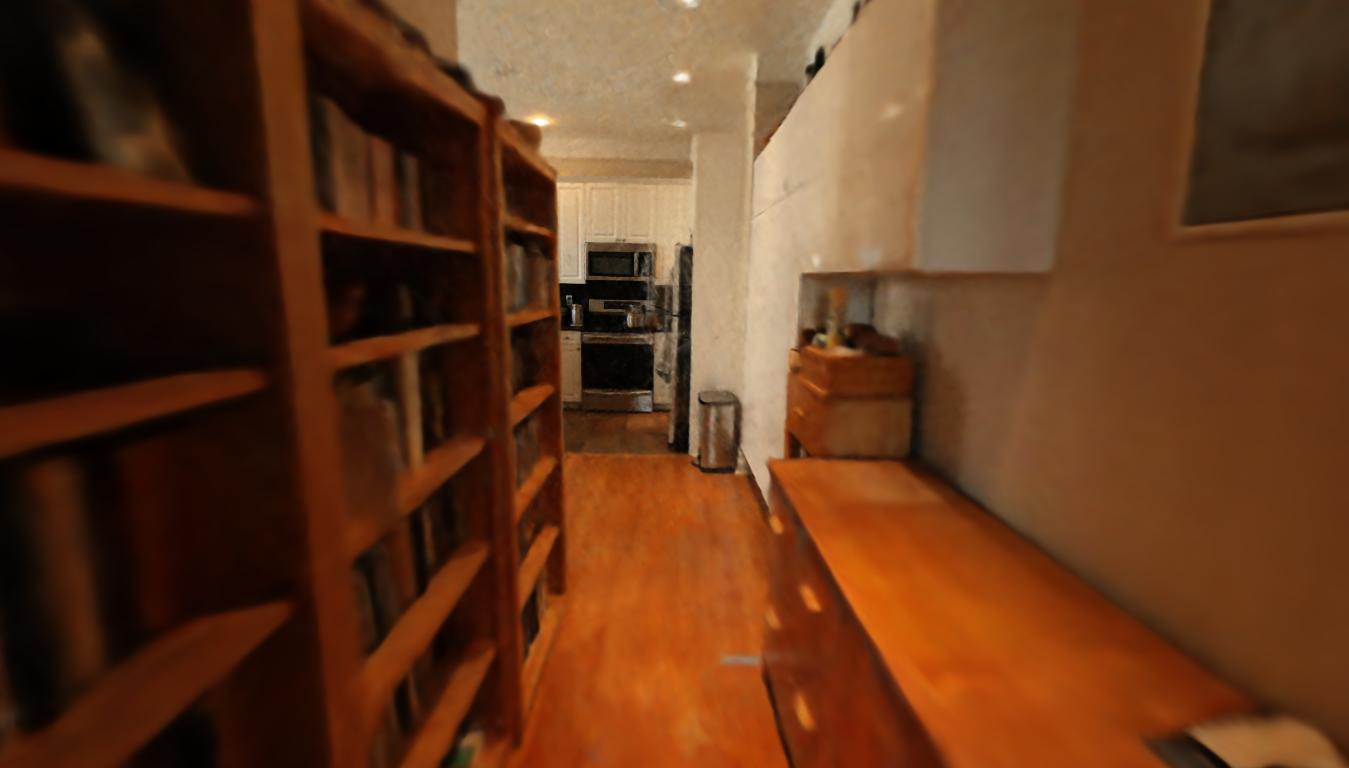}\\

    \end{tabular}
    \caption{\textbf{Additional visualizations of novel view synthesis results on Zip-NeRF~\cite{zip-nerf} dataset.}}
    \label{fig:suppl_qualitative_zipnerf}
\end{figure*}

We present additional qualitative results for both ScanNet++~\cite{scannetpp} and Zip-NeRF~\cite{zip-nerf} datasets in Figure~\ref{fig:suppl_qualitative_scannet++} and Figure~\ref{fig:suppl_qualitative_zipnerf}. 
All six scenes from ScanNet++ and four scenes from the Zip-NeRF dataset are illustrated.
By incorporating our optimal basis placements and geometric regularization into prior scene representations, we achieve a notable reduction of artifacts, particularly in textureless or under-constrained regions where traditional NeRF-based methods often struggle.
Compared to the baselines, our method produces cleaner geometry, and more photorealistic synthesis under novel viewpoints.
This improvement highlights the effectiveness of our method in reducing ambiguity and enabling stable reconstructions across diverse scene layouts.

\clearpage

%

\if 0 % arXiv
\clearpage
\clearpage
\bibliographystyle{IEEEtran}
\bibliography{main} 

@article{llff,
  title={Local Light Field Fusion: Practical View Synthesis with Prescriptive Sampling Guidelines},
  author={Ben Mildenhall and Pratul P. Srinivasan and Rodrigo Ortiz-Cayon and Nima Khademi Kalantari and Ravi Ramamoorthi and Ren Ng and Abhishek Kar},
  journal={ACM Transactions on Graphics (TOG)},
  year={2019},
}

@article{acorn,
  title={ACORN: {Adaptive} coordinate networks for neural scene representation},
  author={Julien N. P. Martel and David B. Lindell and Connor Z. Lin and Eric R. Chan and Marco Monteiro and Gordon Wetzstein},
  journal={ACM Trans. Graph. (SIGGRAPH)},
  volume={40},
  number={4},
  year={2021},
}

@article{nsvf,
  title={Neural Sparse Voxel Fields},
  author={Liu, Lingjie and Gu, Jiatao and Lin, Kyaw Zaw and Chua, Tat-Seng and Theobalt, Christian},
  journal={NeurIPS},
  year={2020}
}

@inproceedings{nerf,
  title={NeRF: Representing Scenes as Neural Radiance Fields for View Synthesis},
  author={Ben Mildenhall and Pratul P. Srinivasan and Matthew Tancik and Jonathan T. Barron and Ravi Ramamoorthi and Ren Ng},
  year={2020},
  booktitle={ECCV},
}

@article{nerf++,
  title={Nerf++: Analyzing and improving neural radiance fields},
  author={Zhang, Kai and Riegler, Gernot and Snavely, Noah and Koltun, Vladlen},
  journal={arXiv preprint arXiv:2010.07492},
  year={2020}
}

@article{instant-ngp,
    author = {Thomas M\"uller and Alex Evans and Christoph Schied and Alexander Keller},
    title = {Instant Neural Graphics Primitives with a Multiresolution Hash Encoding},
    journal = {ACM Trans. Graph.},
    issue_date = {July 2022},
    volume = {41},
    number = {4},
    month = jul,
    year = {2022},
    pages = {102:1--102:15},
    articleno = {102},
    numpages = {15},
    url = {https://doi.org/10.1145/3528223.3530127},
    doi = {10.1145/3528223.3530127},
    publisher = {ACM},
    address = {New York, NY, USA},
}

@inproceedings{plenoxels,
      title={Plenoxels: Radiance Fields without Neural Networks}, 
      author={Sara Fridovich-Keil and Alex Yu and Matthew Tancik and Qinhong Chen and Benjamin Recht and Angjoo Kanazawa},
      year={2022},
      booktitle={CVPR},
}

@inproceedings{plenoctrees,
      title={{PlenOctrees} for Real-time Rendering of Neural Radiance Fields},
      author={Alex Yu and Ruilong Li and Matthew Tancik and Hao Li and Ren Ng and Angjoo Kanazawa},
      year={2021},
      booktitle={ICCV},
}

@article{zip-nerf,
    title={Zip-NeRF: Anti-Aliased Grid-Based Neural Radiance Fields},
    author={Jonathan T. Barron and Ben Mildenhall and 
            Dor Verbin and Pratul P. Srinivasan and Peter Hedman},
    journal={ICCV},
    year={2023}
}

@article{mip-nerf360,
    title={Mip-NeRF 360: Unbounded Anti-Aliased Neural Radiance Fields},
    author={Jonathan T. Barron and Ben Mildenhall and 
            Dor Verbin and Pratul P. Srinivasan and Peter Hedman},
    journal={CVPR},
    year={2022}
}

@InProceedings{egonerf,
    author    = {Choi, Changwoon and Kim, Sang Min and Kim, Young Min},
    title     = {Balanced Spherical Grid for Egocentric View Synthesis},
    booktitle = {Proceedings of the IEEE/CVF Conference on Computer Vision and Pattern Recognition (CVPR)},
    month     = {June},
    year      = {2023},
    pages     = {16590-16599}
}

@inproceedings{nerfstudio,
	title        = {Nerfstudio: A Modular Framework for Neural Radiance Field Development},
	author       = {
		Tancik, Matthew and Weber, Ethan and Ng, Evonne and Li, Ruilong and Yi, Brent
		and Kerr, Justin and Wang, Terrance and Kristoffersen, Alexander and Austin,
		Jake and Salahi, Kamyar and Ahuja, Abhik and McAllister, David and Kanazawa,
		Angjoo
	},
	year         = 2023,
	booktitle    = {ACM SIGGRAPH 2023 Conference Proceedings},
	series       = {SIGGRAPH '23}
}

@inproceedings{nelf-pro,
      author = {Zinuo You and Andreas Geiger and Anpei Chen},
      title = {NeLF-Pro: Neural Light Field Probes for Multi-Scale Novel View Synthesis},
      booktitle = {Conference on Computer Vision and Pattern Recognition (CVPR)},
      year = {2024}
}

@article{f2nerf,
  title={F2-NeRF: Fast Neural Radiance Field Training with Free Camera Trajectories},
  author={Wang, Peng and Liu, Yuan and Chen, Zhaoxi and Liu, Lingjie and Liu, Ziwei and Komura, Taku and Theobalt, Christian and Wang, Wenping},
  journal={CVPR},
  year={2023}
}

@inproceedings{localrf,
  author    = {Meuleman, Andreas and Liu, Yu-Lun and Gao, Chen and Huang, Jia-Bin and Kim, Changil and Kim, Min H. and Kopf, Johannes},
  title     = {Progressively Optimized Local Radiance Fields for Robust View Synthesis},
  booktitle = {CVPR},
  year      = {2023},
}

@InProceedings{mega-nerf,
    author    = {Turki, Haithem and Ramanan, Deva and Satyanarayanan, Mahadev},
    title     = {Mega-NERF: Scalable Construction of Large-Scale NeRFs for Virtual Fly-Throughs},
    booktitle = {Proceedings of the IEEE/CVF Conference on Computer Vision and Pattern Recognition (CVPR)},
    month     = {June},
    year      = {2022},
    pages     = {12922-12931}
}

@article{block-nerf,
  title={Block-NeRF: Scalable Large Scene Neural View Synthesis},
  author={Matthew Tancik and Vincent Casser and Xinchen Yan and Sabeek Pradhan and Ben Mildenhall and Pratul P. Srinivasan and Jonathan T. Barron and Henrik Kretzschmar},
  journal={arXiv},
  year={2022}
}

@inproceedings{nerfvs,
  title={Nerfvs: Neural radiance fields for free view synthesis via geometry scaffolds},
  author={Yang, Chen and Li, Peihao and Zhou, Zanwei and Yuan, Shanxin and Liu, Bingbing and Yang, Xiaokang and Qiu, Weichao and Shen, Wei},
  booktitle={Proceedings of the IEEE/CVF Conference on Computer Vision and Pattern Recognition},
  pages={16549--16558},
  year={2023}
}

@inproceedings{nerf-xl,
  title={{NeRF-XL}: Scaling NeRFs with Multiple {GPUs}},
  author={Ruilong Li and Sanja Fidler and Angjoo Kanazawa and Francis Williams},
  year={2024},
  booktitle={European Conference on Computer Vision (ECCV)},
}

@inproceedings{ddp,
    title={Dense Depth Priors for Neural Radiance Fields from Sparse Input Views}, 
    author={Barbara Roessle and Jonathan T. Barron and Ben Mildenhall and Pratul P. Srinivasan and Matthias Nie{\ss}ner},
    booktitle={Proceedings of the IEEE/CVF Conference on Computer Vision and Pattern Recognition (CVPR)},
    month={June},
    year={2022}
}

@inproceedings{scade,
      title = {SCADE: NeRFs from Space Carving with Ambiguity-Aware Depth Estimates},
      author = {Mikaela Angelina Uy and Ricardo Martin-Brualla and Leonidas Guibas and Ke Li},
      booktitle = {Conference on Computer Vision and Pattern Recognition (CVPR)},
      year = {2023}
}

@InProceedings{dsnerf,
    author    = {Deng, Kangle and Liu, Andrew and Zhu, Jun-Yan and Ramanan, Deva},
    title     = {Depth-supervised {NeRF}: Fewer Views and Faster Training for Free},
    booktitle = {Proceedings of the IEEE/CVF Conference on Computer Vision and Pattern Recognition (CVPR)},
    month     = {June},
    year      = {2022}
}

@InProceedings{regnerf,
          author    = {Michael Niemeyer and Jonathan T. Barron and Ben Mildenhall and Mehdi S. M. Sajjadi and Andreas Geiger and Noha Radwan},  
          title     = {RegNeRF: Regularizing Neural Radiance Fields for View Synthesis from Sparse Inputs},
          booktitle = {Proc. IEEE Conf. on Computer Vision and Pattern Recognition (CVPR)},
          year      = {2022},
}

@article{sparsenerf,
    title={SparseNeRF: Distilling Depth Ranking for Few-shot Novel View Synthesis},
    author={Guangcong and Zhaoxi Chen and Chen Change Loy and Ziwei Liu},
    journal={IEEE/CVF International Conference on Computer Vision (ICCV)},
    year={2023}
}

@article{vip-nerf,
    title = {{ViP-NeRF}: Visibility Prior for Sparse Input Neural Radiance Fields},
    author = {Somraj, Nagabhushan and Soundararajan, Rajiv},
    booktitle = {ACM Special Interest Group on Computer Graphics and Interactive Techniques (SIGGRAPH)},
    journal = {ACM Trans. Graph.},
    month = {August},
    year = {2023},
    doi = {10.1145/3588432.3591539},
}

@inproceedings{activenerf,
  title={ActiveNeRF: Learning Where to See with Uncertainty Estimation},
  author={Pan, Xuran and Lai, Zihang and Song, Shiji and Huang, Gao},
  booktitle={Computer Vision--ECCV 2022: 17th European Conference, Tel Aviv, Israel, October 23--27, 2022, Proceedings, Part XXXIII},
  pages={230--246},
  year={2022},
  organization={Springer}
}

@inproceedings {progressive_camera_placement,
		booktitle = {Vision, Modeling, and Visualization},
		editor = {Guthe, Michael and Grosch, Thorsten},
		title = {{Improving NeRF Quality by Progressive Camera Placement for Free-Viewpoint Navigation}},
		author = {Kopanas, Georgios and Drettakis, George},
		year = {2023},
		publisher = {The Eurographics Association},
		ISBN = {978-3-03868-232-5},
		DOI = {10.2312/vmv.20231222}
}

@InProceedings{NeRF_Director,
    author    = {Xiao, Wenhui and Santa Cruz, Rodrigo and Ahmedt-Aristizabal, David and Salvado, Olivier and Fookes, Clinton and Lebrat, Leo},
    title     = {NeRF Director: Revisiting View Selection in Neural Volume Rendering},
    booktitle = {Proceedings of the IEEE Conference on Computer Vision and Pattern Recognition (CVPR)},
    month     = {June},
    year      = {2024}
}

@inproceedings{nope-nerf,
  title={Nope-nerf: Optimising neural radiance field with no pose prior},
  author={Bian, Wenjing and Wang, Zirui and Li, Kejie and Bian, Jia-Wang and Prisacariu, Victor Adrian},
  booktitle={Proceedings of the IEEE/CVF Conference on Computer Vision and Pattern Recognition},
  pages={4160--4169},
  year={2023}
}

@INPROCEEDINGS{tensorf,
  author = {Anpei Chen and Zexiang Xu and Andreas Geiger and Jingyi Yu and Hao Su},
  title = {TensoRF: Tensorial Radiance Fields},
  booktitle = {European Conference on Computer Vision (ECCV)},
  year = {2022}
}

@inproceedings{neural_visibility_field,
  title={Neural Visibility Field for Uncertainty-Driven Active Mapping},
  author={Xue, Shangjie and Dill, Jesse and Mathur, Pranay and Dellaert, Frank and Tsiotra, Panagiotis and Xu, Danfei},
  booktitle={Proceedings of the IEEE/CVF Conference on Computer Vision and Pattern Recognition},
  pages={18122--18132},
  year={2024}
}

@Article{3dgs,
      author       = {Kerbl, Bernhard and Kopanas, Georgios and Leimk{\"u}hler, Thomas and Drettakis, George},
      title        = {3D Gaussian Splatting for Real-Time Radiance Field Rendering},
      journal      = {ACM Transactions on Graphics},
      number       = {4},
      volume       = {42},
      month        = {July},
      year         = {2023},
      url          = {https://repo-sam.inria.fr/fungraph/3d-gaussian-splatting/}
}

@Article{hierarchical_3dgs,
      author       = {Kerbl, Bernhard and Meuleman, Andreas and Kopanas, Georgios and Wimmer, Michael and Lanvin, Alexandre and Drettakis, George},
      title        = {A Hierarchical 3D Gaussian Representation for Real-Time Rendering of Very Large Datasets},
      journal      = {ACM Transactions on Graphics},
      number       = {4},
      volume       = {43},
      month        = {July},
      year         = {2024},
      url          = {https://repo-sam.inria.fr/fungraph/hierarchical-3d-gaussians/}
}

@inproceedings{FSGS,
  title={Fsgs: Real-time few-shot view synthesis using gaussian splatting},
  author={Zhu, Zehao and Fan, Zhiwen and Jiang, Yifan and Wang, Zhangyang},
  booktitle={European conference on computer vision},
  pages={145--163},
  year={2024},
  organization={Springer}
}

@article{dngaussian,
   title={DNGaussian: Optimizing Sparse-View 3D Gaussian Radiance Fields with Global-Local Depth Normalization},
   author={Li, Jiahe and Zhang, Jiawei and Bai, Xiao and Zheng, Jin and Ning, Xin and Zhou, Jun and Gu, Lin},
   journal={arXiv preprint arXiv:2403.06912},
   year={2024}
}

@inproceedings{omnidata,
  title={Omnidata: A Scalable Pipeline for Making Multi-Task Mid-Level Vision Datasets From 3D Scans},
  author={Eftekhar, Ainaz and Sax, Alexander and Malik, Jitendra and Zamir, Amir},
  booktitle={Proceedings of the IEEE/CVF International Conference on Computer Vision},
  pages={10786--10796},
  year={2021}
}

@inproceedings{volsdf,
  title={Volume rendering of neural implicit surfaces},
  author={Yariv, Lior and Gu, Jiatao and Kasten, Yoni and Lipman, Yaron},
  booktitle={Thirty-Fifth Conference on Neural Information Processing Systems},
  year={2021}
}

@article{monosdf,
  author    = {Yu, Zehao and Peng, Songyou and Niemeyer, Michael and Sattler, Torsten and Geiger, Andreas},
  title     = {MonoSDF: Exploring Monocular Geometric Cues for Neural Implicit Surface Reconstruction},
  journal   = {Advances in Neural Information Processing Systems (NeurIPS)},
  year      = {2022},
}

@inproceedings{marching_cubes,
author = {Lorensen, William E. and Cline, Harvey E.},
title = {Marching cubes: A high resolution 3D surface construction algorithm},
year = {1987},
isbn = {0897912276},
publisher = {Association for Computing Machinery},
address = {New York, NY, USA},
url = {https://doi.org/10.1145/37401.37422},
doi = {10.1145/37401.37422},
booktitle = {Proceedings of the 14th Annual Conference on Computer Graphics and Interactive Techniques},
pages = {163–169},
numpages = {7},
series = {SIGGRAPH '87}
}

@inproceedings{colmap_sfm,
    author={Sch\"{o}nberger, Johannes Lutz and Frahm, Jan-Michael},
    title={Structure-from-Motion Revisited},
    booktitle={Conference on Computer Vision and Pattern Recognition (CVPR)},
    year={2016},
}

@inproceedings{colmap_mvs,
    author={Sch\"{o}nberger, Johannes Lutz and Zheng, Enliang and Pollefeys, Marc and Frahm, Jan-Michael},
    title={Pixelwise View Selection for Unstructured Multi-View Stereo},
    booktitle={European Conference on Computer Vision (ECCV)},
    year={2016},
}

@ARTICLE {midas,
    author  = "Ren\'{e} Ranftl and Katrin Lasinger and David Hafner and Konrad Schindler and Vladlen Koltun",
    title   = "Towards Robust Monocular Depth Estimation: Mixing Datasets for Zero-Shot Cross-Dataset Transfer",
    journal = "IEEE Transactions on Pattern Analysis and Machine Intelligence",
    year    = "2022",
    volume  = "44",
    number  = "3"
}

@article{dpt,
	author    = {Ren\'{e} Ranftl and Alexey Bochkovskiy and Vladlen Koltun},
	title     = {Vision Transformers for Dense Prediction},
	journal   = {ICCV},
	year      = {2021},
}

@inproceedings{scannetpp,
  title={ScanNet++: A High-Fidelity Dataset of 3D Indoor Scenes},
  author={Yeshwanth, Chandan and Liu, Yueh-Cheng and Nie{\ss}ner, Matthias and Dai, Angela},
  booktitle = {Proceedings of the International Conference on Computer Vision ({ICCV})},
  year={2023}
}

@article{opencv,
  title={The opencv library.},
  author={Bradski, Gary},
  journal={Dr. Dobb's Journal: Software Tools for the Professional Programmer},
  volume={25},
  number={11},
  pages={120--123},
  year={2000},
  publisher={Miller Freeman Inc.}
}

@article{tanks,
    author    = {Arno Knapitsch and Jaesik Park and Qian-Yi Zhou and Vladlen Koltun},
    title     = {Tanks and Temples: Benchmarking Large-Scale Scene Reconstruction},
    journal   = {ACM Transactions on Graphics},
    volume    = {36},
    number    = {4},
    year      = {2017},
}

@article{kitti-360,
   title =  {{KITTI}-360: A Novel Dataset and Benchmarks for Urban Scene Understanding in 2D and 3D},
   author = {Yiyi Liao and Jun Xie and Andreas Geiger},
   journal = {Pattern Analysis and Machine Intelligence (PAMI)},
   year = {2022},
}

@article{open3d,
    author    = {Qian-Yi Zhou and Jaesik Park and Vladlen Koltun},
    title     = {{Open3D}: {A} Modern Library for {3D} Data Processing},
    journal   = {arXiv:1801.09847},
    year      = {2018},
}

@InProceedings{diffnerf,
    author    = {Ehret, Thibaud and Mar{\'\i}, Roger and Facciolo, Gabriele},
    title     = {A Generic and Flexible Regularization Framework for NeRFs},
    booktitle = {Proceedings of the IEEE/CVF Winter Conference on Applications of Computer Vision (WACV)},
    month     = {January},
    year      = {2024},
    pages     = {3088-3097}
}

@misc{MAPAnything,
  title={{MapAnything}: Universal Feed-Forward Metric {3D} Reconstruction},
  author={Nikhil Keetha and Norman M\"{u}ller and Johannes Sch\"{o}nberger and Lorenzo Porzi and Yuchen Zhang and Tobias Fischer and Arno Knapitsch and Duncan Zauss and Ethan Weber and Nelson Antunes and Jonathon Luiten and Manuel Lopez-Antequera and Samuel Rota Bul\`{o} and Christian Richardt and Deva Ramanan and Sebastian Scherer and Peter Kontschieder},
  note={arXiv preprint arXiv:2509.13414},
  year={2025}
}

@inproceedings{vggt,
  title={VGGT: Visual Geometry Grounded Transformer},
  author={Wang, Jianyuan and Chen, Minghao and Karaev, Nikita and Vedaldi, Andrea and Rupprecht, Christian and Novotny, David},
  booktitle={Proceedings of the IEEE/CVF Conference on Computer Vision and Pattern Recognition},
  year={2025}
}

@inproceedings{LongSplat,
 title={Longsplat: Robust unposed 3d gaussian splatting for casual long videos},
 author={Lin, Chin-Yang and Sun, Cheng and Yang, Fu-En and Chen, Min-Hung and Lin, Yen-Yu and Liu, Yu-Lun},
 booktitle={Proceedings of the IEEE/CVF International Conference on Computer Vision},
 pages={27412--27422},
 year={2025}
}

@InProceedings{Mip-Splatting,
    author    = {Yu, Zehao and Chen, Anpei and Huang, Binbin and Sattler, Torsten and Geiger, Andreas},
    title     = {Mip-Splatting: Alias-free 3D Gaussian Splatting},
    booktitle = {Proceedings of the IEEE/CVF Conference on Computer Vision and Pattern Recognition (CVPR)},
    month     = {June},
    year      = {2024},
    pages     = {19447-19456}
}

@inproceedings{Scaffold-GS,
  title={Scaffold-gs: Structured 3d gaussians for view-adaptive rendering},
  author={Lu, Tao and Yu, Mulin and Xu, Linning and Xiangli, Yuanbo and Wang, Limin and Lin, Dahua and Dai, Bo},
  booktitle={Proceedings of the IEEE/CVF Conference on Computer Vision and Pattern Recognition},
  pages={20654--20664},
  year={2024}
}

@article{SGGS,
  title={SGGS: Semantic-Guided 3D Gaussian Splatting With Adaptive Rendering},
  author={Zhou, Annan and Wang, Li and Li, Jian and Huang, Jing and Li, Li and Yao, Jian},
  journal={IEEE Transactions on Visualization and Computer Graphics},
  year={2026},
  publisher={IEEE}
}

@inproceedings{IndoorGS,
  title={Indoorgs: Geometric cues guided gaussian splatting for indoor scene reconstruction},
  author={Ruan, Cong and Wang, Yuesong and Guan, Tao and Zhang, Bin and Ju, Lili},
  booktitle={Proceedings of the Computer Vision and Pattern Recognition Conference},
  pages={844--853},
  year={2025}
}

@article{PlanarGS,
  title={PlanarGS: High-Fidelity Indoor 3D Gaussian Splatting Guided by Vision-Language Planar Priors},
  author={Jin, Xirui and Jin, Renbiao and Li, Boying and Zou, Danping and Yu, Wenxian},
  journal={arXiv preprint arXiv:2510.23930},
  year={2025}
}

@inproceedings{dust3r,
      title={DUSt3R: Geometric 3D Vision Made Easy}, 
      author={Shuzhe Wang and Vincent Leroy and Yohann Cabon and Boris Chidlovskii and Jerome Revaud},
      booktitle = {CVPR},
      year = {2024}
}

@misc{mast3r,
      title={Grounding Image Matching in 3D with MASt3R}, 
      author={Vincent Leroy and Yohann Cabon and Jerome Revaud},
      booktitle = {ECCV},
      year = {2024}
}

@article{FisherRF,
      title={FisherRF: Active View Selection and Uncertainty Quantification for Radiance Fields using Fisher Information},
      author={Wen Jiang and Boshu Lei and Kostas Daniilidis},
      journal={arXiv},
      year={2023}
  }
\fi

\end{document}